\documentclass[twoside]{article}

\usepackage[preprint]{aistats2027}
\usepackage{amsmath,amssymb,amsfonts,amsthm,mathtools}
\usepackage{graphicx}
\usepackage{booktabs}
\usepackage{enumitem}
\usepackage{xcolor}
\usepackage{colortbl}
\AddToHook{env/table/begin}{\setlength{\belowcaptionskip}{6pt}}
\AddToHook{env/table*/begin}{\setlength{\belowcaptionskip}{6pt}}
\usepackage[round]{natbib}

\usepackage{algorithm,algpseudocode}
\usepackage[colorlinks=true,linkcolor=blue,citecolor=blue,urlcolor=blue]{hyperref}
\usepackage{capt-of}
\usepackage{placeins}
\newcommand{\wip}[1]{\par\noindent{\small\color{purple}\textbf{[WIP:} #1\textbf{]}}\par}
\graphicspath{{experiments/figures/}}
\setlist{itemsep=0.1em,topsep=0.2em,leftmargin=1.2em}

\newtheorem{theorem}{Theorem}
\newtheorem{proposition}[theorem]{Proposition}

\newtheorem{corollary}[theorem]{Corollary}

\newtheorem{example}[theorem]{Example}
\theoremstyle{remark}
\newtheorem{remark}[theorem]{Remark}

\DeclareMathOperator{\rank}{rank}
\DeclareMathOperator{\spanop}{span}
\DeclareMathOperator{\diag}{diag}
\DeclareMathOperator{\im}{im}
\DeclareMathOperator{\ri}{ri}

\newcommand{\R}{\mathbb R}

\newcommand{\cC}{\mathcal C}

\newcommand{\cH}{\mathcal H}

\newcommand{\cP}{\mathcal P}

\newcommand{\frz}{\mathrm{frz}}

\newcommand{\g}[1]{{\color{purple}[#1]}}
\definecolor{setupbg}{RGB}{238,243,249}
\definecolor{definitionhighlight}{RGB}{255,242,179}
\newtheoremstyle{highlightdefinition}
  {\topsep}{\topsep}{\itshape}{}{\bfseries}{}{.5em}
  {{\setlength{\fboxsep}{2pt}\colorbox{definitionhighlight}{\strut
    \thmname{#1}\thmnumber{ #2}\thmnote{ \normalfont(#3)}.}}}
\theoremstyle{highlightdefinition}
\newtheorem{maindefinition}[theorem]{Definition}
\theoremstyle{remark}

\newcommand{\todo}[1]{\textcolor{red}{\textbf{TODO:} #1}}
\newcommand{\gc}[1]{\textcolor{orange}{\textbf{[G:} #1\textbf{]}}}
\newcommand{\toedit}[1]{{\color{teal}\textbf{[to edit:} #1\textbf{]}}}

\newcommand{\spn}{\spanop}
\newif\ifpaperreview
\paperreviewfalse

\AtBeginDocument{\ifpaperreview\else
  \renewcommand{\g}[1]{}\renewcommand{\gc}[1]{}\renewcommand{\toedit}[1]{}\renewcommand{\wip}[1]{}\renewcommand{\todo}[1]{}
\fi}
\newcommand{\PaperAuthorsPlain}{Giulia Lanzillotta and Thomas Hofmann}
\newcommand{\PaperRunningAuthors}{Giulia Lanzillotta, Thomas Hofmann}
\newcommand{\PaperAuthors}{Giulia Lanzillotta \And Thomas Hofmann}
\newcommand{\PaperAffiliations}{ETH AI Center, ETH Zurich \And ETH Zurich}

\hypersetup{pdftitle={Flat oceans: vast convex invariance at the end of training},pdfauthor={\PaperAuthorsPlain}}
\begin{document}

\runningtitle{Flat oceans: vast convex invariance at the end of training}
\runningauthor{\PaperRunningAuthors}

\twocolumn[
\aistatstitle{Flat oceans: vast convex invariance at the end of training}
\aistatsauthor{\PaperAuthors}
\aistatsaddress{\PaperAffiliations}
]

\begin{abstract}\emergencystretch=1em
Flatness plays a central role in neural network optimization, Bayesian inference and generalization, yet commonly used characterizations are numerically delicate and limited to local approximations.
We discover and characterize \emph{flat oceans}: large, high-dimensional convex regions in parameter space within which a network’s outputs on a given dataset remain exactly or approximately constant. Our analysis starts from an exact decomposition of the non-linear terms in the network response into a polynomial capturing cross-layer interactions and an activation-dependent residual. Differently from prior work, we approach the question of functional invariance constructively. We design procedures to describe flat oceans around any parameter which run in polynomial-time and use simple linear algebra. Our construction is precise and works across several architectures and activation functions. Experiments show that large invariant spaces emerge and persist during training, with sampled directions remaining approximately invariant far beyond generic parameter directions. 
\end{abstract}

\section{INTRODUCTION}
\label{sec:intro}
\begin{figure}[!htbp]
\centering
\includegraphics[width=\linewidth]{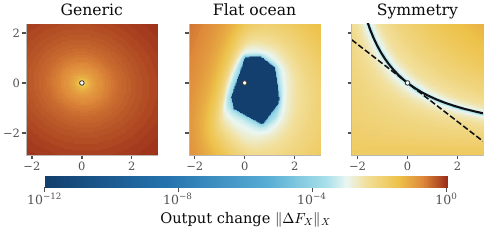}
\caption{Two-dimensional slices of a trained ReLU network in parameter space. Left: two generic directions. Middle: two flat ocean directions. Right: an exact rescaling symmetry orbit (solid) and its straight tangent (dashed). Each axis measures displacement along a unit parameter direction; axis limits and the output-change color scale are shared. White dots mark the reference parameters. \looseness-1}
\label{fig:teaser}\label{fig:sym-vs-ocean}
\end{figure}

Neural network parameter spaces have many redundancies, as the same function can often be encoded by infinitely many parameters. 
The study of \emph{neural network symmetries} and \emph{parameter identifiability} is concerned
with describing these redundancies inherent in the network parametrizations. Substantial progress has been made on this front in recent years. Complete characterizations of functional equivalence have been obtained for classes of networks with tanh and related activations, shallow ReLU networks, and generic three-layer ReLU bottleneck architectures \citep{vlacic2020affine,ramakrishnan2026complete,gegenfurtner2026symmetries}.

These studies primarily concern parameters that represent the same function for all inputs in the domain. When considering instead a specific dataset, the class of redundancies of the \emph{empirical network function} (i.e., measured on the dataset) can be much larger. Prior work studies conditions for identifiability from sample outputs and the dimensionality of the parameter changes that preserve those outputs locally \citep{stock2021embedding,bonapellissier2022local,grigsby2025functional,bonapellissier2024geometry}.  Their descriptions of these changes are generally implicit, and their results depend on assumptions about the network architecture and activation functions.\looseness-1

Closely related questions arise in the study of neural network loss landscapes. In fact, the flat regions of the loss landscape are intimately tied to the invariances of the empirical network function (Appendix~\ref{app:hessian}). Locally, geometry is commonly characterized through the network-output Jacobian and curvature matrices such as the loss Hessian and generalized Gauss--Newton matrix \citep{sagun2017hessian,immer2021linearization}. These objects are at the basis of several methods in optimization, Bayesian inference and continual learning, which rely on approximations of the loss landscape around a given parameter  \citep{martens2015kfac,khan2018vogn,daxberger2021laplace,kirkpatrick2017ewc,farajtabar2020ogd}. 
Despite their widespread use, these local approximations offer limited guidance about flatness at finite distances, and estimating their flat directions is costly and numerically unreliable for most realistic architectures.

Characterizations of empirical functional invariance that remain valid at finite distances have thus countless applications, ranging from optimization to uncertainty quantification and privacy.
Despite this, the question of sample-dependent geometry beyond the infinitesimal reach of gradients and Hessians remains largely open. 
In this work, we develop a constructive approach to empirical functional invariance at large finite distances. Given reference parameters $\theta_0\in\mathbb R^p$, we seek an orthonormal basis $Q\in\mathbb R^{p\times k}$ and a radius $r>0$ defining a region
\begin{align*}
\mathcal O_{Q,r}=\{\theta_0+Qu:u\in\mathbb R^k,\ \|u\|_2\le r\}
\end{align*}
of approximate empirical functional invariance, which we call \emph{flat ocean}, with large dimension $k$ and radius $r$. Our construction is at the same time precise, easy to operationalize, and applies to various network architectures, and activation functions. 

\paragraph{Our contributions.}
We introduce an exact decomposition of the network response into a frozen polynomial and an activation remainder. This yields a constructive characterization of flat oceans: we describe their structure and dimensionality, develop efficient procedures to build them, and bound their finite extent in the true network. Experiments reveal large invariant spaces that emerge and persist during training; in the width-64 ReLU example, the median measured radius along invariant-subspace directions is about \(300\times\) that of generic directions.
We discuss applications to constrained adaptation, posterior exploration and compression in Appendix~\ref{app:finite-invariance-uses}.

\section{SETUP AND DEFINITIONS}
\label{sec:setup}

\subsection{Flat oceans}
\label{sec:oceans-def}
\paragraph{Notation}
Let $X=(x_i)_{i=1}^s$, with $x_i\in\R^{n_0}$, and consider an MLP with $L$ hidden layers of widths $n_1,\ldots,n_L$ and $d$ outputs. We include each bias as the last column of its weight matrix. For a vector $v$, write $\bar v=(v,1)$, so multiplication by the extended matrix adds the bias. 
The layer parameter matrices are $W_\ell\in\R^{n_\ell\times(n_{\ell-1}+1)}$ and $V\in\R^{d\times(n_L+1)}$.
For pointwise activation $\sigma$, the forward recursion is
\begin{align*}
 h_{0,i}&=x_i,&z_{\ell,i}&=W_\ell\bar h_{\ell-1,i},\\
 h_{\ell,i}&=\sigma(z_{\ell,i}),&f_i&=V\bar h_{L,i}.
\end{align*}

The concatenated parameter vector $\theta\in\R^p$ has dimension
$p=\sum_{\ell=1}^L n_\ell(n_{\ell-1}+1)+d(n_L+1)$.
Throughout $F_X(\theta)=(f_i)_{i=1}^s$ and
$\|(v_i)_i\|_X^2=s^{-1}\sum_i\|v_i\|_2^2$.
Parameter norms are Euclidean: the squared norm of a parameter vector is the sum of the squared Frobenius norms of its layer components.
In the text and figures, we abbreviate numbered layer references as L1, L2, and so on, counting from the input; the readout is named explicitly.

We study $F_X$ around a base parameter $\theta_0=(W_1^0,\ldots,W_L^0,V^0)$.
The \emph{empirical fiber} through $\theta_0$ is $F_X^{-1}(F_X(\theta_0))=\{\theta\in\R^p:F_X(\theta)=F_X(\theta_0)\}$, the set of all parameters producing exactly the same outputs on every sample in $X$.
We study convex subsets of this fiber. Convexity ensures that any convex combination of points in the set is still in the set.

\begin{maindefinition}[flat ocean]
A convex set $\mathcal O\subseteq\R^p$ is an $\varepsilon$-flat ocean through $\theta_0$ if $\theta_0\in\ri(\mathcal O)$, its direction space $M=\spn(\mathcal O-\theta_0)$ has positive dimension, and
\[
 \sup_{\theta\in\mathcal O}\|F_X(\theta)-F_X(\theta_0)\|_X\le\varepsilon.
\]
It is exact when $\varepsilon=0$. The condition $\theta_0\in\ri(\mathcal O)$ means that, for some $r>0$, every $u\in M$ with $\|u\|<r$ satisfies $\theta_0+u\in\mathcal O$.\footnote{Equivalently, $\theta_0+(M\cap B_r(0))\subseteq\mathcal O$, where $B_r(0)=\{u\in\R^p:\|u\|<r\}$. See Appendix~\ref{app:setting} for the general relative-interior definition.}
\end{maindefinition}

Our probing experiments use a working tolerance $\varepsilon=10^{-3}$. Numerical ranks use a relative singular-value threshold of $10^{-9}$ (Appendix~\ref{app:experiments}).

\begin{figure}[!t]
\noindent{\setlength{\fboxsep}{6pt}\colorbox{setupbg}{\parbox{\dimexpr\linewidth-2\fboxsep\relax}{\hypertarget{experimental-setup}{}\textbf{Experimental setup.} The baseline network is a biased ReLU MLP with PCA input dimension 256, three hidden layers of width 128 and 10 outputs, trained on 60,000 MNIST images using MSE and Adam.  A fixed 10,000-image training subset $X$ is used for analysis, unless stated otherwise. Five training stages are considered: initialization, early fitting (error $\le10\%$), high accuracy (error $\le0.1\%$), interpolation (MSE $<10^{-4}$), and epoch 500, using the first saved checkpoint meeting each criterion. Description of all ablations, seeds and numerical checks: Appendix~\ref{app:experiments}.}}}
\end{figure}

\paragraph{Relation to structural symmetries.}
The convexity requirement distinguishes flat oceans from general sets of functionally equivalent parameters. 
For example, common network symmetries are hidden-unit permutations and, for homogeneous activations, positive rescalings \citep{simsek2021geometry,shewchuk2024geometry}. Their orbits lie in the empirical fiber for any given $X$, but continuous rescaling orbits are generally curved (right panel in Figure~\ref{fig:sym-vs-ocean}), and permutations are discrete transformations and therefore not convex. Affine symmetry families, such as common-coordinate shifts before LayerNorm \citep{ba2016layernorm}, give rise to flat oceans (Appendix~\ref{app:layernorm}), although a flat ocean need not arise from a structural symmetry.

\subsection{The frozen decomposition}
\label{sec:frozen}
\begin{figure}[!htbp]
\centering
\includegraphics[width=\linewidth]{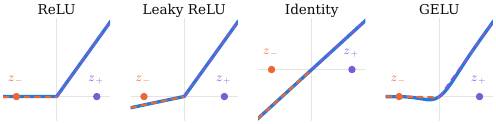}
\caption{The four activations tested in this paper. Orange and purple dashed lines show affine tangents at the marked inputs $z_-=-3$ and $z_+=3$, respectively. They coincide with the corresponding linear pieces for ReLU, leaky ReLU and identity; for GELU they approximate the two tails. Deviation from the base tangent contributes to the transition residual.}
\label{fig:tails}
\end{figure}
The frozen model keeps the usual linear preactivations and replaces each activation by its affine tangent at the base network. Run the network at $\theta_0$ to obtain $z_{\ell,i}^0,h_{\ell,i}^0$ and set $S_{\ell,i}=\diag\sigma'(z_{\ell,i}^0)$. Define the sample- and layer-specific affine map
\[
 \sigma^{\mathrm{lin}}_{\ell,i}(z)
 =\sigma(z_{\ell,i}^0)+S_{\ell,i}(z-z_{\ell,i}^0).
\]
For an edit $w=(A_1,\ldots,A_L,B)$, including bias columns, start with $h_{0,i}^{\frz}=x_i$ and use
\begin{align}
 &z_{\ell,i}^{\frz}=(W_\ell^0+A_\ell)\bar h_{\ell-1,i}^{\frz},\nonumber\\
 &h_{\ell,i}^{\frz}=\sigma^{\mathrm{lin}}_{\ell,i}(z_{\ell,i}^{\frz}).\label{eq:frozen}
\end{align}

Here $\bar h^{\frz}=(h^{\frz},1)$ incorporates each bias exactly as in the original network. At zero edit this recursion recovers the base outputs; stacking them defines $F_X^{\frz}(w)$.

The frozen map is affine in the parameters of each layer separately and has degree at most $L+1$. Let $J_X=DF_X(\theta_0)$ and define
\begin{align*}
 P_X^{\frz}(w)&=F_X^{\frz}(w)-F_X(\theta_0)-J_Xw,\\
 A_X^{\rm tr}(w)&=F_X(\theta_0+w)-F_X^{\frz}(w)
\end{align*}
The following decomposition is exact:
\begin{equation}
 \boxed{\Delta F_X(w)=
 \underbrace{\colorbox{blue!8}{$J_Xw+P_X^{\frz}(w)$}}_{\text{frozen response}}
 +\underbrace{\colorbox{orange!15}{$A_X^{\rm tr}(w)$}}_{\text{activation transition}}}
 \label{eq:decomp}
\end{equation}

Here $\Delta F_X(w)=F_X(\theta_0+w)-F_X(\theta_0)$ is the \emph{network response} to the edit $w$, $J_Xw$ is the first-order response, $P_X^{\frz}$ collects the higher-order interactions with slopes fixed, and $A_X^{\rm tr}$ is the \emph{activation transition potential}. Hereafter, we call $\Delta F_X^{\frz}(w):=F_X^{\frz}(w)-F_X(\theta_0)=J_Xw+P_X^{\frz}(w)$ the \emph{frozen response}.

\begin{figure}[!htbp]
\centering
\includegraphics[width=\linewidth]{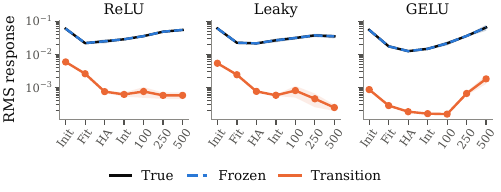}
\caption{True, frozen, and activation-transition response norms during training of width-128 MNIST MLPs. Full-network edits have layer-relative radius $10^{-2}$. Curves show means of within-seed medians over 256 matched directions; bands show one standard deviation across three seeds. Init, Fit, HA and Int denote initialization, early fitting, high accuracy and interpolation. Further radii and layer attribution: Appendix~\ref{app:transition}.}
\label{fig:transition}
\end{figure}

A typical approximation of the network function is the first-order response $J_Xw$, which is linear in $w$ and ignores all nonlinear terms in the true response.
\emph{Our decomposition separates the nonlinear effects due to cross-layer interactions and those due to the activation functions being nonlinear.} This simple distinction is surprisingly powerful, since we find that the cross-layer interactions dominate the nonlinear part in the response throughout training (Figure~\ref{fig:transition}). Appendix~\ref{app:transition} extends the comparison across radii, layers and architectures, including CNNs and BatchNorm. For ReLU and leaky-ReLU MLPs, the separation also grows substantially from initialization to the end of training.

This is good news since the cross-layer interactions form a polynomial in $w$ which is still easy to manipulate algebraically, compared to the full response. This decomposition offers a useful new tool for studying neural network functions, which sits between the true function and its first-order approximation, gaining substantially in precision, while keeping the complexity low. The decomposition may therefore be useful beyond the questions studied here.

\subsection{Construction strategy}
\label{sec:branches}

Using the decomposition in Equation~\ref{eq:decomp}, our goal is to find a nonzero subspace $M$ of parameter edits and a radius $r>0$ such that
\[
 \sup_{\substack{w\in M\\\|w\|\le r}}
 \|\Delta F_X^{\frz}(w)+A_X^{\rm tr}(w)\|_X\le\varepsilon.
\]
The resulting region $\theta_0+\{w\in M:\|w\|\le r\}$ is a flat ocean. We want both its dimension and its radius to be large.

Our approach has two steps: find where the frozen response is zero, then determine how far we can move while keeping the activation remainder small. We therefore seek
\begin{align}
 &\Delta F_X^{\frz}(w)=0\quad\text{for every }w\in M,
 \label{eq:strategy-frozen}\\
 &\sup_{\substack{w\in M\\\|w\|\le r}}\|A_X^{\rm tr}(w)\|_X\le\varepsilon.
 \label{eq:strategy-radius}
\end{align}
We call a linear space satisfying Equation~\ref{eq:strategy-frozen} a \emph{frozen invariant subspace} (shortened to invariant subspace in Sections~\ref{sec:atomic}--\ref{sec:layerwise}). 

For smooth activations, exact oceans could also arise through cancellation between the frozen response and the activation remainder. Our construction controls the two terms separately. Appendix~\ref{app:smooth-exact} presents the conjecture that every exact ocean for tanh, sigmoid or softplus MLPs spans a frozen invariant subspace, proving it in a nondegenerate one-unit setting. Our results do not depend on that conjecture.

The polynomial form of the frozen response explains why it is enough to look for subspaces. We call a convex set $\mathcal O_{\frz}$ of edits a \emph{frozen ocean} if $0\in\ri(\mathcal O_{\frz})$ and $\Delta F_X^{\frz}(w)=0$ throughout the set. Any such region extends to a whole invariant subspace, as the following result shows.
\begin{theorem}[Frozen oceans and invariant subspaces]\label{thm:correspondence}
If the frozen response vanishes on a relative neighborhood of zero in a linear space $M$, it vanishes everywhere on $M$. Thus every frozen ocean spans a frozen invariant subspace, and every frozen invariant subspace is itself a frozen ocean. Every such subspace satisfies $M\subseteq\ker J_X$.
\end{theorem}
The proof is in Appendix~\ref{app:coefficient-test}. This lets us study the subspace first and its true-network radius afterwards. Several maximal invariant subspaces can coexist, and their union need not be a subspace.

\begin{maindefinition}[Layerwise ocean]\label{def:layerwise-ocean}
A frozen invariant subspace $M$ is \emph{layerwise} if
\(
 M=\bigoplus_{\ell=1}^{L+1}M_\ell,
\)
where $M_\ell$ occupies only layer $\ell$, with the readout counted as layer $L+1$. The layer edits can be chosen independently from their component spaces: every combined edit preserves the frozen outputs. A flat ocean whose direction space is such an $M$ is a \emph{layerwise ocean}.
\end{maindefinition}

Our study focuses on layerwise oceans. We also discuss \emph{coupled oceans}, whose layer edits cannot all be chosen independently, in Section~\ref{sec:coupled}. Our measurements suggest that little additional dimensionality remains beyond the constructed layerwise oceans. 

\textbf{Paper structure}\;\;\;
Section~\ref{sec:atomic} first characterizes the invariant subspaces within each layer, which give the \emph{atomic oceans}. Section~\ref{sec:layerwise} then determines which of these subspaces can be combined: its compatibility tests and constructions ensure that simultaneous edits also preserve the frozen outputs. Finally, Section~\ref{sec:radius} returns to the true network and the radius condition in Equation~\ref{eq:strategy-radius}.

\section{ATOMIC OCEANS}
\label{sec:atomic}

\subsection{The feature-null ocean}
\label{sec:featnull}

We start with the simplest ocean structure. Add to layer $\ell$ a matrix $A_\ell$ annihilating each input feature:
\[
 A_\ell\bar h_{\ell-1,i}^0=0\qquad\text{for every }i.
\]
The layer's preactivations stay unchanged on $X$, hence so does every layer above it. Such edits compose across layers and preserve the function at any size.

For each layer, define its input-feature span $\cH_\ell=\spn\{\bar h_{\ell-1,i}^0:i\le s\}$ and rank $r_\ell=\dim\cH_\ell$.
Include the readout by setting $A_{L+1}=B$ and $n_{L+1}=d$.
\begin{samepage}
\begin{theorem}[Feature-null ocean]\label{thm:feature-ocean}
The feature-null edit space is the linear subspace
\[
 M_{\rm feat}=\bigoplus_{\ell=1}^{L+1}\{A_\ell:A_\ell\cH_\ell=0\}.
\]
The affine set $\theta_0+M_{\rm feat}$ is a flat ocean. Every point in this space satisfies $F_X(\theta_0+w)=F_X(\theta_0)$ for every $w\in M_{\rm feat}$ and any pointwise activation. Its dimension is $\dim M_{\rm feat}=\sum_{\ell=1}^{L+1}n_\ell(n_{\ell-1}+1-r_\ell)$. 
\end{theorem}
\par\end{samepage}

Although exact and easy to construct this subspace is typically small. 
In the baseline network described in the setup box in Section~\ref{sec:oceans-def}, this space contains none of the L2 invariant subspace and at most $15\%$ of the L3 invariant subspace. The analysis below describes the remaining dimensions.

\subsection{The full atomic ocean}
\label{sec:fullatomic} 
\begin{figure}[!htbp]
\centering
\includegraphics[width=\linewidth]{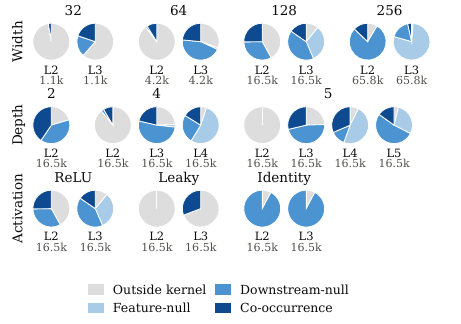}
\caption{Atomic invariant subspace decomposition in MNIST MLPs at epoch 500, seed 0. Each pie represents all parameters of one hidden layer; the three blue components sum to its atomic kernel, and gray is the complement. Downstream-null and co-occurrence components account for invariance beyond feature-null directions. Rows vary width, depth and activation; layer sizes appear below the pies. Details in Appendix~\ref{app:atomic-full}.}
\label{fig:dimensions}
\end{figure}

We call \emph{atomic} an edit which changes only the parameters of one layer. In this case the frozen response reduces to one first-order term. For hidden layer $\ell$, define the downstream map $D_{\ell,i}:\R^{n_\ell}\to\R^d$ that carries a preactivation change to the output with all later weights and slopes fixed:\footnote{The subscript $\mathrm{lin}$ removes the bias column when propagating a hidden-state change. The edit $A_\ell$ still includes its bias column.}
\[
 D_{L,i}=V_{\mathrm{lin}}^0S_{L,i},\qquad
 D_{\ell,i}=D_{\ell+1,i}W_{\ell+1,\mathrm{lin}}^0S_{\ell,i}.
\]
Here the recursion runs backward over $\ell=L-1,\ldots,1$. An atomic edit $A_\ell$ changes the layer's preactivation by $A_\ell\bar h_{\ell-1,i}^0$; applying $D_{\ell,i}$ gives its frozen output response, $D_{\ell,i}A_\ell\bar h_{\ell-1,i}^0$. For the readout, the same formula holds with $D_{L+1,i}=I_d$. Stacking these vectors across samples gives the layer Jacobian $J_\ell$, which measures which edits affect the empirical output:  
$J_\ell\operatorname{vec}(A_\ell)=(D_{\ell,i}A_\ell\bar h_{\ell-1,i}^0)_{i=1}^s$\footnote{Here $\operatorname{vec}$ stacks matrix columns.}.
The kernel of this operator, which we hereafter refer to as the \emph{atomic kernel} $K_\ell$, is the space of atomic edits with zero frozen response, i.e. the atomic invariant subspace. 
Below we identify its components and dimensionality.

Alongside the feature span $\cH_\ell$, define the downstream covector span $\cC_\ell=\sum_i\operatorname{row-span}(D_{\ell,i})$ and its dimension $r_{d,\ell}=\dim\cC_\ell$.

\begin{theorem}[Atomic dimension decomposition]
\label{thm:atomic}
Define the \emph{co-occurrence deficit} by
\[
 \delta_{\rm joint,\ell}=r_\ell r_{d,\ell}-\rank J_\ell\ge0.
\]
The complete atomic frozen invariant subspace is $K_\ell$, with
\begin{equation}
 \dim K_\ell=\underbrace{n_\ell(n_{\ell-1}+1-r_\ell)}_{\text{feature-null}}+\underbrace{r_\ell(n_\ell-r_{d,\ell})}_{\text{downstream-null}}+\delta_{\rm joint,\ell}.
 \label{eq:atomic}
\end{equation}
\end{theorem}
 Proof: Appendix~\ref{app:atomic-proof}.

Pairings between the input-feature and downstream-covector spans define constraints in the system of equations $(D_{\ell,i}A_\ell\bar h_{\ell-1,i}^0)_{i=1}^s=0$. However, the actual equations use only the pairings realized on the same samples, and thus $\rank J_\ell\le r_\ell r_{d,\ell}$. We refer to any rank shortfall as co-occurrence deficit.

Measured atomic kernels (Figure~\ref{fig:dimensions} and Appendix~\ref{app:atomic-full}) substantially exceed the feature-null subspace in dimensionality (L2--L3: $16$--$64\%$ versus $0$--$7\%$ of layer parameters at high accuracy and interpolation, three seeds). The co-occurrence deficit accounts for large portions of the ReLU and leaky-ReLU atomic invariant subspaces: $76$--$91\%$ of L2's kernel and $32$--$64\%$ of L3's for ReLU, and $100\%$ of L3's for leaky ReLU. GELU has no resolved exact kernel in these measurements, but a similarly large approximate kernel can be extracted given a threshold (Figure~\ref{fig:atomic-tails}). Training on SVHN or raw-pixel MNIST gives similar decompositions (Appendix~\ref{app:activations}).

\subsection{Inside the co-occurrence deficit}
\label{sec:cooccurrence}
\begin{figure}[!htbp]\centering
\includegraphics[width=\linewidth]{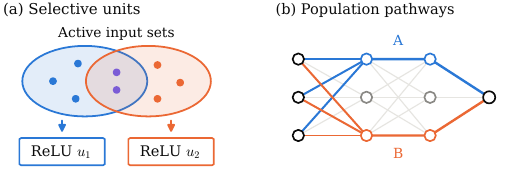}
\caption{Co-occurrence mechanisms. (a) ReLU units activate on overlapping input sets. (b) Populations A and B activate different routes from a common input layer; gray edges are inactive.}
\label{fig:cooccurrence-mechanisms}
\end{figure}

While the feature-null and downstream-null spaces can readily be constructed based on the input features and downstream map, it is less clear what leads to a co-occurrence deficit. In this section we propose two mechanisms that create such a deficit (Figure~\ref{fig:cooccurrence-mechanisms}).

\textbf{A linear network has no deficit.} Its downstream map is the same on every sample, so the Jacobian row space $\operatorname{row}(J_\ell)=\spn_{i,a}\{\bar h_{\ell-1,i}^0\otimes D_{\ell,i}^\top e_a\} = \cH_\ell\otimes\cC_\ell$, where $e_a$ is the $a$th output-coordinate vector (to see it, simply take out the shared downstream map from the samplewise span calculation). Hence $\rank J_\ell=r_\ell r_{d,\ell}$ and $\delta_{\rm joint,\ell}=0$.\footnote{Note that this does not mean linear networks have smaller atomic subspaces because the marginal ranks also differ between networks.}

\textbf{Mechanism 1: Selective units.}
For ReLU, an edit to a unit's incoming weights need only annihilate features of samples on which the unit is active. Its active-feature span is $\cH_{\ell j}^{\rm act}(X)=\spn\{\bar h_{\ell-1,i}^0:z_{\ell,i,j}^0>0\}$. This defines a space containing the feature-null space, which we call \emph{selective}:
\begin{align*}
 G_\ell^{\rm sel}&=\{A:S_{\ell,i}A\bar h_{\ell-1,i}^0=0\ \forall i\},\\
 \dim G_\ell^{\rm sel}
 &=\sum_{j=1}^{n_\ell}\bigl(n_{\ell-1}+1-\dim\cH_{\ell j}^{\rm act}(X)\bigr).
\end{align*}
Notice the difference from the feature-null constraint: $G_\ell = \{A_\ell:A_\ell\cH_\ell=0\}$. If a unit is inactive on a subspace of $\cH_\ell$, the incoming weights can be changed within it without inducing a frozen response.   Like the feature-null subspace, the selective subspaces compose across layers because each edit preserves the corresponding frozen postactivation.
\begin{proposition}[Selective-unit composition]\label{prop:selective-composition}
The direct sum of the hidden selective-unit spaces and the feature-null readout is a frozen invariant subspace. 
\end{proposition}
At high accuracy and interpolation, after removing overlap with the marginal-null spaces, selective spaces explain nearly all of the width-128 ReLU deficit (L2: $99.8$--$100\%$; L3: $92.7$--$100\%$, across three seeds), leaving a small cross-unit remainder (Appendix~\ref{app:methods-branches}). 
The CNN L5 (\texttt{conv5}) kernel is also predominantly selective patch space ($93$--$99\%$; Appendix~\ref{app:mnist-cnn}).

\textbf{Mechanism 2: Clustering.} Different input populations can use different active pathways \citep{lakshminarayanan2020paths}, leaving cross-population feature--downstream pairings unobserved. For a partition $\mathcal G$ of samples by downstream-map similarity, let $\mathcal H_{\ell,G}$ and $\mathcal C_{\ell,G}$ be the feature and downstream covector spans within group $G$ and define $\mathcal P_\ell(\mathcal G)=\sum_{G\in\mathcal G}\mathcal H_{\ell,G}\otimes\mathcal C_{\ell,G}$. For $\delta_{\rm joint,\ell}>0$, the fraction explained by missing cross-group pairings is
\begin{align*}
 \text{explained fraction}&=\frac{r_\ell r_{d,\ell}-\dim\mathcal P_\ell(\mathcal G)}{\delta_{\rm joint,\ell}}.
\end{align*}
At epoch 500, ten downstream-map groups explain $62.2\%$ of the width-128 ReLU L3 deficit, versus $7.2\%$ for size-matched random groups; 80 groups explain $99.96\%$ (Appendix~\ref{app:mechanism}).

\subsection{Ablations}\label{sec:ablations}

\paragraph{Architecture}
The atomic dimension decomposition in Theorem~\ref{thm:atomic} also applies to convolutional layers after summing each sample's output change over spatial positions (proof in Appendix~\ref{app:cnn-theory}). CNNs trained on MNIST develop large late-layer kernels, again much larger than the feature-null subspace. 
We test the same constructions in larger architectures. In particular, we analyze the final layers and readouts of a ViT-B/16 and a ResNet-50 (Appendices~\ref{app:vit} and \ref{app:resnet-common}). Due to the large dimensionality of these layers, we cannot directly compute the atomic kernel. However, we find large feature-null and LayerNorm-induced downstream-null constructions in the ViT-B/16 (Figure~\ref{fig:native-vit-sampled}b), and a large selective subspace in the ResNet-50 (Table~\ref{tab:resnet-common}).  Extensions to attention and compatibility across other architectural components are discussed in Appendix~\ref{app:future-work}.

\begin{figure}[!htbp]\centering
\includegraphics[width=\linewidth]{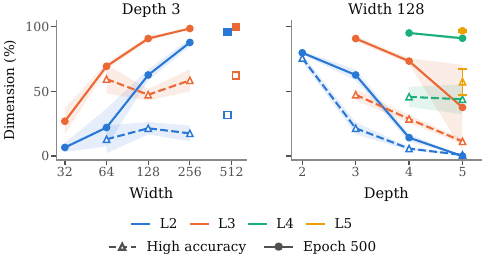}
\caption{Invariant atomic subspace fractions; at width 512 the kernel computation did not converge, squares show the selective space dimensionality. Colors identify layers; linestyle the training stage. Means and one standard deviation over three seeds. Width 32 never reaches high accuracy. Details: Appendix~\ref{app:scaling}.}
\label{fig:scaling-dimensions}
\end{figure}

\paragraph{Effects of scaling on atomic-ocean dimension.}
At epoch 500, increasing width enlarges the full atomic kernels, while increasing depth reduces their dimensionality in early layers (Figure~\ref{fig:scaling-dimensions}). The L2 invariant subspace grows from $6.7\%$ of layer parameters at width 32 to $87.9\%$ at width 256. At width 128, increasing depth from two to five hidden layers reduces the L2 fraction from $79.9\%$ to zero. Appendix~\ref{app:scaling} reports the full counts.

\paragraph{Training data and new inputs}
\label{sec:otherdata}
So far we have studied atomic invariant subspaces measured on a fixed subset of the training data (\hyperlink{experimental-setup}{experimental setup box}). We ask what portion survives on a new, unseen dataset. With the trained network fixed, we evaluate the atomic invariant subspace on additional inputs. At epoch 500, $93.8$--$99.0\%$ of the hidden-layer kernel survives when we double the 10,000-image analysis set with new inputs from the same distribution, using either additional training images or test images (Table~\ref{tab:newdata-main}). 

Most of the hidden-layer invariant space also survives when clean MNIST test images are interpolated toward pure uniform noise. At pure noise, $99.9\%$ of the original L2 space and $98.9\%$ of L3 survive on average across three seeds; intermediate mixtures preserve slightly less. The surviving readout fraction is smaller and varies more across seeds (Figure~\ref{fig:newinput-shift}). Similarly, adding SVHN images preserves $96.7$--$99.4\%$ of the hidden-layer invariant spaces across three seeds (Table~\ref{tab:newinput-svhn}). Thus, most of the atomic invariant space survives the tested distribution shifts. 

Increasing training data has a different effect because it changes the learned network. Interestingly, evaluated on the same 1,000 images, models trained on all 60,000 MNIST images have larger invariant subspaces than models trained on those 1,000 alone. Thus more training data can produce greater functional redundancy on a fixed evaluation set. 
Full details are in Appendix~\ref{app:training-data}.

\paragraph{Persistence of the ocean}
\label{sec:persistence}
We close the atomic analysis with a striking observation: much of the ocean persists during training, while its dimension grows. We measure the mean squared projection of an earlier atomic subspace onto a later one. A subspace extracted at a high-accuracy point retains $99.3$--$99.9\%$ of its mass at epoch 500 across hidden layers and seeds  (Table~\ref{tab:persistence-main}, Appendix~\ref{app:persistence}). These measurements suggest that invariant subspaces grow during training by accretion: many earlier invariant directions persist as new ones emerge. How optimization produces this behavior is an open question (Appendix~\ref{app:future-work}).
\begin{table}[!htbp]
\centering
\caption{Retained atomic invariant subspace mass (\%) in MNIST MLPs: mean $\pm$ one sample standard deviation over three seeds. HA and Int denote high accuracy and interpolation; numbers denote epochs.}
\label{tab:persistence-main}
\small\setlength{\tabcolsep}{3pt}
\begin{tabular}{lrr}\toprule
Earlier $\to$ later & L2 & L3\\\midrule
HA $\to$ Int & $89.75 \pm 9.20$ & $97.39 \pm 0.29$\\
Int $\to$ 100 & $93.41 \pm 0.82$ & $97.85 \pm 0.53$\\
100 $\to$ 250 & $97.70 \pm 0.95$ & $98.96 \pm 0.32$\\
250 $\to$ 500 & $99.56 \pm 0.42$ & $99.29 \pm 0.34$\\
\midrule
HA $\to$ 500 & $99.73 \pm 0.24$ & $99.37 \pm 0.11$\\
\bottomrule\end{tabular}
\end{table}

\section{COMPOSITION ACROSS DEPTH}
\label{sec:layerwise}

Every layerwise frozen ocean $M=\bigoplus_{\ell=1}^{L+1}M_\ell$ lies within the direct sum of the atomic invariant subspaces: $M_\ell\subseteq K_\ell$ for each layer, since each layer edit must independently satisfy Equation~\ref{eq:strategy-frozen}. If $\bigoplus_\ell K_\ell$ also satisfies Equation~\ref{eq:strategy-frozen}, it is the unique largest layerwise frozen ocean (Proposition~\ref{prop:app-atomic-ceiling}).

\begin{figure}[!htbp]\centering
\includegraphics[width=\linewidth]{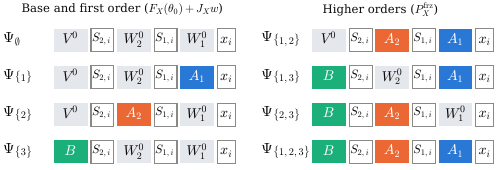}
\caption{Frozen expansion for two hidden layers and a readout, in bias-free homogeneous notation. The left panel contains the base output and first-order response; the right contains the mixed-layer terms retained in $P_X^{\frz}(w)$. Colored matrices are edits, gray matrices are base weights, and the slopes $S_{\ell,i}$ stay fixed.}
\label{fig:frozen-terms}
\end{figure}

The combination of atomic invariant subspaces across layers is not necessarily an invariant subspace. To see why, expand $F_X^{\frz}$ over subsets of edited layers $I\subseteq\{1,\ldots,L+1\}$. Each term $\Psi_I$ collects the factors edited in layers $I$, keeping the other weights at their base values. The empty subset gives the base output, singleton subsets sum to $J_Xw$, and the remaining terms sum to $P_X^{\frz}(w)$.\footnote{\citet{laurent2018multilinear} establish the related piecewise multilinear structure for ReLU and leaky-ReLU networks. Our frozen expansion applies to any pointwise activation for which the base affine tangents are defined, including smooth activations.} Figure~\ref{fig:frozen-terms} shows the expansion for two hidden layers and a readout. For invariance, the mixed-layer terms must vanish along with the single-layer responses.

We first give an exact test for whether chosen atomic subspaces can be combined, then use it to construct layerwise frozen oceans greedily.

\subsection{A layerwise ocean test}
\label{sec:lwtest}

The test (Algorithm~\ref{alg:branch-test}) proceeds layer by layer, from input to output.
For sample $i$, $R_{\ell-1,i}$ spans the frozen hidden-state changes reaching layer $\ell$ from the earlier edits.
The received-input space at layer $\ell$ is then
\[
 U_{\ell,i}=\spn\{\bar h_{\ell-1,i}^0\}+(R_{\ell-1,i}\times\{0\}).
\]
Hidden-state changes are augmented with zero because the appended constant remains one. Bias edits are still included through the last column of $A_\ell$ acting on the base augmented input. For spaces $M,U$, we write their product $MU=\spn\{Au:A\in M,u\in U\}$. 
For the readout, $D_{L+1,i}=I_d$.

\begin{algorithm}[!htbp]
\caption[Layerwise ocean test]{\protect\colorbox{definitionhighlight}{\strut Layerwise ocean test}}\label{alg:branch-test}
\small
\begin{algorithmic}[1]
\Require Candidate spaces $M_1,\ldots,M_{L+1}$ and base network on $X$.
\State $R_{0,i}\gets\{0\}$ for every sample $i$.
\For{$\ell=1,\ldots,L+1$}
 \State For each $i$: $U_{\ell,i}\gets\spn\{\bar h_{\ell-1,i}^0\}+(R_{\ell-1,i}\times\{0\})$.
 \State \textbf{if} any $D_{\ell,i}(M_\ell U_{\ell,i})\ne\{0\}$ \textbf{return false}.
 \State For $\ell\le L$, update each received-change space:
 \Statex \hspace{\algorithmicindent}$R_{\ell,i}\gets S_{\ell,i}(W_{\ell,\mathrm{lin}}^0R_{\ell-1,i}+M_\ell U_{\ell,i})$.
\EndFor
\State \Return true
\end{algorithmic}
\end{algorithm}
At layer $\ell$, the test checks interactions between the earlier layers' edits (contained in $U_{\ell,i}$) and the layer candidate subspace $M_\ell$, keeping the upper layers fixed ($D_{\ell,i}$). If the edits are compatible, the propagated output will be zero. Conveniently, this layerwise procedure is exact and complete (proof in Theorem~\ref{thm:composition} in Appendix~\ref{app:composition-proof}), and its cost is only polynomial in the supplied basis sizes, widths, depth and sample count (Appendix~\ref{app:composition-proof}).

\subsection{Constructing layerwise oceans}
\label{sec:construction}
We use the compatibility test to construct a layerwise invariant subspace that cannot be enlarged without losing invariance. The construction proceeds greedily, selecting the full compatible space at each layer. The resulting subspace is maximal by inclusion, although another choice of layer spaces may yield a larger total dimension.

The bottom-up construction starts at the input and adds one layer at a time. Initially $M_j=\{0\}$ for every layer. At layer $\ell$, the spaces selected before it ($M_1, \dots, M_{\ell-1}$) determine $R_{\ell-1,i}$ and $U_{\ell,i}$.
Choose all layer edits whose action on these inputs is invisible to the fixed downstream network:
\begin{equation}
 M_\ell=\{A:D_{\ell,i}Au=0\ \text{for every }i\text{ and }u\in U_{\ell,i}\}.
 \label{eq:bottom-up-constraints}
\end{equation}
These are homogeneous linear constraints on $A$, which preserve the frozen output and remove interactions with the layer edits already selected. Update $R_{\ell,i}$ as in Algorithm~\ref{alg:branch-test} and continue to the next layer. The construction also applies in other layer orders, with additional constraints involving already selected higher layers; Appendix~\ref{app:composition-proof} gives the general equations, algorithm and validity proof.

The algorithm is greedy in dimensionality, because at each layer it selects the largest subspace compatible with the constraints. Choosing a smaller intermediate $M_\ell$ reduces the constraints on later layers, potentially yielding a larger invariant subspace overall.
On the measured width-64 MNIST MLPs, the atomic invariant subspaces compose into a layerwise invariant subspace, so every tested order recovers their full direct sum (Appendix~\ref{app:selective}). Order matters when training only on 1,000 samples: bottom-up and top-down return $6{,}719$ and $7{,}999$ dimensions, respectively, against an invalid atomic sum of $13{,}034$ (Table~\ref{tab:subset-order}).

\textbf{Maximum dimension.} 
The procedure to construct invariant subspaces gives us insight into the variety and dimensionality of invariant subspaces. A natural question is what is the dimensionality of the largest layerwise invariant subspace. This is a difficult question to answer, but we can already say something in this regard.
A layerwise invariant subspace satisfies
\[
 \dim M\le\min\{\dim\ker J_X,\sum_{\ell=1}^{L+1}\dim K_\ell\}.
\]
The greedy construction supplies a lower bound on the maximum. 
For a bias-free depth-two linear network,  the maximum layerwise invariant subspace dimension has a closed-form solution (Appendix~\ref{app:deep-linear}). For linear networks of any depth, we derive a Bellman recursion maximizing the remaining dimension over the propagated hidden subspace (Appendix~\ref{app:deep-linear}). Its state includes the subspace's orientation, so the exact recursion is a continuous optimization with no established finite polynomial-time solution. The maximum dimension for arbitrary-depth nonlinear networks remains open.

\subsection{Beyond layerwise invariant subspaces}
\label{sec:coupled}

The construction procedure above applies to layerwise oceans. What about coupled ones?
Coupled invariant subspaces tie edits across layers, so their layer components cannot all vary independently. 
Based on the condition in Equation~\ref{eq:strategy-frozen}, we can write a general test for both coupled and layerwise invariant subspaces (Appendix~\ref{app:coefficient-test}). The number of coefficient conditions to verify is combinatorial in the layers' dimension and network depth. Although this test characterizes coupled invariant subspaces exactly, the number of conditions makes it impractical for constructing them. An efficient general test remains open. 

We can nevertheless bound how many independent directions coupled subspaces could add.  Every invariant subspace through a differentiable base point is a subspace of $\ker J_X$ (Remark~\ref{rem:exact-ocean-tangent}). Given a constructed layerwise subspace $M$, at most $\dim\ker J_X-\dim M$ additional independent directions remain. Some may be tangents to curved invariant families rather than directions of finite straight invariance (Remark~\ref{rem:exact-ocean-tangent}).

\noindent{\setlength{\fboxsep}{6pt}\colorbox{setupbg}{\parbox{\dimexpr\linewidth-2\fboxsep\relax}{\textbf{How much remains beyond layerwise invariance?} Across three width-64 ReLU MLPs at high accuracy, the constructed layerwise invariant subspaces account for $94.3$--$94.7\%$ of the numerical tangent-kernel dimension. Adding the rescaling-symmetry tangent span leaves only $0.18$--$1.23\%$ unaccounted for (Tables~\ref{tab:tangent-ocean-seeds} and~\ref{tab:c-symmetry-intersection}). 
}}}

\section{OCEAN RADII}
\label{sec:radius}

The previous sections identify a subspace $M$ where the frozen response vanishes identically (Equation~\ref{eq:strategy-frozen}). In this section, we seek a radius $r$ which controls the true network response as follows: 
\[\sup_{\substack{w\in M\\\|w\|\le r}}\|A_X^{\rm tr}(w)\|_X\le\varepsilon.\]
By construction, on a frozen invariant subspace $\Delta F_X=A_X^{\rm tr}$. For piecewise-linear activations, the remainder is zero inside the base activation chamber.\footnote{The base activation chamber is the connected component containing $\theta_0$ in which every sampled preactivation stays in its base affine piece.} 
Beyond the chamber, or for smooth activations, the radius depends on the activation residuals and their propagation to the output.

\subsection{The activation-transition identity}
\label{sec:radius-bounds}
Let $\Delta z_{\ell,i}(w)$ be the true preactivation change at $\theta_0+w$. Its local activation residual is the linearization error
\[
 \rho_{\ell,i}(w)=\sigma(z_{\ell,i}^0+\Delta z_{\ell,i}(w))
                  -h_{\ell,i}^0-S_{\ell,i}\Delta z_{\ell,i}(w).
\]
For ReLU, local residuals begin at gate crossings and equal the overshoot past zero. For smooth activations, the residual measures how accurately the base tangent predicts the new activation value: a more accurate linearization gives a smaller residual. 
\begin{theorem}[Exact activation transition decomposition]
\label{thm:transition}
Let $T_{\ell,i}(w)$ propagate a residual inserted after layer $\ell$ through the edited downstream weights with the base activation slopes held fixed. For every edit,
\begin{equation}
 A_i^{\rm tr}(w)=\sum_{\ell=1}^L T_{\ell,i}(w)\rho_{\ell,i}(w).
 \label{eq:transition}
\end{equation}
\end{theorem}
This simple decomposition shows that local activation residuals only tell half of the story. Downstream maps ($T_{\ell,i}(w)$) can suppress these residuals, and their contributions can cancel. In both cases, the activation transition term would stay low, despite linearization errors. In Appendix~\ref{app:radius-proof} we give the proof of the decomposition and build a conservative uniform bound on the activation transition term. 

\subsection{Estimating a safe radius}
\label{sec:radius-measured}
Even without a uniform bound, we can ask what fraction of directions remain safe up to a given radius. Let $S(M)$ be the unit sphere of a fixed invariant subspace $M$. At radius $r$ and output tolerance $\varepsilon$, define the failure probability
\[
 p(r,\varepsilon)=\Pr_{v\sim\operatorname{Unif}(S(M))}\!\left[
 \sup_{0\le t\le r}\|A_X^{\rm tr}(tv)\|_X>\varepsilon\right].
\]
We seek a radius with $p(r,\varepsilon)\le\delta$: at least $1-\delta$ of directions then preserve the outputs within tolerance along their entire segment from the center to radius $r$.

In practice, we approximate the supremum using a radius grid $0<r_1<\cdots<r_m$. Write $p_j(\varepsilon)$ for the probability of exceeding tolerance at any tested radius through $r_j$. 
For $n$ independent directions sampled uniformly from $S(M)$, the failure count is binomial:
\[
 \begin{aligned}
 \sum_{i=1}^n\mathbf1\!\left\{
 \max_{h\le j}\|A_X^{\rm tr}(r_hv_i)\|_X>\varepsilon\right\} \sim\operatorname{Binomial}\bigl(n,p_j(\varepsilon)\bigr).
 \end{aligned}
\]
Inverting the binomial tail gives an upper confidence bound $\overline p_j(\varepsilon)$ \citep{clopper1934confidence}. We use a total error probability $\alpha=0.05$, giving $95\%$ simultaneous confidence.\footnote{The bounds are calibrated so that, with probability at least $1-\alpha$, they all hold simultaneously. This permits selecting the largest safe radius after inspecting the scan.} We certify a radius $r_j$ as safe on the tested grid whenever $\overline p_j(\varepsilon)\le\delta$. Appendix~\ref{app:radius-confidence} details the construction.

\begin{figure}[!t]
\centering
\includegraphics[width=\linewidth]{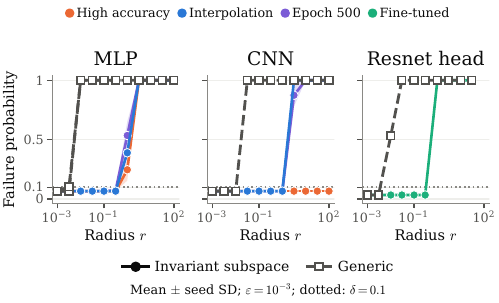}
\caption{Upper bounds on grid-based failure probability at $\varepsilon=10^{-3}$: mean $\pm$ sample SD across seeds. Below the dotted line, the mean failure probability is at most $10\%$. ResNet uses selective edits of \texttt{layer4.2.conv2}.}
\label{fig:radius-grid}
\end{figure}
Figure~\ref{fig:radius-grid} shows mean upper confidence bounds on the failure probability, with standard deviations across seeds. At $\varepsilon=10^{-3}$, and $\delta=0.1$, each measured MLP checkpoint has a certified safe radius of $0.316$ and each late-stage CNN checkpoint has radius $1$. These are at least $100\times$ the corresponding certified generic radii (Appendix~\ref{app:radius-certificates}).

Leaky-ReLU and BatchNorm MLPs show the same separation from generic directions (Figure~\ref{fig:radius-variants}). Direct output-response curves, including GELU approximate tails and wider MLPs, are given in Appendices~\ref{app:radius-full} and \ref{app:radius-responses}.

Feature-null edits alone preserve the true outputs at every radius. In a single-layer test, their component can be removed before certifying the remaining space. For multilayer edits, this removal also requires that the feature-null component annihilate the features changed by earlier layers.

\section{From function to loss}
\label{sec:discussion}
Flat oceans describe regions of approximate functional invariance in the parameter space. In the previous sections we have observed that flat oceans often occupy a large part of the parameter space. In this final section we ask whether they also explain a large part of the loss flatness observed in parameter space.

Functional invariance is directly connected to loss invariance (also called \emph{flatness} in the literature, \citep{sagun2017hessian}). For squared loss $\mathcal L(\theta)=\|F_X(\theta)-Y\|_X^2$, let $Y$ denote the targets and $e_0=F_X(\theta_0)-Y$ the residual at $\theta_0$. Writing $\Delta\mathcal L$ for the loss change under an edit, $\Delta\mathcal L=2\langle e_0,\Delta F_X\rangle_X+\|\Delta F_X\|_X^2$.
An ocean with output tolerance $\varepsilon$ therefore guarantees
\[
 |\Delta\mathcal L|\le2\|e_0\|_X\varepsilon+\varepsilon^2.
\]
For a fixed residual this is $O(\varepsilon)$; at exact interpolation it becomes $\varepsilon^2$.

Not all loss invariance is explained by functional invariance, however. Every direction of exact finite loss invariance must satisfy the first-order condition $
 \nabla_\theta\mathcal L^\top v
 =\nabla_F\mathcal L^\top J_Xv=0.
$
This includes both functionally invisible directions ($J_Xv=0$) and output changes tangent to a loss level set. Writing $u=J_Xv$ for the first-order output change, the loss-invisible changes are loss dependent, e.g.:
\[
 \begin{aligned}
 \text{MSE:}&\quad \langle e_0,u\rangle_X=0,\\
 \text{Cross-entropy:}&\quad \langle\Pi_0-Y,u\rangle_X=0,
 \end{aligned}
\]
where $\Pi_0$ contains the softmax probabilities at $\theta_0$ and $Y$ the one-hot targets. In the function space, MSE level sets are spheres around the targets, while cross-entropy also admits arbitrary straight logit shifts $\Delta f_i=a_i\mathbf1$ for each sample $i$. 

\textbf{Flatness coverage.} \;\;
Intrinsically loss-invisible directions can also yield flat loss regions without functional invariance. We therefore ask: how much of the observed local loss flatness is captured by the constructed functional invariances? We compare the constructed layerwise invariant subspace with the numerical kernels of the loss Hessian and the generalized Gauss--Newton (GGN) matrix. These matrices describe local curvature and coincide at exact interpolation for MSE. We measure \emph{coverage}, the average squared projection of the numerical-kernel directions onto the constructed invariant space.

In three width-64 MNIST MLPs at epoch 500, the constructed flat oceans capture almost all of the numerical Hessian kernel and $95.21\pm0.94\%$ of the numerical GGN kernel (Table~\ref{tab:exact-kernel-coverage}). We use $|\lambda|\le10^{-14}$ to identify numerical zeros in the Hessian and GGN eigenvalues, near the eigensolver's resolution. Hessian coverage remains above $99.9\%$ at $10^{-12}$. Keeping the ocean fixed while increasing the eigenvalue threshold reduces coverage (Figure~\ref{fig:fixed-ocean-coverage}).

The numerical Hessian kernel aligns almost entirely with the flat ocean, whose activation-remainder bounds control finite output and loss changes (Section~\ref{sec:radius}). This close alignment suggests that most numerical Hessian-null directions in these networks reflect functional invariance. \emph{Consequently, our results provide evidence that the loss flatness identified by the Hessian kernel extends over finite distances, with guarantees supplied by the  flat-ocean construction.} Direct ray tests show comparable median visibility radii for the ocean and cumulative bands at $\tau=10^{-8}$, spanning about $34\%$ (Hessian) and $54\%$ (GGN) of all parameters (Figure~\ref{fig:curvature-cumulative-rays}). At $\tau=10^{-4}$, these fractions reach $91\%$ and $95\%$, but the output tolerance is crossed earlier.

\begin{table}\centering
\caption{Constructed-ocean dimension and numerical-kernel coverage: mean and sample SD across three width-64 MNIST seeds. Coverage is in percent at $|\lambda|\le10^{-14}$. The tiny Hessian shortfall is reported numerically, not as an error bound.}
\label{tab:exact-kernel-coverage}
\small\begin{tabular}{lrr}\toprule
Quantity & Mean & SD\\\midrule
$\dim M$ & 4,088 & 477.20\\
Hessian coverage & $100-1.324\!\times\!10^{-9}$ & $1.324\!\times\!10^{-9}$\\
GGN coverage & $95.21$ & $0.94$\\
\bottomrule\end{tabular}
\end{table}

\textbf{Convexity and linear mode connectivity.}\;\;
Convexity ensures that the segment between any two points in an exact ocean preserves the loss; an approximate ocean controls loss variation throughout that segment. Sibling networks (i.e., networks that share an initial training trajectory) are often connected by low-loss segments \citep{frankle2020linear}. In our preliminary MLP and CNN experiments, siblings also share a large portion of their constructed invariant subspaces (Appendix~\ref{app:lmc-shared}). These observations suggest that \emph{flat oceans may offer a useful framework for studying phenomena such as linear mode connectivity and model merging.} We leave these connection to future work.

\section{RELATED WORK}
\label{sec:related}

\paragraph{Global functional equivalence.}
These studies ask which weight changes preserve a network's function on all inputs. Complete answers exist for tanh-type networks, shallow ReLU networks and generic three-layer ReLU bottlenecks \citep{vlacic2020affine,ramakrishnan2026complete,gegenfurtner2026symmetries}. Deep ReLU networks can have redundancies beyond neuron rescaling and permutation, although these exhaust the ambiguity at some parameter settings \citep{grigsby2023hidden,grillo2026identifiable}. Other results characterize equivalence through architecture-changing rewrites \citep{zhang2026identification}, or restrict transformations defined across parameter space to scaling and permutation under regularity assumptions \citep{deng2026covering}. We seek the additional freedom obtained by preserving only a dataset's outputs, and construct finite convex invariant regions.

\paragraph{Empirical equivalence and local dimension.}
\citet{stock2021embedding} give conditions under which finitely many sample outputs locally determine a deep ReLU network's parameters up to rescaling. Their path-product representation removes this rescaling freedom and makes output constraints locally linear in the new coordinates.
\citet{bonapellissier2022local} describe parameters preserving sample outputs locally by intersecting observation constraints with the set of realizable path products, and derive rank conditions for identifiability. Functional-dimension studies also use evaluation Jacobians to describe locally available output changes and parameter redundancy \citep{grigsby2025functional,bonapellissier2024geometry}. Compared to this line of work, our approach is constructive: our goal is to explicitly and cheaply describe the parameter set where the empirical network function is constant.

\paragraph{Finite invariance from network embeddings.}
Embedding a narrower network into a wider one also creates finite affine invariant sets. \citet{fukumizu2019semiflat} study the resulting flat minima and saddles; \citet{simsek2021geometry} describe connected collections of affine subspaces arising from neuron replication. These constructions preserve outputs for all inputs. We characterize additional sample-dependent invariance at a given checkpoint. Our construction works for several architectures and does not require specific constructions, such as a narrower-network.

\paragraph{Algebraic representations and response approximations.}
ReLU networks have piecewise multilinear parameter dependence \citep{laurent2018multilinear}, expressible through activation gates and path-weight products \citep{lakshminarayanan2020paths}. Algebraic methods also characterize realizable-function dimensions for polynomial networks \citep{kileel2019expressive}. For ReLU, our frozen network agrees with the fixed-gate polynomial. Our affine activation tangents extend this construction to smooth activations, with an exact remainder accounting for activation changes. Retaining mixed-layer products, beyond whole-network parameter linearization, allows us to approximate the network function to a higher fidelity, without increasing complexity.

\section{CONCLUSION}
\label{sec:conclusion}
In this work we have taken a constructive approach to the characterization of sample dependent invariance. Given reference parameters $\theta_0\in\mathbb R^p$, we provide methods to identify an orthonormal basis $Q\in\mathbb R^{p\times k}$ and a radius $r>0$ defining a region
\begin{align*}
\mathcal O_{Q,r}=\{\theta_0+Qu:u\in\mathbb R^k,\ \|u\|_2\le r\}
\end{align*}
of approximate empirical functional invariance, which we call \emph{flat ocean}, with large dimension $k$ and radius $r$. We studied layerwise invariant subspaces, which give exact invariance in the frozen network, and measured approximate invariance of the true outputs at large sampled radii.

We also introduced a new, simple decomposition of the network response function, which separates a dominant polynomial term from an activation-dependent residual. This allows us to study significantly more complex behaviors than a common first-order approximation while keeping the complexity of the terms very low. 

Our analysis leaves empirical phenomena such as persistence through training unexplained. A dynamical account of how training creates, preserves and enlarges these regions is therefore a central next step. Appendix~\ref{app:future-work} collects broader theoretical questions and applications.

\subsubsection*{AI Use Statement}
The discovery of flat oceans, the idea of the network-response decomposition, and the structure of the mathematical analysis originated with the authors. The authors determined the research questions, the experimental objectives, and the organization of the paper. Generative AI tools assisted in developing and checking proofs, writing and validating experimental code, and analyzing results. Each mathematical proof was reviewed by the authors. For the writing, the authors provided initial drafts and used AI assistance to polish wording and structure through repeated rounds of revision. The authors reviewed the resulting text and remain responsible for all claims, proofs, experiments and conclusions.

\subsubsection*{Acknowledgments}
We thank Jacopo Graldi, Rodrigo Casado Noguerales, Alexander Bakarsky and Weronika Ormaniec for their support and early discussions, which helped shape the intuition behind flat oceans, and Razvan Pascanu for useful discussions on loss landscapes, continual learning and linear mode connectivity. We acknowledge funding from the ETH AI Center and support from the Department of Computer Science at ETH Zurich.

\bibliographystyle{plainnat}
\bibliography{shared/flat_ocean_references,early_draft_references}

\clearpage
\appendix
\onecolumn

\makeatletter
\begingroup
\setcounter{tocdepth}{3}
\def\@pnumwidth{1.8em}
\def\@tocrmarg{2.2em}
\renewcommand*{\l@section}[2]{\addvspace{.7em}{\bfseries\@dottedtocline{1}{0em}{1.6em}{#1}{#2}}}
\renewcommand*{\l@subsection}[2]{\@dottedtocline{2}{.4em}{2.8em}{#1}{#2}}
\renewcommand*{\l@subsubsection}[2]{\@dottedtocline{3}{.9em}{3.4em}{#1}{#2}}
\begin{center}{\Large\bfseries Appendix contents}\end{center}
\vspace{.5em}
\noindent
\begin{minipage}[t]{.485\linewidth}
\fontsize{10}{12}\selectfont
\@starttoc{apxleft}
\end{minipage}\hfill
\begin{minipage}[t]{.485\linewidth}
\fontsize{10}{12}\selectfont
\@starttoc{apxright}
\end{minipage}
\endgroup
\clearpage

\newcommand{\AppendixContentsRecord}[2]{%
  \ifnum\value{section}<3\def\apxcontentsfile{apxleft}%
  \else\def\apxcontentsfile{apxright}\fi
  \addtocontents{\apxcontentsfile}{\protect\contentsline{#1}%
    {\protect\numberline{\csname the#1\endcsname}#2}{\thepage}{\@currentHref}}}
\let\AppendixOriginalSection\section
\let\AppendixOriginalSubsection\subsection
\let\AppendixOriginalSubsubsection\subsubsection
\renewcommand{\section}[1]{\AppendixOriginalSection{#1}\AppendixContentsRecord{section}{#1}}
\renewcommand{\subsection}[1]{\AppendixOriginalSubsection{#1}\AppendixContentsRecord{subsection}{#1}}
\renewcommand{\subsubsection}[1]{\AppendixOriginalSubsubsection{#1}\AppendixContentsRecord{subsubsection}{#1}}
\makeatother

\section{Theoretical analysis}
\label{app:proofs}

This appendix gives the full theoretical treatment of the main paper's results, together with additional results and proofs. It introduces the notation and assumptions from the beginning so it can be read as a standalone document.

\theoremstyle{plain}
\newtheorem*{appstatement}{\appstatementheading}
\newenvironment{restated}[1]{\def\appstatementheading{#1}\begin{appstatement}}{\end{appstatement}}
\newtheorem*{appconjecture}{Conjecture}
\theoremstyle{remark}
\newtheorem*{appremark}{\appstatementheading}
\newenvironment{restatedremark}[1]{\def\appstatementheading{#1}\begin{appremark}}{\end{appremark}}

\subsection{Setting and notation}
\label{app:setting}
\paragraph{Network and data.}
Let $X=(x_i)_{i=1}^s$, with $x_i\in\R^{n_0}$, and consider an MLP with $L$ hidden layers of widths $n_1,\ldots,n_L$ and $d$ outputs. We include each bias as the last column of its weight matrix. For a vector $v$, write $\bar v=(v,1)$, so multiplication by the extended matrix adds the bias. The layer parameter matrices are
\[
 W_\ell\in\R^{n_\ell\times(n_{\ell-1}+1)},\qquad
 V\in\R^{d\times(n_L+1)}.
\]
Write $W_\ell=[W_{\ell,\mathrm{lin}}\;b_\ell]$ and $V=[V_{\mathrm{lin}}\;b_{L+1}]$ to separate the linear weights from the bias vectors. For pointwise activation $\sigma$, the forward recursion is
\begin{align*}
 h_{0,i}&=x_i,& z_{\ell,i}&=W_\ell\bar h_{\ell-1,i},\\
 h_{\ell,i}&=\sigma(z_{\ell,i}),& f_i&=V\bar h_{L,i}.
\end{align*}
Here $i=1,\ldots,s$ indexes samples and $\ell=1,\ldots,L$ indexes hidden layers. The concatenated parameter vector $\theta\in\R^p$ has dimension
$p=\sum_{\ell=1}^L n_\ell(n_{\ell-1}+1)+d(n_L+1)$.
Throughout $F_X(\theta)=(f_i)_{i=1}^s$ and
$\|(v_i)_i\|_X^2=s^{-1}\sum_i\|v_i\|_2^2$.
We identify stacked outputs with $\R^{sd}$. Parameter norms are Euclidean: the squared norm of a parameter vector is the sum of the squared Frobenius norms of its layer components.
In the text and figures, we abbreviate numbered layer references as L1, L2, and so on, counting from the input; the readout is named explicitly. For a matrix, $\|\cdot\|_F$ is the square root of the sum of squared entries and $\|\cdot\|_{\rm op}$ is its Euclidean operator norm.

\paragraph{Base point and edits.}
Fix $\theta_0=(W_1^0,\ldots,W_L^0,V^0)$ and use superscript $0$ for its states. Write an edit as $w=(A_1,\ldots,A_L,B)$, with
\[
 A_\ell=[A_{\ell,\mathrm{lin}}\;a_\ell],\qquad
 B=[B_{\mathrm{lin}}\;a_{L+1}].
\]
The vector $a_\ell$ changes bias $b_\ell$. To include the readout in layer-indexed formulas, set $A_{L+1}=B$ and $n_{L+1}=d$. A hidden-state change augments as $(v,0)$, because the constant bias coordinate does not change.

Assume $\sigma$ is differentiable at every base preactivation and define
\[
 S_{\ell,i}=\diag\sigma'(z_{\ell,i}^0),\qquad J_X=DF_X(\theta_0).
\]
The operator $\diag$ puts the listed entries on a diagonal, and $D$ in $DF_X$ denotes differentiation. For ReLU, differentiability at every base preactivation means $z_{\ell,i,j}^0\ne0$ for every sample, layer and unit. We call the region of parameters with those same signs the \textbf{empirical activation chamber} $\cC_X(\theta_0)$.

\paragraph{The empirical fiber and flat oceans.}
The \textbf{fiber} of a map over an output is the set of all inputs mapped to that output, also called its level set \citep[Ch.~4]{lee2013smooth}. For the empirical output map, the fiber through $\theta_0$ is
\[
 F_X^{-1}(F_X(\theta_0))=\{\theta\in\R^p:F_X(\theta)=F_X(\theta_0)\}.
\]
We call a convex set $\mathcal O\subseteq\R^p$ an \textbf{$\varepsilon$-flat ocean} through $\theta_0$, for $\varepsilon\ge0$, if $\theta_0\in\ri(\mathcal O)$, its \textbf{direction space} $M=\spn(\mathcal O-\theta_0)$ has positive dimension, and
\[
 \sup_{\theta\in\mathcal O}\|F_X(\theta)-F_X(\theta_0)\|_X\le\varepsilon.
\]
Here $\spn$ means linear span. For a nonempty convex set $C$, let $L_C=\spn(C-C)$ be its direction space. Its \textbf{relative interior} is
\[
 \ri(C)=\{x\in C:\exists r>0,\ x+(L_C\cap B_r(0))\subseteq C\},
 \qquad B_r(0)=\{u:\|u\|<r\}.
\]
This is the usual interior relative to the affine hull $x+L_C$ \citep{boyd2004convex}. We call oceans \textbf{exact} when they have $\varepsilon=0$. Convexity keeps every segment between their points inside the ocean; relative interior supplies a neighborhood of the base point in their direction space.

\begin{remark}[Exact oceans lie in the tangent kernel]\label{rem:exact-ocean-tangent}
Assume $F_X$ is differentiable at $\theta_0$. If $\mathcal O$ is an exact flat ocean through $\theta_0$, then its direction space $M=\spn(\mathcal O-\theta_0)$ satisfies
\[
 M\subseteq\ker J_X,\qquad \dim M\le\dim\ker J_X.
\]
\end{remark}
\begin{proof}
By relative interior, some $r>0$ satisfies $\theta_0+(M\cap B_r(0))\subseteq\mathcal O$. Fix $v\in M$. For every sufficiently small $t$, exact invariance gives $F_X(\theta_0+tv)=F_X(\theta_0)$. Differentiability therefore yields
\[
 J_Xv=\lim_{t\to0}\frac{F_X(\theta_0+tv)-F_X(\theta_0)}{t}=0.
\]
Since $v$ was arbitrary, $M\subseteq\ker J_X$.
\end{proof}
For an invariant differentiable curve through $\theta_0$, the same argument places its tangent in $\ker J_X$; its finite chords need not lie there. For an $\varepsilon$-flat ocean with fixed $\varepsilon>0$, exact tangent nullity does not follow: the bound $\|F_X(\theta_0+tv)-F_X(\theta_0)\|_X\le\varepsilon$ does not give a vanishing difference quotient. For instance, $F(t)=t$ on $[-\varepsilon,\varepsilon]$ satisfies that bound at zero and has derivative one.

\paragraph{Downstream maps and atomic kernels.}
We call an edit \textbf{atomic} if it changes only the parameters of one layer. Define its \textbf{base downstream map} by
\[
 D_{L,i}=V_{\mathrm{lin}}^0S_{L,i},\qquad
 D_{\ell,i}=D_{\ell+1,i}W_{\ell+1,\mathrm{lin}}^0S_{\ell,i}
 \quad(\ell=L-1,\ldots,1).
\]
Thus $D_{\ell,i}:\R^{n_\ell}\to\R^d$; set $D_{L+1,i}=I_d$, the identity on the output space. The \textbf{layer Jacobian} $J_\ell:\R^{n_\ell(n_{\ell-1}+1)}\to\R^{sd}$ and its \textbf{atomic kernel}, written in matrix form, are
\[
 J_\ell\operatorname{vec}(A_\ell)
   =(D_{\ell,i}A_\ell\bar h_{\ell-1,i}^0)_{i=1}^s,
 \qquad K_\ell=\{A_\ell:J_\ell\operatorname{vec}(A_\ell)=0\},
\]
where $\operatorname{vec}$ stacks matrix columns. We identify matrices and their vectorizations when writing $\ker J_\ell$. Consequently $J_Xw=\sum_{\ell=1}^{L+1}J_\ell\operatorname{vec}(A_\ell)$.

For later residual calculations, define the \textbf{edited downstream map} $T_{\ell,i}(w):\R^{n_\ell}\to\R^d$ by
\begin{align*}
 T_{L,i}(w)&=V_{\mathrm{lin}}^0+B_{\mathrm{lin}},\\
 T_{\ell,i}(w)&=T_{\ell+1,i}(w)S_{\ell+1,i}
             (W_{\ell+1,\mathrm{lin}}^0+A_{\ell+1,\mathrm{lin}}).
\end{align*}
The second recursion runs backward for $\ell=L-1,\ldots,1$. In particular, $D_{\ell,i}=T_{\ell,i}(0)S_{\ell,i}$. Zero activation slopes prevent recovering $T_{\ell,i}(0)$ by inverting this relation.

\begin{table}[!htbp]\centering
\caption{Notation used throughout the theoretical analysis. All states and maps are evaluated on the fixed sample $X$ unless another sample is specified.}\label{tab:appendix-notation}
\small\renewcommand{\arraystretch}{1.12}
\begin{tabular}{p{.25\linewidth}>{\columncolor{black!8}}p{.68\linewidth}}\toprule
Symbol & Meaning\\\midrule
$X=(x_i)_{i=1}^s$ & Sample of $s$ inputs in $\R^{n_0}$.\\
$L,n_\ell,d,p$ & Hidden-layer count, widths, output dimension and parameter count.\\
$W_\ell,V;\ \theta_0$ & Bias-augmented hidden/readout matrices; base parameter.\\
$z_{\ell,i},h_{\ell,i},f_i$ & Preactivation, hidden state and output on sample $i$; superscript $0$ denotes base values.\\
$w=(A_1,\ldots,A_L,B)$ & Parameter edit, with $A_{L+1}=B$ and $n_{L+1}=d$.\\
$\bar v=(v,1)$ & Bias-augmented vector; its change is augmented with zero.\\
$F_X(\theta),\ \|\cdot\|_X$ & Stacked outputs and sample root-mean-square Euclidean norm.\\
$\mathcal O,M$ & Convex ocean and its linear direction space.\\
$S_{\ell,i},\ \cC_X(\theta_0)$ & Base activation slope matrix and empirical activation chamber.\\
$J_X,J_\ell,K_\ell$ & Full Jacobian, layer Jacobian and atomic matrix kernel.\\
$D_{\ell,i}$ & Base derivative from a preactivation displacement to the output.\\
$T_{\ell,i}(w)$ & Edited-weight, base-slope map from a postactivation residual to the output.\\\bottomrule
\end{tabular}
\end{table}

\subsection{The frozen polynomial and its invariant subspaces}
\label{app:coefficient-test}
\paragraph{The frozen model.}
We separate the response to a parameter edit into changes that retain the base activation slopes and a remainder due to activation changes. Run the original network at $\theta_0$ and hold its states $z_{\ell,i}^0$, $h_{\ell,i}^0$ and $f_i^0$ fixed. We define the \textbf{frozen model} by replacing each activation with its affine linearization:
\[
 \sigma(z_{\ell,i}^0+u)\quad\longmapsto\quad h_{\ell,i}^0+S_{\ell,i}u,
 \qquad u\in\R^{n_\ell}.
\]
The activation value at the base point is retained; its slope is frozen at $S_{\ell,i}$. The edited weights remain variable, so their interactions across layers are retained. Let $\delta z_{\ell,i}$ and $\delta h_{\ell,i}$ denote the resulting preactivation and hidden-state changes. Starting from $\delta h_{0,i}=0$, we obtain
\begin{align}
 \delta z_{\ell,i}&=(W_{\ell,\mathrm{lin}}^0+A_{\ell,\mathrm{lin}})\delta h_{\ell-1,i}+A_{\ell,\mathrm{lin}}h_{\ell-1,i}^0+a_\ell,\nonumber\\
 \delta h_{\ell,i}&=S_{\ell,i}\delta z_{\ell,i},\label{eq:app-frozen}\\
 F_i^{\frz}(w)&=f_i^0+(V_{\mathrm{lin}}^0+B_{\mathrm{lin}})\delta h_{L,i}+B_{\mathrm{lin}}h_{L,i}^0+a_{L+1}.\nonumber
\end{align}
Write $h_{\ell,i}^{\frz}=h_{\ell,i}^0+\delta h_{\ell,i}$ for the frozen hidden state and stack the outputs into $F_X^{\frz}(w)$. Its argument is an edit, so $F_X^{\frz}(0)=F_X(\theta_0)$.

\paragraph{Polynomial structure and response decomposition.}
\begin{proposition}[Frozen polynomial]
The map $F_X^{\frz}$ is affine in the parameters of each layer separately and has total degree at most $L+1$. Its derivative at zero is $J_X$.
\end{proposition}
\begin{proof}
In \eqref{eq:app-frozen}, each layer multiplies a polynomial from earlier layers by its own edited weights. Its edit appears once and has not appeared in the preceding state. Induction over the $L$ hidden layers and readout proves separate affinity and the degree bound. Differentiating the recursion at zero gives the chain rule for the original network, because $S_{\ell,i}=\diag\sigma'(z_{\ell,i}^0)$. Thus the linear term is $J_Xw$.
\end{proof}
For a subset $I\subseteq\{1,\ldots,L+1\}$, let $\Psi_I$ collect the polynomial terms using precisely those edited layers. These terms are multilinear in the selected edits. The constant term is $\Psi_\varnothing=F_X(\theta_0)$, singleton terms sum to $J_Xw$, and the higher-degree terms contain interactions between edits.

We call the higher-degree part the \textbf{frozen interaction polynomial} $P_X^{\frz}$, and the true-minus-frozen response the \textbf{activation transition potential} $A_X^{\rm tr}$:
\begin{align*}
 P_X^{\frz}(w)&=F_X^{\frz}(w)-F_X(\theta_0)-J_Xw,\\
 A_X^{\rm tr}(w)&=F_X(\theta_0+w)-F_X^{\frz}(w).
\end{align*}
For the output drift $\Delta F_X(w)=F_X(\theta_0+w)-F_X(\theta_0)$, adding the definitions gives
\begin{equation}
 \Delta F_X(w)=J_Xw+P_X^{\frz}(w)+A_X^{\rm tr}(w).
 \label{eq:app-decomp}
\end{equation}
The first term is linear in the edit, and the second retains cross-layer interactions at fixed activation slopes. The third measures the output change caused by departures of the activations from their base affine tangents: gate switches for ReLU and curvature for smooth activations. For piecewise-linear activations, true and frozen states agree whenever each preactivation stays in its base affine piece; induction through the network then gives $A_X^{\rm tr}(w)=0$. Smooth activations can have a nonzero transition term arbitrarily near the base point.

\paragraph{Frozen oceans and invariant subspaces.}
We call a convex set $\mathcal O_{\frz}$ of edits a \textbf{frozen ocean} if $0\in\ri(\mathcal O_{\frz})$ and $F_X^{\frz}(w)=F_X^{\frz}(0)$ for every $w\in\mathcal O_{\frz}$. A linear parameter subspace $M$ is a \textbf{frozen invariant subspace} if this identity holds for every $w\in M$. We call it \textbf{layerwise} if $M=\bigoplus_{\ell=1}^{L+1}M_\ell$, where each matrix space $M_\ell$ occupies only layer $\ell$; $\bigoplus$ denotes the direct sum on these disjoint coordinates. We call other invariant subspaces \textbf{coupled}. The zero invariant subspace is allowed, while a positive-dimensional ocean requires a nonzero direction space.

\begin{restated}{Theorem~\ref{thm:correspondence} (Frozen oceans and invariant subspaces)}
If the frozen response vanishes on a relative neighborhood of zero in a linear space $M$, it vanishes everywhere on $M$. Thus every frozen ocean spans a frozen invariant subspace, and every frozen invariant subspace is itself a frozen ocean. Every such subspace satisfies $M\subseteq\ker J_X$.
\end{restated}
\begin{proof}
Suppose the frozen response vanishes on a relative neighborhood of zero in $M$. There is then $r>0$ such that it vanishes for every $v\in M$ with $\|v\|<r$. This includes the frozen-ocean case: take $M=\spn(\mathcal O_{\frz})$ and use $0\in\ri(\mathcal O_{\frz})$. We show that constancy near zero extends along every line in $M$.
Fix any nonzero $w\in M$ and consider the one-variable vector polynomial
\[
 q_w(t)=F_X^{\frz}(tw)-F_X^{\frz}(0),\qquad t\in\R.
\]
For $|t|<r/\|w\|$, the edit $tw$ lies in this neighborhood, so $q_w(t)=0$. Each scalar coordinate of $q_w$ therefore vanishes on an interval. A nonzero one-variable polynomial has only finitely many roots, so every coordinate is identically zero \citep{cox2015ideals}. Evaluating at $t=1$ gives
\[
 F_X^{\frz}(w)=F_X^{\frz}(0).
\]
This holds for every $w\in M$, including $w=0$, and proves that $M$ is an invariant subspace.

Conversely, a linear invariant subspace is convex, contains zero in its relative interior, and has constant frozen output, so it is a frozen ocean. Finally, for every $u\in M$, differentiating its constant output along the line $tu$ yields
\[
 J_Xu=\left.\frac{d}{dt}F_X^{\frz}(tu)\right|_{t=0}=0.
\]
Hence $M\subseteq\ker J_X$.
\end{proof}

\begin{remark}[Piecewise-linear activations]\label{rem:chamber-branch}
Suppose each base preactivation lies inside an affine piece of the activation. Every exact true ocean through $\theta_0$ spans a frozen invariant subspace, and every nonzero invariant subspace gives an exact true ocean on a sufficiently small subspace ball.
\end{remark}
\begin{proof}
Continuity gives a neighborhood of $\theta_0$ where all preactivations stay in their base pieces and the frozen and true networks agree. Intersect a true ocean with a sufficiently small ball in that neighborhood. Its direction space is unchanged, because $\theta_0$ is in its relative interior. In edit coordinates the intersection is a frozen ocean, and the preceding theorem applies. Conversely, a sufficiently small ball in an invariant subspace is convex, contains zero in its relative interior, and preserves the true outputs. Constancy also holds on the entire invariant subspace--chamber intersection, which need not be convex.
\end{proof}

\paragraph{A general coefficient test.}
Theorem~\ref{thm:correspondence} reduces the frozen-ocean question to constancy on a linear subspace. To test a candidate subspace, express every edit in a basis and collect the coefficients of its frozen output change. This gives finitely many equations that enforce constancy for every linear combination of the basis directions.

Choose a basis $u_1,\ldots,u_k$ of the candidate subspace $M$, where $k=\dim M$. Every edit in $M$ has a unique expression $w=\sum_{j=1}^k\alpha_j u_j$. Thus $\alpha\in\R^k$ gives the coordinates of the edit $w$ in this basis. Substitute this expression into the frozen map and subtract the base output:
\[
 Q_M(\alpha)=F_X^{\frz}\left(\sum_{j=1}^k\alpha_j u_j\right)-F_X(\theta_0)
 =\sum_{1\le |\nu|\le L+1}c_\nu\alpha^\nu.
\]
Here $\nu\in\mathbb N_0^k$ is a tuple of nonnegative integers, $|\nu|=\sum_j\nu_j$, $\alpha^\nu=\prod_j\alpha_j^{\nu_j}$, and $c_\nu\in\R^{sd}$ is the coefficient of that monomial. For example, with two basis directions the monomials include $\alpha_1$, $\alpha_2$, $\alpha_1^2$, $\alpha_1\alpha_2$ and $\alpha_2^2$. The invariant subspace condition requires $Q_M(\alpha)=0$ for \emph{every} choice of $\alpha\in\R^k$: every linear combination of the basis edits must leave every sampled output unchanged. Checking the basis directions alone is insufficient. For example, the scalar polynomial $q(\alpha_1,\alpha_2)=\alpha_1\alpha_2$ vanishes on both coordinate axes but equals one at $(1,1)$. Thus the test must check every monomial coefficient, including the mixed coefficients that describe simultaneous motion along several basis directions.
\begin{proposition}[General coefficient test]\label{prop:coefficients}
The candidate $M$ is a frozen invariant subspace exactly when every nonconstant coefficient $c_\nu$ vanishes. There are at most
\[
 sd\left[\binom{k+L+1}{L+1}-1\right]
\]
scalar coefficient conditions.
\end{proposition}
\begin{proof}
The basis map $\alpha\mapsto\sum_j\alpha_j u_j$ covers $M$, so invariant subspace constancy is equivalent to $Q_M(\alpha)=0$ for all $\alpha$. Distinct monomials are linearly independent as polynomial functions. Applying this fact to each output coordinate makes constancy equivalent to $c_\nu=0$ for every monomial.

The standard stars-and-bars count gives $\binom{k+L+1}{L+1}$ monomials of degree at most $L+1$ in $k$ variables. Removing the constant monomial, whose coefficient is already zero, and multiplying by $sd$ output coordinates gives the stated bound.
\end{proof}

The test is general because it applies to atomic, layerwise and coupled candidate spaces. For coupled edits, terms from different layer subsets can contribute to the same monomial and must be summed before testing. Independent layer coordinates in a layerwise space permit the more efficient test in Appendix~\ref{app:composition-proof}, also used in the main paper.

\subsection{Smooth exact oceans}\label{app:smooth-exact}
Our construction first makes the frozen response vanish on an invariant subspace and then controls the activation-transition term. Requiring these contributions to vanish separately is a sufficient route to an exact ocean. The definition of an exact ocean only requires their sum to vanish. For smooth activations, frozen interactions and activation curvature could therefore cancel each other.

Assume $F_X$ is analytic near $\theta_0$, meaning that it equals a convergent Taylor series there. Decompose the frozen interaction polynomial and the transition term by total degree:
\[
 P_X^{\frz}(w)=\sum_{q=2}^{L+1}P_q(w),\qquad
 A_X^{\rm tr}(w)=\sum_{q=2}^{\infty}A_q(w).
\]
Here $P_q,A_q:\R^p\to\R^{sd}$ are homogeneous polynomial maps of degree $q$: scaling their argument by $t$ scales their value by $t^q$. Both series start at degree two because the true and frozen models have the same value and Jacobian at zero edit. Set $P_q=0$ for $q>L+1$.

\begin{proposition}[Necessary Taylor conditions]
If $F_X$ is constant on a relative ball about $\theta_0$ in $\theta_0+M$, then
\[
 J_X|_M=0,\qquad (P_q+A_q)|_M=0\quad(q\ge2),
\]
where restriction to $M$ means evaluation at every $w\in M$.
\end{proposition}
\begin{proof}
Fix $w\in M$ and substitute $tw$ into the drift decomposition. Constancy for all sufficiently small $t$ forces every coefficient of the convergent one-variable power series to vanish. This holds for every $w$.
\end{proof}
Thus an exact smooth ocean could, in principle, have nonzero frozen coefficients canceled by activation curvature.

The following conjecture asks whether the frozen construction captures every exact local ocean for the stated smooth MLPs. The invariant subspace tests and radius bounds proved in this paper hold independently of this conjecture.
\begin{appconjecture}
For MLPs with tanh, sigmoid or softplus activations, outside an exceptional set of data and base parameters of Lebesgue measure zero, the direction space of every exact local ocean is a frozen invariant subspace.
\end{appconjecture}
The scope matches the architecture analyzed here and the activation families covered by the one-unit calculation below. That calculation supplies evidence for these families; it establishes no general classification for other activations or architectures. The reciprocal-affine counterexample below shows why an unrestricted claim over smooth activations would be false. Cancellations among several units remain open even within the stated MLP setting.

\begin{example}[The conjecture in a one-unit model]
Consider scalar output $f(t)=(v+tc)\sigma(z+t)$ on an interval containing zero. Here $z,v$ are the base preactivation and readout, $c$ is the readout change per unit of $t$, and a nonzero preactivation velocity has been absorbed into $t$. For tanh, sigmoid or softplus, this function cannot be a nonzero constant. Thus this nondegenerate one-unit setting admits no exact line sustained by cancellation between frozen interactions and activation curvature.
\end{example}
\begin{proof}
Suppose $f$ is a nonzero constant. Then $v\ne0$, and its first two derivatives at zero give
\[
 c\sigma(z)+v\sigma'(z)=0,\qquad
 2c\sigma'(z)+v\sigma''(z)=0.
\]
Multiply the second identity by $\sigma(z)$ and use the first to eliminate $c$. Since $v\ne0$, a necessary condition is $\sigma(z)\sigma''(z)-2(\sigma'(z))^2=0$. For $\sigma(z)=\tanh z$, sigmoid $\sigma(z)=(1+e^{-z})^{-1}$, and softplus $\sigma(z)=\log(1+e^z)$, this expression is, respectively,
\[
 -2\operatorname{sech}^2z,\qquad
 -\frac{e^{2z}}{(1+e^z)^3},\qquad
 \frac{e^z}{(1+e^z)^2}[\log(1+e^z)-2e^z].
\]
Here $\operatorname{sech}z=1/\cosh z$. Each expression is strictly negative; for the last, use $\log(1+a)<a$ for $a>0$. None of these activations therefore admits such a one-unit line.
\end{proof}
The activation restriction matters: for fixed scalars $C\ne0,a\ne0,b$, the activation $\sigma(z)=C/(az+b)$ and affine readout $a(z+t)+b$ give constant output $C$ wherever the denominator is nonzero. These calculations do not exclude cancellations between several units.

\subsection{Structural symmetries and affine invariance}
Following the standard parameter-space notion \citep{godfrey2022symmetries}, a \textbf{parameter symmetry} is an invertible transformation $T$ with $f_{T(\theta)}(x)=f_\theta(x)$ for every parameter $\theta$ and input $x$, where $f_\theta$ denotes the network function. Its \textbf{orbit} through $\theta_0$ is the set of parameters reachable by transformations in the chosen symmetry group. Every point of this orbit has the same sampled outputs.

\paragraph{Known structural symmetries.}
The following transformations preserve the function \citep{godfrey2022symmetries}:
\begin{itemize}
\item \textbf{Hidden-unit permutations.} For a permutation matrix $P$, replace $(W_\ell,W_{\ell+1,\mathrm{lin}})$ by $(PW_\ell,W_{\ell+1,\mathrm{lin}}P^{-1})$. Coordinatewise activation satisfies $\sigma(Pz)=P\sigma(z)$, so the permutation cancels at the next linear map. The augmented matrix $PW_\ell$ also permutes the bias of layer $\ell$; the bias of the following layer stays unchanged. At the final hidden layer use $V_{\mathrm{lin}}$ as the next matrix.
\item \textbf{Positive rescaling.} If $\sigma(cz)=c\sigma(z)$ for $c>0$, the same formula works for a positive diagonal matrix $P$. For one hidden unit with incoming row $a$ (including bias), outgoing column $b$, and preceding augmented feature $\bar h$,
\[
 b\sigma(a\bar h)=(b/c)\sigma(ca\bar h).
\]
\end{itemize}

\begin{remark}[A symmetry tangent need not define an ocean]
The invariant scaling path $(a,b)\mapsto(e^t a,e^{-t}b)$ has tangent $(a,-b)$ at zero. Its straight tangent edit instead gives, for $|t|<1$,
\[
 (1-t)b\,\sigma((1+t)a\bar h)=(1-t^2)b\,\sigma(a\bar h).
\]
Whenever this second-order change reaches the sampled output, the straight direction belongs to no exact ocean through the base point. An exact ocean would contain a small segment in that direction and require constant outputs there. Thus tangency to an invariant orbit alone does not establish ocean membership. If the unit is in the final hidden layer, a nonzero contribution $b\sigma(a\bar h)$ on any sample already ensures the output changes.
\end{remark}

Define the \textbf{symmetry tangent space} $T_{\rm sym}$ as the span of derivatives at $\theta_0$ of differentiable symmetry paths through $\theta_0$. Differentiating their invariant outputs gives $T_{\rm sym}\subseteq\ker J_X$. For an invariant subspace $M$, the quotient dimension $\dim M-\dim(M\cap T_{\rm sym})$ counts its directions beyond these tangents. Tangent dimensions are evaluated at the actual base parameters, where degeneracies can reduce them.

Some architectural components also admit exact affine symmetry families. Appendix~\ref{app:layernorm} treats the shifts removed by LayerNorm and the resulting affine invariance.

\subsection{Atomic oceans}
\label{app:atomic-proof}
\subsubsection{Elements of the atomic ocean}
An \textbf{atomic edit} changes a single parameter layer and leaves every other layer fixed. Write $w_\ell(A)$ for the full parameter edit with matrix $A$ in layer $\ell$ and zero in all other layers. Separate affinity of the frozen model gives the exact identity
\[
 F_X^{\frz}(w_\ell(A))-F_X^{\frz}(0)
 =J_\ell\operatorname{vec}(A)
 =(D_{\ell,i}A\bar h_{\ell-1,i}^0)_{i=1}^s.
\]
\textbf{The atomic kernel $K_\ell$ is therefore the complete frozen constancy space for one moving layer.} Every edit in $K_\ell$ preserves the frozen output at every magnitude; every edit outside it changes at least one sampled output. No higher-order terms survive when only one layer moves. We now decompose this space, starting with edits that preserve the features themselves.

\paragraph{Feature-preserving edits.}
For each layer, define its \textbf{input-feature span} and rank by
\[
 \cH_\ell=\spn\{\bar h_{\ell-1,i}^0:i\le s\},\qquad
 r_\ell=\dim\cH_\ell.
\]
We define the \textbf{feature-null space} as $G_\ell=\{A:A\cH_\ell=0\}$. 
\begin{restated}{Theorem~\ref{thm:feature-ocean} (Feature-null ocean)}
The direct sum of the feature-null spaces is
\[
 M_{\rm feat}=\bigoplus_{\ell=1}^{L+1}\{A_\ell:A_\ell\cH_\ell=0\},
 \qquad
 \dim M_{\rm feat}=\sum_{\ell=1}^{L+1}n_\ell(n_{\ell-1}+1-r_\ell).
\]
The affine set $\theta_0+M_{\rm feat}$ is an exact flat ocean. For every $w\in M_{\rm feat}$ and any pointwise activation,
$F_X(\theta_0+w)=F_X(\theta_0)$, without a radius restriction.
\end{restated}
\begin{proof}
We prove by induction that every true hidden state stays at its base value. The input is unchanged: $h_{0,i}(\theta_0+w)=x_i=h_{0,i}^0$. Suppose the claim holds through layer $\ell-1$. The input to layer $\ell$ is then still $\bar h_{\ell-1,i}^0\in\cH_\ell$, so
\[
 z_{\ell,i}(\theta_0+w)
 =(W_\ell^0+A_\ell)\bar h_{\ell-1,i}^0
 =W_\ell^0\bar h_{\ell-1,i}^0
 =z_{\ell,i}^0.
\]
Applying the same pointwise activation to equal preactivations gives $h_{\ell,i}(\theta_0+w)=h_{\ell,i}^0$. Induction reaches layer $L$. The readout then satisfies
\[
 (V^0+B)\bar h_{L,i}^0=V^0\bar h_{L,i}^0=f_i^0,
\]
because $B\cH_{L+1}=0$. These equalities hold for edits of every magnitude.

Each row of $A_\ell$ may be any vector orthogonal to $\cH_\ell$, a space of dimension $n_{\ell-1}+1-r_\ell$. The $n_\ell$ rows are independent, giving $n_\ell(n_{\ell-1}+1-r_\ell)$ dimensions in layer $\ell$. Summing over the disjoint layers proves the dimension formula.
\end{proof}

\paragraph{The complete atomic dimension.}
Define the \textbf{downstream covector span} and its dimension by
\[
 \cC_\ell=\sum_i\operatorname{row}(D_{\ell,i}),\qquad
 r_{d,\ell}=\dim\cC_\ell.
\]
Here $\operatorname{row}$ denotes row space and a sum of spaces means their linear span. Write $\cC_\ell^\perp$ for its Euclidean orthogonal complement. The intermediate space
\[
 \mathcal N_\ell=\{A:A\cH_\ell\subseteq\cC_\ell^\perp\}
\]
sends every realized feature to a downstream-invisible vector. Consequently $G_\ell\subseteq\mathcal N_\ell\subseteq K_\ell$.

\begin{restated}{Theorem~\ref{thm:atomic} (Exact atomic dimension)}
Define the \emph{co-occurrence deficit} by
\[
 \delta_{\rm joint,\ell}=r_\ell r_{d,\ell}-\rank J_\ell\ge0.
\]
The complete atomic frozen invariant subspace is $K_\ell$, with
\begin{equation}
 \dim K_\ell=\underbrace{n_\ell(n_{\ell-1}+1-r_\ell)}_{\text{feature-null}}
 +\underbrace{r_\ell(n_\ell-r_{d,\ell})}_{\text{downstream-null}}
 +\underbrace{\delta_{\rm joint,\ell}}_{\text{co-occurrence}}.
\label{eq:app-atomic}
\end{equation}
The last two terms are the successive quotient dimensions
$\dim(\mathcal N_\ell/G_\ell)$ and $\dim(K_\ell/\mathcal N_\ell)$, so the count avoids double-counting overlapping mechanisms.
\end{restated}
\begin{proof}
For column-wise vectorization, an output coordinate of $D_{\ell,i}A\bar h_{\ell-1,i}^0$ has coefficient vector $\bar h_{\ell-1,i}^0\otimes D_{\ell,i}^\top e_a$, where $e_a$ is the $a$th standard basis vector of $\R^d$. The tensor product $\cH_\ell\otimes\cC_\ell$ is the span of such products with the two factors chosen independently from the two spaces. Thus
\[
 \operatorname{row}(J_\ell)
 =\spn_{i,a}\{\bar h_{\ell-1,i}^0\otimes D_{\ell,i}^\top e_a\}
 \subseteq\cH_\ell\otimes\cC_\ell.
\]
Products of basis vectors form a basis of the tensor product, giving dimension $r_\ell r_{d,\ell}$ and proving nonnegativity of the deficit. Rank--nullity, $\dim\ker J_\ell=n_\ell(n_{\ell-1}+1)-\rank J_\ell$, gives \eqref{eq:app-atomic}.

A map in $G_\ell$ is arbitrary on a complement of $\cH_\ell$ and zero on $\cH_\ell$. A map in $\mathcal N_\ell$ additionally sends its $r_\ell$-dimensional feature span into the $(n_\ell-r_{d,\ell})$-dimensional downstream complement. These choices add $r_\ell(n_\ell-r_{d,\ell})$ dimensions. Subtracting from $\dim K_\ell$ leaves the deficit as the final quotient dimension. The frozen response to one layer is linear, so no further coefficient constraints exist.
\end{proof}
For the readout, $D_{L+1,i}=I_d$ and $r_{d,L+1}=d$; its kernel is entirely feature-null. For any piecewise-linear hidden layer, the atomic frozen space preserves true outputs as long as the edit remains in the base chamber.

\subsubsection{The co-occurrence deficit}
\label{app:cooccurrence-deficit}
\paragraph{Why sample pairings matter.}
The marginal spans describe which feature directions and downstream covectors occur anywhere in the data. The Jacobian pairs them only when they occur on the same sample. Its row space can therefore be smaller than the full tensor product of the two marginal spans.

\textbf{A linear network has no deficit.} Its activation derivatives are constant, so the downstream map is the same on every sample. Every sample feature is then paired with every row of this common map, giving the full tensor-product span. The same conclusion holds whenever a layer has a sample-independent downstream map:
\begin{corollary}[Constant downstream map]
\label{cor:app-constant-downstream}
If $D_{\ell,i}$ is the same for every sample, then $\rank J_\ell=r_\ell r_{d,\ell}$ and $\delta_{\rm joint,\ell}=0$.
\end{corollary}
\begin{proof}
Write $D_\ell$ for the common map. The Jacobian rows pair every sample feature with every row of $D_\ell$, so
\[
 \operatorname{row}(J_\ell)
 =\spn_i\{\bar h_{\ell-1,i}^0\}\otimes\operatorname{row}(D_\ell)
 =\cH_\ell\otimes\cC_\ell.
\]
Its dimension is $r_\ell r_{d,\ell}$, proving the claim.
\end{proof}
A linear network can still have a large atomic ocean through deficient feature or downstream ranks. With nonlinear activations, sample-dependent slopes can both enlarge the downstream span and leave some feature--downstream pairings absent. The following example isolates the latter effect.

\begin{example}[Full marginal spans with missing pairings]
Consider an edited matrix $A\in\R^{2\times2}$ and a scalar output. Let $e_1,e_2$ be the standard basis of $\R^2$. Take two augmented input features and downstream maps
\[
 \bar h_{\ell-1,1}^0=e_2,\quad D_{\ell,1}=e_1^\top,
 \qquad
 \bar h_{\ell-1,2}^0=e_1+e_2,\quad D_{\ell,2}=e_2^\top.
\]
Both feature vectors have last coordinate one, as required for the bias convention. Their span and the downstream span are both $\R^2$, so $r_\ell=r_{d,\ell}=2$. Yet the two output changes are only
\[
 J_\ell\operatorname{vec}(A)=(A_{12},\ A_{21}+A_{22}).
\]
These are two independent constraints on four parameters. Thus
\[
 K_\ell=\left\{\begin{pmatrix}a&0\\b&-b\end{pmatrix}:a,b\in\R\right\},
 \qquad \delta_{\rm joint,\ell}=4-2=2.
\]
The entire atomic ocean in this example comes from missing pairings.
\end{example}

Selective ReLU units and sample clustering describe two ways that these pairings can be missing. The selective-unit construction uses ReLU's zero slopes. The cluster decomposition applies to any activation differentiable at the base preactivations.

\paragraph{Selective ReLU units and composition.}
For hidden unit $j=1,\ldots,n_\ell$, define its \textbf{active-feature span}
\[
 \cH_{\ell j}^{\rm act}(X)
 =\spn\{\bar h_{\ell-1,i}^0:z_{\ell,i,j}^0>0\}.
\]
We call $G_\ell^{\rm sel}=\{A:S_{\ell,i}A\bar h_{\ell-1,i}^0=0\ \forall i\}$ the \textbf{selective-unit space}. On an active sample, the ReLU slope is one, so row $j$ must annihilate $\bar h_{\ell-1,i}^0$. On an inactive sample the slope is zero and imposes no constraint on that row. Each row can therefore be any vector orthogonal to $\cH_{\ell j}^{\rm act}(X)$, independently of the other rows, giving
\[
 \dim G_\ell^{\rm sel}
 =\sum_{j=1}^{n_\ell}\bigl(n_{\ell-1}+1-\dim\cH_{\ell j}^{\rm act}(X)\bigr).
\]
A unit contributes whenever its active features span a proper subspace, even if it is active on many samples. Feature-null edits are selective, and every selective edit is atomically invisible, so $G_\ell\subseteq G_\ell^{\rm sel}\subseteq K_\ell$. The next result explains why selective spaces also compose across layers.

\begin{restated}{Proposition~\ref{prop:selective-composition} (Selective-unit composition)}
The direct sum of the hidden selective-unit spaces and the feature-null readout is a frozen invariant subspace. Its dimension is the sum of the hidden counts above plus $d(n_L+1-r_{L+1})$ for the readout. It preserves the true ReLU network locally within the base chamber.
\end{restated}
\begin{proof}
We prove by induction that every frozen hidden-state change is zero. The base case is $\delta h_{0,i}=0$, since the inputs are fixed. Suppose $\delta h_{\ell-1,i}=0$ for every sample. Substituting into the frozen recursion \eqref{eq:app-frozen} gives
\begin{align*}
 \delta z_{\ell,i}
 &=(W_{\ell,\mathrm{lin}}^0+A_{\ell,\mathrm{lin}})\underbrace{\delta h_{\ell-1,i}}_{=0}
   +A_\ell\bar h_{\ell-1,i}^0
   =A_\ell\bar h_{\ell-1,i}^0,\\
 \delta h_{\ell,i}
 &=S_{\ell,i}A_\ell\bar h_{\ell-1,i}^0=0,
\end{align*}
where the last equality is the defining selective-unit constraint. Thus the frozen postactivation remains fixed even though the preactivation may move on inactive units. In particular, no hidden-state change reaches the next layer to interact with its edit.

Induction gives $\delta h_{L,i}=0$. The feature-null readout then gives
\[
 F_i^{\frz}(w)-f_i^0
 =(V_{\mathrm{lin}}^0+B_{\mathrm{lin}})\delta h_{L,i}
   +B\bar h_{L,i}^0=0.
\]
The direct sum is therefore an invariant subspace. Its layer coordinates are disjoint, so its dimension is the sum of their dimensions. Inside the base chamber, the frozen and true ReLU networks agree, proving local true invariance.
\end{proof}

\begin{remark}[Selective invariance and gate crossings]
Selective edits can change the preactivations of inactive units while leaving their frozen postactivations zero. If an inactive unit crosses zero, its true ReLU output can change. The selective invariant subspace therefore has a local true-network guarantee inside the base chamber. Feature-null edits preserve the preactivations themselves and have the global guarantee of Theorem~\ref{thm:feature-ocean}.
\end{remark}

\paragraph{Cluster decomposition.}
Sample clustering describes missing pairings for general activations through their base features and downstream derivatives. Choose any partition of the sample indices into nonempty, disjoint groups, and write $X_g$ for the samples in group $g$. Let $\cH_\ell(X_g)$, $\cC_\ell(X_g)$ and $J_\ell(X_g)$ be the feature span, downstream span and layer Jacobian restricted to that group. Define
\[
 \cP=\sum_g\cH_\ell(X_g)\otimes\cC_\ell(X_g).
\]
\begin{proposition}[Cluster decomposition of the deficit]
The co-occurrence deficit splits into two nonnegative terms:
\[
 \delta_{\rm joint,\ell}
 =\underbrace{r_\ell r_{d,\ell}-\dim\cP}_{\text{across clusters}}
 +\underbrace{\dim\cP-\rank J_\ell}_{\text{within clusters}}.
\]
If the downstream map is constant within each group, the within-cluster term is zero.
\end{proposition}
\begin{proof}
Stacking the sample constraints gives
\[
 \operatorname{row}(J_\ell)=\sum_g\operatorname{row}(J_\ell(X_g))
 \subseteq\cP\subseteq\cH_\ell\otimes\cC_\ell.
\]
These inclusions prove nonnegativity. Add and subtract $\dim\cP$ in the definition of $\delta_{\rm joint,\ell}$ to obtain the identity. When $D_{\ell,i}$ is constant within each group, Corollary~\ref{cor:app-constant-downstream} applied to $X_g$ gives $\operatorname{row}(J_\ell(X_g))=\cH_\ell(X_g)\otimes\cC_\ell(X_g)$. Summing makes $\operatorname{row}(J_\ell)=\cP$.
\end{proof}

Call a partition \textbf{meaningful} when its within-cluster deficit is zero, so missing pairings across groups explain the entire co-occurrence deficit. A \textbf{minimal meaningful partition} uses the fewest groups among all meaningful partitions. The singleton partition always has zero within-cluster deficit; minimizing the group count asks for the most compact explanation. Constant downstream maps within groups are sufficient for meaningfulness, although the same dimension equality can hold for varying maps.

In trained networks, we adopt an approximate version of the constant-downstream criterion: we group downstream maps whose normalized Frobenius distances are at most a tolerance $\varepsilon$. The late-training groups in Appendix~\ref{app:mechanism}, Table~\ref{tab:dmap-threshold}, are nearly class-pure across ReLU, leaky ReLU and GELU. This establishes approximate downstream concentration. Whether such a partition explains the deficit is measured separately by its within-cluster term; a distance tolerance alone does not certify zero deficit or minimality.

\subsection{Layerwise invariant subspaces at arbitrary depth}
\label{app:composition-proof}
Atomic invariant subspaces can fail to compose: two edits that are individually invisible may change the output through their cross-layer product. This section gives an exact test, a construction certified by that test, and the dimension results available for specific network classes.

\subsubsection{An efficient test for layerwise invariant subspaces}
Fix a candidate $M=\bigoplus_{\ell=1}^{L+1}M_\ell$, with each $M_\ell$ a matrix space in layer $\ell$. For a matrix space $N$ and vector space $U$, write $NU=\spn\{Au:A\in N,u\in U\}$. Starting from $R_{0,i}=\{0\}$, the recursion
\begin{align}
 U_{\ell,i}&=\spn\{\bar h_{\ell-1,i}^0\}+(R_{\ell-1,i}\times\{0\}),\label{eq:received}\\
 R_{\ell,i}&=S_{\ell,i}\bigl(W_{\ell,\mathrm{lin}}^0R_{\ell-1,i}+M_\ell U_{\ell,i}\bigr),\nonumber
\end{align}
collects the coefficients of the frozen hidden-state change in $R_{\ell,i}$ and the directions received by edit $\ell$ in $U_{\ell,i}$. The appended zero keeps the bias coordinate fixed. At the readout only the received span is needed, and $D_{L+1,i}=I_d$. All spaces in the algorithm are stored as bases; each sum or image is reduced to a basis.

\begin{algorithm}[H]
\caption{Exact layerwise invariant subspace test}\label{alg:app-branch-test}
\small
\begin{algorithmic}[1]
\Require Bases of $M_1,\ldots,M_{L+1}$; base states, weights, slopes and downstream maps on $X$.
\State $R_{0,i}\gets\{0\}$ for every sample $i$.
\For{$\ell=1,\ldots,L+1$}
 \For{each sample $i$}
  \State $U_{\ell,i}\gets\spn\{\bar h_{\ell-1,i}^0\}+(R_{\ell-1,i}\times\{0\})$.
  \If{$D_{\ell,i}Au\ne0$ for any basis pair $A\in M_\ell$, $u\in U_{\ell,i}$}
   \State \Return false.
  \EndIf
  \If{$\ell\le L$}
   \State $R_{\ell,i}\gets S_{\ell,i}(W_{\ell,\mathrm{lin}}^0R_{\ell-1,i}+M_\ell U_{\ell,i})$.
  \EndIf
 \EndFor
\EndFor
\State \Return true.
\end{algorithmic}
\end{algorithm}

\begin{theorem}[Exactness and completeness of the layerwise test]
\label{thm:composition}
In exact arithmetic, Algorithm~\ref{alg:app-branch-test} returns true if and only if $M$ is a frozen invariant subspace. Equivalently, every received span satisfies
\[
 D_{\ell,i}M_\ell U_{\ell,i}=\{0\}
 \qquad(1\le\ell\le L+1,\ 1\le i\le s).
\]
\end{theorem}
\begin{proof}[Proof of Theorem~\ref{thm:composition}]
Choose independent basis coordinates in every $M_\ell$. The frozen hidden-state change obeys
\[
 \delta h_{\ell,i}=S_{\ell,i}\left(
 W_{\ell,\mathrm{lin}}^0\delta h_{\ell-1,i}
 +A_\ell \bar h_{\ell-1,i}^0
 +A_{\ell,\mathrm{lin}}\delta h_{\ell-1,i}\right).
\]
The three groups of monomials respectively omit slot $\ell$, use only slot $\ell$, and use slot $\ell$ together with lower slots. Their independent coefficient vectors span exactly the spaces in \eqref{eq:received}. Induction from $R_{0,i}=0$ thus identifies $R_{\ell,i}$ as the span of every nonconstant coefficient in $\delta h_{\ell,i}$.

Any nonconstant output monomial has a unique highest perturbed slot. If it is hidden slot $\ell$, its coefficient is $D_{\ell,i}A u$, where $A$ is a basis matrix of $M_\ell$ and $u$ is the base input or a lower coefficient augmented with zero. These vectors span $U_{\ell,i}$. If the highest slot is the readout, the coefficient is $Bu$ with $u\in U_{L+1,i}$. The conditions in the theorem therefore kill every coefficient. Conversely, if the output polynomial is constant, independence of the layer coordinates forces each such coefficient to vanish, hence also every vector in their span. These are exactly the zero tests in Algorithm~\ref{alg:app-branch-test}, proving both implications.

\end{proof}

\begin{remark}[Cost of the test]
After bases have been formed, the number of scalar zero tests is at most
\[
 N_{\rm test}=d\sum_{i=1}^s\sum_{\ell=1}^{L+1}k_\ell u_{\ell,i},
 \qquad k_\ell=\dim M_\ell,\quad u_{\ell,i}=\dim U_{\ell,i}.
\]
Since $u_{\ell,i}\le n_{\ell-1}+1$, this is at most $sd\sum_\ell k_\ell(n_{\ell-1}+1)$. For bounded widths and basis sizes, the number of tests grows linearly with the sample count and depth. Each test involves matrix--vector products, and forming the bases requires additional linear algebra; their total arithmetic cost is polynomial in the supplied dimensions. 

To compare this cost with the general coefficient test (Proposition~\ref{prop:coefficients}), choose independent coordinates in each layer space. Direct enumeration then has up to $sd[\prod_\ell(1+k_\ell)-1]$ scalar conditions, since a monomial can use either no coordinate or one of $k_\ell$ coordinates from each layer. This count can grow exponentially with depth; the received-span recursion avoids enumerating those monomials. Numerical implementations must additionally state their rank and residual tolerances.
\end{remark}

\subsubsection{Constructing layerwise invariant subspaces}
The test also constructs an invariant subspace by choosing compatible matrix spaces one layer at a time. The following greedy procedure takes the full compatible space at each step. Any ordering is allowed, with each layer visited once.

Let $\mathcal S$ be the set of layers already selected and let $\mathcal B_j$ be a basis of $M_j$ for each $j\in\mathcal S$. For the new layer $\ell$, the admissible space is
\begin{equation}
\begin{split}
 M_\ell=\{A:\;&\Psi_{I\cup\{\ell\}}(A,(A_j)_{j\in I})=0\\
 &\forall I\subseteq\mathcal S,\quad
   \forall A_j\in\mathcal B_j\ (j\in I)\}.
\end{split}\label{eq:greedy-constraints}
\end{equation}
Each argument of $\Psi$ occupies its indicated layer slot, with $A$ in slot $\ell$. The empty set $I$ gives the atomic constraint. Every equation is homogeneous and linear in $A$, because a layer appears at most once in each multilinear term. In bottom-up order these equations reduce to Eq.~\eqref{eq:bottom-up-constraints}.

\paragraph{Computing the constraints by backward propagation.}
Equation~\eqref{eq:greedy-constraints} can be evaluated without enumerating subsets. Compute $U_{\ell,i}$ by the received-space recursion using the already selected lower layers. For the higher layers, put $W_{L+1}^0=V^0$ and start with the output covector space $\mathcal Z_{L+1,i}=\R^d$; recursively set, for $j=L+1,\ldots,\ell+1$,
\begin{equation}
 \mathcal Z_{j-1,i}
 =\spn\{cW_{j,\mathrm{lin}}^0S_{j-1,i},\ cB_{\mathrm{lin}}S_{j-1,i}:
 c\in\mathcal Z_{j,i},\ B\in\mathcal B_j\cup\{0\}\}.
 \label{eq:app-backward-constraints}
\end{equation}
Here covectors are written as rows, and an unselected layer has $\mathcal B_j=\{0\}$. The compatible space is precisely
\[
 M_\ell=\{A:cAu=0\quad\forall i,\ c\in\mathcal Z_{\ell,i},\ u\in U_{\ell,i}\}.
\]
The forward recursion spans all lower coefficient vectors, and the backward recursion spans all higher coefficient covectors. Their products are the coefficients containing the new edit $A$, so these equations are equivalent to~\eqref{eq:greedy-constraints}. Orthonormalizing at each step bounds both propagated spaces by their layer widths; testing their basis pairs gives a linear system in $A$ of polynomial size.

\begin{algorithm}[H]
\caption{Greedy construction in any layer order}\label{alg:app-branch-build}
\small
\begin{algorithmic}[1]
\Require Base network on $X$ and a permutation $\pi$ of $\{1,\ldots,L+1\}$.
\State $M_j\gets\{0\}$ for all layers $j$.
\For{$t=1,\ldots,L+1$}
 \State $\ell\gets\pi(t)$; keep $M_j$, $j\ne\ell$, fixed.
 \State Set $\mathcal S\gets\{\pi(1),\ldots,\pi(t-1)\}$ and choose bases $\mathcal B_j$ of the selected $M_j$.
 \State Compute $U_{\ell,i}$ forward and $\mathcal Z_{\ell,i}$ backward by Eq.~\eqref{eq:app-backward-constraints}.
 \State Impose $cAu=0$ for every sample and basis pair $c\in\mathcal Z_{\ell,i}$, $u\in U_{\ell,i}$.
 \State $M_\ell\gets$ the full nullspace of these constraints.
\EndFor
\State \Return $\bigoplus_\ell M_\ell$.
\end{algorithmic}
\end{algorithm}

\begin{proposition}[Validity of greedy layer addition]
At each step of Algorithm~\ref{alg:app-branch-build}, the admissible matrices form a linear space. Adding that entire space preserves a frozen invariant subspace. The returned space is therefore a frozen invariant subspace for every ordering $\pi$.
\end{proposition}
\begin{proof}
Assume the previously selected direct sum is an invariant subspace. Every new monomial contains the unknown layer matrix $A$ exactly once, because the frozen map is affine in that layer. Its coefficient is linear in $A$. Vanishing of all new coefficients imposes homogeneous linear equations, whose solution space is the stated nullspace. Every linear combination of admissible matrices satisfies those equations, so admitting the entire space preserves the invariant subspace. The initial zero space is an invariant subspace, and induction completes the proof.
\end{proof}
In bottom-up order, the constraints are $D_{\ell,i}Au=0$ for every sample and received basis vector. For other orders, already selected higher layers impose further mixed-coefficient constraints. These must be included when forming the linear system.

\paragraph{Order and multiplicity.}
Keeping earlier choices fixed can change the final invariant subspace and its dimension. The reduced-data experiment in Table~\ref{tab:subset-order} gives different dimensions for bottom-up and top-down orders. In contrast, the measured full-data MLP atomic kernels compose, so all tested orders recover the same sum (Appendix~\ref{app:selective}). The order is a discrete choice; allowing smaller compatible spaces also introduces continuous choices of orientation.

\subsubsection{Dimensions and maximal layerwise invariant subspaces}
\label{app:deep-linear}\label{app:sample-gates}
An earlier edit can enlarge the feature span received by later layers and reduce their available dimensions. Choosing the largest current space therefore need not maximize the sum of all layer dimensions. Write $k_{\max}^{\rm lw}$ for the largest dimension of a layerwise frozen invariant subspace. An invariant subspace is \textbf{inclusion-maximal} if no larger layerwise invariant subspace contains it; different inclusion-maximal invariant subspaces may have different dimensions.

\begin{proposition}[Atomic ceiling and its attainment]\label{prop:app-atomic-ceiling}
Every layerwise invariant subspace has $M_\ell\subseteq K_\ell$ and
\[
 \dim M\le\min\left\{\dim\ker J_X,\sum_{\ell=1}^{L+1}\dim K_\ell\right\}.
\]
If the full atomic sum $\bigoplus_\ell K_\ell$ passes Algorithm~\ref{alg:app-branch-test}, it contains every layerwise invariant subspace. It is then the unique inclusion-maximal and unique maximum-dimensional layerwise invariant subspace.
\end{proposition}
\begin{proof}
For $A\in M_\ell$, setting every other layer to zero yields an edit in $M$. It must preserve the frozen outputs, so $A\in K_\ell$. Thus every layerwise invariant subspace lies in the full atomic sum. If this sum is an invariant subspace, it contains every other one, proving both uniqueness claims. The dimension bounds follow from the layer inclusions and $M\subseteq\ker J_X$.
\end{proof}
When the atomic sum fails the test, Algorithm~\ref{alg:app-branch-build} gives a constructive lower bound on $k_{\max}^{\rm lw}$. The general ceiling does not by itself determine the maximum when atomic spaces are incompatible. For two matrices, linear activations give a closed-form maximum, while sample-dependent gates lead to an exact optimization over subspaces. We then give a subspace recursion for linear networks of arbitrary depth.

\paragraph{Two matrices with linear activations.}
We first consider one hidden linear layer without biases. In this paragraph, identify the sample $X$ with the matrix $[x_1\ \cdots\ x_s]\in\R^{n_0\times s}$ and arrange the outputs as columns. The model is $F_X(W_1,W_2)=W_2W_1X$, where $W_1\in\R^{n_1\times n_0}$ and $W_2\in\R^{d\times n_1}$ are fixed base matrices; $W_2$ is the linear readout $V^0$. Their edits are $A\in\R^{n_1\times n_0}$ and $B\in\R^{d\times n_1}$. Put $H_0=\operatorname{im}X$, $H_1=W_1H_0$, $r_j=\dim H_j$ for $j=0,1$, $\mathcal W=\ker W_2$, $\kappa=\dim\mathcal W$, and $c=\dim(\mathcal W\cap H_1)$.
\begin{theorem}[Depth-two maximum]\label{thm:depth-two}
Writing $k_{\max}^{\rm lw}$ for the maximum dimension over layerwise invariant subspaces, its value is
\[
 k_{\max}^{\rm lw}=n_1(n_0-r_0)+d(n_1-r_1)+r_0c
                  +\max\{0,(r_0-d)(\kappa-c)\}.
\]
\end{theorem}
\begin{proof}
The response to $(A,B)$ is $W_2AX+BW_1X+BAX$. For a candidate $M_1\oplus M_2$, independent layer coordinates force all three terms to vanish separately: set either layer to zero for the linear terms, then subtract them to isolate the mixed term. Let $R=\spn\{Ah:A\in M_1,h\in H_0\}$. Then $R\subseteq\mathcal W$, and every invariant subspace is contained in
\[
 \{A:AH_0\subseteq R\}\oplus\{B:B(H_1+R)=0\}.
\]
Conversely, each such sum is an invariant subspace. With $q=\dim R$ and $j=\dim(R\cap H_1)$, its dimension is
\[
 n_1(n_0-r_0)+d(n_1-r_1)+r_0j+(r_0-d)(q-j).
\]
The intersection $R\cap H_1$ lies in $\mathcal W\cap H_1$, so $j\le c$. The image of $R$ in the quotient $\mathcal W/(\mathcal W\cap H_1)$ has dimension $q-j$, giving $q-j\le\kappa-c$. If $r_0>d$, both coefficients $r_0$ and $r_0-d$ are positive and the two upper bounds give the maximum. If $r_0\le d$, the last term is nonpositive and the expression is at most its constant part plus $r_0c$. It is attained by $R=\mathcal W$ if $r_0>d$, and by $R=\mathcal W\cap H_1$ otherwise.
\end{proof}

\paragraph{Two matrices with sample-dependent gates.}
Consider one hidden layer without biases, with fixed base matrices $W_1\in\R^{n_1\times n_0}$ and $W_2=V^0\in\R^{d\times n_1}$. Assume the frozen activations have zero affine offset, as for ReLU or leaky ReLU, and let $S_i\in\R^{n_1\times n_1}$ be the diagonal base slope matrix on $x_i$. The frozen model is then $f_i=W_2S_iW_1x_i$, for $i=1,\ldots,s$. Its two edited layers have matrix spaces $M_1\subseteq\R^{n_1\times n_0}$ and $M_2\subseteq\R^{d\times n_1}$. Set $H=\spn_i(S_iW_1x_i)$ and $\mathcal W=\ker W_2$. For a target $T\subseteq\mathcal W$, define
\[
 \mathcal A(T)=\{A\in\R^{n_1\times n_0}:S_iAx_i\in T\ \forall i\},\qquad
 \mathcal B(T)=\{B\in\R^{d\times n_1}:B(H+T)=0\}.
\]
Let $P_{T^\perp}$ be the orthogonal projector onto $T^\perp$ and stack the matrices $x_i^\top\otimes(P_{T^\perp}S_i)$ vertically into $\Gamma(T)\in\R^{sn_1\times n_1n_0}$. Thus $\mathcal A(T)=\ker\Gamma(T)$ under column-wise vectorization. 
\begin{proposition}[Depth-two maximum with sample-dependent gates]
The maximum layerwise invariant subspace dimension is
\[
 \max_{T\subseteq\mathcal W}
 \left[n_1n_0-\rank\Gamma(T)+d\{n_1-\dim(H+T)\}\right].
\]
\end{proposition}
\begin{proof}
Expanding the frozen output in the independent edits $A\in M_1$ and $B\in M_2$ gives three terms. Their vanishing requires $W_2S_iAx_i=0$, $BS_iW_1x_i=0$, and $BS_iAx_i=0$. For any invariant subspace $M_1\oplus M_2$, collect its gated images into $T=\spn\{S_iAx_i:A\in M_1,\ i=1,\ldots,s\}\subseteq\mathcal W$; the invariant subspace lies in $\mathcal A(T)\oplus\mathcal B(T)$. Conversely, every such sum satisfies the three equations. Vectorization and rank--nullity give $\dim\mathcal A(T)=n_1n_0-\rank\Gamma(T)$. Each of the $d$ rows of $B$ can range independently over $(H+T)^\perp$, giving $\dim\mathcal B(T)=d[n_1-\dim(H+T)]$. Maximizing their sum proves the formula.
\end{proof}
The samplewise gates remain in $\Gamma(T)$, so a formula using only marginal spaces generally loses information.

\paragraph{Arbitrary depth: exact dimension and dynamic programming.}
Write $X=[x_1\ \cdots\ x_s]$ for the data matrix. For a bias-free deep linear model with $L+1$ matrices $W_\ell^0\in\R^{n_\ell\times n_{\ell-1}}$, set $n_{L+1}=d$ and $W_{L+1}^0=V^0$, $H_0=\operatorname{im}X$, $H_\ell=W_\ell^0H_{\ell-1}$, and $\mathcal W_\ell=\ker(W_{L+1}^0\cdots W_{\ell+1}^0)$, with $\mathcal W_{L+1}=\{0\}$ because the downstream map at the readout is the identity. Here $\operatorname{im}X$ is the column span of the data matrix, $H_\ell$ is the base feature span, and $\mathcal W_\ell$ is the kernel of the downstream product after layer $\ell$. Choose targets $T_\ell\subseteq\mathcal W_\ell$, with $T_{L+1}=0$. The target $T_\ell$ is a subspace of $\R^{n_\ell}$. Starting at $R_0=\{0\}$, set
\[
 U_\ell=H_{\ell-1}+R_{\ell-1},\quad
 R_\ell=W_\ell^0R_{\ell-1}+T_\ell,\quad
 u_\ell=\dim U_\ell,\quad t_\ell=\dim T_\ell.
\]
$R_\ell$ records the nonconstant feature-coefficient span and $U_\ell$ the span received by layer $\ell$. Require $T_\ell=0$ whenever $U_\ell=0$. 
\begin{theorem}[Deep-linear maximum over target subspaces]
The maximum layerwise invariant subspace dimension is
\[
 k_{\max}^{\rm lw}=\max_{(T_\ell)}\sum_{\ell=1}^{L+1}
       [n_\ell(n_{\ell-1}-u_\ell)+t_\ell u_\ell].
\]
\end{theorem}
\begin{proof}
For any tuple, the spaces $\{A:AU_\ell\subseteq T_\ell\}$ form an invariant subspace and attain the displayed count. Conversely, for any invariant subspace $\bigoplus_\ell M_\ell$, extract $T_\ell=M_\ell U_\ell$, where the product denotes the span of all images as defined in Appendix~\ref{app:composition-proof}. The recursive test places it in $\mathcal W_\ell$, and filling out the allowed maps preserves the same coefficient spans while containing the original invariant subspace. This proves equality.
\end{proof}

For $\ell=1,\ldots,L+1$ and an incoming coefficient span $R\subseteq\R^{n_{\ell-1}}$, let $V_\ell(R)$ be the maximum sum of the remaining layer dimensions, optimizing over subsequent admissible targets. Here $V_\ell(R)$ is a scalar value function; the subscript and argument distinguish it from the readout matrix $V$. There are no remaining layers after the readout, so set $V_{L+2}(R)=0$ for $R\subseteq\R^d$. 
\begin{corollary}[Backward recursion for the dimension maximum]
The value functions satisfy
\[
 V_\ell(R)=\max_{T\subseteq\mathcal W_\ell}
 \{n_\ell(n_{\ell-1}-u)+u\dim T+V_{\ell+1}(W_\ell^0R+T)\},
 \quad u=\dim(H_{\ell-1}+R),
\]
with the same zero-input convention and maximum $V_1(0)$.
\end{corollary}
\begin{proof}
For each admissible current target $T$, the current layer contributes $n_\ell(n_{\ell-1}-u)+u\dim T$ dimensions. The next coefficient span is $W_\ell^0R+T$, so the largest remaining contribution is $V_{\ell+1}(W_\ell^0R+T)$. Maximizing their sum over $T$ gives the recursion. The terminal value is zero because no layers remain.
\end{proof}

This recursion characterizes the maximum for linear networks of any depth, but does not yet provide an efficient finite computation. Each maximization ranges over subspaces of $\mathcal W_\ell$, including both their dimensions and their continuously varying orientations. Equal-dimensional targets can have different intersections with the base features and different effects on later layers. A finite reduction of this search has not been established here. Determining the maximum layerwise invariant subspace dimension for general nonlinear networks of arbitrary depth remains open.

\subsection{Radius}
\label{app:radius-proof}
On a frozen invariant subspace the polynomial response vanishes, so the true output change is entirely due to activation residuals. We first derive their exact propagation identity. Uniform norm bounds then give sufficient conditions for a whole invariant subspace ball to remain approximately flat.

\subsubsection{Exact propagation and a uniform bound}
For an edit $w$, let $\widetilde z_{\ell,i}$ and $\widetilde h_{\ell,i}$ be the true states at $\theta_0+w$. Define the true preactivation displacement $\Delta z_{\ell,i}(w)=\widetilde z_{\ell,i}-z_{\ell,i}^0$ and the \textbf{activation residual}
\[
 \rho_{\ell,i}(w)=\sigma(z_{\ell,i}^0+\Delta z_{\ell,i}(w))
              -h_{\ell,i}^0-S_{\ell,i}\Delta z_{\ell,i}(w).
\]
Here $\Delta z_{\ell,i}$ is a true-network displacement; the frozen displacement $\delta z_{\ell,i}$ was defined in \eqref{eq:app-frozen}. The maps $T_{\ell,i}(w)$ from Appendix~\ref{app:setting} propagate a postactivation residual through later edited weights and base slopes.

\begin{proposition}[Exact activation-residual propagation]\label{prop:app-exact-transition}
For every edit $w$, the sample-$i$ component of the activation transition potential satisfies
\begin{equation}
 A_i^{\rm tr}(w)=\sum_{\ell=1}^L T_{\ell,i}(w)\rho_{\ell,i}(w).
 \label{eq:app-transition}
\end{equation}
If $w$ belongs to a frozen invariant subspace, this expression also equals the true output drift $F_i(\theta_0+w)-F_i(\theta_0)$.
\end{proposition}
\begin{proof}
Subtract the frozen hidden state $h_{\ell,i}^{\frz}=h_{\ell,i}^0+\delta h_{\ell,i}$ from the true state and write $e_{\ell,i}=\widetilde h_{\ell,i}-h_{\ell,i}^{\frz}$. The inputs agree, so $e_{0,i}=0$. The preactivations differ by $(W_{\ell,\mathrm{lin}}^0+A_{\ell,\mathrm{lin}})e_{\ell-1,i}$ because the biases cancel. Using the definition of $\rho_{\ell,i}$ gives
\[
 e_{\ell,i}=S_{\ell,i}(W_{\ell,\mathrm{lin}}^0+A_{\ell,\mathrm{lin}})e_{\ell-1,i}+\rho_{\ell,i}(w).
\]
Repeated substitution propagates each residual through the later edited weights and base slopes. Multiplying by the edited linear readout gives \eqref{eq:app-transition}. On an invariant subspace the frozen output equals its base value, proving the final assertion.
\end{proof}
This identity retains the direction of every propagated residual and all cancellation between layers. The following bound replaces those vectors by uniform scalar bounds.

\begin{corollary}[Uniform drift bound]\label{cor:app-uniform-drift}
Let $M$ be a frozen invariant subspace and $B_M(r)=\{w\in M:\|w\|_2\le r\}$ its ball of radius $r\ge0$. Suppose $\chi_\ell(r)$ and $\Omega_\ell(r)$ satisfy, for every $w\in B_M(r)$,
\[
 \max_i\|T_{\ell,i}(w)\|_{\rm op}\le \chi_\ell(r),\qquad
 \| (\rho_{\ell,i}(w))_{i=1}^s\|_X\le\Omega_\ell(r).
\]
Then
\begin{equation}
 \sup_{w\in B_M(r)}\|\Delta F_X(w)\|_X
       \le\sum_{\ell=1}^L \chi_\ell(r)\Omega_\ell(r).
 \label{eq:app-bound}
\end{equation}
Any radius for which the right-hand side is at most $\varepsilon$ certifies an $\varepsilon$-flat invariant subspace ball.
\end{corollary}
\begin{proof}
Proposition~\ref{prop:app-exact-transition}, the triangle inequality and the operator-norm bound give
\begin{align*}
 \|\Delta F_X(w)\|_X
 &\le\sum_{\ell=1}^L\left(s^{-1}\sum_{i=1}^s\|T_{\ell,i}(w)\rho_{\ell,i}(w)\|_2^2\right)^{1/2}\\
 &\le\sum_{\ell=1}^L \chi_\ell(r)\Omega_\ell(r).
\end{align*}
The bound is uniform over $B_M(r)$, proving \eqref{eq:app-bound}.
\end{proof}
Proposition~\ref{prop:app-exact-transition} proves the identity in Theorem~\ref{thm:transition}; Corollary~\ref{cor:app-uniform-drift} supplies the conservative bound. Only the propagation identity is an equality; the bound can be loose because it discards cancellations and residual alignment. For an approximately constant frozen map, add $\sup_{w\in B_M(r)}\|F_X^{\frz}(w)-F_X(\theta_0)\|_X$ to the bound.

\subsubsection{Smooth activations and explicit depth dependence}
For smooth activations, Taylor's remainder bounds each residual using activation curvature and preactivation displacement.
\begin{corollary}[Smooth activation bound]\label{cor:app-smooth-bound}
Let $M$ be a frozen invariant subspace and let $\chi_\ell(r)$ bound the downstream gains as in Corollary~\ref{cor:app-uniform-drift}. Suppose $\sigma$ is twice continuously differentiable on every scalar interval between $z_{\ell,i,j}^0$ and $z_{\ell,i,j}^0+\Delta z_{\ell,i,j}(w)$, for all samples, units and $w\in B_M(r)$. Choose $d_\ell(r)$ and $\beta_\ell(r)$ such that
\[
 \|\Delta z_{\ell,i}(w)\|_2\le d_\ell(r),\qquad
 |\sigma''(u)|\le\beta_\ell(r)
\]
for all these edits, samples and traversed scalar values $u$. Then
\begin{equation}
 \sup_{w\in B_M(r)}\|\Delta F_X(w)\|_X
 \le\tfrac12\sum_{\ell=1}^L \chi_\ell(r)\beta_\ell(r)d_\ell(r)^2.
 \label{eq:smooth}
\end{equation}
\end{corollary}
\begin{proof}
For a scalar base value $z$ and change $v$, Taylor's integral remainder is
\[
 \sigma(z+v)-\sigma(z)-\sigma'(z)v
       =v^2\int_0^1(1-t)\sigma''(z+tv)\,dt.
\]
Apply this identity to coordinates $j=1,\ldots,n_\ell$. Taking the Euclidean norm gives
\[
 \|\rho_{\ell,i}(w)\|_2
 \le\tfrac12\beta_\ell(r)\left(\sum_{j=1}^{n_\ell}|\Delta z_{\ell,i,j}(w)|^4\right)^{1/2}
 \le\tfrac12\beta_\ell(r)d_\ell(r)^2,
\]
since $(\sum_j v_j^4)^{1/2}\le\sum_jv_j^2$. Substitution into \eqref{eq:app-bound} proves \eqref{eq:smooth}.
\end{proof}
The quantities in this bound control the entire ball. Sampled RMS displacements or base-point gains give a proxy without a uniform guarantee.

\paragraph{What controls the bound.}
The following estimates express the gains and displacements in terms of the base weights, base features and the layers that the invariant subspace allows to move. For a nonzero invariant subspace define
\[
 \gamma_\ell=\sup_{\substack{w\in M\\\|w\|_2=1}}\|A_\ell(w)\|_{\rm op},\qquad
 \gamma_V=\sup_{\substack{w\in M\\\|w\|_2=1}}\|B_{\mathrm{lin}}(w)\|_{\rm op}.
\]
Here $A_\ell(w)$ is the bias-augmented layer component of edit $w$. These constants measure the largest layer operator norm in a unit edit in the invariant subspace; a fixed layer has zero exposure. Put
\[
 \omega_\ell=\|W_{\ell,\mathrm{lin}}^0\|_{\rm op},\qquad
 \omega_V=\|V_{\mathrm{lin}}^0\|_{\rm op},\qquad
 h_\ell^{\max}=\max_i\|\bar h_{\ell,i}^0\|_2\quad(0\le\ell\le L).
\]
For this calculation take $d_\ell(r)=\max_i\sup_{w\in B_M(r)}\|\Delta z_{\ell,i}(w)\|_2$, and let $e_\ell(r)$ be the corresponding largest true hidden-state displacement. Suppose $\lambda_\ell$ bounds $|\sigma'|$ on all traversed intervals. Starting from $e_0(r)=0$, the forward recursion gives
\begin{align}
 d_\ell(r)&\le(\omega_\ell+\gamma_\ell r)e_{\ell-1}(r)
                  +\gamma_\ell rh_{\ell-1}^{\max},\label{eq:app-radius-forward}\\
 e_\ell(r)&\le\lambda_\ell d_\ell(r).\nonumber
\end{align}
Indeed, the exact true preactivation change is
\[
 \Delta z_{\ell,i}(w)
 =(W_{\ell,\mathrm{lin}}^0+A_{\ell,\mathrm{lin}})\bigl(\widetilde h_{\ell-1,i}-h_{\ell-1,i}^0\bigr)
   +A_\ell\bar h_{\ell-1,i}^0.
\]
Taking norms proves the first inequality, and the scalar mean value theorem proves the second. Similarly, multiplying downstream operator norms gives the admissible gain bound
\begin{equation}
 \chi_\ell(r)=(\omega_V+\gamma_Vr)
          \prod_{j=\ell+1}^L\lambda_j(\omega_j+\gamma_jr).
 \label{eq:app-gain-product}
\end{equation}
The empty product at $\ell=L$ is one. These expressions exhibit the dependence on feature magnitudes, trained weight norms, activation slopes and layer exposures of the invariant subspace.

\begin{proposition}[Large-radius envelope]\label{prop:app-large-radius}
Let $M$ be a nonzero frozen invariant subspace. Suppose the slopes and curvatures admit finite bounds $\lambda_\ell$ and $\beta_\ell$ on every interval reached by $M$, independent of the radius. Define
\begin{align*}
 c_1&=\gamma_1h_0^{\max},\\
 c_\ell&=(\omega_\ell+\gamma_\ell)\lambda_{\ell-1}c_{\ell-1}
                    +\gamma_\ell h_{\ell-1}^{\max}\quad(2\le\ell\le L),\\
 \bar\chi_\ell&=(\omega_V+\gamma_V)\prod_{j=\ell+1}^L\lambda_j(\omega_j+\gamma_j),\qquad
 C_M=\tfrac12\sum_{\ell=1}^L\bar\chi_\ell\beta_\ell c_\ell^2.
\end{align*}
For $r\ge1$,
\[
 \sup_{w\in B_M(r)}\|\Delta F_X(w)\|_X\le C_Mr^{2L+1}.
\]
If $\gamma_V=0$ (the linear readout is fixed), the exponent can be replaced by $2L$.
\end{proposition}
\begin{proof}
Induction in \eqref{eq:app-radius-forward} gives $d_\ell(r)\le c_\ell r^\ell$ for $r\ge1$. Equation~\eqref{eq:app-gain-product} gives $\chi_\ell(r)\le\bar\chi_\ell r^{L-\ell+1}$. Substituting both into \eqref{eq:smooth} yields
\[
 \sup_{w\in B_M(r)}\|\Delta F_X(w)\|_X
 \le\tfrac12\sum_{\ell=1}^L\bar\chi_\ell\beta_\ell c_\ell^2r^{L+\ell+1}
 \le C_Mr^{2L+1}.
\]
When $\gamma_V=0$, one moving downstream factor is removed, giving $r^{L+\ell}$ in each summand and the bound $C_Mr^{2L}$.
\end{proof}
The bound is controlled by base feature magnitudes, trained weight norms, layer exposures of the invariant subspace, activation slopes and curvatures. A residual contains a squared preactivation displacement and is amplified by the later weight matrices. For fixed depth, the resulting envelope is polynomial in the radius $r$, with degree at most $2L+1$ (or $2L$ when $\gamma_V=0$). The degree grows linearly with depth, but this is not a polynomial bound in depth: $C_M$ itself contains products of layer norms. Any radius satisfying this upper bound at the desired drift tolerance is a lower bound on the true flat radius; cancellations and smaller realized gains may allow a larger radius.

\paragraph{Concrete smooth-tail examples.}
Sigmoid, softplus and tanh illustrate how decaying curvature can produce a small activation transition potential in smooth networks. They are examples of the mechanism, rather than an exhaustive list.
Let $q(u)=(1+e^{-u})^{-1}$ denote the sigmoid. The second derivatives of sigmoid, softplus and tanh are
\[
 q''(u)=q(u)(1-q(u))(1-2q(u)),\qquad
 \frac{d^2}{du^2}\log(1+e^u)=q(u)(1-q(u)),\qquad
 \frac{d^2}{du^2}\tanh u=-2\tanh u\,\operatorname{sech}^2u.
\]
Since $q(u)(1-q(u))\le e^{-|u|}$ and $\operatorname{sech}^2u\le4e^{-2|u|}$, their curvatures decay exponentially in the tails. If the entire interval between $z$ and $z+v$ lies outside $[-a,a]$, with $a\ge0$, Taylor's formula gives
\[
 |\sigma(z+v)-\sigma(z)-\sigma'(z)v|
 \le
 \begin{cases}
  \tfrac12e^{-a}v^2,&\sigma=\text{sigmoid or softplus},\\
  4e^{-2a}v^2,&\sigma=\tanh.
 \end{cases}
\]
Thus one may use $\beta_\ell(r)\le e^{-a}$ or $8e^{-2a}$, respectively, when every scalar interval reached by the invariant subspace ball remains in those tails. 

These estimates require the entire traversed interval to stay in a tail. A large base preactivation alone is insufficient if the edit returns it to the transition region. The exponential factor multiplies the squared displacement; the residual also depends on how far the preactivation moves.

\subsubsection{Piecewise-linear activations and gate distance}
For a piecewise-linear activation the transition residual stays zero while every preactivation remains in its base affine piece. The distance to the first activation boundary therefore gives a conservative exact radius. For ReLU and leaky ReLU, the boundary is at zero. The following formula treats the simplified case in which every frozen preactivation is affine in the invariant subspace coordinates, so each boundary is a hyperplane.
\begin{proposition}[Gate distance for affine preactivations]
Let $M$ be a ReLU or leaky-ReLU frozen invariant subspace and let the columns of $Q$ form an orthonormal basis of $M$, so $w=Qy$ with $y\in\R^{\dim M}$. Assume that every frozen preactivation is affine in $y$. Index all sample--layer--unit triples by $j$ and write
\[
 z_j^{\frz}(y)=z_j^0+a_j^\top y,\qquad a_j\in\R^{\dim M},\quad z_j^0\ne0.
\]
The distance to the first gate boundary is
\[
 r_{\rm gate}=\min_{j:a_j\ne0}\frac{|z_j^0|}{\|a_j\|_2},
\]
with empty minimum $+\infty$. For every $\|y\|_2<r_{\rm gate}$, $F_X(\theta_0+Qy)=F_X(\theta_0)$.
\end{proposition}
\begin{proof}
By Cauchy--Schwarz, $|a_j^\top y|\le\|a_j\|_2\|y\|_2$ preserves every sign for $\|y\|_2<r_{\rm gate}$. When the minimum is finite, a minimizing index reaches zero at $y=-z_j^0a_j/\|a_j\|_2^2$. Within the sign-preserving region, induction over layers makes the frozen and true states identical, so the invariant subspace is exact there.
\end{proof}
The affine assumption holds, for example, when only the parameters of one layer are edited: later frozen layers apply fixed affine maps to its response. With several edited layers, products of edits generally make the preactivations polynomial, and their zero sets can be curved. The hyperplane formula then no longer applies. Measuring the first crossing along one sampled direction also gives no guarantee for a whole ball, since another direction may reach a boundary sooner.

\subsubsection{Probabilistic radius guarantees}
\label{app:radius-confidence}
Fix the data, reference parameters and a nonzero subspace $M$. Write
$D_v(t)=\|F_X(\theta_0+tv)-F_X(\theta_0)\|_X$ for the output change along a unit direction $v$. For an invariant subspace this equals $\|A_X^{\rm tr}(tv)\|_X$. The directional failure probability is
\[
 p(r,\varepsilon)=\Pr_{v\sim\operatorname{Unif}(S(M))}
 \left[\sup_{0\le t\le r}D_v(t)>\varepsilon\right].
\]
Thus $p(r,\varepsilon)\le\delta$ means that at least a fraction $1-\delta$ of directions preserve the outputs within tolerance along their entire segment of length $r$.

For fixed radii $0<r_1<\cdots<r_m$, let
\[
 Y_j(v)=\max_{h\le j}D_v(r_h),\qquad
 p_j(\varepsilon)=\Pr_v[Y_j(v)>\varepsilon].
\]
The grid probability satisfies $p_j(\varepsilon)\le p(r_j,\varepsilon)$. A statistical upper bound on $p_j$ therefore certifies the tested grid; extending it to the continuous ray requires a bound between grid points. For example, if $|D_v(t)-D_v(t')|\le L|t-t'|$ uniformly in $v$ and $\Delta$ is the largest grid spacing (including $r_0=0$), then
\[
 p(r_j,\varepsilon)\le p_j(\varepsilon-L\Delta/2)
 \qquad (\varepsilon\ge L\Delta/2).
\]
Indeed, every point lies within $\Delta/2$ of a tested point or the origin, where $D_v(0)=0$.

\begin{proposition}[Simultaneous directional certificates]
\label{prop:probabilistic-radius}
Consider $K$ fixed checkpoint--space--radius comparisons at a fixed tolerance. In each comparison, sample $n$ independent uniform directions and observe $k$ failures. For total confidence error $\alpha\in(0,1)$, put $\eta=\alpha/K$ and
\[
 \overline p(k,n)=
 \begin{cases}
 \operatorname{Beta}^{-1}(1-\eta;k+1,n-k),&k<n,\\
 1,&k=n.
 \end{cases}
\]
Here $\operatorname{Beta}^{-1}(q;a,b)$ is the $q$-quantile of the beta distribution with shape parameters $a,b$. With probability at least $1-\alpha$ over the sampled directions, every reported grid failure probability is at most its bound $\overline p$.
\end{proposition}
\begin{proof}
For each comparison the indicators are independent Bernoulli variables with failure probability $p_j$, so their sum is binomial. For $k<n$, the stated bound solves
\[
 \Pr_{Z\sim\operatorname{Binomial}(n,\overline p)}[Z\le k]=\eta.
\]
Inverting this binomial test gives the one-sided Clopper--Pearson bound with coverage at least $1-\eta$ \citep{clopper1934confidence}. A union bound over the $K$ comparisons gives simultaneous coverage at least $1-K\eta=1-\alpha$. Independence between comparisons is unnecessary.
\end{proof}

For an allowed failure fraction $\delta$, define
\[
 R^{\rm grid}_{M,\delta}(\varepsilon)
 =\max\bigl(\{0\}\cup\{r_j:p_j(\varepsilon)\le\delta\}\bigr),\qquad
 \widehat R_{M,\delta}(\varepsilon)
 =\max\bigl(\{0\}\cup\{r_j:\overline p_j\le\delta\}\bigr).
\]
On the simultaneous event, $\widehat R_{M,\delta}\le R^{\rm grid}_{M,\delta}$, including when the radius is selected after viewing the scan. The guarantee concerns random directions in the fixed space; it does not assert preservation along every direction or on unseen data.

\subsection{Extension to other architectures}\label{app:architectures}
The preceding theory was developed for MLPs. Its polynomial construction also applies to CNNs because convolution is linear in its trainable kernel. LayerNorm has exact affine invariances of its own, while restricting edits to selected parameters gives a way to study one part of a larger fixed network.

\subsubsection{Convolutional networks}\label{app:cnn-theory}
A convolution is a linear operator whose matrix reuses the same kernel entries at several spatial positions; biases can be included through an augmented input coordinate. Because one kernel edit acts at every position, its output response sums the downstream contributions from all those positions.

Fix one convolutional layer, with $n$ output channels and augmented patch dimension $m$ including its bias coordinate. Let $A\in\R^{n\times m}$ be its kernel edit, let $q_{i,\xi}\in\R^m$ be the base input patch at position $\xi$ on sample $i$, and let $D_{i,\xi}\in\R^{d\times n}$ map a preactivation change at that position through the frozen downstream network. Here $\Lambda$ is the set of spatial output positions of this layer, and $\xi\in\Lambda$. All other layers are fixed. The single-layer frozen response and its sample Jacobian are
\[
 \mathcal E(A)_i=\sum_{\xi\in\Lambda} D_{i,\xi}Aq_{i,\xi},\qquad
 J_i=\sum_{\xi\in\Lambda} q_{i,\xi}^\top\otimes D_{i,\xi},
 \qquad \mathcal E(A)_i=J_i\operatorname{vec}(A).
\]
Each sum includes every spatial output position, since the same kernel edit acts at all of them. Let $J$ be the vertical stack of $J_1,\ldots,J_s$, so the atomic kernel is $\{A:J\operatorname{vec}(A)=0\}$. Define
\[
 H=\spn_{i,\xi}\{q_{i,\xi}\},\quad
 C=\sum_{i,\xi}\operatorname{row}(D_{i,\xi}),\quad
 r_h=\dim H,\quad r_d=\dim C.
\]
These are the convolutional counterparts of $\mathcal H_\ell,\mathcal C_\ell,r_\ell,r_{d,\ell}$, with $m$ replacing $n_{\ell-1}+1$ and $n$ replacing $n_\ell$. 
\begin{proposition}[Atomic dimension with shared convolutional weights]
The single-layer frozen kernel has dimension
\[
 \dim\ker J=n(m-r_h)+r_h(n-r_d)+\delta,
 \qquad \delta=r_hr_d-\rank J\ge0.
\]
\end{proposition}
\begin{proof}
Each row of $J$ is a sum of patch--downstream tensor products and therefore lies in $H\otimes C$. Its dimension $r_hr_d$ bounds $\rank J$ and proves $\delta\ge0$. Rank--nullity gives $\dim\ker J=nm-\rank J$; adding and subtracting $r_hr_d$ yields the three terms.
\end{proof}
The first two terms count feature-null and downstream-null directions as for a dense layer; the last counts additional directions left unconstrained within the marginal tensor support. Contributions from different spatial positions can cancel even when none is individually zero, so the kernel constraint must be imposed after summing them. The deficit $\delta$ can therefore include cancellations induced by parameter sharing.

\subsubsection{LayerNorm and exact affine parameter invariance}
\label{app:layernorm}
Structural symmetries can have affine parameter orbits. Layer normalization provides examples, complementing the curved rescaling orbits discussed above. For $x\in\R^q$, let $\mathbf1_q$ denote the all-ones vector, $\bar x=q^{-1}\mathbf1_q^\top x$, and fix scale and bias vectors $\gamma,\beta\in\R^q$. With numerical stabilization $\epsilon_{\rm LN}>0$, \textbf{LayerNorm}~\citep{ba2016layernorm} is
\[
 \operatorname{LN}(x)=\gamma\odot\frac{x-\bar x\mathbf1_q}{s_{\rm LN}(x)}+\beta,
 \qquad s_{\rm LN}(x)=\sqrt{q^{-1}\|x-\bar x\mathbf1_q\|^2+\epsilon_{\rm LN}},
\]
where $\odot$ denotes coordinatewise multiplication.

\paragraph{Affine invariances around normalization.}
\begin{proposition}[LayerNorm shift and feature constraints]
For an affine map $Wh+b_{\rm in}$, with $h\in\R^m$, $W\in\R^{q\times m}$ and $b_{\rm in}\in\R^q$, the edits
\[
 \Delta W=\mathbf1_qv^\top,\qquad \Delta b_{\rm in}=c\mathbf1_q,
 \qquad (v,c)\in\R^m\times\R
\]
leave $\operatorname{LN}(Wh+b_{\rm in})$ unchanged for every input, giving $m+1$ independent directions.

If all $\gamma_j\ne0$, define $\eta\in\R^q$ by $\eta_j=1/\gamma_j$. For any normalized feature $u=\operatorname{LN}(x)$ and following affine map $Au+b_{\rm out}\in\R^n$, with $A\in\R^{n\times q}$ and $b_{\rm out}\in\R^n$, the edits
\[
 \Delta A=t\eta^\top,\qquad \Delta b_{\rm out}=-t(\eta^\top\beta),\qquad t\in\R^n
\]
preserve its output and give $n$ independent directions. Both invariances hold at arbitrary amplitudes with the normalization parameters fixed.
\end{proposition}
\begin{proof}
Adding $c\mathbf1_q$ to $x$ changes neither its centered vector nor its variance, so $\operatorname{LN}(x+c\mathbf1_q)=\operatorname{LN}(x)$. The first edits add $(v^\top h+c)\mathbf1_q$ before normalization, and the parameterization by $(v,c)$ is injective. This proves the first identity and dimension.

For the second, the definition of normalization gives
\[
 \eta^\top(u-\beta)=\frac{\mathbf1_q^\top(x-\bar x\mathbf1_q)}{s_{\rm LN}(x)}=0.
\]
The output change is consequently $t[\eta^\top u-\eta^\top\beta]=0$. Since $\eta\ne0$, the map from $t$ to its parameter edit is injective, proving the dimension.
\end{proof}
\subsubsection{Restricting edits to selected parameters}\label{app:restricted-parameters}
An invariant subspace can be studied in any chosen linear space of parameter edits while the remaining parameters stay fixed. Let $E\in\R^{p\times q}$ have orthonormal columns spanning the allowed edit space, so an allowed edit is $w=E\eta$ with $\eta\in\R^q$. Define the restricted true and frozen maps by
\[
 F_{X,E}(\eta)=F_X(\theta_0+E\eta),\qquad
 F_{X,E}^{\frz}(\eta)=F_X^{\frz}(E\eta).
\]
Their Jacobian at zero is $J_XE$, and the same decomposition holds after substituting $w=E\eta$. If $N\subseteq\R^q$ is an invariant subspace of the restricted frozen map, then $EN$ is an invariant subspace of the full frozen map. Conversely, every invariant subspace contained in $\im E$ is obtained this way. The general coefficient test therefore applies without change.

For a selection of whole layers, set $M_\ell=\{0\}$ on fixed layers and apply Algorithm~\ref{alg:app-branch-test}. It still propagates through fixed layers, whose weights and activation slopes affect the constraints. If only some entries within a layer may move, intersect the admissible matrix space with that coordinate restriction when forming the constraints. Relations tying edits in different layers generally produce a coupled space and require the general coefficient test.

\providecommand{\g}[1]{\textcolor{purple}{[#1]}}
\providecommand{\toedit}[1]{\textcolor{teal}{\textbf{[to edit:} #1\textbf{]}}}
\newcommand{\exptable}[1]{{\vspace*{0.5em}\footnotesize\setlength{\tabcolsep}{4pt}\resizebox{\ifdim\width>\linewidth\linewidth\else\width\fi}{!}{\input{experiments/tables/#1}}}}

\FloatBarrier
\section{Experimental methods}
\label{app:experiments}
The experiments start from the MNIST baseline in Appendix~\ref{app:protocol}. Each comparison below states what changes, which parameters are measured, and how the resulting spaces are checked. Results follow in Appendices~\ref{app:theoretical-support}--\ref{app:additional-material}.

\paragraph{Computing infrastructure.}\label{app:compute}
Experiments used a shared Linux server with two AMD EPYC 9554 processors (128 physical cores, 256 hardware threads), approximately 377\,GiB of host memory and four NVIDIA RTX A6000 GPUs with 48\,GB each. Training and network evaluation used GPUs; dense Jacobian factorizations also used CPU linear algebra. Geometry calculations used float64, with thread limits and memory checks for large factorizations.

\paragraph{Software.}\label{app:software}
The implementation uses PyTorch \citep{paszke2019pytorch} and TorchVision \citep{torchvision}, with NumPy \citep{harris2020numpy} and SciPy \citep{virtanen2020scipy} for numerical calculations, scikit-learn \citep{pedregosa2011sklearn} for clustering comparison scores, and Matplotlib \citep{hunter2007matplotlib} for figures.

\subsection{Jacobians, SVDs and numerical validation}
\label{app:sanity}
\paragraph{Jacobians and ranks.}
We differentiate the outputs on the fixed analysis set with respect to one affine layer or the whole parameter vector. For an MLP layer, the explicit Jacobian combines each augmented input feature with its downstream derivative. For a convolutional layer, contributions from all spatial positions are summed in the shared kernel coordinates. Explicit rows are checked against automatic differentiation.

Dense Jacobians are reduced by QR, followed by an SVD of the triangular factor. This avoids squaring the condition number by forming a Gram matrix. The same rank rule is applied to feature and downstream matrices to obtain the atomic decomposition.

For the largest layers, storing and factorizing the full Jacobian is too expensive. We instead estimate its range by randomized subspace iteration, using products with the Jacobian. Write $J\in\R^{sd\times p_\ell}$, where $s$ is the number of analysis samples, $d$ the output dimension and $p_\ell$ the number of parameters in the edited layer. We draw $k$ Gaussian parameter directions as columns of $\Omega\in\R^{p_\ell\times k}$ and compute
\[
 Y=J\Omega,\qquad Q=\operatorname{orth}(Y),\qquad B=Q^\top J.
\]
Here $\operatorname{orth}$ returns an orthonormal basis, and $k$ is the number of sampled directions. Products with $J$ and $J^\top$ are evaluated in chunks, without storing the full Jacobian. Repeated products refine $Q$; the singular values of the smaller matrix $B$ estimate the resolved spectrum. The residual $\|(I-QQ^\top)J\|_{\rm op}$ measures what lies outside the recovered image. Its numerical upper estimate is checked against the rank threshold before reporting a resolved rank. If the allowed value of $k$ is exhausted first, only a rank lower bound is reported. We use two subspace iterations with QR after each multiplication and double the sketch size from $k=32$ up to the memory cap (8,192 for the width-512 hidden layers); a full-dimension sketch uses no power iteration. Six fresh Gaussian probes $g_a$ give the omitted-range bound $\eta=10\sqrt{2/\pi}\max_{a\le6}\|(I-QQ^\top)Jg_a\|_2$ with failure probability at most $10^{-6}$ per check \citep[Lemma 4.1]{halko2011randomized}; rank $r$ is resolved only if $\hat\sigma_r>10^{-9}(\hat\sigma_1+\eta)$ and $\hat\sigma_{r+1}+\eta\le10^{-9}\hat\sigma_1$, including a $10^{-14}\hat\sigma_1$ rounding allowance in $\eta$ and a union bound over checks.

\begin{table}[!htbp]\centering
\caption{Numerical conventions. Additional construction cutoffs and unresolved ranks are stated with the affected results.}\label{tab:early-method-numerical}
\small
\begin{tabular}{p{.28\linewidth}p{.65\linewidth}}\toprule
Quantity & Convention\\\midrule
Analysis precision & Float64; stored bases checked for orthonormality and Jacobian residuals.\\
Rank & Singular values above $10^{-9}\sigma_1$, with checks at $10^{-6}\sigma_1$ and $10^{-12}\sigma_1$; $\sigma_1$ is the largest singular value.\\
Output norm & $\|V\|_X^2=s^{-1}\sum_i\|v_i\|_2^2$ for stacked outputs $V=(v_i)_{i=1}^s$; Jacobians normalized by $1/\sqrt{s}$.\\
Invariant subspace validation & Received-space constraints and frozen-polynomial residual, tolerance $10^{-9}$.\\
Output visibility & First sampled radius with $\|\Delta F_X\|_X>10^{-3}$ in activation-family scans; relative metrics for reduced-data and ViT studies are specified below.\\
Data consistency & Checkpoint, preprocessing, ordered samples and parameter coordinates matched before reusing a factor.\\\bottomrule
\end{tabular}
\end{table}

\paragraph{Gauss--Newton spectra and leading directions.}
The full-network spectrum in the motivation experiment uses the width-64 MNIST model. Its Gauss--Newton eigenvalues are proportional to the squared Jacobian singular values; a relative singular-value threshold must therefore be squared when expressed as an eigenvalue threshold. The leading three directions are recovered with a matrix-free eigensolver applying $v\mapsto J_X^\top(J_Xv)$ through automatic differentiation. We check their orthonormality, eigenpair residuals and agreement with the stored spectrum, then probe both signs of each direction. These are checkpoint-specific directions.

The reported spectral measurements concern the Jacobian and Gauss--Newton operator. For MSE, the full loss Hessian additionally contains a residual-weighted second-derivative term. That term vanishes at exact interpolation, where the Gauss--Newton operator equals the loss Hessian.

\paragraph{Validation and numerical limits.}
We check equality of the true and frozen networks at the base point, the response decomposition, and independent propagation of activation residuals. Float64 recomputation of affected stored bases removes the float32 ray floor without changing their ranks. Most ReLU and leaky-ReLU ranks are stable across the two tighter thresholds. The ReLU L3 and CNN conv5 exceptions are retained in the results, and late GELU deficiencies without a spectral gap are described as numerical kernels.

\subsection{Baseline model and training}
\label{app:protocol}
To keep the parameter dimension tractable for Jacobian factorization, the reference MNIST experiment \citep{lecunmnist} trains a biased ReLU MLP on PCA inputs. We center images using the training set, retain 256 principal components and divide by one global scale, without whitening. The same fitted transform is used for evaluation. The raw-pixel ablation checks the effect of this dimensionality reduction. Table~\ref{tab:early-method-baseline} collects the baseline settings. MSE averages over samples and output coordinates: $\mathcal L=(sd)^{-1}\sum_i\|F_i-y_i\|_2^2$.

Training and test accuracies at the analysis stages are reported for every retained ablation in Appendix~\ref{app:performance}. Each table identifies the checkpoint epochs and available seeds; an unreached milestone is left empty.

\begin{table}[!htbp]\centering
\caption{MNIST baseline. The width-64 companion uses the same training recipe and supports the full-network, invariant subspace and application analyses.}\label{tab:early-method-baseline}
\small\begin{tabular}{p{.28\linewidth}p{.65\linewidth}}\toprule
Setting & Value\\\midrule
Reference architecture & $256\to128\to128\to128\to10$, biases in every affine map.\\
Companion architecture & $256\to64\to64\to64\to10$.\\
Training data & All 60,000 training images; one-hot targets.\\
Optimization & MSE, Adam, learning rate $10^{-3}$, batch size 256, 500 epochs, no schedule.\\
Replication & Three seeds; separate initialization and minibatch-order generators.\\
Geometry data & Fixed stratified training subset $X$ of 10,000 images; disjoint comparison set $X'$ of the same size.\\\bottomrule
\end{tabular}
\end{table}

Checkpoints are selected by training quantities: initialization, early fitting, high accuracy, MSE interpolation and the late endpoint. Their thresholds and attained epochs are given in Table~\ref{tab:stages}. Test performance is reported after selection. Unreached stages remain missing rather than being replaced by the endpoint. In the reference model, the analysis set supplies more output constraints than there are parameters, so counting alone forces no kernel.

\begin{table}[!htbp]
\caption{Training-defined stages of the three reference MNIST runs. Early fitting uses training error at most $10\%$, high accuracy uses the first saved checkpoint after error reaches $0.1\%$, and interpolation uses the first training MSE below $10^{-4}$. A unit is dead when inactive on all of $X$.}
\label{tab:stages}
\centering\begin{tabular}{llrrrrrr}
\toprule
stage & seed & epoch & train errors & train MSE & test acc. & $\|\theta\|$ & dead units (L1, L2, L3) \\
\midrule
initialization & 0 & 0 & 55{,}344 & 0.8 & 0.076 & 28.0 & 0, 0, 0 \\
 & 1 & 0 & 54{,}117 & 1.3 & 0.099 & 27.8 & 0, 0, 0 \\
 & 2 & 0 & 56{,}351 & 0.92 & 0.061 & 27.8 & 0, 0, 0 \\
\addlinespace
early fitting & 0 & 1 & 5{,}411 & 0.027 & 0.908 & 27.1 & 0, 0, 1 \\
 & 1 & 2 & 3{,}830 & 0.019 & 0.930 & 26.5 & 0, 0, 1 \\
 & 2 & 1 & 5{,}740 & 0.029 & 0.902 & 27.0 & 0, 0, 0 \\
\addlinespace
high accuracy & 0 & 40 & 60 & $2.4\!\times\!10^{-4}$ & 0.977 & 32.9 & 0, 2, 21 \\
 & 1 & 50 & 44 & $1.4\!\times\!10^{-4}$ & 0.973 & 33.6 & 0, 6, 39 \\
 & 2 & 40 & 50 & $2.0\!\times\!10^{-4}$ & 0.975 & 32.2 & 0, 3, 20 \\
\addlinespace
interpolation & 0 & 61 & 30 & $9.8\!\times\!10^{-5}$ & 0.976 & 37.1 & 0, 3, 39 \\
 & 1 & 59 & 28 & $9.2\!\times\!10^{-5}$ & 0.974 & 35.4 & 0, 6, 43 \\
 & 2 & 50 & 33 & $9.7\!\times\!10^{-5}$ & 0.975 & 33.8 & 0, 9, 23 \\
\addlinespace
epoch 500 & 0 & 500 & 5 & $5.2\!\times\!10^{-5}$ & 0.976 & 76.9 & 0, 42, 78 \\
 & 1 & 500 & 8 & $3.2\!\times\!10^{-5}$ & 0.975 & 76.8 & 0, 48, 86 \\
 & 2 & 500 & 1 & $2.2\!\times\!10^{-5}$ & 0.973 & 76.0 & 0, 52, 75 \\
\bottomrule
\end{tabular}

\end{table}

\subsection{Activation and architecture variants}
\label{app:methods-activations}

\paragraph{Activation functions.}
The matched activation comparison changes the hidden activation while keeping the reference architecture and training recipe. The frozen model uses the activation's value and derivative at each base preactivation. This retains the affine offset required for smooth activations. Table~\ref{tab:early-method-activations} lists the variants; the additional leaky-ReLU invariant subspace comparison uses the width-64 companion.

\begin{table}[!htbp]\centering
\caption{Activation variants on the three-hidden-layer MNIST MLP.}\label{tab:early-method-activations}
\small\begin{tabular}{lll}\toprule
Activation & Definition & Hidden width\\\midrule
ReLU & $\max(0,z)$ & 128 (reference), 64 (companion)\\
Leaky ReLU & $\max(z,0.1z)$ & 128 and 64\\
GELU & $z\Phi(z)$, $\Phi$ the standard normal CDF & 128\\
Identity & $z$ & 128\\\bottomrule
\end{tabular}
\end{table}

\paragraph{Convolutions.}
The MNIST CNN uses five convolutional layers, three average-pooling stages and a linear readout. Images are zero-padded and standardized per channel. The base channel widths are $(16,16,32,32,64)$: all convolutions have $3\times3$ kernels, stride one and padding one, with $2\times2$ stride-two average pooling after L2, L4 and L5, followed by flattening to $64\cdot4\cdot4$ features. Training uses the baseline MSE/Adam settings (learning rate $10^{-3}$, batch 256, 500 epochs); width multipliers appear in Table~\ref{tab:early-method-scaling}. Tied-weight Jacobians sum spatial contributions before factorization. Selective-patch spaces are computed from patches at active positions of each channel; the full atomic calculation also permits cancellations across positions.

\paragraph{BatchNorm.}
The width-64 normalization ablation inserts BatchNorm between each hidden affine map and ReLU, with the plain width-64 model as its paired control. Normalization parameters are trained, then held fixed with the running statistics during geometry analysis. The evaluation-mode scale is carried through every derivative and frozen forward map. Perturbations and radii stay in the original weight-and-bias coordinates.

\paragraph{Fine-tuned ViT.}\label{app:methods-vit}
We fine-tune the full ViT-B/16 \citep{dosovitskiy2021vit} on NWPU-RESISC45 \citep{cheng2017resisc} with a 45-class readout. Initialization uses TorchVision's \texttt{IMAGENET1K\_V1} weights \citep{torchvision}, pretrained on ImageNet-1K \citep{deng2009imagenet}. The subsequent geometry analysis fixes the preceding transformer and normalization parameters, restricting edits to the final encoder MLP and readout. More specifically, with the preceding transformer fixed, let $h\in\R^{768}$ be the class-token state entering the final encoder MLP. The analyzed map is
\[
 u=L_2(h),\quad a=\operatorname{GELU}(W_0u+b_0),\quad
 z=h+W_3a+b_3,\quad F=V L_f(z)+b_V,
\]
where $W_0\in\R^{3072\times768}$, $W_3\in\R^{768\times3072}$ and $V\in\R^{45\times768}$, with matching biases. The fixed LayerNorm maps $L_2,L_f$ have nonzero scales. The setup therefore tests whether ocean calculations can be applied to selected layers of a trained transformer. The fine-tuning settings and predictive performance appear in Table~\ref{tab:native-vit-training}. 

\begin{table}[!htbp]\centering
\caption{ViT-B/16 fine-tuning on RESISC45. Training accuracy is the recorded online accuracy over augmented training minibatches during the selected epoch; validation and test accuracy are evaluated after that epoch.}\label{tab:native-vit-training}
\small\begin{tabular}{p{.28\linewidth}p{.64\linewidth}}\toprule
Setting & Value\\\midrule
Initialization & ImageNet-1K ViT-B/16; new 45-class linear readout.\\
Training & All parameters; cross-entropy; AdamW, learning rate $3\times10^{-5}$, weight decay $0.05$; 12 epochs, batch 64.\\
Schedule and images & One warmup epoch, cosine decay, gradient clipping at 1; ImageNet resize/crop/normalization and horizontal/vertical flips.\\
Selection & Best accuracy on a held-out validation split of 50 images per class; three seeds.\\
Geometry & 10,000 cached class-token states; float64; final encoder MLP and readout; normalization fixed, dropout disabled.\\\bottomrule
\end{tabular}
\par\smallskip
\begin{tabular}{rrrrr}\toprule
Seed & Selected epoch & Train (\%) & Validation (\%) & Test (\%)\\\midrule
0 & 10 & 99.982 & 95.644 & 95.206\\
1 & 8 & 99.850 & 95.333 & 95.397\\
2 & 11 & 99.994 & 95.333 & 95.048\\
\bottomrule\end{tabular}
\end{table}
\paragraph{Fine-tuned ResNet.}\label{app:methods-resnet}
We fine-tune all parameters of an ImageNet-pretrained ResNet-50 on RESISC45, replacing the readout with 45 outputs. The seed-0 run uses cross-entropy and AdamW for 15 epochs, with learning rate $10^{-4}$, weight decay $0.05$ and batch size 64. A warmup epoch is followed by cosine decay; images use the same resize, crop, normalization and flips as the ViT study. Validation selects epoch 12, with test accuracy $95.56\%$. Geometry analysis uses 10,000 images and restricts edits to the nine convolutions in the last three bottlenecks and the readout. BatchNorm parameters and statistics are fixed in evaluation mode, and downstream maps include the residual shortcuts.

\paragraph{Common layerwise protocol for ViT and ResNet.}\label{app:methods-common-families}
We apply the feature-null, downstream-null and selective constructions of Section~\ref{sec:atomic} separately to each selected layer. With $H$ containing observed input features (convolutional patches for ResNet), let $U_H$ span the feature space and let $U_{H^\perp}$ span its complement. Let $U_{C^\perp}$ span the kernel of the stacked downstream maps. The two marginal families are
\[
 A=Z U_{H^\perp}^{\top},\qquad A=U_{C^\perp}Z U_H^{\top},
\]
where $Z$ is arbitrary of the appropriate size; the second family excludes feature-null overlap. Selective edits use the active-feature or active-patch constraints. All operators include the intervening activations, normalization and residual paths. We use the numerical rank rule of Appendix~\ref{app:sanity}; the LayerNorm shift is also verified algebraically. For approximate comparisons, we retain factor rates $\sigma/\sqrt{s}\le10^{-3}$, with $s=10,000$, and remove exact components. This operator cutoff is distinct from the finite-output tolerance. For ResNet, spatial positions enter the operators while the normalization remains by the number of images. Near-zero Gram eigenvalues are resolved by direct row-matrix SVD. The large full atomic Jacobians are not factorized.

\paragraph{ViT joint-edit comparisons.}
The supplementary joint study combines first-layer and readout feature-null edits, with or without the second-layer LayerNorm shifts. For approximate comparisons, feature tails use relative cutoffs $10^{-2}$ and $10^{-3}$ in the first and second MLP maps; downstream tails use $10^{-2}$, or retain 722 or 567 second-map directions. Joint approximate edits combine the two downstream tails at $10^{-2}$ after removing exact components. Radius metrics and normalization are specified in Appendix~\ref{app:methods-radius}.

\subsection{Width and depth scaling}
\label{app:methods-scaling}
The width and depth sweeps ask how the number of invariant parameter directions changes as a network becomes larger. Each sweep varies one architectural size while retaining the training protocol and pairing initialization and minibatch-order seeds across models. Table~\ref{tab:early-method-scaling} lists the architectures.

We apply the same definitions of atomic kernels, selective subspaces and multilayer compatibility throughout the size sweeps. Their measurements have different computational coverage. On the fixed 10,000-image analysis set, dense Jacobian factorizations measure full atomic kernels for widths 32, 64, 128 and 256 and hidden depths 2, 4 and 5; Table~\ref{tab:scaling-coverage} lists the stages. We also compute selective dimensions for every MLP size on both these 10,000 images and all 60,000 training images. The latter tests whether the explicit selective directions survive the additional data constraints.

Selective dimensions count a specific part of an atomic kernel: each unit's incoming edit annihilates its active input features. They give lower bounds on full atomic dimensions and can omit directions that require cancellation across units. Likewise, single-layer dimensions do not establish a compatible multilayer invariant subspace. Full multilayer constructions in the width/depth comparison have been validated for width 64; extending them to the other MLP sizes remains unfinished.

At width 512, the hidden-layer Jacobians are large enough that we use the randomized procedure in Appendix~\ref{app:sanity} to estimate their ranks without storing the full matrices. At epoch 500, the third hidden layer has a resolved numerical rank in seeds 0 and 1. For the second hidden layer, all 8,192 recovered singular directions exceed the rank threshold, and the computation reaches its sampling limit before resolving the rest of the spectrum. This establishes $\operatorname{rank}J_2\ge8{,}192$; the exact numerical rank remains open.

For CNNs, dense atomic measurements cover the base architecture. The width sweep uses the analogous selective-patch construction, in which a channel's kernel edit annihilates every input patch at its active spatial positions. The double-width conv5 layer has selective-dimension measurements, with full atomic factorization still pending. Counting these directions and assembling a basis that passes the multilayer invariant subspace checks are separate computational steps; Table~\ref{tab:cnn-width} identifies which results include both.

\begin{table}[!htbp]\centering
\caption{Scaling grids. Depth counts hidden layers; CNN scaling changes all channel widths together and keeps the pooling pattern.}\label{tab:early-method-scaling}
\small\begin{tabular}{lll}\toprule
Sweep & Varied values & Fixed setting\\\midrule
MLP width & $32,64,128,256,512$ & Three hidden layers\\
MLP depth & $2,3,4,5$ & Width 128\\
CNN width & $0.5,1,2$ times base channels & Base: $(16,16,32,32,64)$\\\bottomrule
\end{tabular}
\end{table}

\begin{table}[!htbp]\centering
\caption{Dense atomic measurements in the MLP size sweeps. Each listed checkpoint covers every affine layer, including the readout, on 10,000 analysis images. All rows use three seeds, subject to the stated milestone availability.}\label{tab:scaling-coverage}
\small\begin{tabular}{p{.23\linewidth}p{.68\linewidth}}\toprule
Architecture & Measured stages\\\midrule
Width 32 & Initialization, early fitting, epochs 100, 250 and 500.\\
Width 64 & Initialization, high accuracy and epoch 500; interpolation in seed 1.\\
Width 128, depth 3 & Initialization, early fitting, high accuracy, interpolation, epochs 100, 250 and 500.\\
Width 256 & High accuracy and epoch 500.\\
Depth 2, width 128 & Initialization, early fitting, high accuracy, epochs 100, 250 and 500. For seed 2, high accuracy is the epoch-250 checkpoint.\\
Depth 4, width 128 & Initialization, early fitting, high accuracy, interpolation, epochs 100, 250 and 500.\\
Depth 5, width 128 & Initialization, early fitting, high accuracy, interpolation and epoch 500.\\\bottomrule
\end{tabular}\end{table}

Comparisons at matched training stages ask whether networks with similar fitting accuracy have similar geometry. Comparisons at epoch 500 instead give every network the same training budget. Both are useful because narrow and shallow networks can take longer to reach the accuracy milestones. Appendix~\ref{app:performance} gives the attained epochs and train/test accuracies.

\subsection{Training data and representation ablations}
\label{app:methods-data}
\paragraph{Sample count.}
We compare retraining on a smaller sample with restricting the analysis set of a model trained on all MNIST images. Both conditions use the same evaluation subsets, but only the first changes the learned network. Hidden-layer invariant subspaces fix the input layer. Adding the disjoint set $X'$ to a fixed checkpoint separately tests how many directions survive additional constraints.

\paragraph{Input representation and dataset.}
The raw-pixel ablation tests whether the PCA reduction used for the baseline affects the measured oceans. We train paired MLPs with three width-64 hidden layers, using either all 784 centered pixels or 256 principal components. Each representation uses one global scale. During geometry analysis, we hold the input layer fixed and edit L2, L3 and the readout, so the compared parameter spaces have the same dimensions.

SVHN \citep{netzer2011svhn} provides a second digit-classification dataset that can be studied with these small networks. The baseline comparison retains the MNIST hidden architecture, optimizer and 500-epoch budget. A second SVHN recipe tests a model selected for validation performance: we reserve 500 training images per class, fit PCA on the remaining training images, add Adam's coupled L2 penalty of $10^{-4}$ to all weights and biases, and select the checkpoint with the best validation accuracy. The baseline uses PCA fitted on the full training set, zero weight decay and the epoch-500 checkpoint. The revised recipe changes these choices together, so its comparison measures their combined effect.

For RESISC45, we fine-tune ViT-B/16 and ResNet-50 and study their final layers with the preceding network fixed (Appendix~\ref{app:methods-activations}). Table~\ref{tab:early-method-data} summarizes the comparisons.

\begin{table}[!htbp]\centering
\caption{Data comparisons and their scope. All recipe changes belong to the indicated comparison rather than to the MNIST baseline.}\label{tab:early-method-data}
\small\begin{tabular}{p{.24\linewidth}p{.32\linewidth}p{.34\linewidth}}\toprule
Comparison & Settings & Analysis or selection\\\midrule
Reduced MNIST & $N=1{,}000,2{,}000,5{,}000$; three seeds & Retraining and fixed-model restriction; input layer fixed.\\
Raw versus PCA & 784 pixels versus 256 components & Paired width-64 models; L2, L3 and readout.\\
SVHN baseline & PCA-256 MLP; matched-width CNN & Equal-budget endpoints; fitting quality reported.\\
SVHN revised recipe & Fit-set PCA; 500 validation images per class; Adam L2 penalty $10^{-4}$ & Validation-selected checkpoint; comparison changes several factors.\\
ViT / ResNet & RESISC45; full-network fine-tuning & Final MLP / final three bottlenecks and readout; preceding layers fixed.\\\bottomrule
\end{tabular}
\end{table}

\paragraph{Constraints added by new inputs.}
At a fixed network, let $J_\ell(x)$ be the layer output Jacobian on input $x$. For a sample set $S$ and distribution $\mu$, define
\begin{align*}
 K_\ell(S)&=\{v:J_\ell(x)v=0\text{ for all }x\in S\},\\
 K_\ell(\mu)&=\{v:J_\ell(x)v=0\text{ for }\mu\text{-almost every }x\}.
\end{align*}
Adding inputs imposes the old and new constraints together:
\[
 K_\ell(X\cup X')=K_\ell(X)\cap K_\ell(X').
\]
For ReLU selective spaces, each incoming row must annihilate its unit's active-feature span. Hence the number of dimensions removed is
\[
 \sum_j[\dim\cH_{\ell j}^{\rm act}(X\cup X')-\dim\cH_{\ell j}^{\rm act}(X)].
\]
These identities follow from intersecting kernels and from rank--nullity on each row, respectively. They hold with the network fixed; retraining changes the Jacobians and need not preserve either inclusion.

\paragraph{Additional inputs at a fixed network.}\label{app:methods-newinput}
For three width-128 ReLU MNIST networks at epoch 500, let $X$ be the original 10,000-image analysis set. Each condition adds a separate batch $Y$ of 10,000 images and measures $\dim K_\ell(X\cup Y)/\dim K_\ell(X)$, the fraction of the original atomic space that survives. For clean-to-noise interpolation, $y_i(\alpha)=(1-\alpha)y_i+\alpha z_i$, with clean MNIST test images $y_i$ and $z_i\sim\operatorname{Unif}([0,1]^{784})$, at $\alpha=0,0.1,\ldots,1$. A single noise draw is reused across levels and seeds. SVHN test images are converted to grayscale with BT.601 weights and resized by adaptive average pooling to $28\times28$. All conditions use the saved MNIST centering, PCA projection and scaling, without refitting. Conditions add inputs independently rather than cumulatively.

Numerically unresolved ranks retain the sensitivity checks of Appendix~\ref{app:sanity}. For interpolation, 13 of 99 cells, all in L3, are cutoff-sensitive. Recomputing original and union ranks at the same alternative cutoff preserves the L3 minimum at $\alpha=0.3$, while shifting its mean survival from $95.5\%$ to $94.9\%$. This cutoff variation is separate from variation across seeds.

\subsection{Persistence tests}
\label{app:methods-persistence}
Persistence asks whether a parameter direction that leaves outputs unchanged at one checkpoint retains that property later in training. We compare the same parameter layer at checkpoints $t$ and $t'$ on the same ordered analysis samples. Let $B_t\in\R^{p_\ell\times k_t}$ and $B_{t'}\in\R^{p_\ell\times k_{t'}}$ be orthonormal bases of their kernels, and let $P_{t'}=B_{t'}B_{t'}^\top$. For $k_t>0$, retained mass and effective lost dimension are
\[
 Q(K_t,K_{t'})=\frac{\|B_{t'}^\top B_t\|_F^2}{k_t},\qquad
 d_{\rm lost}=\|(I-P_{t'})B_t\|_F^2.
\]
Here $Q=1$ means that every direction in the earlier kernel lies in the later kernel, and $Q=0$ means that the two spaces are orthogonal. The effective lost dimension weights each earlier direction by how far it moves out of the later kernel. The singular values of $B_{t'}^\top B_t$ are principal-angle cosines. Missing dimensions, if $k_t>k_{t'}$, count as right angles. Thus $d_{\rm lost}$ is the sum of squared sines, including those missing dimensions.

Let $V_{t'}$ contain the right singular vectors of the later Jacobian above the rank threshold. Since $I-P_{t'}=V_{t'}V_{t'}^\top$, we compute
\[
 d_{\rm lost}=\|V_{t'}^\top B_t\|_F^2
\]
directly. Computing $k_t-\|B_{t'}^\top B_t\|_F^2$ instead would subtract two nearly equal values when almost everything persists, amplifying rounding error in the small difference.

The calculation is repeated after removing dead-unit coordinates. Applying the later Jacobian to $(I-P_{t'})B_t$ measures the sensitivity of the lost component, with generic directions as a reference. Parent--sibling comparisons use the same geometry at independently continued checkpoints. New-batch comparisons instead keep the checkpoint fixed and change its sample constraints.

To determine whether directions lost from the kernel become strongly visible, let $U$ be an earlier kernel or one of its component subspaces, with orthonormal basis $B_U$. We measure the output sensitivity per unit norm of its lost component and compare it with a generic direction:
\[
 \frac{\|\widehat J_{t'}(I-P_{t'})B_U\|_F}
      {\|(I-P_{t'})B_U\|_F},\qquad
 g_{\rm generic}=\frac{\|\widehat J_{t'}\|_F}{\sqrt{p_\ell}},
\]
when the first denominator is nonzero. Here $\widehat J_{t'}=J_{t'}/\sqrt{s}$ and $P_{t'}$ projects onto its kernel.

For the component comparison, $C_t$ is the coordinate space supported on weights entering or leaving units inactive on every analysis image. We compare the full kernel with $S_t=K_t\cap C_t^\perp$ and with the canonical co-occurrence component of the atomic decomposition. For a projector $P$ and kernel basis $Q$, the split is accepted only when every eigenvalue of $Q^\top P Q$ is within $10^{-6}$ of zero or one and the expected dimension identities hold. Quotients by a contained subspace $N$ are represented by the orthogonal complement $K\cap N^\perp$; the co-occurrence quotient is obtained through $\Pi_\ell$ below. For GELU, the numerical dead-coordinate comparison additionally varies the maximum derivative threshold; unresolved splits are omitted.

The positive spectral tail $E_t^+(\varepsilon)$ contains right singular vectors with singular values of $J_t/\sqrt{s}$ between the numerical rank threshold and $\varepsilon$. Its overlap with a later kernel uses the same projected-mass statistic. The backward mass $Q(K_{t'},K_t)$ measures how much of the later kernel was already present earlier; the projector distance $\|P_t-P_{t'}\|_F/\sqrt{2}$ also reflects dimension growth. These measurements compare checkpoints and do not determine the kernel's continuous path between them.

\subsection{Invariant subspace construction and validation}
\label{app:methods-branches}
An atomic candidate is a parameter subspace that leaves the frozen outputs unchanged when one layer is edited on its own. We obtain the full atomic space from that layer's Jacobian nullspace. Feature-null and selective constructions give explicit subspaces by requiring edits to annihilate, respectively, all input features or the features seen by each active unit. To assemble a multilayer invariant subspace, we propagate received spaces and solve the compatible linear constraints in the chosen layer order, as in Appendix~\ref{app:composition-proof}. When the full atomic sum passes, it is used directly; otherwise bottom-up and top-down constructions can give different spaces.

Validation checks orthonormality, the received-space constraints and the frozen output on mixed layer edits. If a received-space singular value is close to the construction cutoff, changing the sampled basis can change whether that direction is retained. We rebuild such spans with fresh random samples and validate against those independent reconstructions. 

\paragraph{Marginal overlap and unit-level measurements.}
Let $\Pi_\ell(A)=P_{\mathcal C_\ell}AP_{\mathcal H_\ell}$, whose kernel is the marginal-null space $\mathcal N_\ell$. The selective percentage in Section~\ref{sec:cooccurrence} is $100\,\dim\Pi_\ell(G_\ell^{\rm sel})/\delta_{\rm joint,\ell}$: we project an orthonormal selective basis and count singular values above $10^{-9}$, removing overlap with both marginal components.

For any activation, the within-unit operator is $J_{\ell j}=\operatorname{stack}_i[D_{\ell,i}e_j(\bar h_{\ell-1,i}^0)^\top]$, where $D_{\ell,i}$ already includes that unit's derivative. Soft selectivity records $k_j(\varepsilon)=\#\{\sigma_a(J_{\ell j})/\sqrt{s}\le\varepsilon\}$ and $n_j(\tau)=\#\{i:|\sigma'(z_{\ell,i,j}^0)|>\tau\}$, with counting proxy $\sum_j(n_{\ell-1}+1-n_j(\tau))_+$ and unit-level share $\sum_j k_j(\varepsilon)/\#\{\sigma_a(J_\ell)/\sqrt{s}\le\varepsilon\}$; normalized margins are $m_{\ell,i,j}=z_{\ell,i,j}^0/(\|W_\ell^0[j,:]\|_2\|\bar h_{\ell-1,i}^0\|_2)$.

\paragraph{Mechanism comparisons.}
The co-occurrence deficit counts atomic directions that remain after accounting for the feature-null and downstream-null components. To understand where it comes from, we group samples by their labels, input features, downstream maps, joint feature--downstream embeddings or gate patterns, and compare against random groups of the same sizes. These groupings test whether the missing constraints are associated with class structure or with finer similarities between samples. A pairing shuffle replaces the samplewise pairs $(\bar h_{\ell-1,i}^0,D_{\ell,i})$ by $(\bar h_{\ell-1,i}^0,D_{\ell,\pi(i)})$, where $\pi$ is a permutation of all samples or a permutation within each class. It preserves the marginal spans and sample count. A change in deficit therefore measures sensitivity to the specific pairings. The counting floor caused by too few output constraints remains unchanged by the shuffle, although surviving deficit can also have other causes.

A downstream map $D_{\ell,i}$ sends a preactivation change at layer $\ell$ on sample $i$ to an output change in the frozen network. Clustering these maps asks which samples transmit changes through the rest of the network in similar ways. At a fixed checkpoint and layer, define the common scale and pairwise distance
\[
 c=\frac1s\sum_i\|D_{\ell,i}\|_F,\qquad
 d(i,j)=\frac{\|D_{\ell,i}-D_{\ell,j}\|_F}{c},\quad c>0.
\]
Complete linkage merges groups only when the largest distance between their members is at most $\varepsilon$. Every resulting group $G$ therefore satisfies $\max_{i,j\in G}d(i,j)\le\varepsilon$. The scalar $c$ is shared by all downstream maps $D_{\ell,1},\ldots,D_{\ell,s}$ at this checkpoint and layer. Dividing by it expresses distances relative to a typical downstream-map norm while preserving differences in magnitude between samples.

We measure how closely a partition follows the class labels by its purity. Let $\mathcal G$ be the partition of the $s$ samples and let $c_i\in\{1,\ldots,d\}$ be the class index of sample $i$. Define
\[
 \operatorname{purity}(\mathcal G)
 =\frac1s\sum_{G\in\mathcal G}\max_{a\in\{1,\ldots,d\}}
       \bigl|\{i\in G:c_i=a\}\bigr|.
\]
Each group is assigned its majority label, and purity is the fraction of samples agreeing with those assignments. A value of one means that every group contains only one class. Small groups can have high purity by chance, so we compare with randomly permuted labels while retaining the same groups and sizes. Purity measures label agreement; the rank-based explained deficit measures how many constraints are missing across groups. The latter is defined in Appendix~\ref{app:mechanism}. For $c$ classes, centered class means $M$, within-/between-class covariances $\Sigma_W,\Sigma_B$, and row-normalized means $\bar M$, we measure $\mathrm{NC1}=\operatorname{tr}(\Sigma_W\Sigma_B^+)/c$, $\mathrm{NC2}=\|\bar M\bar M^\top/\|\bar M\bar M^\top\|_F-(I-\mathbf1\mathbf1^\top/c)/\sqrt{c-1}\|_F$ and the coefficient of variation of class-mean norms. NC3 is $\|V/\|V\|_F-M/\|M\|_F\|_F$ at the penultimate layer; NC4 is nearest-class-center disagreement with labels (or predictions where specified), with results in Appendix~\ref{app:nc}.

\paragraph{Symmetry controls.}
For ReLU MLPs and CNNs, unit or channel rescaling multiplies incoming parameters and divides outgoing weights by the same factor. We check finite transformations on training and random inputs, then compare their tangent span with the measured kernel and invariant subspace. GELU supplies an activation control for this rescaling identity.

\subsection{Radius analyses}

\label{app:methods-radius}
\subsubsection{Ray tests}
Ray tests measure output changes along $\theta_0+rv$ for unit directions $v$ and a fixed grid of radii. We record actual and frozen responses using the same edits. Unless stated otherwise, radius is Euclidean parameter length and drift is $\|F_X(\theta_0+rv)-F_X(\theta_0)\|_X$. Generic controls edit the same layers. A layer-relative scan instead scales each layer by its base parameter norm; captions identify this convention.

Independent uniform directions are obtained as $v=Qg/\|g\|_2$, with $g\sim\mathcal N(0,I)$ and $Q$ an orthonormal basis of the space. Each direction is reused at every radius. The multilayer MLP/CNN batteries use 128 directions per checkpoint and space; atomic activation comparisons use 32 per seed, and transition/GELU-tail comparisons use 256. Fixed-angle scans in selected two-dimensional planes are descriptive slices and are excluded from the certificates below.

The first tested output change above $10^{-3}$ defines visibility; a scan with no crossing is censored at its last radius. ReLU and leaky-ReLU scans also record the first gate crossing. Reduced-data scans use relative dataset-RMS drift. These choices are stated with the corresponding results.

\paragraph{Curvature-band rays.}\label{app:methods-curvature-rays}
For the three width-64 ReLU MNIST models at epoch 500, we apply the ray protocol to cumulative Hessian and GGN eigenspaces $|\lambda_T|\le\tau$, $T\in\{H,G\}$, at $\tau=10^{-14},10^{-12},10^{-10},10^{-8},10^{-6},10^{-4}$. We sample 256 normalized Gaussian mixtures of the retained eigenvectors per space and seed, and evaluate the original network in float64 on the same 10,000 analysis images. The radius grid is $0.001,0.003,0.01,0.03,0.1,0.2,0.3,0.5,0.7,1,1.5,2,3,5,7,10,15,20,30,50,100$. Both operators use identical constructed-ocean and generic controls. Curves report the mean and sample SD of the three within-seed medians; these summaries do not give uniform subspace guarantees.

\paragraph{GELU approximate-kernel rays.}
The extended width-128 GELU comparison edits L3 only, using the right singular subspace of $J_3/\sqrt{s}$ with singular values at most $10^{-3}$, including its numerical kernel. For each of three seeds and the ReLU-matched high-accuracy, interpolation and epoch-500 checkpoints, we reuse 256 directions on the grid $0.001,0.003,0.01,0.03,0.1,0.3,1,3,10,30,100,300,1000$, with matched generic L3 edits and float64 forwards.

\paragraph{ViT and ResNet family rays.}
The common family studies use 256 unit-Frobenius directions per nonempty family, dimension-matched random-subspace controls and generic edits. Approximate families and their controls undergo the same removal of exact components. We measure relative dataset-RMS logit drift with tolerance $10^{-3}$ at radii $0.001,0.003,0.01,0.03,0.1,0.3,1,3,10,30$. ViT extends the same directions to $100,300,1000$ and uses float64 throughout. ResNet uses float64 operators and float32 actual forwards, checked against float64; responses at the measured precision floor are marked accordingly. Its selective battery spans 10,170 of the family's 10,171 directions.

\paragraph{Residuals, bounds and slices.}
For an edit $w$, let $\rho_{\ell,i}(w)$ be the activation residual and $T_{\ell,i}(w)$ its edited frozen tail, as in Appendix~\ref{app:radius-proof}. Define $v_\ell(w)=(T_{\ell,i}(w)\rho_{\ell,i}(w))_{i=1}^s$ and $g_\ell(w)=\max_i\|T_{\ell,i}(w)\|_{\rm op}$. We evaluate the chain
\[
 \|A_X^{\rm tr}(w)\|_X
 =\Big\|\sum_\ell v_\ell(w)\Big\|_X
 \le\sum_\ell\|v_\ell(w)\|_X
 \le\sum_\ell g_\ell(w)\|\rho_\ell(w)\|_X.
\]
The first inequality measures cancellation across layers; the second replaces samplewise propagation by its largest operator norm. Replacing $g_\ell(w)$ by the product of edited downstream weight norms and base-slope norms gives the further weight-norm bound. These quantities are evaluated at the actual edit $w$.

The GELU plot additionally compares the transition with the diagnostic expression
\[
 \widehat B(w)=\frac12\sum_\ell \chi_\ell^0\beta_\ell
          \widehat d_\ell(w)^2,\qquad
 \widehat d_\ell(w)^2=\frac1s\sum_i\|\Delta z_{\ell,i}(w)\|_2^2,
 \quad \chi_\ell^0=\max_i\|T_{\ell,i}(0)\|_{\rm op}.
\]
Here $\beta_\ell$ bounds activation curvature; for GELU the calculation uses $\beta_\ell=\sqrt{2/\pi}$. This expression approximates the smooth-radius bound using measured RMS preactivation displacements and gains at the unperturbed network. Those substitutions remove the theorem's uniform guarantee: $\widehat B$ is a diagnostic to compare with the observed transition, rather than a certified upper bound on an invariant subspace ball.

Angle--radius scans use selected two-dimensional slices. Independent rays support the directional certificates in Section~\ref{sec:radius-measured}. The fixed-angle slices are descriptive; neither sampling scheme establishes the worst direction in a full subspace ball.

\paragraph{Margin-regularized training.}\label{app:methods-margin}
We test whether pushing ReLU preactivations away from zero delays gate crossings and extends the distance over which outputs remain nearly unchanged. These experiments use the width-64 MNIST model and the baseline training budget, with an additional margin penalty in the objective.

For a minibatch $\mathcal B$, let $y_i\in\R^d$ be the one-hot target and write the unregularized loss as $\mathcal L_{\rm MSE}(\theta;\mathcal B)=(|\mathcal B|d)^{-1}\sum_{i\in\mathcal B}\|F_i(\theta)-y_i\|_2^2$. Let $L$ be the number of hidden layers, $n_\ell$ the width of layer $\ell$, and $z_{\ell,i,j}(\theta)$ the preactivation of unit $j$ on sample $i$. The margin-regularized objective is

\[
 \mathcal L_{\rm MSE}(\theta;\mathcal B)
 +\frac{\lambda}{L}\sum_{\ell=1}^L\frac{1}{|\mathcal B|n_\ell}
    \sum_{i\in\mathcal B}\sum_{j=1}^{n_\ell}
       \log\!\left(1+e^{\tau-|z_{\ell,i,j}(\theta)|}\right).
\]
The coefficient $\lambda$ sets the penalty strength, and $\tau$ sets its preactivation scale. Each layer contributes its mean penalty over samples and units, and the layer means are averaged. We set $\tau=1$ in raw preactivation units. The penalty is largest near zero and encourages a wider gap from the activation boundary; it does not constrain the eventual output drift.

\begin{table}[!htbp]\centering
\caption{Margin-regularized training settings. Each arm shares initialization, minibatch-order seeds and baseline budget with its control.}\label{tab:early-method-interventions}
\small\begin{tabular}{lll}\toprule
Arm & Parameters & Replication\\\midrule
Margin & $\tau=1$, $\lambda\in\{0.01,0.1\}$ & Three seeds per strength\\
\bottomrule
\end{tabular}
\end{table}

For each trained network, we construct and validate its invariant subspace before interpreting radius probes as motion within that invariant subspace. Appendix~\ref{app:controlled} reports the margin-radius comparison. Train/test accuracy, unregularized MSE and parameter norms are given in Appendix~\ref{app:performance}, because the margin penalty can also change fitting quality and parameter scale.

\subsubsection{Radius certification}
\label{app:methods-certificates}
Write $D_{v_i}(r_h)=\|F_X(\theta_0+r_hv_i)-F_X(\theta_0)\|_X$ for the output drift on ray $v_i$ at radius $r_h$.
We apply Proposition~\ref{prop:probabilistic-radius} to the saved independent rays. For each checkpoint and space:
\begin{enumerate}
\item At each radius $r_j$, compute the running maximum $Y_{ij}=\max_{h\le j}D_{v_i}(r_h)$ along each ray.
\item Count $k_j=\sum_i\mathbf1\{Y_{ij}>\varepsilon\}$ at the fixed absolute dataset-RMS tolerance $\varepsilon=10^{-3}$.
\item For $K$ reported comparisons, evaluate the one-sided Clopper--Pearson upper bound with error probability $0.05/K$. The largest grid radius with upper bound at most $\delta=0.1$ certifies at least $90\%$ safe directions on that grid.
\end{enumerate}
The radius grids and comparisons are fixed before evaluating failures. Figure~\ref{fig:radius-grid} uses $K=328$ comparisons, and Figure~\ref{fig:radius-certificates-all} uses its own stated family. Reusing rays across radii is permitted by the simultaneous guarantee. Curves summarize per-checkpoint bounds by mean and sample SD across available seeds; rays are never pooled. The mean bounds the mean failure probability of those checkpoints, while SD describes seed variation. We make no continuity assumption between tested radii.

The ResNet panel uses the final convolutional layer \texttt{layer4.2.conv2}; its absolute drift is taken from the same battery whose relative drift is reported in Appendix~\ref{app:resnet-common}. The ViT panels likewise use absolute drift for certificates, keeping their relative-drift family comparisons separate. GELU uses 256 independent atomic L3 rays at epoch 500; its earlier L3 kernels are empty. The two SVHN recipes retain their respective endpoint and validation-selected checkpoints.

\subsection{Linear mode connectivity analyses}
\label{app:methods-lmc}
Linear mode connectivity asks whether the straight parameter segment between two trained networks maintains low loss. We obtain related endpoint networks by copying a trained MLP or CNN checkpoint, called the parent, and continuing two copies with different minibatch orders. The resulting networks are siblings. Both copies inherit the parent's Adam moment estimates as well as its parameters. We evaluate a fixed grid, including the midpoint, on the segment between endpoints, recording MSE, classification error, distance to the endpoint functions, and the frozen/transition decomposition. Table~\ref{tab:early-method-lmc} collects the comparison rules.

Low loss along a segment can coexist with changes in the output vectors. We therefore measure functional change separately. Let $\theta_A,\theta_B$ be the endpoint parameters, let $F_A=F_X(\theta_A)$ and $F_B=F_X(\theta_B)$, and write $\theta_\alpha=(1-\alpha)\theta_A+\alpha\theta_B$ for $\alpha\in[0,1]$. The mean distance to the endpoint functions is
\[
 E(\alpha)=\frac12\left(
 \|F_X(\theta_\alpha)-F_A\|_X+
 \|F_X(\theta_\alpha)-F_B\|_X\right).
\]
This measures how far the interpolated network's outputs lie from the two endpoint outputs. Even at an endpoint it equals half the distance between the endpoint functions, so its value includes their difference.

We also measure departure from a linear interpolation \emph{in output space}. The chord-deviation vector and its magnitude are
\[
 D_\alpha=F_X(\theta_\alpha)-\bigl((1-\alpha)F_A+\alpha F_B\bigr),
 \qquad \|D_\alpha\|_X.
\]
This quantity is zero at both endpoints and throughout the segment if the outputs interpolate linearly. A segment can consequently have zero chord deviation while its outputs vary substantially. The plots identify which of these two measurements they show.

To relate connectivity to ocean geometry, we compare the complete constructed endpoint spaces using their continuous sharing fraction and the four displacement projections defined in Appendix~\ref{app:lmc-shared}. Parent--child and sibling intersection dimensions are separate measurements, using the principal-angle tolerance and residual checks in Table~\ref{tab:early-method-lmc}. For orthonormal bases $Q_0,\ldots,Q_m$, we form $E=\operatorname{stack}_{j=1}^m[(I-Q_jQ_j^\top)Q_0]$ and retain $Q_0v$ for right singular vectors of $E$ below the declared residual cutoff. Triple intersections are also formed successively from the sibling intersection and the parent; whole-span membership residuals and cutoff gaps are checked, with unresolved dimensions reported as ranges.

\begin{table}[!htbp]\centering
\caption{Connectivity measurements. Displayed siblings inherit Adam moments; each cohort retains its recorded spawn epochs.}\label{tab:early-method-lmc}
\small\begin{tabular}{p{.30\linewidth}p{.62\linewidth}}\toprule
Measurement & Procedure\\\midrule
Segment & 30 equally spaced points plus the midpoint on 10,000 samples; children continued to epoch 500.\\
Loss connectivity & MSE barrier below ten times the mean endpoint MSE.\\
Intersection dimensions & Principal-angle tolerance $0.1^\circ$, followed by kernel and invariant subspace residual checks.\\
Displacement alignment & Full-space and sequential projections; 256 globally normalized Gaussian controls and a separate layer-energy-matched reference.\\
Finite-distance geometry & Endpoint rays and planes through the parent and both children.\\\bottomrule
\end{tabular}
\end{table}
\FloatBarrier

\section{Theoretical support, extended version}
\label{app:theoretical-support}\label{app:atomic-full}
We report empirical tests of the theoretical constructions across architectures, training stages and data conditions. After the symmetry comparison, we examine atomic dimensions and test the corresponding directions at finite distances. We then compare multilayer constructions and their radius, relate output drift to activation transitions, and study margin-regularized training. Connectivity applications are discussed in Appendix~\ref{app:applications}.

\subsection{Structural symmetries}
\label{app:symmetries}
A rescaling symmetry gives a familiar way to change network parameters while preserving its function. For a ReLU or leaky-ReLU hidden unit, choose any positive constant $c$. Multiplying the incoming weights and bias by $c$ multiplies the activation by $c$; dividing the outgoing weights by $c$ cancels this change. The same argument applies to a convolutional channel, because average pooling commutes with positive scaling.

To compare this curved family of equivalent parameters with a linear ocean, differentiate the rescaling at $c=1$. For hidden unit $j$ of layer $\ell$, the nonzero layer components of the resulting tangent direction are
\[
 \bigl(W_\ell[j,:],\ b_\ell[j],\ -W_{\ell+1}[:,j]\bigr).
\]
Here $j$ indexes the unit whose incoming and outgoing parameters are rescaled. The finite tests use positive factors $e$ and $e^{-1}$; the symmetry holds for every $c>0$.

We test the ReLU MLPs of widths 64 and 128, the width-128 leaky-ReLU and GELU models, and the base MNIST CNN, each with three seeds. The positively homogeneous models preserve outputs under finite rescaling, and their generator ranks equal the number of hidden units or channels. GELU supplies a control: the same parameter transformation changes its outputs. Table~\ref{tab:c-symmetry-summary} summarizes the numerical ranks, Jacobian responses and finite-rescaling errors across seeds and measured training stages.

\begin{table}[!htbp]\centering
\caption{Rescaling checks pooled over the measured seed--stage pairs for each architecture. Cells give the arithmetic mean and sample standard deviation; this describes variation across both training stages and seeds. The Jacobian response is divided by generator norm. Finite-change columns summarize each record\textquotesingle s maximum absolute output change under the tested positive rescalings. GELU is a nonhomogeneous control.}\label{tab:c-symmetry-summary}
\footnotesize\setlength{\tabcolsep}{4pt}\begin{tabular}{lrrrr}\toprule
Model & Generator rank & Jacobian response & Analysis-set change & Random-input change\\\midrule
ReLU, w128 & $384.0 \pm 0.0$ & $(1.26 \pm 0.39)\,10^{-16}$ & $(6.52 \pm 2.07)\,10^{-15}$ & $(6.02 \pm 3.30)\,10^{-15}$\\
ReLU, w64 & $192.0 \pm 0.0$ & $(1.91 \pm 0.91)\,10^{-16}$ & $(9.38 \pm 3.81)\,10^{-15}$ & $(9.90 \pm 5.34)\,10^{-15}$\\
leaky ReLU, w128 & $384.0 \pm 0.0$ & $(2.67 \pm 0.98)\,10^{-16}$ & $(1.05 \pm 0.48)\,10^{-14}$ & $(6.12 \pm 3.16)\,10^{-15}$\\
GELU, w128 & $384.0 \pm 0.0$ & $(6.36 \pm 3.09)\,10^{-2}$ & $(9.24 \pm 3.08)\,10^{-1}$ & $(9.20 \pm 2.41)\,10^{-1}$\\
CNN, base & $160.0 \pm 0.0$ & $(7.57 \pm 5.29)\,10^{-17}$ & $(3.45 \pm 1.65)\,10^{-15}$ & $(4.34 \pm 3.30)\,10^{-15}$\\
\bottomrule\end{tabular}\end{table}
\begin{table}[!htbp]\centering
\caption{Joint symmetry--invariant subspace comparisons for width-64 MLPs at high accuracy on 10,000 images, using the relative rank cutoff $10^{-9}$. $T$ is the rescaling-symmetry tangent span and $M$ the constructed layerwise invariant subspace. Remainder is $\dim\ker J_X-\dim(M+T)$.}\label{tab:c-symmetry-intersection}
\small\setlength{\tabcolsep}{4pt}\begin{tabular}{lrrrrrrr}\toprule
Model & Seed & $\dim\ker J_X$ & $\dim M$ & $\dim T$ & $\dim(M\cap T)$ & $\dim(M+T)$ & Remainder\\\midrule
Leaky ReLU & 0 & 1211 & 866 & 192 & 0 & 1058 & 153\\
ReLU & 0 & 3221 & 3036 & 192 & 25 & 3203 & 18\\
ReLU & 1 & 3283 & 3102 & 192 & 17 & 3277 & 6\\
ReLU & 2 & 3824 & 3621 & 192 & 36 & 3777 & 47\\
\bottomrule\end{tabular}\end{table}
\FloatBarrier

The layerwise invariant-subspace comparison with the full tangent kernel is available for all three width-64 ReLU seeds (Table~\ref{tab:tangent-ocean-seeds}). The saved constructions and kernel factorizations use matching checkpoints, preprocessing and analysis images. Each construction passes its recorded compatibility validation. The seed-0 kernel rank is cutoff-sensitive: its gap is 1.74, and tightening the relative cutoff from $10^{-9}$ to $10^{-12}$ reduces the kernel dimension from 3,221 to 3,220. The seed-1 and seed-2 gaps at the registered cutoff are 10.8 and $3.9\times10^6$, respectively.
\begin{table}[!htbp]\centering
\caption{Layerwise invariant-subspace and full-network tangent-kernel dimensions at high accuracy, width-64 ReLU MNIST MLPs on 10,000 images. Numerical ranks use the relative singular-value cutoff $10^{-9}$. The ratio compares dimensions.}
\label{tab:tangent-ocean-seeds}
\small\begin{tabular}{rrrrr}\toprule
Seed & Epoch & $\dim M$ & $\dim\ker J_X$ & Ratio\\\midrule
0 & 305 & 3,036 & 3,221 & $94.26\%$\\
1 & 188 & 3,102 & 3,283 & $94.49\%$\\
2 & 313 & 3,621 & 3,824 & $94.69\%$\\
\bottomrule\end{tabular}
\end{table}

Comparing the full tangent kernel, the certified invariant subspace and the symmetry tangent span, we find that the invariant subspace and symmetries together explain most of the tangent kernel in the tested width-64 ReLU and leaky-ReLU networks (Table~\ref{tab:c-symmetry-intersection}). At high accuracy, the combined span accounts for $98.8$--$99.8\%$ of the numerical tangent kernel across the three ReLU seeds, leaving 18, 6 and 47 dimensions, respectively. The leaky-ReLU comparison uses seed 0. Their overlap with the invariant subspace differs between the two activations, so the dimension of their combined span is computed using the measured intersection.

\subsection{Atomic decomposition}
\label{app:atomic}\label{app:mnist-cnn}
An atomic kernel consists of parameter edits to one affine layer that leave the frozen outputs unchanged. The three-component decomposition asks why these edits are invisible: some act on unused input-feature directions, some are annihilated by the downstream network, and the remainder comes from constraints missing in the observed feature--downstream pairings. All dimensions below use the numerical rank convention in Appendix~\ref{app:sanity}.

Table~\ref{tab:atomic} gives the reference width-128 MNIST ReLU MLP in absolute dimensions. The panels that follow extend the same accounting to every layer and recorded stage of the width and depth sweeps, activation variants, raw-pixel and BatchNorm models, and SVHN models. For CNNs, the plots distinguish total atomic dimensions from a full component decomposition. Reduced-data dimensions are compared in Appendix~\ref{app:training-data}. Margin-regularized models are discussed separately in Appendix~\ref{app:controlled}.
\begin{table}[!htbp]
\caption{Reference atomic decomposition: width-128, three-hidden-layer MNIST ReLU MLP on 10,000 analysis images, three seeds. At relative threshold $10^{-9}$, $\dim K_\ell$ is the sum of feature-null, downstream-null and co-occurrence dimensions. ``Dead'' counts units inactive on the analysis set; ``gap'' is the singular-value ratio at the reported rank.}\label{tab:atomic}
\centering\begin{tabular}{llrrrrrrrrrrrrr}
\toprule
 & & \multicolumn{6}{c}{L2 ($p_\ell=16{,}512$)} & \multicolumn{6}{c}{L3 ($p_\ell=16{,}512$)} & readout \\
\cmidrule(lr){3-8}\cmidrule(lr){9-14}
stage & seed & $\dim K$ & f.-null & d.-null & deficit & dead & gap & $\dim K$ & f.-null & d.-null & deficit & dead & gap & $\dim K$ \\
\midrule
high accuracy & 0 & 2{,}691 & 0 & 258 & 2{,}433 & 2 & $1.7\!\times\!10^{8}$ & 8{,}174 & 256 & 2{,}667 & 5{,}251 & 21 & $4.7\!\times\!10^{6}$ & 210 \\
 & 1 & 3{,}711 & 0 & 774 & 2{,}937 & 6 & $6.6\!\times\!10^{9}$ & 8{,}327 & 768 & 4{,}797 & 2{,}762 & 39 & $3.3\!\times\!10^{6}$ & 400 \\
 & 2 & 4{,}256 & 0 & 387 & 3{,}869 & 3 & $1.0\!\times\!10^{10}$ & 7{,}007 & 384 & 2{,}520 & 4{,}103 & 20 & $2.5\!\times\!10^{7}$ & 210 \\
\addlinespace
interpolation & 0 & 2{,}929 & 0 & 387 & 2{,}542 & 3 & $3.6\!\times\!10^{7}$ & 10{,}554 & 384 & 4{,}914 & 5{,}256 & 39 & $6.4\!\times\!10^{5}$ & 400 \\
 & 1 & 4{,}002 & 0 & 774 & 3{,}228 & 6 & $1.3\!\times\!10^{9}$ & 8{,}936 & 768 & 5{,}289 & 2{,}879 & 43 & $4.8\!\times\!10^{6}$ & 450 \\
 & 2 & 4{,}864 & 0 & 1{,}161 & 3{,}703 & 9 & $4.3\!\times\!10^{8}$ & 7{,}789 & 1{,}152 & 2{,}760 & 3{,}877 & 23 & $2.4\!\times\!10^{6}$ & 240 \\
\addlinespace
epoch 500 & 0 & 9{,}583 & 0 & 5{,}418 & 4{,}165 & 42 & $1.9\!\times\!10^{8}$ & 14{,}696 & 5{,}376 & 6{,}786 & 2{,}534 & 78 & 2.1 & 780 \\
 & 1 & 10{,}543 & 0 & 6{,}321 & 4{,}222 & 48 & $1.6\!\times\!10^{7}$ & 15{,}331 & 6{,}400 & 6{,}794 & 2{,}137 & 86 & $1.3\!\times\!10^{6}$ & 880 \\
 & 2 & 10{,}936 & 0 & 6{,}708 & 4{,}228 & 52 & $3.6\!\times\!10^{8}$ & 15{,}005 & 6{,}784 & 5{,}700 & 2{,}521 & 75 & $2.0\!\times\!10^{6}$ & 760 \\
\bottomrule
\end{tabular}

\end{table}
\FloatBarrier

\paragraph{What changes across networks.}
The late hidden layers and readouts often contain the largest invariant fractions, but the balance among components depends on the architecture and training condition. ReLU models can leave many feature and downstream directions unused. Leaky-ReLU models retain a co-occurrence kernel even with nonzero slopes on both sides of zero. The width and depth comparisons also reveal substantial input-layer kernels in some larger networks, where the input-layer selective construction remains empty. Thus the full atomic measurement is essential for identifying directions beyond selectivity.

Each panel below keeps its model, layer and stage explicit. Averaging is only across seeds of that setting. A model whose stage is reached in one seed contributes that one measurement, and the stage label reports the count. The data-ablation figures in Appendix~\ref{app:training-data} separately compare training on a subset with applying fewer constraints to a fixed network.

\begin{figure}[!htbp]\centering
\includegraphics[width=6.272in]{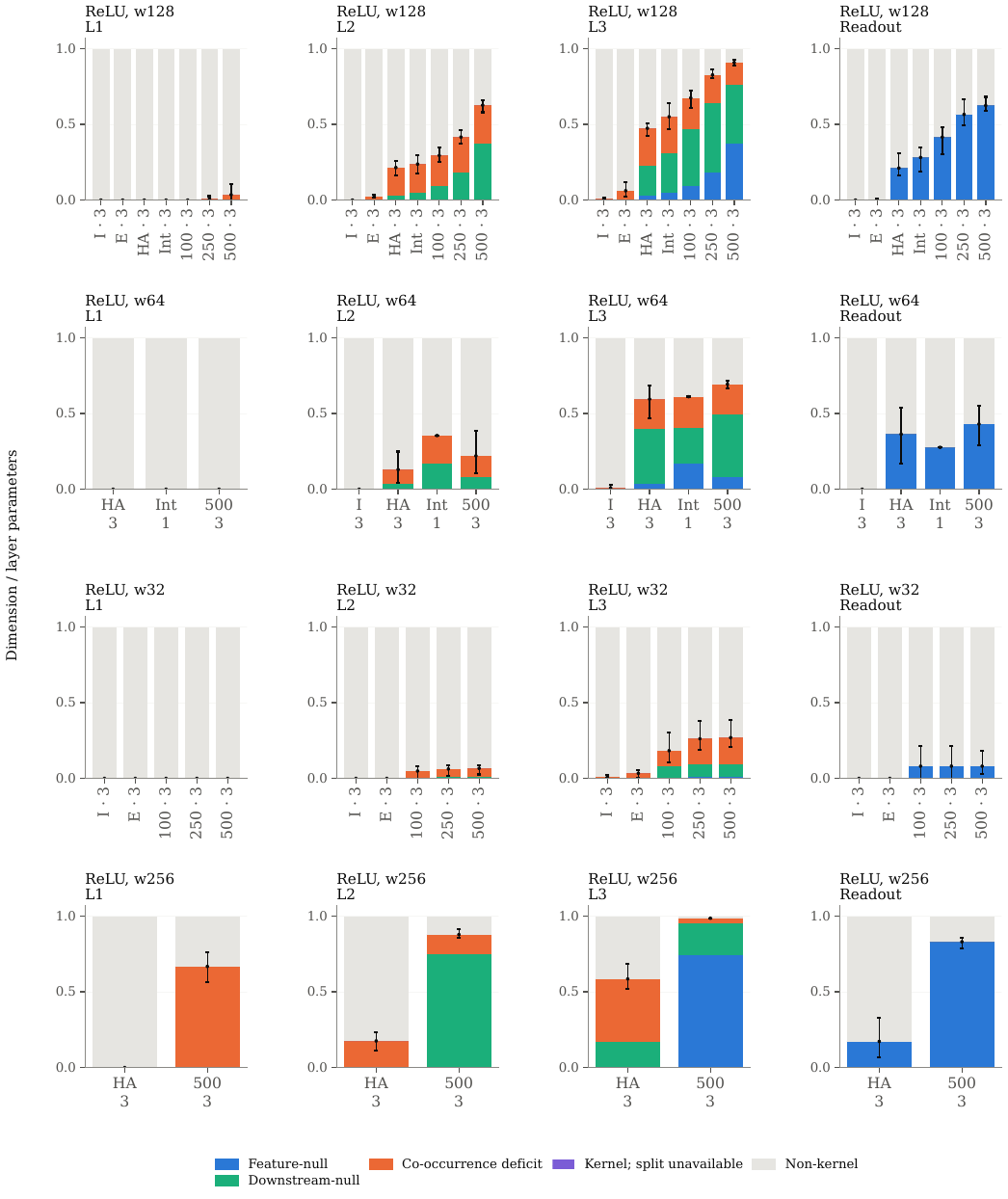}
\caption{Atomic decomposition across the indicated models and layers. Models use MNIST unless labeled SVHN; w and d denote width and hidden depth. Panels show a layer across stages, or all editable layers at the stated endpoint. Heights are fractions of layer parameters; bars average seeds and whiskers span their kernel fractions. Seed counts follow each stage label, on a second line or after a dot. I, E, HA, Int and Val denote initialization, early fitting, high accuracy, interpolation and validation selection; numbers denote fixed epochs. Stages are grouped by milestone and then fixed epoch. For reduced-data models, $n$ is the retraining or analysis-set size as titled. Hatched kernels have no measured component split. Counts use the $10^{-9}$ rank rule; GELU late-stage counts and layers without a spectral gap remain threshold-dependent.}\label{fig:c-atomic-0}
\end{figure}
\begin{figure}[!htbp]\centering
\includegraphics[width=6.272in]{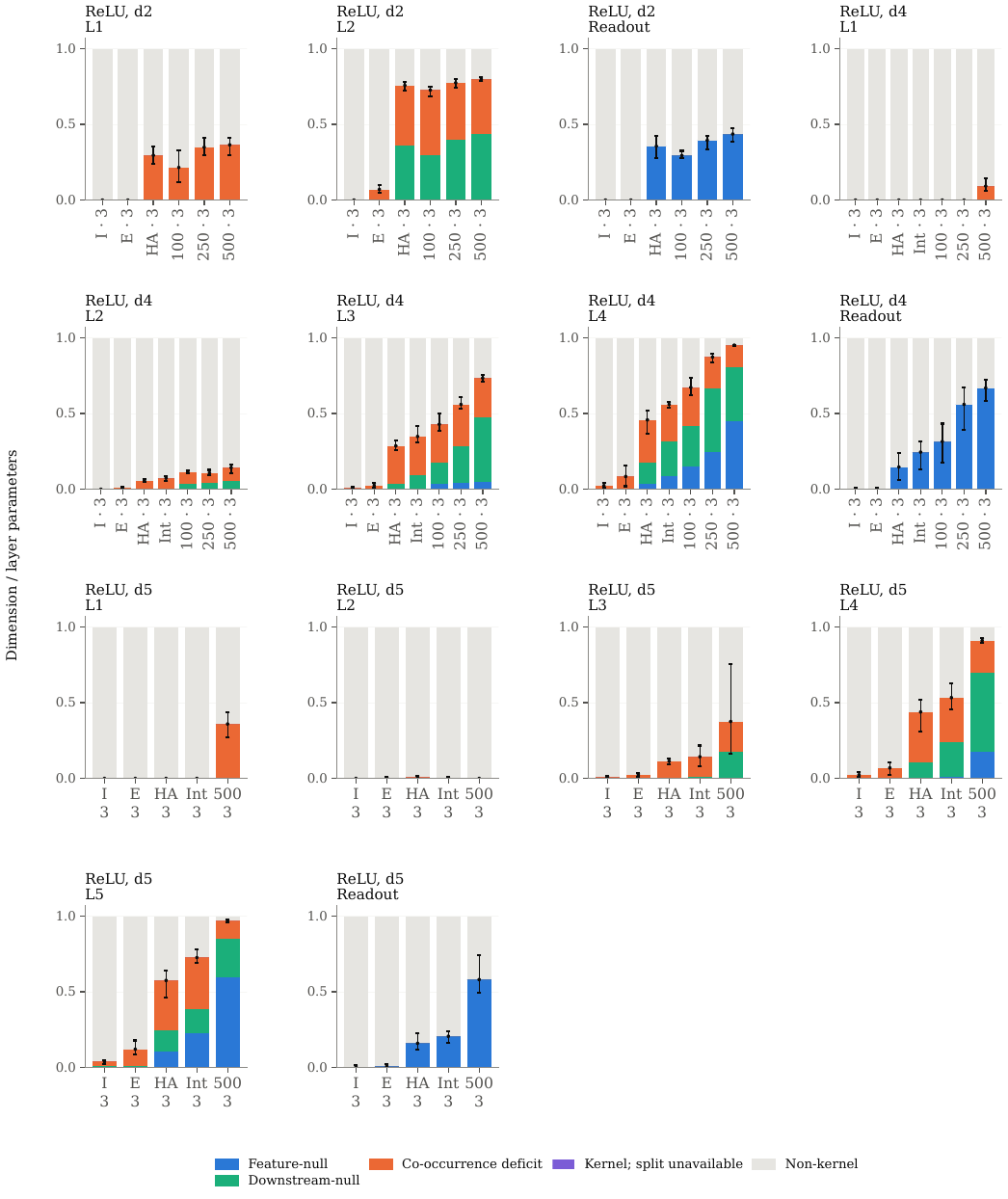}
\caption{Atomic decomposition across the indicated models and layers. Models use MNIST unless labeled SVHN; w and d denote width and hidden depth. Panels show a layer across stages, or all editable layers at the stated endpoint. Heights are fractions of layer parameters; bars average seeds and whiskers span their kernel fractions. Seed counts follow each stage label, on a second line or after a dot. I, E, HA, Int and Val denote initialization, early fitting, high accuracy, interpolation and validation selection; numbers denote fixed epochs. Stages are grouped by milestone and then fixed epoch. For reduced-data models, $n$ is the retraining or analysis-set size as titled. Hatched kernels have no measured component split. Counts use the $10^{-9}$ rank rule; GELU late-stage counts and layers without a spectral gap remain threshold-dependent.}\label{fig:c-atomic-1}
\end{figure}
\begin{figure}[!htbp]\centering
\includegraphics[width=6.272in]{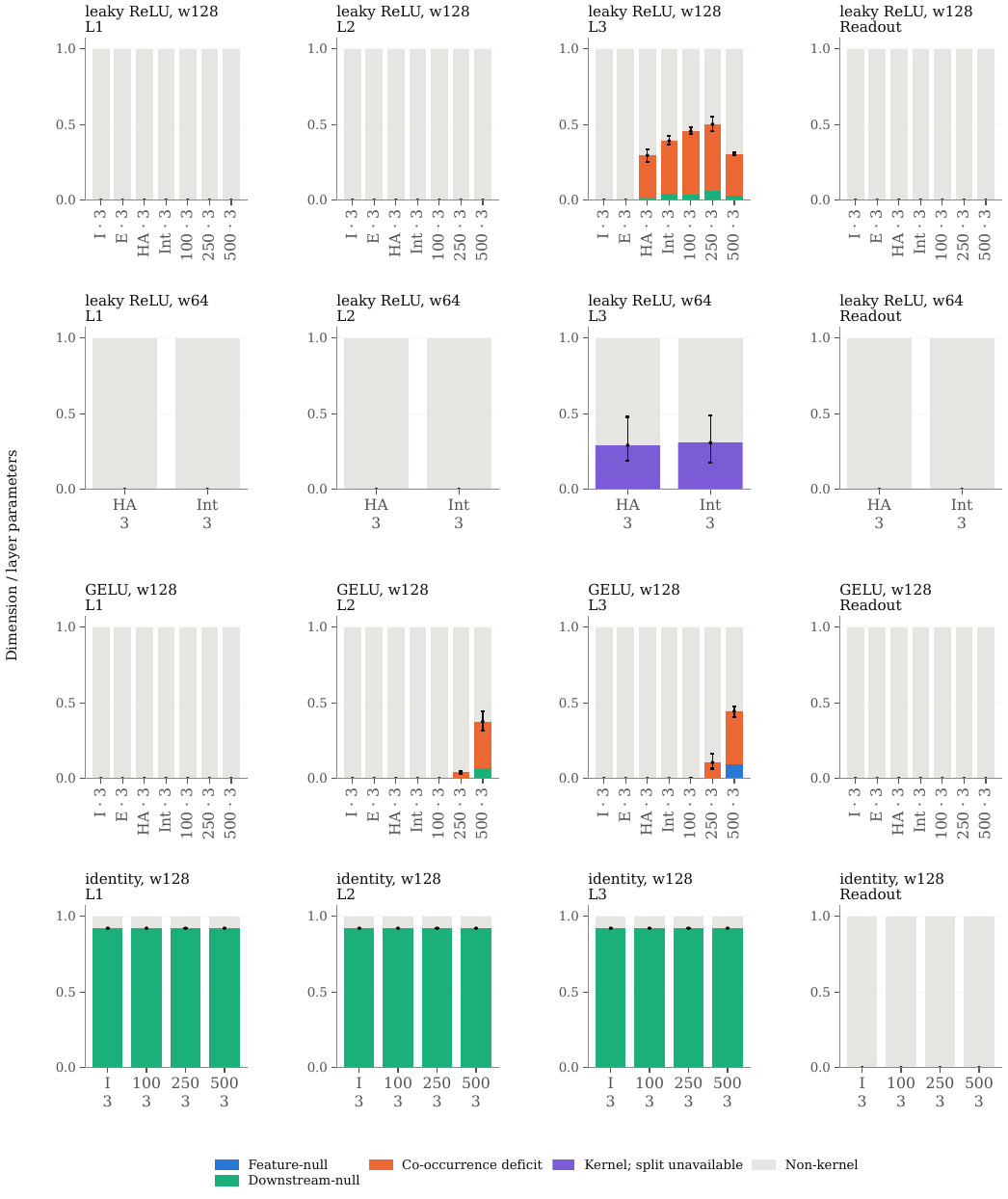}
\caption{Atomic decomposition across the indicated models and layers. Models use MNIST unless labeled SVHN; w and d denote width and hidden depth. Panels show a layer across stages, or all editable layers at the stated endpoint. Heights are fractions of layer parameters; bars average seeds and whiskers span their kernel fractions. Seed counts follow each stage label, on a second line or after a dot. I, E, HA, Int and Val denote initialization, early fitting, high accuracy, interpolation and validation selection; numbers denote fixed epochs. Stages are grouped by milestone and then fixed epoch. For reduced-data models, $n$ is the retraining or analysis-set size as titled. Hatched kernels have no measured component split. Counts use the $10^{-9}$ rank rule; GELU late-stage counts and layers without a spectral gap remain threshold-dependent.}\label{fig:c-atomic-2}
\end{figure}
\begin{figure}[!htbp]\centering
\includegraphics[width=6.272in]{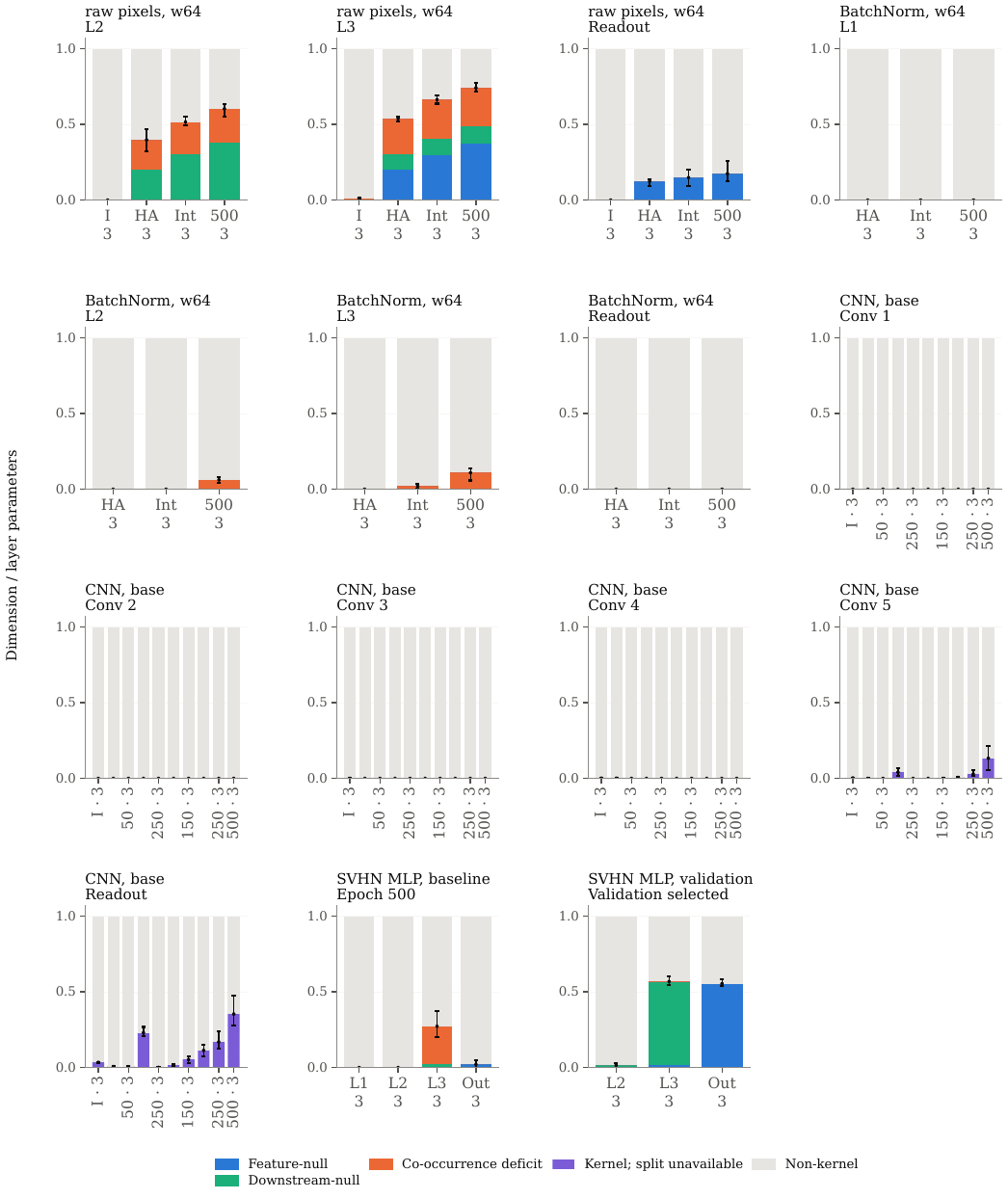}
\caption{Atomic decomposition across the indicated models and layers. Models use MNIST unless labeled SVHN; w and d denote width and hidden depth. Panels show a layer across stages, or all editable layers at the stated endpoint. Heights are fractions of layer parameters; bars average seeds and whiskers span their kernel fractions. Seed counts follow each stage label, on a second line or after a dot. I, E, HA, Int and Val denote initialization, early fitting, high accuracy, interpolation and validation selection; numbers denote fixed epochs. Stages are grouped by milestone and then fixed epoch. For reduced-data models, $n$ is the retraining or analysis-set size as titled. Hatched kernels have no measured component split. Counts use the $10^{-9}$ rank rule; GELU late-stage counts and layers without a spectral gap remain threshold-dependent.}\label{fig:c-atomic-3}
\end{figure}
\FloatBarrier

\paragraph{Activation, normalization and dataset comparisons.}\label{app:activations}\label{app:bn}\label{app:raw}\label{app:svhn}
The same measurements distinguish several ways of obtaining an invariant subspace. Leaky ReLU retains a late hidden-layer kernel through cancellations across units, despite having no exactly inactive negative piece. GELU develops broad approximate tails; its late numerical kernels have no clear spectral gap. BatchNorm leaves substantially smaller kernels, even at matched fitting stages, while the raw-pixel and SVHN comparisons show that representation and fitting conditions also affect their size. The preceding panels include these models, and the invariant subspace and transition comparisons below use the same protocols. Predictive-performance tables are retained in Appendix~\ref{app:performance}; training recipes are in Appendix~\ref{app:methods-activations} and Appendix~\ref{app:methods-data}.

\paragraph{Large layers and numerical resolution.}

For the width-512 MLP, randomized rank calculations resolve the third hidden-layer and readout ranks in two seeds. The second hidden layer has a rank lower bound. Table~\ref{tab:c-wide-ranks} reports each result in that form, so an unresolved rank remains distinct from a measured kernel dimension.
\begin{table}[!htbp]\centering
\caption{Width-512 ReLU MLP at epoch 500, 10,000 analysis images. A resolved numerical rank determines the corresponding nullity; a rank lower bound gives only a nullity upper bound.}\label{tab:c-wide-ranks}
\small\begin{tabular}{llrrr}\toprule
Layer & Seed & Parameters & Rank & Kernel dimension\\\midrule
L2 & 0 & 262656 & $\ge8192$ & $\le254464$\\
L3 & 0 & 262656 & $560$ & $262096$\\
Readout & 0 & 5130 & $360$ & $4770$\\
L2 & 1 & 262656 & $\ge8192$ & $\le254464$\\
L3 & 1 & 262656 & $659$ & $261997$\\
Readout & 1 & 5130 & $430$ & $4700$\\
\bottomrule\end{tabular}\end{table}
\FloatBarrier

The fine-tuned ViT has a different source of exact directions. LayerNorm identities give a compatible invariant subspace in the final MLP and readout, checked in all fine-tuned seeds. Table~\ref{tab:c-native-exact} separates these constructed spaces from counting lower bounds on tangent nullity. The full atomic ranks of the large ViT layers remain unresolved. The architecture is described in Appendix~\ref{app:methods-activations}, and the finite-radius results are in Appendix~\ref{app:native-vit}.
\begin{table}[!htbp]\centering
\caption{ViT-B/16 on RESISC45. Constructed exact dimensions are checked in three fine-tuned seeds. Counting lower bounds use 10,000 images and 45 outputs; they bound tangent nullity without constructing additional invariant subspaces.}\label{tab:c-native-exact}
\small\begin{tabular}{lrrrr}\toprule
Layer & Parameters & Exact space & Kernel lower bound & \shortstack{Remaining kernel\\lower bound}\\\midrule
First MLP affine layer & 2,362,368 & 3,072 & 1,912,368 & 1,909,296\\
Second MLP affine layer & 2,360,064 & 3,073 & 1,910,064 & 1,906,991\\
Readout & 34,605 & 45 & --- & ---\\
\bottomrule\end{tabular}\par\smallskip\footnotesize The remaining-kernel bound subtracts the constructed exact dimension from the full-kernel counting bound.
\end{table}

\paragraph{Approximate atomic directions.}
A numerical kernel counts directions below the tight rank threshold. Larger singular-value thresholds also include directions with a small, nonzero first-order output response. The next panels compare these two counts across the measured training trajectories, using the same relative thresholds across models and keeping every measured layer. The approximate spaces can occupy a substantial fraction of a layer even when its tight-threshold kernel is small. For GELU, a late-stage count without a spectral gap is a threshold-dependent numerical quantity.

\begin{figure}[!htbp]\centering
\includegraphics[width=4.754in]{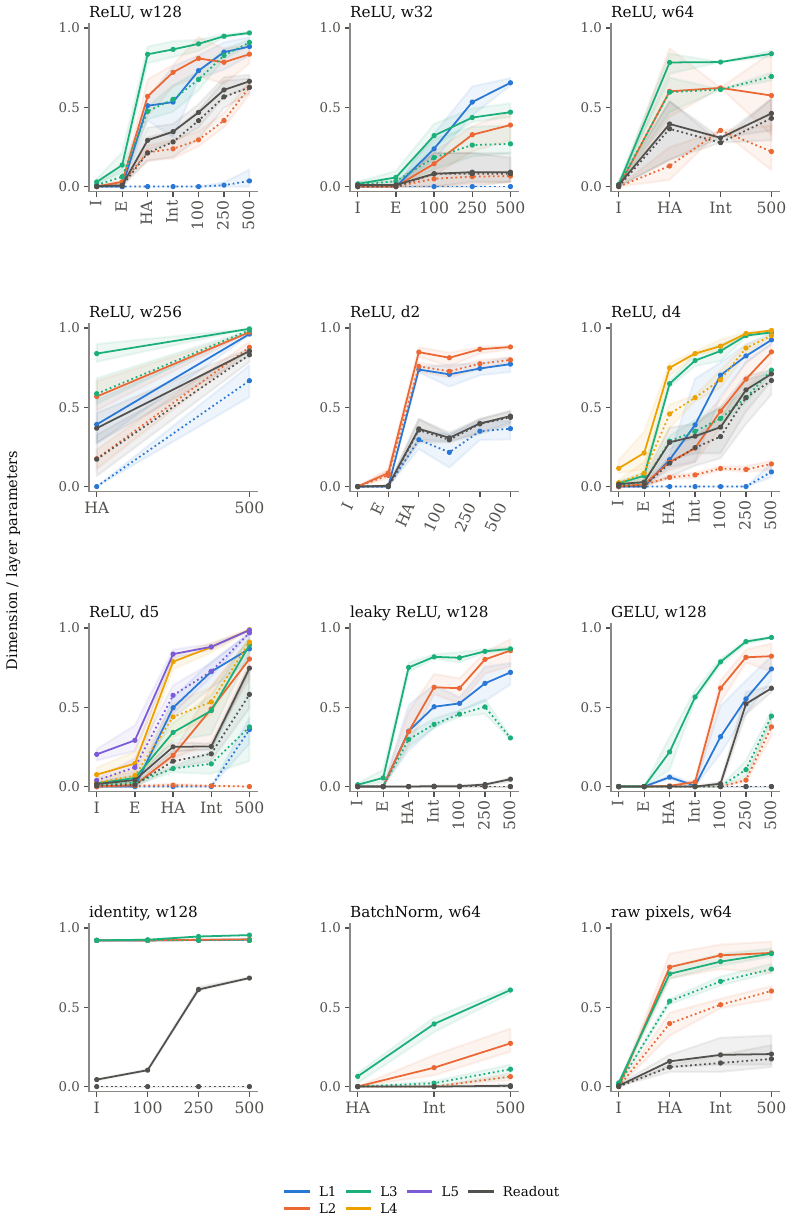}
\caption{Approximate atomic dimensions for models measured at multiple training stages, with all measured layers shown. MNIST is used unless labeled SVHN; layer numbers refer to hidden layers or convolutional layers. Solid curves count singular values at most $10^{-3}\sigma_1$ and dotted curves use $10^{-9}\sigma_1$, each divided by layer parameter count. Lines are seed means and bands are seed ranges. Missing measurements remain gaps. Stage positions are categorical, with abbreviations as in the atomic panels.}
\label{fig:atomic-tails}
\end{figure}
\begin{figure}[!htbp]\centering
\includegraphics[width=1.804in]{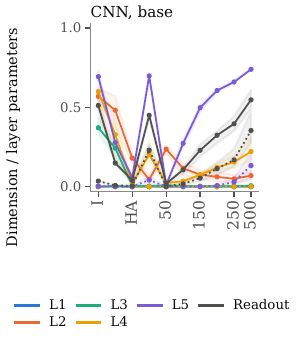}
\caption{Continued. Approximate atomic dimensions for models measured at multiple training stages, with all measured layers shown. MNIST is used unless labeled SVHN; layer numbers refer to hidden layers or convolutional layers. Solid curves count singular values at most $10^{-3}\sigma_1$ and dotted curves use $10^{-9}\sigma_1$, each divided by layer parameter count. Lines are seed means and bands are seed ranges. Missing measurements remain gaps. Stage positions are categorical, with abbreviations as in the atomic panels.}

\end{figure}
\FloatBarrier

\paragraph{Selective units and channels within the atomic kernel.}
For a ReLU unit, an incoming edit that annihilates the features of all samples on which it is active preserves its frozen activation on every sample. This gives an explicit part of the atomic kernel. With $\mathcal A_j$ the active sample set of unit $j$ and $\mathcal H_{\ell j}^{\rm act}(X)$ its augmented active-feature span, its dimension is
\[
 \dim G_\ell^{\rm sel}(X)
 =\sum_j\left(n_{\ell-1}+1-\dim\mathcal H_{\ell j}^{\rm act}(X)\right).
\]
An active set with too few samples leaves free directions by sample count alone. Further directions arise when its features span less than their count permits. The CNN construction imposes the analogous condition on all active patches of a channel.

The selective-space panels include the full MLP width and depth sweeps on both the 10,000-image analysis set and all 60,000 training images, together with the base MNIST and SVHN CNN trajectories. The CNN width endpoints are reported in Appendix~\ref{app:cnn-width}. Selective spaces often account for most of the hidden atomic kernel in fitted ReLU models. The unit-decomposition panels separate directions confined to one unit from the remaining directions that require cancellation across different units. These distinctions matter for leaky ReLU: class selectivity alone does not explain its atomic kernel, because negative-side slopes remain nonzero.

\begin{figure}[!htbp]\centering
\includegraphics[width=6.315in]{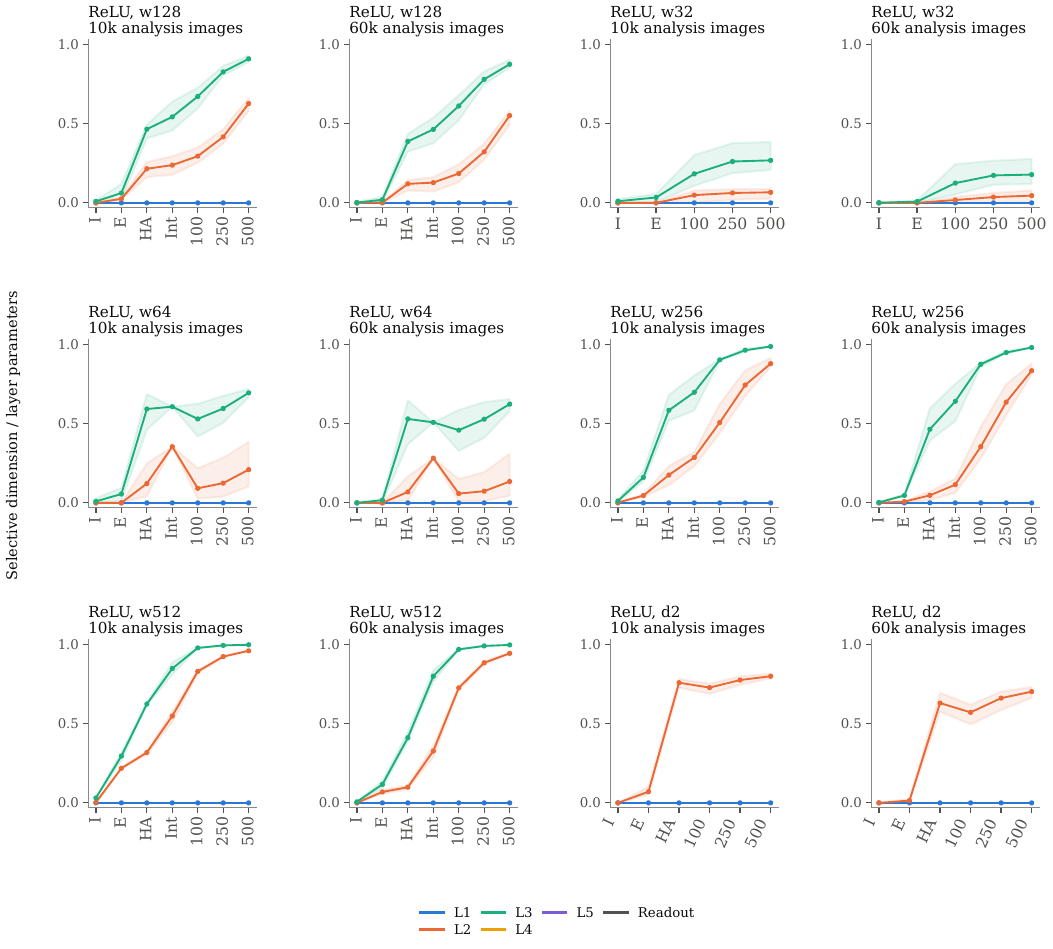}
\caption{Selective dimension as a fraction of layer parameters, across all tested stages and analysis-set sizes. MNIST is used unless labeled SVHN. Colors identify hidden layers or convolutional layers; lines are seed means and bands span seeds. MLP spaces constrain each unit on its active samples; CNN spaces constrain each channel on its active input patches. A zero identifies an empty selective construction; the full atomic kernel can still contain other directions. Stage abbreviations follow the atomic panels.}
\end{figure}
\begin{figure}[!htbp]\centering
\includegraphics[width=4.754in]{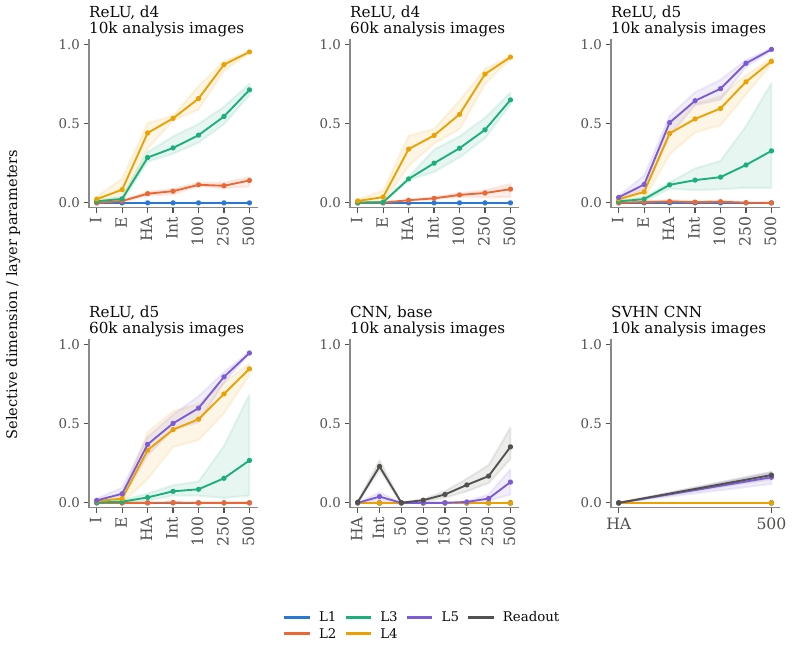}
\caption{Continued. Selective dimension as a fraction of layer parameters, across all tested stages and analysis-set sizes. MNIST is used unless labeled SVHN. Colors identify hidden layers or convolutional layers; lines are seed means and bands span seeds. MLP spaces constrain each unit on its active samples; CNN spaces constrain each channel on its active input patches. A zero identifies an empty selective construction; the full atomic kernel can still contain other directions. Stage abbreviations follow the atomic panels.}
\end{figure}
\begin{figure}[!htbp]\centering
\includegraphics[width=4.762in]{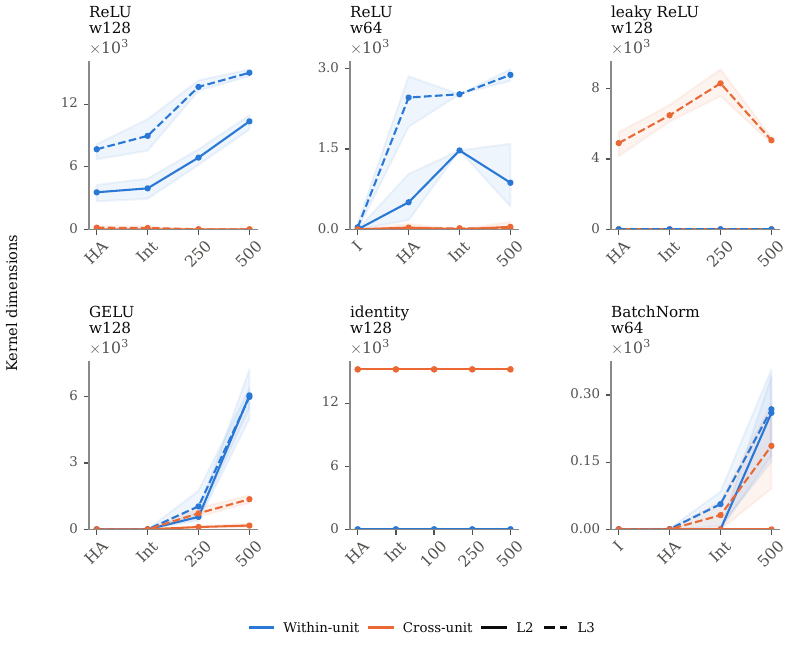}
\caption{Within-unit and cross-unit kernel dimensions for MNIST MLPs. Each panel combines L2 and L3, distinguished by line style. Curves are seed means and bands span seeds. ReLU within-unit directions agree with the selective construction where containment checks pass. Leaky-ReLU and GELU per-unit operators use their actual slopes; late GELU counts remain numerical threshold-dependent quantities.}
\end{figure}
\FloatBarrier

\begin{figure}[!htbp]\centering
\includegraphics[width=.60\linewidth,height=.18\textheight,keepaspectratio]{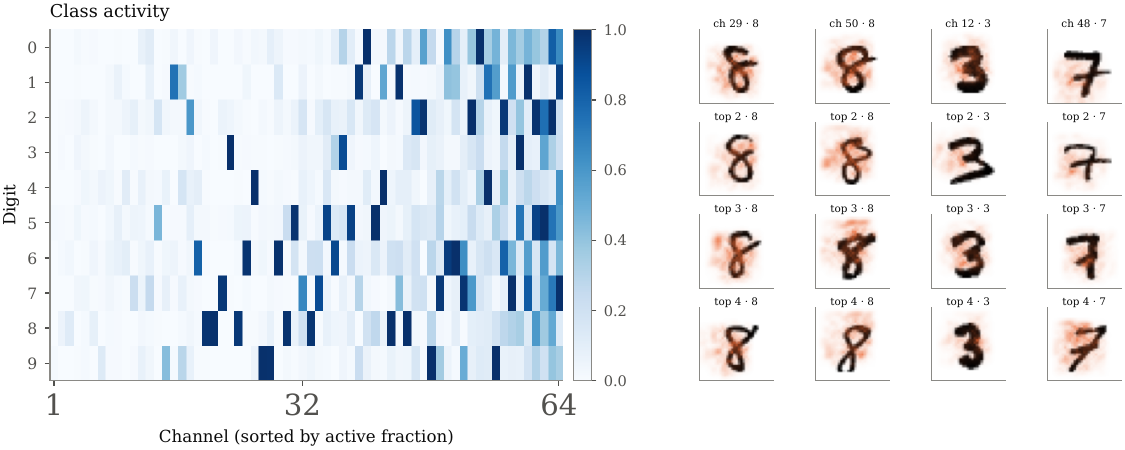}
\caption{Base MNIST CNN, conv5, seed 0 at epoch 500: class selectivity and the four most activating real images per channel, with input-gradient attribution. This qualitative example complements the multi-seed selective-space measurements.}\label{fig:cnn-units}
\end{figure}

The channel-level CNN measurements provide a concrete view of selectivity. Figure~\ref{fig:cnn-units} shows which real inputs activate each late convolutional channel. The images reveal class-specific stroke patterns in this example; the preceding panels provide the corresponding dimensions across seeds and settings.

\FloatBarrier

\subsubsection{Finite-radius tests of atomic spaces}
\label{app:rays}\label{app:radius-full}
A Jacobian kernel describes the first-order response. We test the actual network along its directions to establish how far the frozen invariance persists. Following Appendix~\ref{app:methods-radius}, we compare single-layer kernels, approximate singular tails, generic directions, and edits to dead-unit incoming or outgoing coordinates. We record both the first gate crossing and the first output change exceeding $10^{-3}$.

\begin{figure}[!htbp]\centering
\includegraphics[width=6.195in]{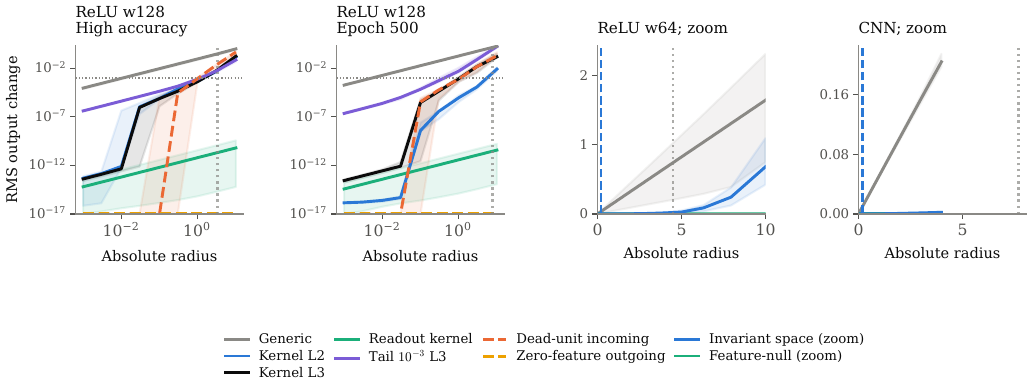}
\caption{Atomic rays and zoomed invariant subspace responses on MNIST. Left pair: width-128 ReLU single-layer kernels, approximate tails and dead-unit controls, with medians of 32 probes per seed and ranges across three seeds; the generic reference edits the whole network. Right pair: seed-0 width-64 MLP and CNN slices, with medians and 10th--90th percentiles over angles; feature-null curves use their own slice axis. These panels use linear axes to resolve visible drift after a first crossing (blue dashed marker). Gray vertical dotted markers give $0.1\|\theta\|_2$; the left panels also mark output visibility at $10^{-3}$. Blue and green retain the invariant-direction and feature-null roles across panels, while the legend distinguishes atomic kernels from multilayer invariant subspaces.}\label{fig:rays-overview}\label{fig:zoom-rays}
\end{figure}

For the width-128 ReLU MLP, the readout kernel and outgoing weights of zero features stay flat throughout the scan. Incoming edits to dead units remain flat until the units reactivate. Hidden-layer directions often cross gates before their output response becomes visible. Figure~\ref{fig:rays-overview} separates these mechanisms, including the dead-unit controls, with only one layer edited at a time. The multilayer invariant subspace scans in Appendix~\ref{app:composition-results} test simultaneous edits to several layers.

\FloatBarrier

\begin{figure}[!htbp]\centering
\includegraphics[width=3.105in]{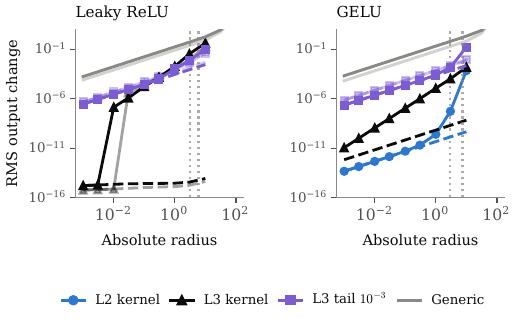}
\caption{Leaky-ReLU and GELU width-128 MNIST MLPs, seed 0, at high accuracy and epoch 500. Single-layer kernel and approximate-tail directions separate frozen invariance from finite-distance output drift. The gray generic reference edits the full network. Gray vertical dotted lines mark $0.1\|\theta\|_2$ at each checkpoint.}\label{fig:c-activation-rays}
\end{figure}

\paragraph{Activation controls.}
Leaky ReLU removes ReLU's exactly zero negative-side slope, while GELU replaces the kink by a smooth transition. Figure~\ref{fig:c-activation-rays} tests how these changes affect single-layer invariance in the width-128 MNIST models. It contrasts numerical kernel directions in L2 and L3 with a L3 approximate tail. Solid curves show the actual response and dashed curves the frozen response; lighter curves denote high accuracy and darker curves epoch 500. The generic curves perturb all layers and serve as a full-network sensitivity reference, so their amplitudes are not a matched single-layer comparison.

\paragraph{Smooth activations and approximate atomic spaces.}
\label{app:gelu-eps}

\begin{figure}[!htbp]\centering
\includegraphics[width=1.924in]{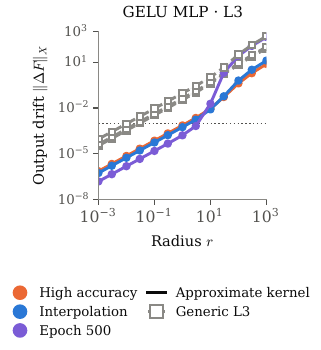}
\caption{Extended GELU L3 rays in width-128 MNIST MLPs. The approximate kernel uses $\sigma(J_3)/\sqrt{s}\le10^{-3}$; generic controls edit the same layer. Lines show the mean and shading one sample SD of the within-seed medians over 256 rays, across three seeds. Colors identify the ReLU-matched training stages. The dotted line marks output drift $10^{-3}$.}\label{fig:gelu-extended-rays}
\end{figure}

For a smooth activation such as GELU, an exact atomic kernel can be empty. Approximate Jacobian kernels instead collect directions with a small first-order response; finite steps can still accumulate drift through activation curvature. Figure~\ref{fig:gelu-extended-rays} tests the L3 approximate kernel at the common sensitivity cutoff $10^{-3}$. All 256 sampled directions per seed and stage stay below output drift $10^{-3}$ through radius 1, while all exceed it by radius 10. The smaller response than generic L3 edits therefore has a finite measured extent.

The diagnostic $\widehat B(w)$ in Figure~\ref{fig:gelu-eps} is defined in Appendix~\ref{app:methods-radius}; it uses measured RMS preactivation displacements and base-point gains in place of uniform quantities. It exceeds measured transition by two to three orders of magnitude; local curvature is much smaller than its global bound along these probes.

\begin{figure}[!htbp]
\centering
\includegraphics[width=4.786in]{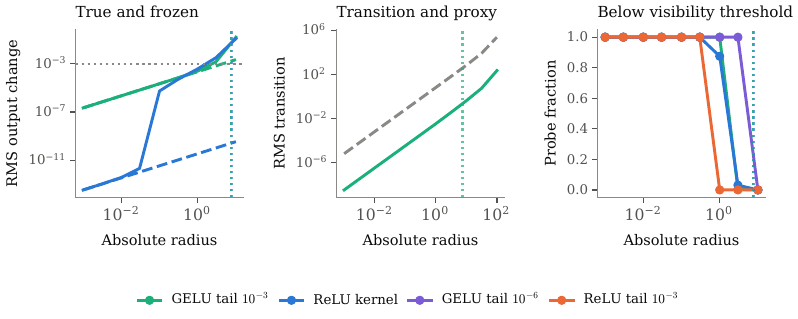}
\caption{GELU approximate flatness and the diagnostic $\widehat B(w)$ from Appendix~\ref{app:methods-radius}. Seed-0 width-128 MNIST models at epoch 500. The left panel compares true (solid) and frozen (dashed) responses; the middle compares the transition norm with the diagnostic proxy (gray dashed); the right gives the fraction of probes below output change $10^{-3}$. Vertical dotted lines mark $0.1\|\theta\|_2$ in each model color. The proxy uses RMS displacements and base-point gains. Numerical late-stage kernels do not certify an exact analytic ocean.}
\label{fig:gelu-eps}
\end{figure}\begin{figure}[!htbp]
\centering
\includegraphics[width=6.194in]{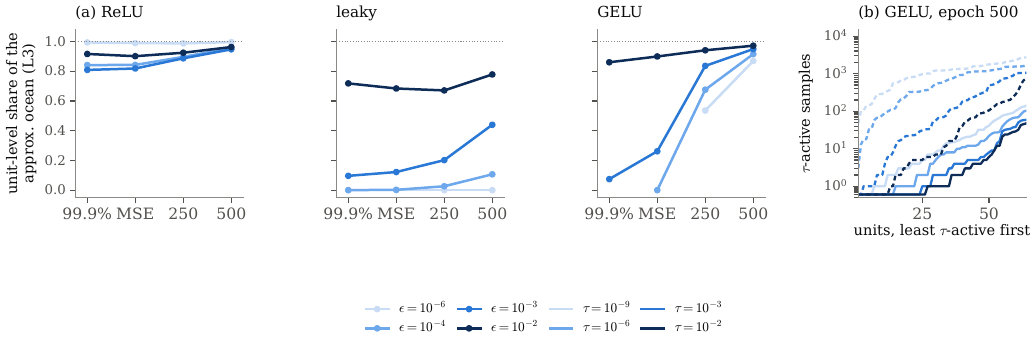}
\caption{Soft selectivity under GELU: weak derivatives and concentrated active-input geometry accompany large approximate tails.}
\label{fig:gelu-soft}
\end{figure}

\FloatBarrier

\subsection{Multilayer invariant subspace construction and validation}
\label{app:composition-results}\label{app:cnn-branch}\label{app:selective}
Single-layer invariance leaves a further question: can several layers be edited at once while preserving the frozen outputs? An early-layer edit changes the inputs received by later layers, so individually invariant edits can interact. The received-space construction checks this compatibility. The construction plots follow the resulting dimensions over training for each attempted layer order and explicit subspace construction. Each curve is one seed. We show models measured at more than one epoch. Crosses mark zero retained dimension, including failed validations; unattempted constructions have no point.

For the tested width-64 full-data MNIST models, the layer orders recover the full atomic sum. CNN invariant subspaces retain almost that sum after removing directions at the numerical rank boundary. The reduced-data experiments in Appendix~\ref{app:training-data} show why compatibility must still be checked: different orders can give different invariant subspace dimensions from the same atomic spaces. We also report the raw-pixel, BatchNorm and SVHN comparisons. Full multilayer constructions have not been computed for every model in the MLP size sweeps, so their atomic dimensions alone do not establish equally large multilayer invariant subspaces.

\begin{figure}[!htbp]\centering
\includegraphics[width=4.762in]{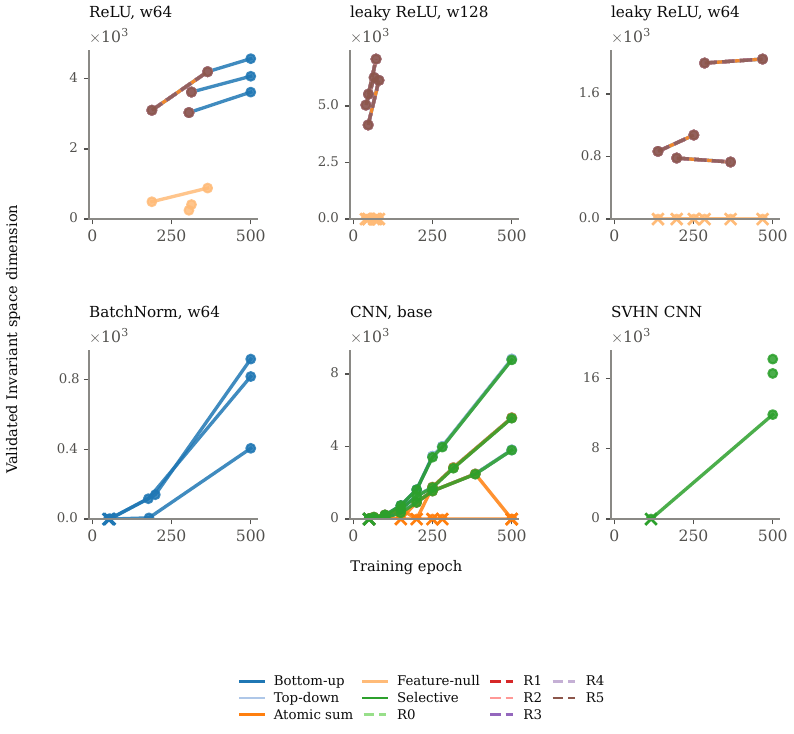}
\caption{Multilayer invariant subspace dimensions versus actual checkpoint epoch. Each curve is one seed; colors identify constructions, and dashed curves denote random layer orders R0--R5. Coincident curves have equal retained dimensions. Crosses mark zero retained dimension, including failed validations; unattempted constructions have no point. MNIST is used unless labeled SVHN. Only models with a trajectory measured within at least one seed are shown.}
\end{figure}
\FloatBarrier

\subsection{Radius tests}
\label{app:radius-results}
\subsubsection{Directional certificates}
\label{app:radius-certificates}
\begin{figure}[!htbp]\centering
\includegraphics[width=6.149in]{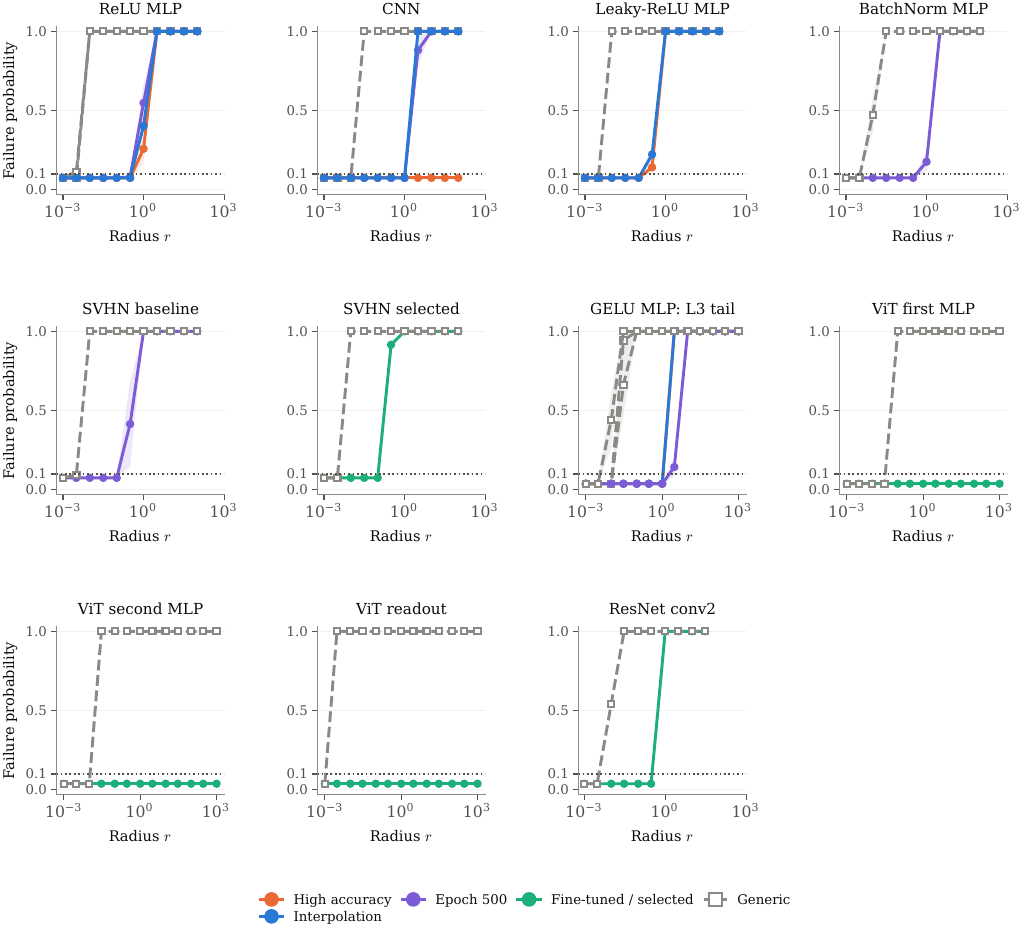}
\caption{Failure-probability upper bounds at absolute output tolerance $10^{-3}$. Solid curves: constructed spaces; dashed curves: generic controls. Colors denote training stages; shading is one sample SD across available seeds. The dotted line marks $10\%$ allowed failures. Bounds hold with $95\%$ simultaneous confidence over all 1126 comparisons in this figure; lines connect tested radii. GELU uses the approximate L3 kernel with $\sigma(J_3)/\sqrt{s}\le10^{-3}$. ViT shows first-MLP and readout feature-null edits and second-MLP downstream-null edits; ResNet shows the final convolution's selective family.}
\label{fig:radius-certificates-all}\label{fig:radius-variants}
\end{figure}

The invariant spaces permit substantially larger certified grid radii than generic directions across the tested architectures (Figure~\ref{fig:radius-certificates-all}). The early CNN spaces remain safe throughout the scan but contain few directions; later spaces are much larger and have shorter measured radii. Exact ViT feature-null and LayerNorm-shift families remain at numerical precision throughout their extended scans.

\FloatBarrier

\subsubsection{Ray responses}\label{app:radius-responses}
The actual and frozen responses show where activation changes become visible along the same edits. The following comparisons include wider networks, different activations and input representations. Their descriptive bands summarize seed variation; the certificates above use independent-ray failure counts.
\begin{figure}[!htbp]\centering
\includegraphics[width=6.313in]{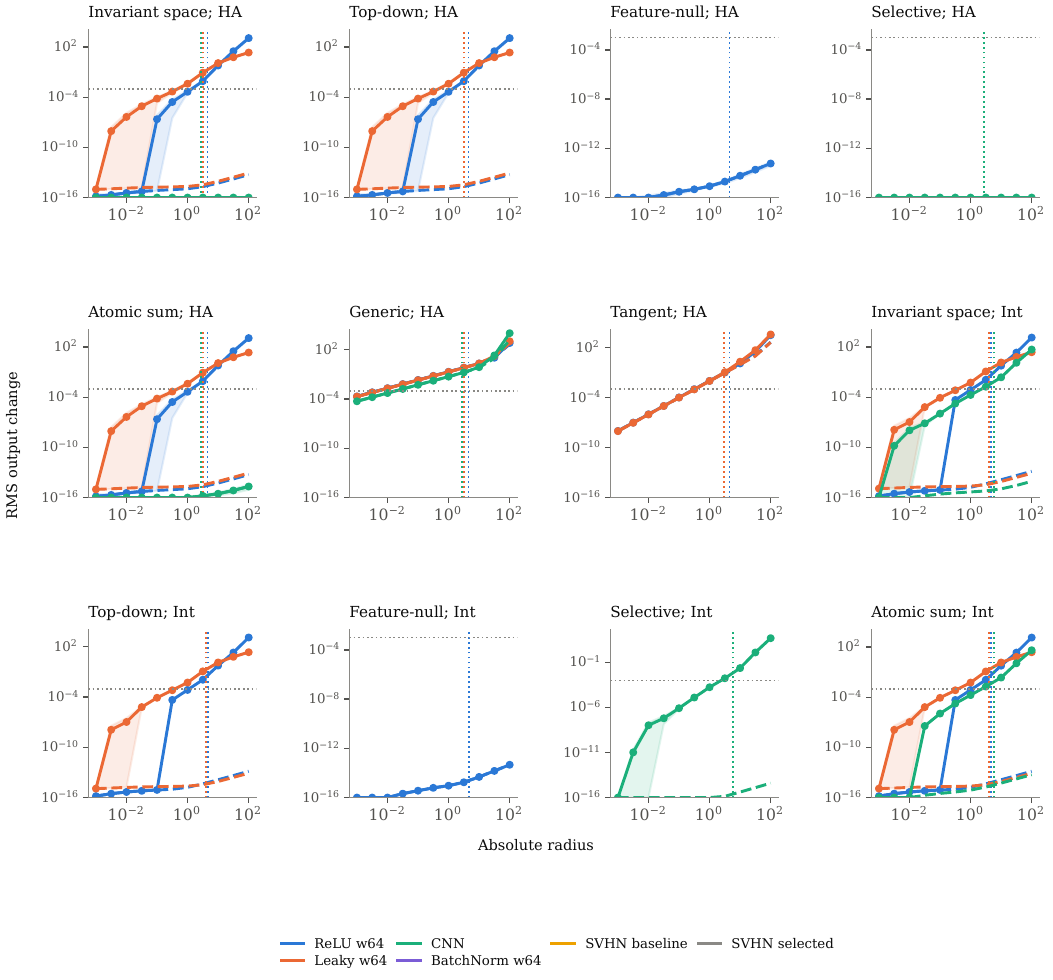}
\caption{Absolute-radius responses across architectures and activations. Each panel fixes the training stage and direction family; only width 64 is shown for leaky ReLU. The Invariant subspace panels combine MLP bottom-up constructions with the CNN received-space construction; Selective panels retain the separate CNN selective candidates. Solid and dashed curves give actual and frozen responses, averaged over seed medians, with seed ranges shaded. Dotted vertical lines mark one tenth of the network norm; the horizontal dotted line marks output change $10^{-3}$. Missing curves indicate unmeasured model--stage combinations.}\label{fig:appendix-ray-comparison}
\end{figure}
\begin{figure}[!htbp]\centering
\includegraphics[width=6.313in]{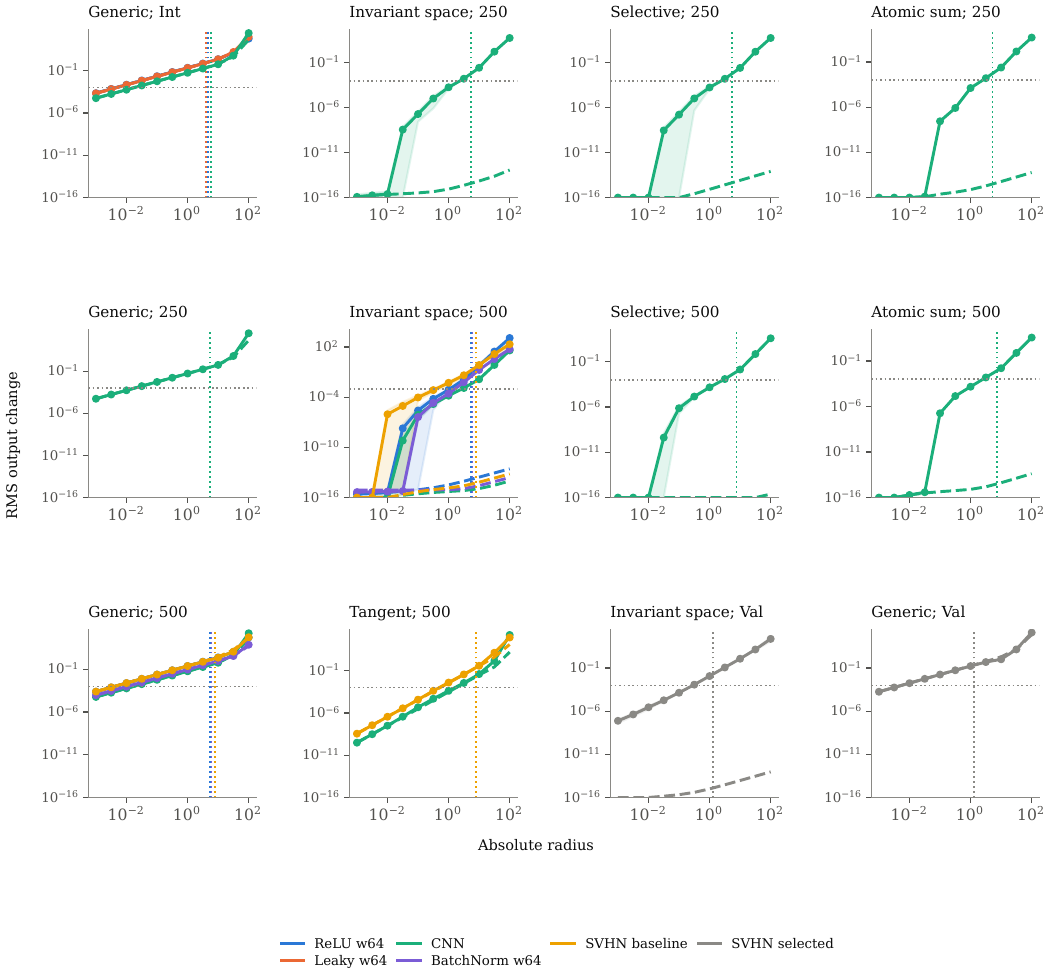}
\caption{Additional stages and direction families, with the conventions of Figure~\ref{fig:appendix-ray-comparison}.}
\end{figure}
\FloatBarrier

\subsection{Activation-transition potential}
\label{app:transition}
We next ask how departures from the base activation pattern contribute to the output response. The local residual $\rho_\ell(w)$ measures departure from the base affine activation at layer $\ell$, and the edited frozen tail $T_\ell(w)$ propagates it through the remaining layers. Their vector contributions add to the activation-transition term,
\[
 A_X^{\rm tr}(w)=\sum_\ell T_\ell(w)\rho_\ell(w).
\]
We measure its magnitude relative to the actual output change, $\|A_X^{\rm tr}(w)\|_X/\|\Delta F_X(w)\|_X$, using 256 paired directions per checkpoint and radius. This ratio can exceed one when the transition and frozen responses partly cancel.

The panels compare width-128 ReLU, leaky-ReLU and GELU MLPs, the base CNN, and width-64 plain and BatchNorm MLPs. Single-layer perturbations identify where the response originates; full-network perturbations also include interactions between edited layers. The figures use absolute radii. The vertical dotted markers identify edits with Euclidean norm equal to 10\% of the network's parameter norm. An absolute radius is the norm of the full edit, including when only one layer moves. Each stage has its own marker color, and positions summarize the three seeds.
\begin{figure}[!htbp]\centering
\includegraphics[width=6.313in]{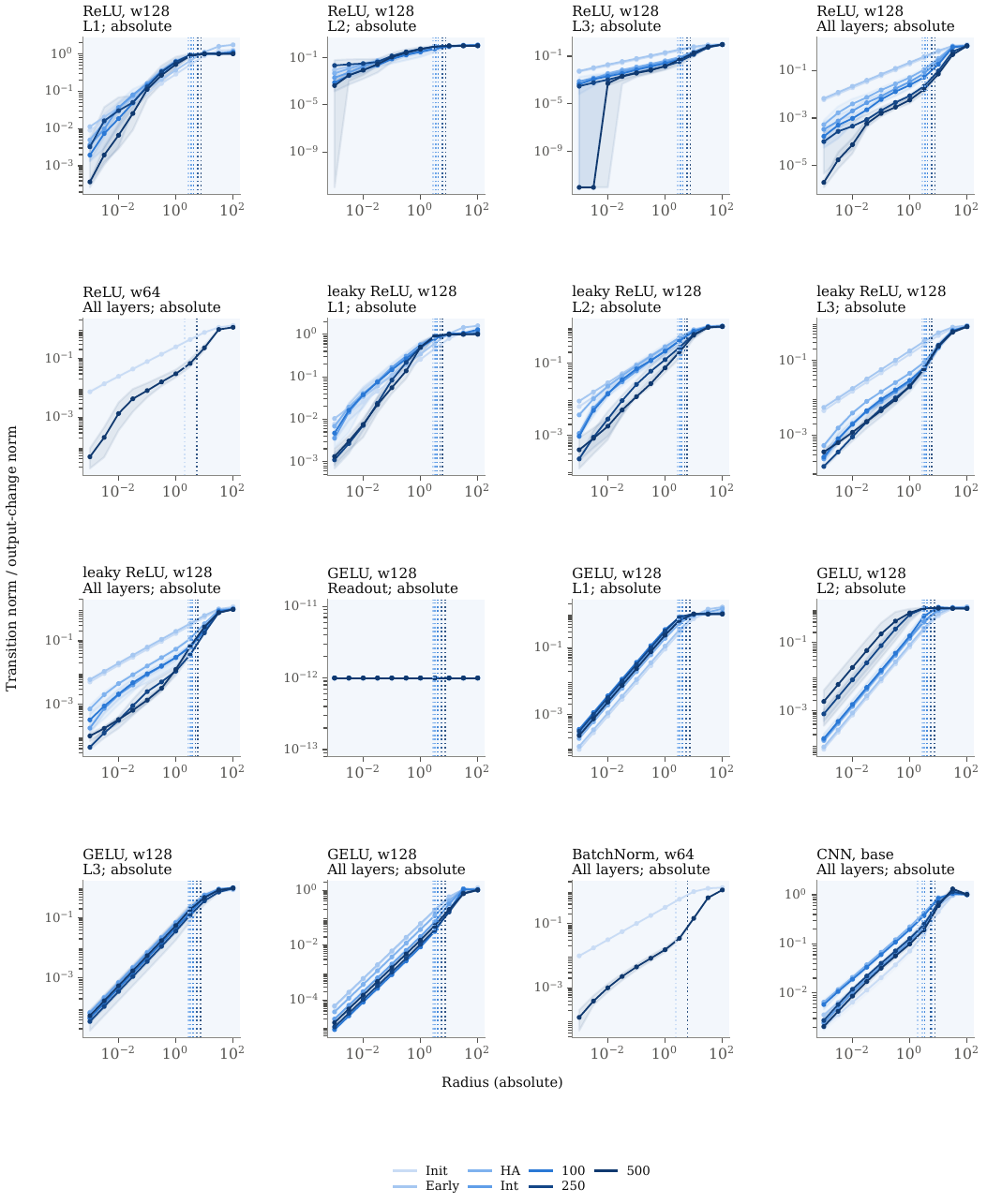}
\caption{Activation-transition share $\|A_X^{\rm tr}\|_X/\|\Delta F_X\|_X$ for the indicated editable layers. Curves show means of within-seed medians over 256 directions; bands span seeds. Dotted vertical lines mark edits of norm $0.1\|\theta\|_2$. Ratios above one indicate cancellation with the frozen response. Values below $10^{-12}$ use that plotting floor.}
\end{figure}
\FloatBarrier

\begin{figure}[!htbp]\centering
\includegraphics[width=6.274in]{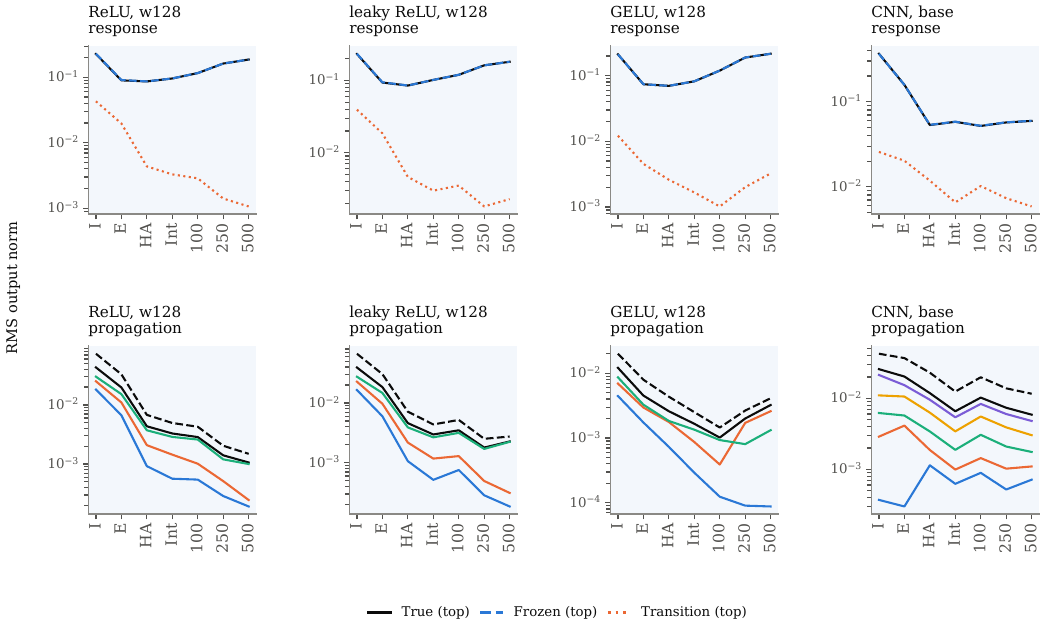}
\caption{Response magnitudes and layerwise propagation on 256 full-network directions per seed at absolute radius 1. Upper panels separate true, frozen and transition norms. Lower panels show each propagated layer residual in the atomic-plot layer colors, their sum of norms (black dashed), and the norm after vector summation (black solid). Their separation measures cancellation across layers. Curves average within-seed medians over three seeds. Values below $10^{-16}$ use that plotting floor.}\label{fig:c-transition-details}
\end{figure}

\paragraph{Response magnitudes and cancellation across layers.}
A large ratio of transition norm to output-change norm can reflect a large activation residual or a small net output change caused by cancellation. Figure~\ref{fig:c-transition-details} distinguishes these cases. Its upper row compares the true, frozen and transition norms at one fixed absolute radius. Its lower row separates the propagated layer residuals and compares their summed norms with the norm of their vector sum. The gap between those two quantities measures cancellation across layers.
\FloatBarrier

\paragraph{Margins and activation curvature.}
The margin distributions show how far preactivations lie from the activation's transition region. ReLU and leaky ReLU are affine on each side of zero; GELU approaches affine behavior in both tails. Sigmoid, tanh and softplus provide further smooth-tail examples in Appendix~\ref{app:radius-proof}. For a fixed preactivation displacement, remaining far enough into an affine or low-curvature region makes the activation residual zero or small. This requires the entire displacement to remain in that region: a large initial margin alone does not protect an edit that returns to the transition region. Downstream propagation then determines how much of the local residual reaches the output.
\begin{figure}[!htbp]\centering
\includegraphics[width=4.606in]{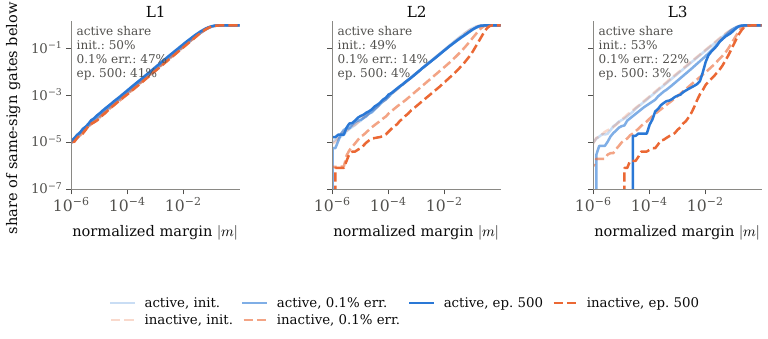}
\caption{Sign-conditional distributions of normalized preactivation margins in the MNIST activation comparison. Solid and dashed lines condition on active and inactive gates; panels also report active fractions.}\label{fig:margins}
\end{figure}

\begin{figure}[!htbp]\centering
\includegraphics[width=4.629in]{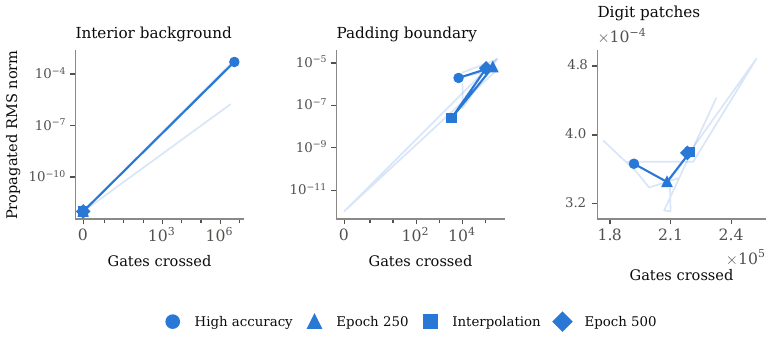}
\caption{Conv1 gate crossings and propagated residuals in the base MNIST CNN at layer-relative radius $0.01$. Each point is a training stage: within-seed medians over 128 matched directions, then means over three seeds; faint paths show individual seeds. Paths follow training order. Background and boundary panels use a horizontal scale linear up to one crossing and logarithmic above it; the digit panel uses linear axes to resolve its narrower range. Zero propagated contributions are drawn at $10^{-12}$.}\label{fig:c-cnn-background}
\end{figure}

\paragraph{CNN background and gate crossings.}
Digit images contain large background regions. A nearly zero preactivation at a background position can change sign while producing only a small activation residual. Counting these crossings therefore need not predict a large output response. Figure~\ref{fig:c-cnn-background} relates the number of conv1 crossings to their propagated contribution, separately for interior background, padding boundaries and digit patches. Each path follows training stages. The large background crossing counts at high accuracy have a small propagated contribution; later checkpoints suppress these crossings. This comparison measures the output importance of crossings rather than treating every changed gate as equally influential.

\FloatBarrier

\subsection{Controlled training: margin regularization}
\label{app:controlled}

\paragraph{Margin regularization.}
The training objective in Appendix~\ref{app:methods-margin} penalizes preactivations near zero, encouraging larger margins from ReLU's switching boundary. We compare two penalty strengths with the unregularized width-64 MNIST control, using three paired seeds and each model's certified invariant subspace at epoch 500.

The penalty empties the near-zero preactivation region and delays the first gate crossing. Output visibility changes much less: its median remains at the control's grid radius in five of the six penalized runs, and moves to the next grid point in the remaining run (Table~\ref{tab:margin-summary}). The radius curves below show the separation between these two events. The penalty also increases parameter norms and worsens fitting, as reported in Appendix~\ref{app:performance}; it therefore changes more than margins alone.
\begin{table}[!htbp]\centering
\caption{Margin arms, seed order 0/1/2. First-crossing radii are medians over probes on a half-decade grid.}\label{tab:margin-summary}
\begin{tabular}{lrrr}\toprule
Arm & first crossing & visible radius & certified dimension\\\midrule
Control & .316/.0316/.0316 &3.16/3.16/3.16&3619/4573/4072\\
$\lambda=.01$ &1/1/1&10/3.16/3.16&4483/4715/4272\\
$\lambda=.1$ &3.16/.316/3.16&3.16/3.16/3.16&4997/5532/5305\\\bottomrule
\end{tabular}\end{table}

\begin{figure}[!htbp]
\centering
\includegraphics[width=2.979in]{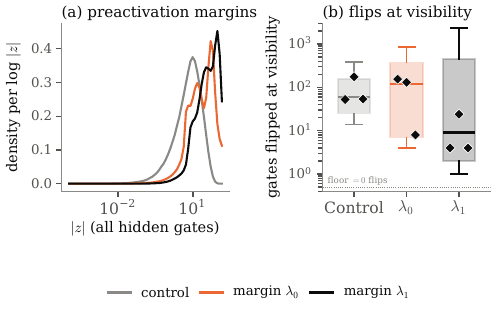}
\caption{Margin penalties shift absolute preactivations away from zero. At first visible output change, penalized models have fewer gate crossings: early output-invisible flips have been removed.}
\label{fig:margin-dists}
\end{figure}

\paragraph{Invariant subspace response under the intervention.}
The next panels compare the control and both margin penalties under absolute edits. Each arm is tested in its own invariant subspace, with a dotted marker at one tenth of its network norm.

\begin{figure}[!htbp]\centering
\includegraphics[width=3.191in]{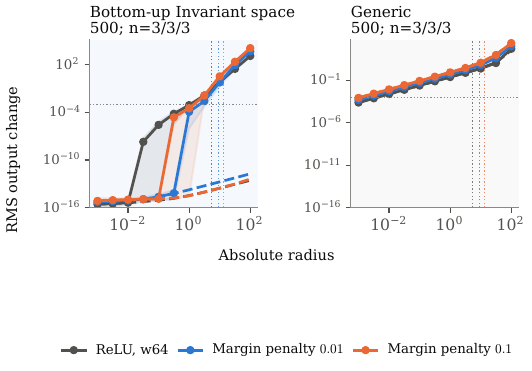}
\caption{Margin-regularized and control networks at epoch 500. Solid/dashed curves show actual/frozen drift, with means and ranges over three seed medians. Dotted lines mark output drift $10^{-3}$ and edits of one tenth of each network's norm.}
\end{figure}
\FloatBarrier

\FloatBarrier

\section{Empirical results, extended version}
\label{app:empirical-results}\label{app:ablations}\label{app:persistence-results}
The empirical comparisons ask how ocean geometry changes with network scale, learned representations, data and continued training. Activation, normalization and dataset ablations are included alongside the corresponding measurements in Appendix~\ref{app:theoretical-support}. Their implementation is described in Appendix~\ref{app:experiments}.

\subsection{Fine-tuned ViT on RESISC45}

\label{app:vit}\label{app:native-vit}

The final ViT layers contain exact invariant families with the same feature- and downstream-based interpretation as in smaller networks. We follow the protocol in Appendix~\ref{app:methods-common-families}. Compatible combinations give 3,117 feature-null directions, or 6,190 directions when LayerNorm shifts are included; these constructions do not exhaust the unresolved full atomic kernels.

\subsubsection{Layerwise results}
\label{app:vit-common}
The common protocol recovers the same 6,190 exact directions in all three seeds (Table~\ref{tab:vit-common}). Feature-null edits account for the first MLP layer and readout; the second MLP layer contributes downstream-null LayerNorm shifts. No additional selective directions are found.

\begin{table}[!htbp]\centering
\caption{ViT family dimensions under the common protocol (Appendix~\ref{app:methods-common-families}). Slash-separated entries give seeds 0/1/2; selective and approximate dimensions exclude exact-family overlap.}
\label{tab:vit-common}
\small\setlength{\tabcolsep}{5pt}
\begin{tabular}{lrrr}\toprule
Family & First MLP layer & Second MLP layer & Readout\\\midrule
Feature-null & 3,072 & 0 & 45\\
Additional downstream-null & 0 & 3,073 & 0\\
Additional selective & 0 & 0 & 0\\\midrule
Feature tail & 0 & 3,068/1,534/2,301 & 0\\
Downstream tail & 768/768/0 & 0 & 0\\\bottomrule
\end{tabular}
\end{table}

\begin{figure}[!htbp]\centering
\includegraphics[width=4.670in]{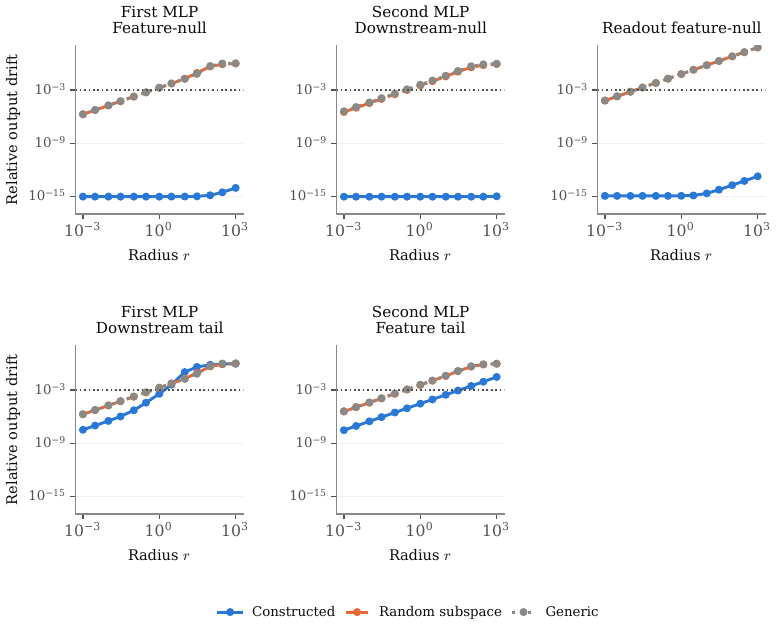}
\caption{ViT common-family ray responses. Curves show the mean of within-seed median relative RMS drift; bands show one sample SD across available seeds. The horizontal line marks $10^{-3}$. Exact-family responses remain at the numerical floor. Controls use the same edited layer and normalization.}
\label{fig:vit-common-rays}\label{tab:vit-common-radii}
\end{figure}

All sampled exact-family edits remain at the numerical floor through radius 1,000. Approximate feature and downstream families also permit larger displacements than matched controls, but show finite crossings (Figure~\ref{fig:vit-common-rays}). The second-layer feature tail crosses by radius 100 in every seed; the first-layer downstream tail crosses by radius 3 where nonempty. The latter is empty in seed 2, with its smallest rate just above the selection cutoff.

\subsubsection{Multilayer combinations and relative-cutoff comparisons}
\begin{figure}[!htbp]\centering
\includegraphics[width=6.122in]{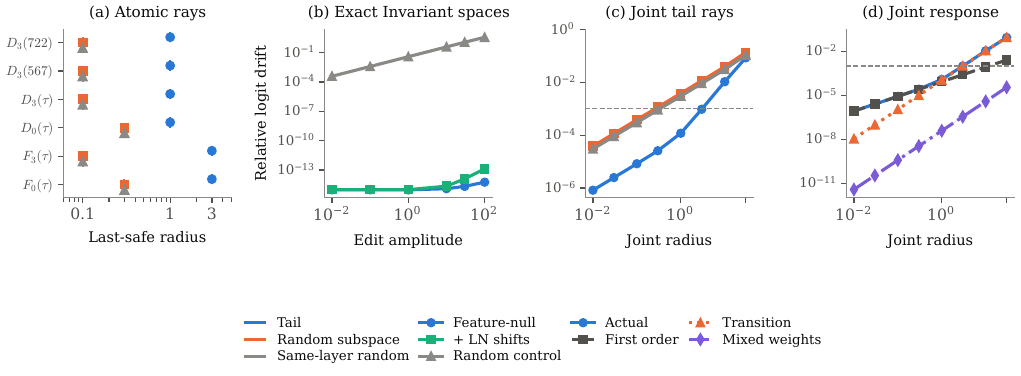}
\caption{ViT joint and relative-cutoff comparisons, following Appendix~\ref{app:methods-common-families}. (a) Last-safe radii for feature ($F$) and downstream ($D$) tails; subscripts identify the two MLP maps. (b) Exact compatible combinations. (c) Joint approximate edits and controls. (d) Their response decomposition. Points and bands show means and ranges of within-seed medians over 256 directions. Horizontal lines mark relative drift $10^{-3}$. Panel (a) uses dataset-RMS drift; (b--d) use median per-image relative drift. Normalizations follow Appendix~\ref{app:methods-radius}.}\label{fig:native-vit-sampled}\label{fig:c-native-approx}
\end{figure}

The compatible exact combinations preserve logits to numerical precision (Figure~\ref{fig:native-vit-sampled}b). The supplementary relative-cutoff study also finds larger sampled radii than controls for single-layer and joint approximate edits (panels a,c). Near their visibility threshold, activation changes dominate the departure from the first-order response, while mixed-weight interactions remain small (panel d).

\subsection{Fine-tuned ResNet on RESISC45}
\label{app:resnet-common}

Every tested feature and common downstream operator has full rank, and all their approximate tails are empty at the stated cutoff. The selective construction gives one substantial exception: a 10,171-dimensional space in \texttt{layer4.2.conv2} (Table~\ref{tab:resnet-common}). These zeros concern the specified families. The full atomic kernels of the convolutional layers remain large by parameter counting, and have not been factorized in this study.

\begin{table}[!htbp]\centering
\caption{ResNet late-layer constructions. FN: feature-null; DN: additional common downstream-null; SEL: additional selective. Both marginal tails are empty in every row. The last column is only the counting lower bound $\max(0,p-45s)$ on the full atomic kernel, with $s=10,000$.}
\label{tab:resnet-common}
\small\begin{tabular}{lrrrrr}\toprule
Layer & Parameters & FN & DN & SEL & Kernel lower bound\\\midrule
\texttt{layer4.0.conv1} & 524,288 & 0 & 0 & 0 & 74,288\\
\texttt{layer4.0.conv2} & 2,359,296 & 0 & 0 & 0 & 1,909,296\\
\texttt{layer4.0.conv3} & 1,048,576 & 0 & 0 & 0 & 598,576\\
\texttt{layer4.1.conv1} & 1,048,576 & 0 & 0 & 0 & 598,576\\
\texttt{layer4.1.conv2} & 2,359,296 & 0 & 0 & 0 & 1,909,296\\
\texttt{layer4.1.conv3} & 1,048,576 & 0 & 0 & 0 & 598,576\\
\texttt{layer4.2.conv1} & 1,048,576 & 0 & 0 & 0 & 598,576\\
\texttt{layer4.2.conv2} & 2,359,296 & 0 & 0 & 10,171 & 1,909,296\\
\texttt{layer4.2.conv3} & 1,048,576 & 0 & 0 & 0 & 598,576\\
Readout & 92,205 & 0 & 0 & 0 & 0\\\bottomrule
\end{tabular}
\end{table}

The selective family comes from five low-activity channels, none entirely inactive (Table~\ref{tab:resnet-selective}).

\begin{table}[!htbp]\centering
\caption{Selective-space dimension in \texttt{layer4.2.conv2}, from the active-patch matrices. The channel indices follow the implementation's zero-based numbering.}
\label{tab:resnet-selective}
\small\begin{tabular}{rrrr}\toprule
Channel & Active patches & Rank & Nullity\\\midrule
118 & 4,164 & 4,162 & 446\\
250 & 2,335 & 2,335 & 2,273\\
275 & 992 & 992 & 3,616\\
413 & 4,059 & 4,059 & 549\\
465 & 1,321 & 1,321 & 3,287\\\midrule
Total & & & 10,171\\\bottomrule
\end{tabular}
\end{table}

\begin{figure}[!htbp]\centering
\includegraphics[width=3.024in]{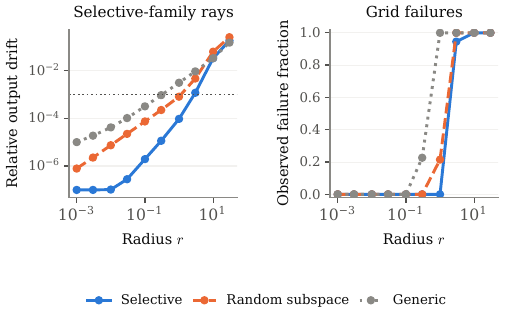}
\caption{ResNet selective-family rays, shown in the same format as the ViT study. Left: median relative RMS drift over 256 directions. Right: fraction exceeding $10^{-3}$ at any tested radius so far. Seed 0; no seed-variation band.}
\label{fig:resnet-common-rays}\label{tab:resnet-radii}
\end{figure}

The selective space preserves outputs over much larger sampled edits than generic directions (Figure~\ref{fig:resnet-common-rays}). All 256 selective probes remain below relative drift $10^{-3}$ through radius 1, while all generic probes have crossed by that radius. The sampled space contains 10,170 of the 10,171 selective directions. These measurements follow the common ray protocol and do not establish a uniform radius.

\subsection{Scaling width and depth}
\label{app:scaling}\label{app:cnn-width}
The full atomic-kernel measurements on the common 10,000-image set are summarized in Figure~\ref{fig:scaling-dimensions} and Table~\ref{tab:atomic-scaling-full}. Wider networks have larger kernel fractions after 500 epochs; at matched high accuracy, the width trend is not monotonic. Increasing depth concentrates the largest fractions toward the final hidden layers. All plotted dimensions use the relative singular-value cutoff $10^{-9}$. Daggers in Table~\ref{tab:atomic-scaling-full} flag full-kernel cases with a gap below ten at this cutoff in at least one seed. Tightening the cutoff to $10^{-12}$ changes any listed per-seed fraction by at most $0.92$ percentage points. At width 512, the full ranks are only partly resolved (Appendix~\ref{app:methods-scaling}). The figure therefore shows selective-space lower bounds on the same 10,000 images as separate square markers, without connecting them to the full-kernel curves. Their epoch-500 means are $96.0\%$ in L2 and $99.8\%$ in L3.

\begin{table}[!htbp]\centering
\caption{Invariant-space dimensions as percentages of layer parameters on 10,000 analysis images: full atomic kernels, except for the selective lower bounds at width 512. Entries give means $\pm$ one sample standard deviation over three seeds at the relative singular-value cutoff $10^{-9}$. Daggers mark a full-kernel gap below ten at this cutoff in at least one seed. A dash means that the layer or milestone does not exist.}
\label{tab:atomic-scaling-full}
\small\setlength{\tabcolsep}{5pt}
\begin{tabular}{llrrrr}\toprule
Architecture & Stage & L2 & L3 & L4 & L5\\\midrule
Width 32 & High accuracy & --- & --- & --- & ---\\
 & Epoch 500 & $6.7\pm3.5$ & $27.0\pm10.0$ & --- & ---\\
Width 64 & High accuracy & $13.0\pm10.7$ & $59.5\pm11.2$ & --- & ---\\
 & Epoch 500 & $22.1\pm14.6$ & $69.5\pm2.5$ & --- & ---\\
Width 128 & High accuracy & $21.5\pm4.8$ & $47.5\pm4.4$ & --- & ---\\
 & Epoch 500 & $62.7\pm4.2$ & $90.9\pm1.9^{\dagger}$ & --- & ---\\
Width 256 & High accuracy & $17.6\pm6.1$ & $58.6\pm8.5^{\dagger}$ & --- & ---\\
 & Epoch 500 & $87.9\pm3.1$ & $98.7\pm0.0$ & --- & ---\\
Width 512 (selective) & High accuracy & $31.8\pm1.1$ & $62.4\pm0.9$ & --- & ---\\
 & Epoch 500 & $96.0\pm0.3$ & $99.8\pm0.0$ & --- & ---\\
Depth 2 & High accuracy & $75.8\pm2.9$ & --- & --- & ---\\
 & Epoch 500 & $79.9\pm1.5$ & --- & --- & ---\\
Depth 4 & High accuracy & $5.8\pm1.1$ & $28.6\pm3.1$ & $45.8\pm7.8$ & ---\\
 & Epoch 500 & $14.3\pm3.1$ & $73.4\pm2.2^{\dagger}$ & $95.2\pm0.4$ & ---\\
Depth 5 & High accuracy & $0.9\pm0.9$ & $11.4\pm2.1$ & $43.9\pm11.5$ & $57.6\pm10.0^{\dagger}$\\
 & Epoch 500 & $0.0\pm0.0$ & $37.6\pm33.0^{\dagger}$ & $91.1\pm1.5^{\dagger}$ & $96.9\pm0.7$\\
\bottomrule\end{tabular}
\end{table}

\begin{figure}[!htbp]\centering
\includegraphics[width=4.588in]{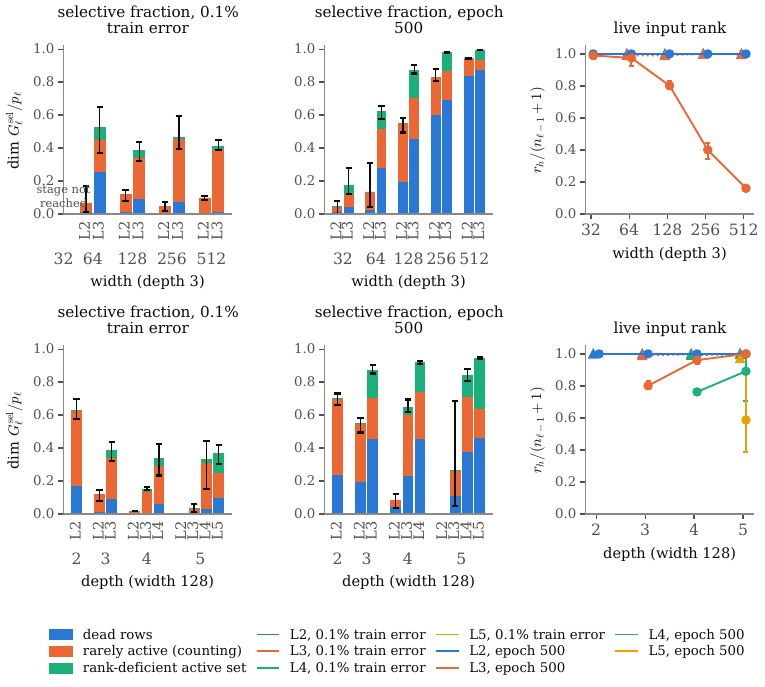}
\caption{Selective-space scaling on all 60,000 training images, complementing the full atomic-kernel measurements on 10,000 images in Figure~\ref{fig:scaling-dimensions}. Rows vary width and depth; columns show selective fractions at high accuracy and epoch 500, then normalized input-feature ranks. Bars group hidden layers in increasing order from L2. Segments separate dead rows, insufficient active-sample counts, and rank-deficient active-feature spans beyond counting. Points and whiskers show seed means and ranges; $r_h$ denotes the input-feature rank $r_\ell$.}
\label{fig:scaling-detail}
\end{figure}

The selective-space sweep additionally covers all 60,000 training images, including width 512 (Figure~\ref{fig:scaling-detail}). It provides constructive lower bounds on the full atomic kernels on that larger set. At epoch 500, its mean L2 dimension rises from $55\%$ of layer parameters at width 128 to $94\%$ at width 512.

The CNN sweep shows the same broad pattern: increasing width enlarges the selective ocean faster than the parameter count, even when test accuracy changes little. At half width, the selective construction leaves only readout directions. Wider models develop late convolutional channels that are active on very few images, leaving many kernel edits unconstrained. The channel-activity plot connects that observation to the measured selective dimensions. Matrix-free counts and fully assembled invariant subspace certificates are distinguished in the table.

\begin{table}[!htbp]
\caption{CNN width sweep at epoch 500. The final column is the mean selective dimension divided by parameter count. Starred cells are matrix-free selective-space dimensions without a full assembled invariant subspace certificate; the remaining cells passed numerical validation.}
\label{tab:cnn-width}
\centering
\vspace{0.45em}
\begin{tabular}{lrrrrrr}
\toprule
network & params & test acc. & seed 0 & seed 1 & seed 2 & ocean/params \\
\midrule
$0.5\times$ & $13{,}922$ & $.990$--$.992$ & $1{,}080$ & $430$ & $620$ & $0.051$ \\
base & $45{,}114$ & $.994$--$.995$ & $3{,}803$ & $8{,}782$ & $5{,}565$ & $0.134$ \\
$2\times$ & $159{,}338$ & $.994$--$.995$ & $45{,}493$ & $64{,}416^{*}$ & $45{,}004^{*}$ & $0.324$ \\
\bottomrule
\end{tabular}
\end{table}
\begin{figure}[!htbp]\centering
\includegraphics[width=4.553in]{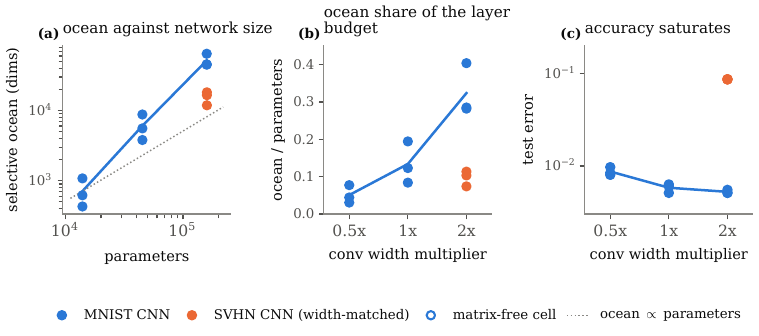}\par\smallskip
\includegraphics[width=3.268in]{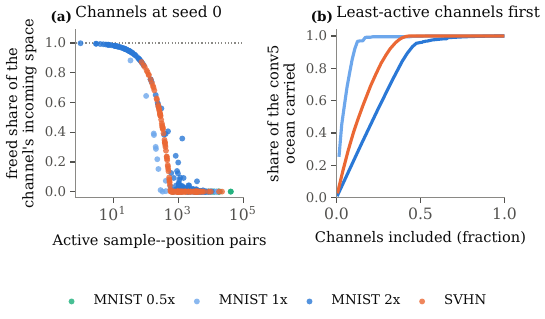}
\caption{CNN width scaling (top) and conv5 channel activity (bottom, seed 0). Selective dimensions grow faster than parameter count while accuracy changes little. Hollow markers denote matrix-free dimensions without assembled certificates; orange points compare the width-matched SVHN CNN. The least-active channels carry most of the conv5 selective space. At half width that space is empty.}\label{fig:early-draft-cnn-width}
\end{figure}

\begin{figure}[!htbp]\centering
\includegraphics[width=1.954in]{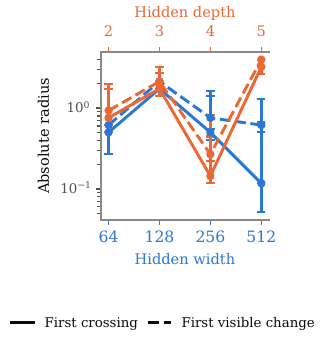}
\caption{Radius versus MLP width (blue, bottom axis) and hidden depth (orange, top axis). Solid lines mark first gate crossings and dashed lines the first output change above $10^{-3}$. Points are medians over sampled slice angles; whiskers span the 10th--90th percentiles. Each architecture uses seed 0 at its tested checkpoint.}\label{fig:radius-sweeps}
\end{figure}

Dimension and radius need not scale together. Figure~\ref{fig:radius-sweeps} compares first gate crossings and first visible output changes on the width/depth slices. Neither radius follows a simple monotone law in these seed-0 measurements. The gap between crossing a boundary and changing outputs remains important when interpreting a larger ocean dimension.

\subsection{Inside the co-occurrence deficit}
\label{app:mechanism}

The deficit measures constraints missing from the observed feature--downstream pairings. We first examine the contribution of selective units, whose exact inactive-side construction uses ReLU's zero slope, then ask how the deficit relates to sample clustering and neural collapse. The grouping and shuffle protocols are defined in Appendix~\ref{app:methods-branches}. Pairing-shuffle results use ReLU seed 0; the clustering comparisons also include leaky ReLU and GELU.

\subsubsection{Selective units and cross-unit cancellations}
Selective units create invariant directions when their active input features leave incoming parameter directions unconstrained. Their contribution can therefore be large even without class-aligned clusters. The full width and depth trajectories and the within-unit decomposition appear in Appendix~\ref{app:atomic}; Figure~\ref{fig:early-draft-mechanism-family} compares their relation to the atomic kernels across model families. In fitted ReLU networks, selective directions account for much of the hidden kernel. Leaky ReLU gives a useful contrast: its nonzero negative-side slopes require cancellations across units.

\begin{figure}[!htbp]\centering
\includegraphics[width=4.558in]{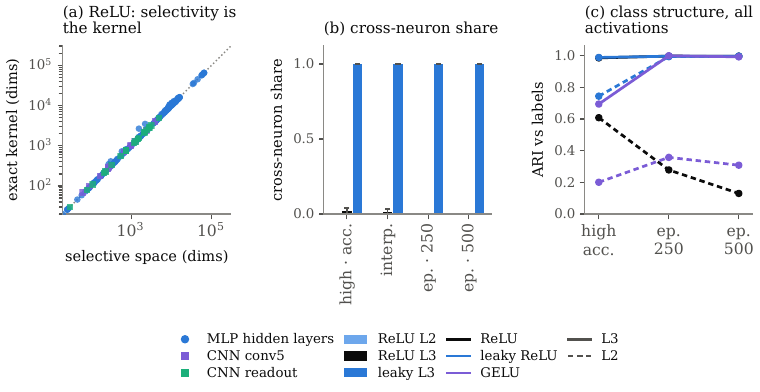}
\caption{Selective and cross-neuron mechanisms across architectures and activations. (a) Selective dimensions against numerical atomic dimensions across measured ReLU widths, depths and CNN layers. (b) Cross-neuron fractions, median and range across seeds. (c) Agreement between clusters of similar downstream maps and the digit labels, measured by the adjusted Rand index: 1 means identical partitions, and 0 is the chance baseline. This panel measures class agreement, not kernel size or the fraction of deficit explained.}\label{fig:early-draft-mechanism-family}
\end{figure}

\begin{table}[!htbp]\centering
\caption{Leaky-ReLU mechanism check, width 128, seed 0 at high accuracy. Selective candidates use the hard activity mask; the additional condition $V_{\mathrm{lin}}^0 A_L\bar h_{L-1,i}^0=0$ for every sample accounts for the nonzero negative-side response. The retained candidate count is compared with the measured numerical kernel.}
\small\begin{tabular}{lrrr}\toprule
Layer & Selective candidates & After readout condition & Atomic kernel\\\midrule
L3 & 6,311 & 5,021 & 5,036\\\bottomrule
\end{tabular}\end{table}

\subsubsection{Sample clustering and missing pairings}
For a partition $\mathcal G$, let $H_G$ be the span of the augmented features in group $G$ and $C_G$ the span of the downstream-map rows in that group. Let $r_h$ and $r_d$ be their respective ranks over all samples, and define $\mathcal P(\mathcal G)=\sum_{G\in\mathcal G}H_G\otimes C_G$. The deficit separates as
\[
 \delta=r_hr_d-\operatorname{rank}J_\ell
 =\underbrace{r_hr_d-\dim\mathcal P(\mathcal G)}_{\text{missing across groups}}
 +\underbrace{\dim\mathcal P(\mathcal G)-\operatorname{rank}J_\ell}_{\text{missing within groups}}.
\]
For a positive deficit $\delta>0$, the explained share is the first term divided by $\delta$. It quantifies how much of the deficit follows from separating samples into these groups. Purity instead measures whether the groups agree with class labels. A class-pure partition can therefore explain little of the rank deficit. Appendix~\ref{app:methods-branches} defines purity and its matched random-label control.

\begin{figure}[!htbp]\centering
\includegraphics[width=3.013in]{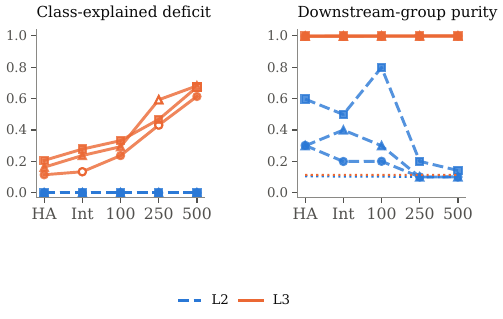}
\caption{Class organization in the width-128 ReLU MLP across five fitted stages. Left: share of the deficit explained by class groups. Right: label purity of ten requested downstream k-means groups; dotted lines give matched random-label purity. Markers distinguish seeds 0--2. Open markers flag tolerance-sensitive partition ranks. HA and Int denote high accuracy and interpolation.}\label{fig:cluster-stages}
\end{figure}

In the three-hidden-layer, width-128 MNIST ReLU MLP, class groups explain an increasing share of the L3 deficit during training, while their L2 share remains zero. Figure~\ref{fig:cluster-stages}  shows the complete five-stage trajectories for all three seeds. This is a comparison between layers of one architecture; the corresponding clustering measurements across network depths have not yet been completed.
\FloatBarrier

\begin{figure}[!htbp]\centering
\includegraphics[width=6.098in]{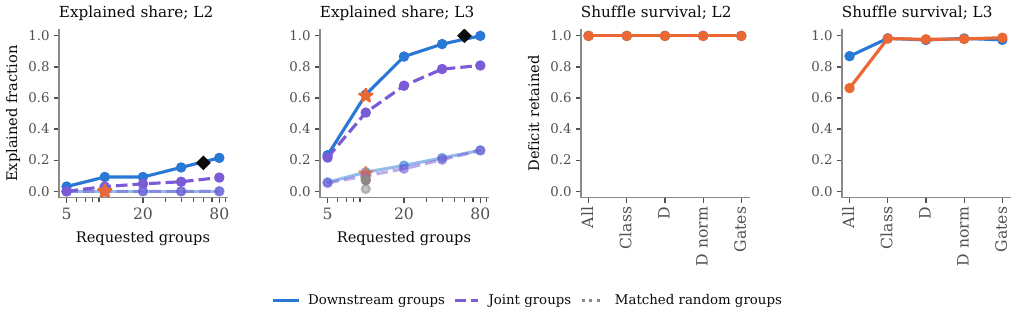}
\caption{Grouping and pairing tests in the width-128 ReLU MLP, seed 0. Left pair: deficit explained as the requested group count changes; light and dark curves denote high accuracy and epoch 500. Stars mark class groups and diamonds suffix-gate patterns, at their actual group count. Right pair: deficit remaining after global shuffles (All) or shuffles within classes, downstream groups (D), normalized downstream groups (D norm), or suffix-gate groups; blue is high accuracy and orange epoch 500. The L2 deficit survives almost entirely; L3 is more sensitive to global shuffling than to shuffling within groups.}\label{fig:mechanism}
\end{figure}

Figure~\ref{fig:mechanism} shows grouping and shuffle tests of the ReLU co-occurrence deficit. Finer downstream groups can explain all of the ReLU L3 endpoint deficit, while matched random groups explain little. A global shuffle retains about two thirds of that deficit, whereas a shuffle restricted to classes retains almost all of it. In L2, both shuffles retain essentially the full deficit. 

\begin{figure}[!htbp]
\centering
\includegraphics[width=3.146in]{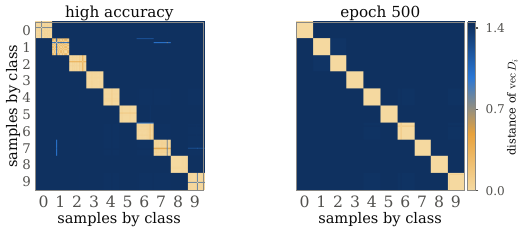}
\caption{Pairwise distances between unit-normalized L3 downstream maps (ReLU width 128, seed 0; 500 samples ordered by class). Sand denotes small distances and navy large distances. The ten class groups become more uniform by epoch 500.}
\label{fig:dmap-pca}
\end{figure}
\FloatBarrier

The distance-threshold analysis groups downstream maps that are close in both direction and magnitude. It uses the common normalization and complete-linkage rule from Appendix~\ref{app:methods-branches}: each group has normalized diameter at most $\varepsilon$. We apply the same threshold grid at each training stage and compare group purity with shuffled-label controls of the same sizes.

Table~\ref{tab:dmap-threshold} shows the L3 results at one common threshold. The late-training groups are nearly class-pure and contain multiple samples; the matched random-label purity is about $0.11$. GELU needs many more groups at high accuracy to satisfy the same distance threshold. Thus the downstream maps become concentrated into a smaller number of class-aligned groups as training proceeds. L2 behaves differently: ReLU and GELU retain overlapping within-class and between-class distance distributions, while leaky ReLU becomes class-separated. These distance measurements describe concentration; the rank decomposition above separately measures its relation to the deficit.

\begin{table}[!htbp]\centering
\caption{L3 downstream groups at normalized diameter threshold $\varepsilon=1.0$, width-128 MNIST MLP, seed 0. Each cell gives number of groups / purity.}
\label{tab:dmap-threshold}
\begin{tabular}{lccc}\toprule
Activation & High accuracy & Epoch 250 & Epoch 500\\\midrule
ReLU & $24/0.997$ & $11/0.998$ & $11/0.999$\\
Leaky ReLU & $24/0.996$ & $10/0.998$ & $10/0.998$\\
GELU & $114/0.998$ & $10/1.000$ & $15/1.000$\\\bottomrule
\end{tabular}
\end{table}

\begin{figure}[!htbp]\centering
\includegraphics[width=6.138in]{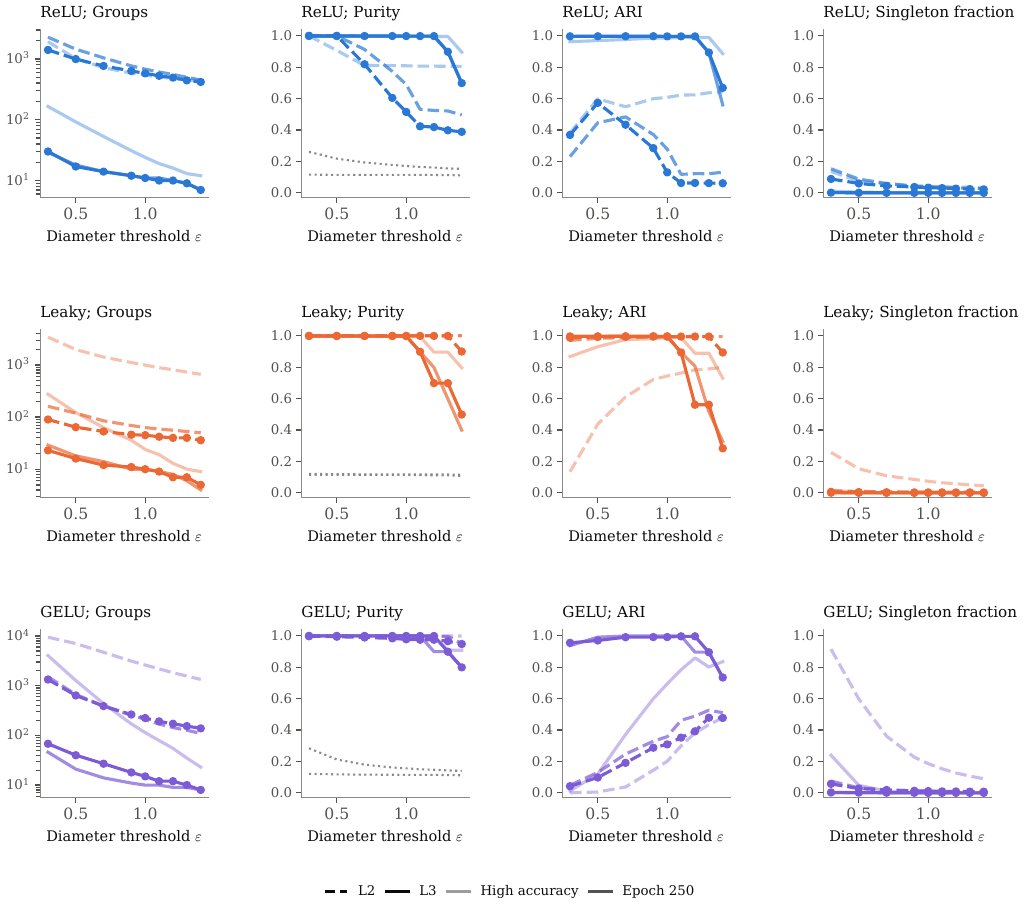}
\caption{Diameter-controlled clustering in the same width-128 networks, seed 0. Rows compare activations; columns show group count, label purity, adjusted Rand index and the fraction of samples in singleton groups. Solid and dashed curves denote L3 and L2. Opacity increases from high accuracy to epoch 250 and epoch 500; endpoint curves have markers. Dotted purity controls shuffle labels at fixed group sizes. L3 forms class-aligned groups for every activation, whereas L2 behavior depends on activation. High purity carried by singletons is distinguished from grouping whole classes.}\label{fig:cluster-diameters}
\end{figure}
\FloatBarrier

Figure~\ref{fig:cluster-diameters} compares both layers at every measured stage. Unlike a fixed ten-group partition, this rule allows more groups when the maps require them. It also reports the fraction of samples in singleton groups, since high purity from isolated samples would provide little evidence of class organization. The adjusted Rand index compares the group and class assignments through sample pairs, correcting for chance agreement: one indicates matching partitions and values near zero indicate chance-level agreement.

\begin{figure}[!htbp]\centering
\includegraphics[width=6.120in]{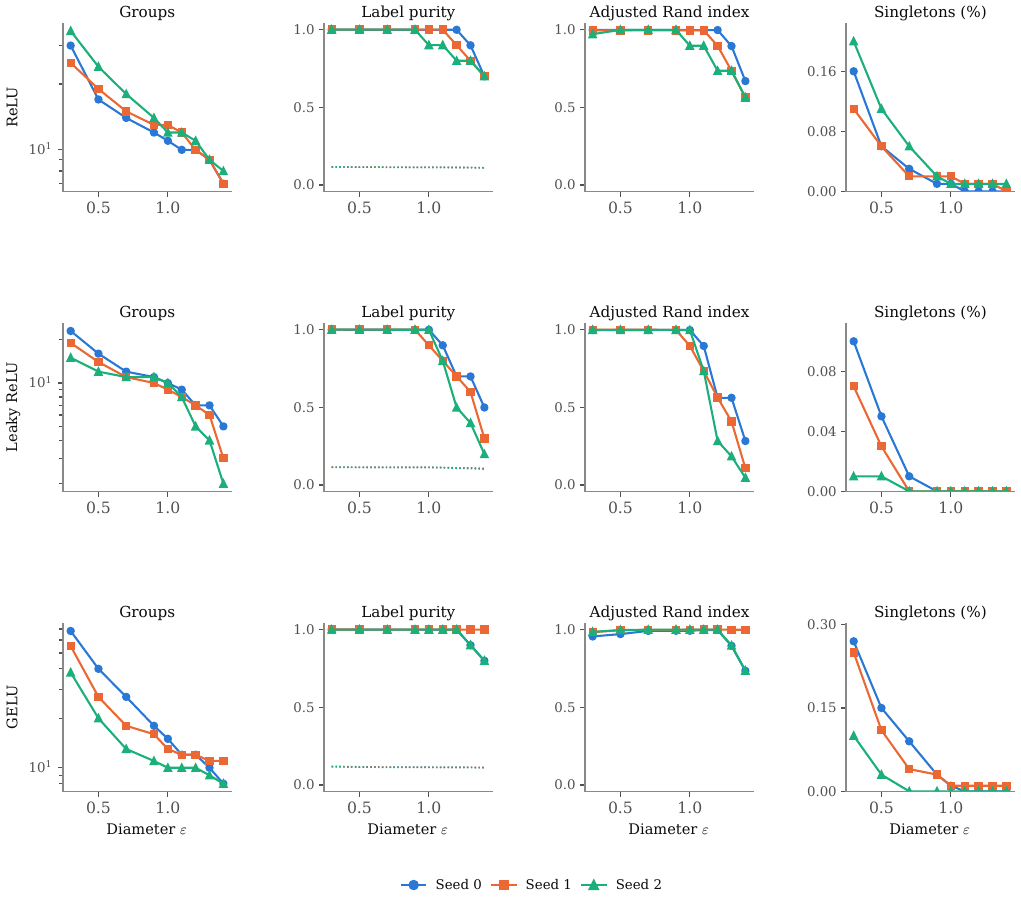}
\caption{Diameter-controlled downstream clustering across seeds 0--2 in width-128 MNIST MLPs, L3 at epoch 500. Rows compare activations; columns show group count, label purity, adjusted Rand index and the percentage of samples in singleton groups. Colors and markers identify seeds; dotted purity curves give matched-size shuffled-label means. All seeds use the same analysis size, distance normalization and diameter grid.}\label{fig:cluster-seeds}
\end{figure}

Figure~\ref{fig:cluster-seeds} extends the L3 endpoint comparison to all three training seeds using the same diameter grid and normalization. The additional seeds cover epoch 500, while the stage and layer comparisons above retain their original coverage.
\FloatBarrier

\subsubsection{Neural collapse and class organization}
\label{app:nc}

Neural collapse describes a collection of geometric patterns that can emerge in a classifier's final features during training: within-class features concentrate, class means approach a regular simplex, classifier weights align with those means, and classification approaches the nearest-class-mean rule \citep{papyan2020collapse}. We use measurements of these patterns to ask whether class organization of the features accompanies class organization of the downstream maps.

\begin{figure}[!htbp]
\centering
\includegraphics[width=6.022in]{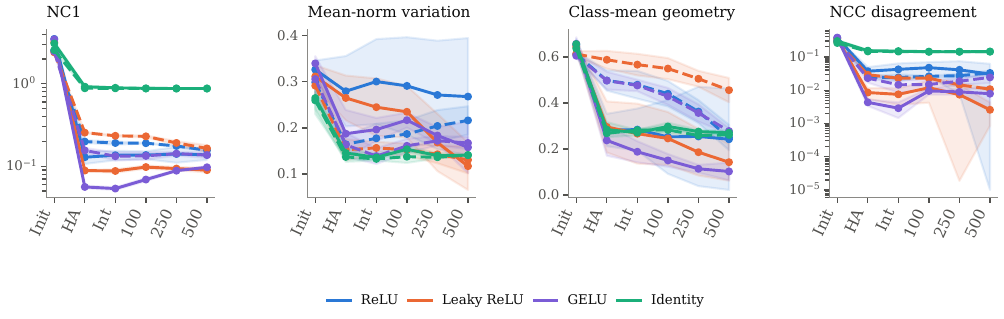}
\caption{Neural-collapse measurements across training, seed means with one sample standard deviation shaded. Solid lines show L3; dashed lines show L2. Bands summarize up to three seeds at each measured stage and use a positive plotting floor of $10^{-5}$ on logarithmic axes.}
\label{fig:nc-full}
\end{figure}

Figure~\ref{fig:nc-full} follows the measured width-128 MNIST ReLU, leaky-ReLU and GELU models, with an identity-activation control. NC1 compares within-class feature variance with between-class variance. The two NC2 measurements quantify variation in class-mean norms and departure of their normalized Gram matrix from an equiangular tight frame. The plotted NC4 diagnostic is the disagreement between nearest-class-mean predictions and labels. All three nonlinear families develop concentrated features during training, although the strength and subsequent evolution depend on activation and layer. The identity control does not show comparable concentration, and these measurements do not establish every neural-collapse property in every architecture.

\begin{figure}[!htbp]\centering
\includegraphics[width=6.147in]{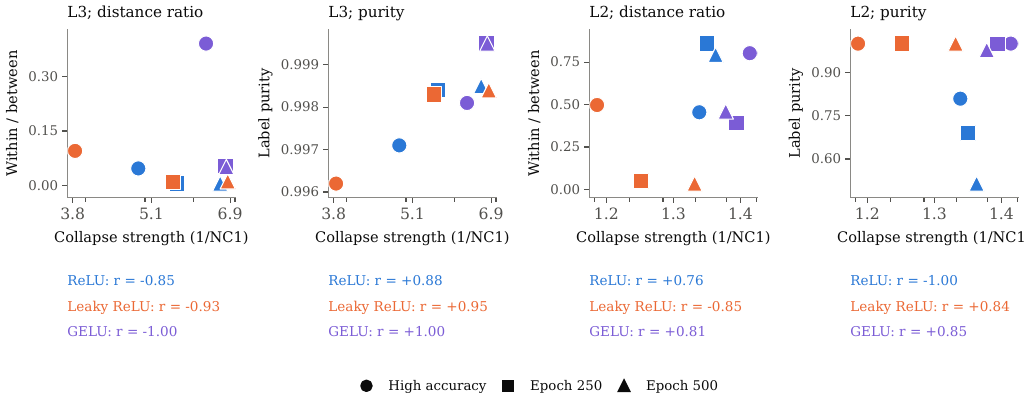}
\caption{Feature concentration versus downstream clustering in width-128 MNIST MLPs, seed 0. Larger $1/\mathrm{NC1}$ means stronger input-feature collapse; the horizontal scale is logarithmic. Points compare within-class/between-class downstream-map distance or diameter-cluster label purity at $\varepsilon=1$, at three matched checkpoints. Left pair: L3 maps and L2 features; right pair: L2 maps and L1 features. Colors identify activations and marker shapes identify stages. Each panel reports the within-activation Pearson correlation with $\log_{10}(1/\mathrm{NC1})$, using only these three points; a constant response has undefined correlation. These descriptive correlations do not establish a causal relationship.}\label{fig:nc-pairings}
\end{figure}

Feature concentration and downstream-map organization can develop together during training. Figure~\ref{fig:nc-pairings} compares inverse NC1 with downstream class separation and label purity at matching checkpoints, separately for L2 and L3. Larger inverse NC1 means stronger feature concentration. The comparisons describe co-occurrence, with variation across activations and layers. They do not establish that either phenomenon causes the other or causes the co-occurrence deficit: downstream maps also depend on later weights and activation slopes. These observations motivate studying the two forms of organization together.

\FloatBarrier

\subsection{Dependence on training and evaluation data}
\label{app:data-dependence}
Data affect ocean geometry in two distinct ways. Changing the training set changes the learned weights and representations. Changing the evaluation set holds the network fixed and changes the constraints defining its invariant directions. We study these effects separately below.

\subsubsection{Training data}
\label{app:training-data}
Training on more images changes the learned representation as well as the number of fitting constraints. We compare models trained from initialization on progressively larger subsets with a model trained on the complete training set, evaluating both on the same smaller subsets. ``Train on $n$'' in the figures denotes the former; ``Full-data, $n$ constraints'' denotes the latter. The architecture and optimization recipe are held fixed, and geometry probes leave the input layer unchanged.

At the smallest matched evaluation set, the model trained on all images has substantially larger hidden atomic kernels than the model trained only on that subset. Training on more data can therefore create a representation with more invariant directions, even though adding evaluation constraints to a fixed representation removes directions. The reduced-data training sweep is not monotone in sample size, so this comparison does not imply a universal growth law.

\begin{figure}[!htbp]\centering
\includegraphics[width=6.156in]{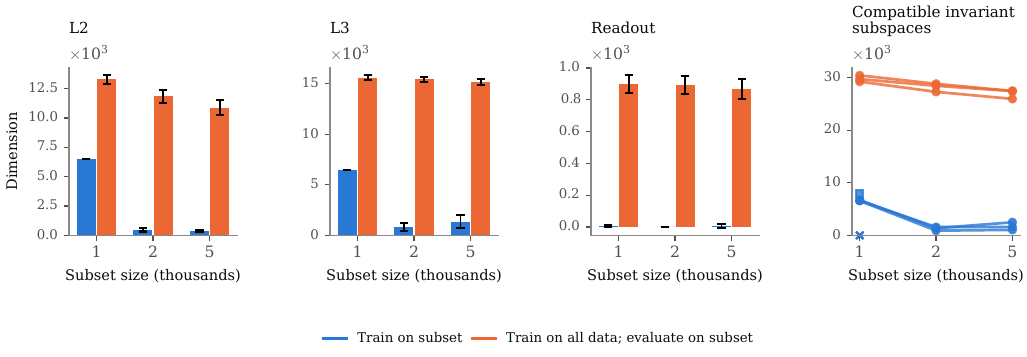}
\caption{Dimensions at epoch 500 versus subset size, width-128 ReLU MLPs. Colors distinguish subset-trained and full-data networks throughout. First three panels: total atomic dimensions, with seed means and sample-standard-deviation whiskers. Fourth: compatible invariant subspace dimensions, with one curve per seed; solid/circles denote bottom-up construction, dashed/squares top-down, and dotted/crosses the atomic sum. Zero means no validated direction, including a failed certificate. Unmeasured constructions have no point. L1 is fixed.}\label{fig:data-atomic}\label{fig:data-branches}
\end{figure}
\FloatBarrier

The small-subset models show why single-layer invariant directions must be checked before combining them. With few samples, the number of output constraints alone can force large hidden atomic kernels, even when the feature and downstream spans have full rank. Each atomic edit preserves the frozen outputs when applied alone. Simultaneous edits can change those outputs through interactions between layers. The rightmost panel of Figure~\ref{fig:data-atomic} compares compatible dimensions across sample counts and construction orders.

Table~\ref{tab:subset-order} gives a concrete example at $N=1{,}000$. The two hidden atomic kernels each have 6,512 directions, yet their full sum fails the invariant subspace check. Bottom-up construction keeps the L2 space and restricts L3; top-down construction does the reverse. Their different totals show that layer order affects how many simultaneously invariant directions the construction retains.
\begin{table}[!htbp]\centering
\caption{Compatibility in the 1,000-sample retraining, seed 0. The input layer is fixed. The atomic sum is a tangent space but fails the invariant subspace check.}\label{tab:subset-order}
\begin{tabular}{lrrrr}\toprule
Construction & L2 & L3 & readout & total\\\midrule
Atomic sum &6512&6512&10&13034\\
Bottom-up &6512&197&10&6719\\
Top-down &1477&6512&10&7999\\\bottomrule
\end{tabular}\end{table}

The radius measurements separate changes in the first gate crossing from changes in output visibility. In the subset-trained models, the first crossing moves outward as training-set size grows, while the visible radius changes much less. Excluded-image responses are already nonzero in the frozen model and decrease as more images constrain the directions. The summary and absolute-radius curves below compare both training conditions.

\begin{figure}[!htbp]\centering
\includegraphics[width=5.057in]{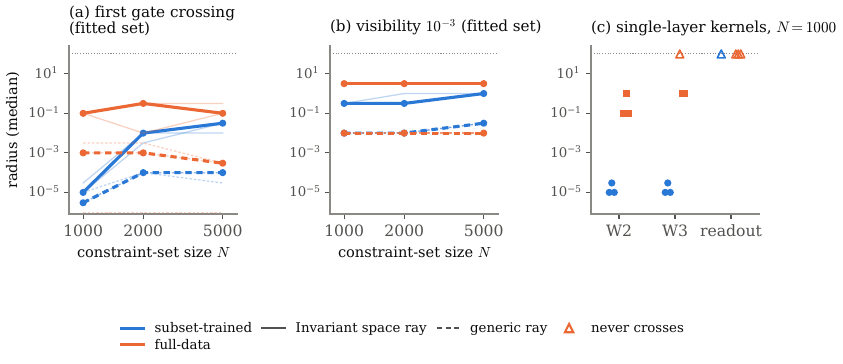}\par\smallskip
\includegraphics[width=4.780in]{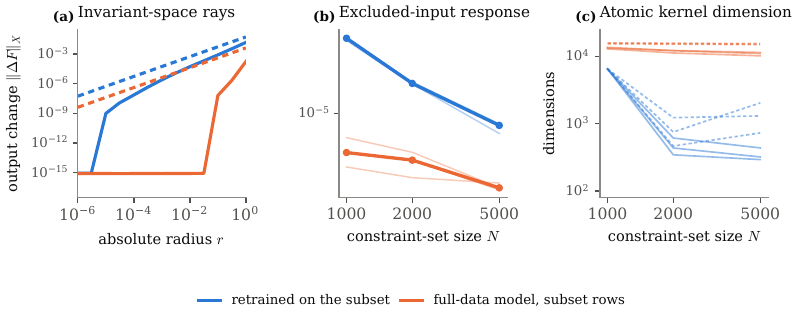}
\caption{Reduced-sample geometry across three seeds, with the input layer fixed. Top: fitted-set first crossing, $10^{-3}$ relative output visibility, and single-layer probes at $N=1{,}000$; markers summarize 128 directions and triangles mark censoring. Bottom: fitted versus excluded response, excluded response at matched radius, and hidden atomic dimensions. Retraining and analysis-set restriction are kept separate. Cap-limited input-layer measurements give $\rank J_1\ge16{,}512$ and $\dim K_1\le16{,}384$; they are excluded from the probes.}\label{fig:early-draft-subset-rays}
\end{figure}

The upper-middle panel of Figure~\ref{fig:early-draft-subset-rays} shows the first sampled radius at which the output change exceeds $10^{-3}$ times the base output norm. Several seed curves coincide because the scan uses a discrete radius grid; equal plotted values mean the threshold was reached at the same tested radius. The lower-left panel follows invariant subspace directions constructed on 1,000 images. Outputs on those fitted images remain near numerical zero over the initial part of the scan, while excluded-image outputs change immediately. This separation shows which invariance depends on the fitted sample set. Colors distinguish subset training from a full-data network throughout the figure. In the upper row, solid and dashed lines denote invariant subspace and generic rays; in the lower-left panel they denote fitted and excluded images.

\begin{figure}[!htbp]\centering
\includegraphics[width=6.313in]{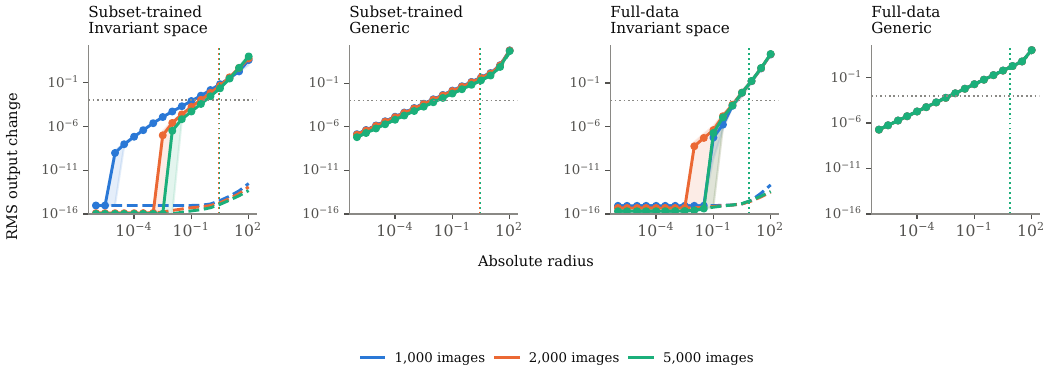}
\caption{Absolute-radius invariant subspace and generic probes at epoch 500, for subset-trained networks (left pair) and full-data networks evaluated on the same subsets (right pair). Colors denote subset size. Solid/dashed lines show actual/frozen response; means and ranges summarize three seed medians. L1 is fixed. Vertical dotted lines mark one tenth of the network norm; the horizontal line marks output change $10^{-3}$.}
\end{figure}
\FloatBarrier

\subsubsection{Evaluation data at a fixed network}
\label{app:newbatch}

How much of an ocean survives when the same trained network is evaluated on more images? We hold its parameters fixed and add either a disjoint training batch or test images to the original analysis set. Every new image adds constraints, so the original kernel can only shrink. The question is how much of its invariance was specific to the original images.

\begin{figure}[!htbp]\centering
\includegraphics[width=1.776in]{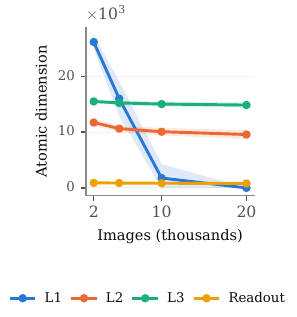}
\caption{Atomic dimensions as more evaluation images constrain the same width-128 MNIST ReLU network at epoch 500. Lines show means over seeds 0 and 1; bands show one sample standard deviation. Every layer has measurements on the same evaluation-size grid.}\label{fig:newbatch}
\end{figure}

The hidden-layer kernels largely survive the additional constraints, including when the analysis set is doubled. The full-training-set selective measurements in Appendix~\ref{app:atomic} also remain substantial. The input layer behaves differently: the measured reference-model kernel disappears on the larger evaluation set. Thus the persistence of hidden directions is not a consequence of every layer being insensitive to new images. Figure~\ref{fig:newbatch} shows the dimension trajectories at epoch 500.

Table~\ref{tab:newdata-main} separates additional training images from unseen test images at three checkpoints. Each model was trained on all 60,000 MNIST training images; the initial analysis set contains 10,000 of them. Both additions remove a smaller fraction of the hidden kernels at epoch 500 than at the earlier fitted stages.
\begin{table}[!htbp]
\centering
\caption{Atomic-kernel dimensions removed by adding 10,000 images to the 10,000-image analysis set. Width-128 ReLU MLP; percentages of the original kernel, with ranges over three seeds. The network is fixed within each comparison. \emph{Training} means additional images seen during training but absent from the initial analysis set; \emph{test} means unseen images.}
\label{tab:newdata-main}
\small\setlength{\tabcolsep}{4pt}
\begin{tabular}{llrr}
\toprule
Stage & Added images & L2 & L3\\
\midrule
High accuracy & Training & $16$--$32\%$ & $4.7$--$7.7\%$\\
 & Test & $16$--$30\%$ & $6.2$--$11.4\%$\\
Interpolation & Training & $18$--$27\%$ & $3.9$--$6.3\%$\\
 & Test & $16$--$26\%$ & $5.3$--$10.0\%$\\
Epoch 500 & Training & $4.0$--$6.2\%$ & $1.0$--$1.4\%$\\
 & Test & $3.7$--$5.6\%$ & $2.1$--$2.9\%$\\
\bottomrule
\end{tabular}
\end{table}

\begin{figure}[!htbp]\centering
\includegraphics[width=1.499in]{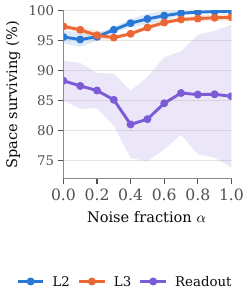}
\caption{Atomic-space survival under clean-to-noise interpolation, mean $\pm$ sample SD across three seeds. L3 has additional rank-cutoff sensitivity (Appendix~\ref{app:methods-newinput}), which preserves the qualitative trend.}
\label{fig:newinput-shift}
\end{figure}

\paragraph{From clean inputs to pure noise.}
\label{app:newinput-interpolation}
We measure how much of each atomic space survives when clean MNIST inputs are gradually replaced by uniform noise, following Appendix~\ref{app:methods-newinput}. Most hidden-layer directions survive throughout. Intermediate mixtures constrain the spaces more than pure noise, while the readout loses a larger and more seed-dependent fraction (Figure~\ref{fig:newinput-shift}). Thus moving farther from the training distribution need not remove more invariant directions.

\paragraph{SVHN inputs.}
Most of the hidden-layer spaces also survive SVHN images passed through the fixed MNIST preprocessing (Table~\ref{tab:newinput-svhn}). L2 survival is higher than for the clean MNIST reference in every seed; L3 and readout effects vary across seeds. Survival therefore depends on the additional feature and downstream constraints, rather than distribution shift alone.
\begin{table}[!htbp]\centering
\caption{Atomic-space survival (\%), mean $\pm$ sample SD across three seeds. SVHN uses the fixed MNIST representation. Rank sensitivity is assessed separately from seed variation.}
\label{tab:newinput-svhn}
\small\begin{tabular}{lccc}\toprule
Added inputs & L2 & L3 & Readout\\\midrule
Clean MNIST & $95.63 \pm 1.07$ & $97.39 \pm 0.42$ & $88.33 \pm 3.30$\\
SVHN & $98.41 \pm 0.91$ & $97.20 \pm 0.58$ & $87.30 \pm 3.04$\\\bottomrule\end{tabular}\end{table}

\FloatBarrier

On smaller analysis subsets, directions invariant on the fitted images can already change excluded-image outputs before any gate switches. The fitted/excluded response comparisons in Figure~\ref{fig:early-draft-subset-rays} make this dependence on the evaluation set explicit. A frozen certificate on one set of images need not extend to another.

\subsection{Kernel persistence}
\label{app:persistence}\label{app:persistence-newdata}
The ReLU ocean grows through training while retaining most of its earlier invariant directions. This is stronger than observing a large kernel at each checkpoint: the spaces remain substantially aligned as learning changes the weights. Removing directions supported on dead-unit coordinates leaves much of the same persistence, so the effect extends to directions involving active units.

\begin{figure}[!htbp]\centering
\includegraphics[width=4.560in]{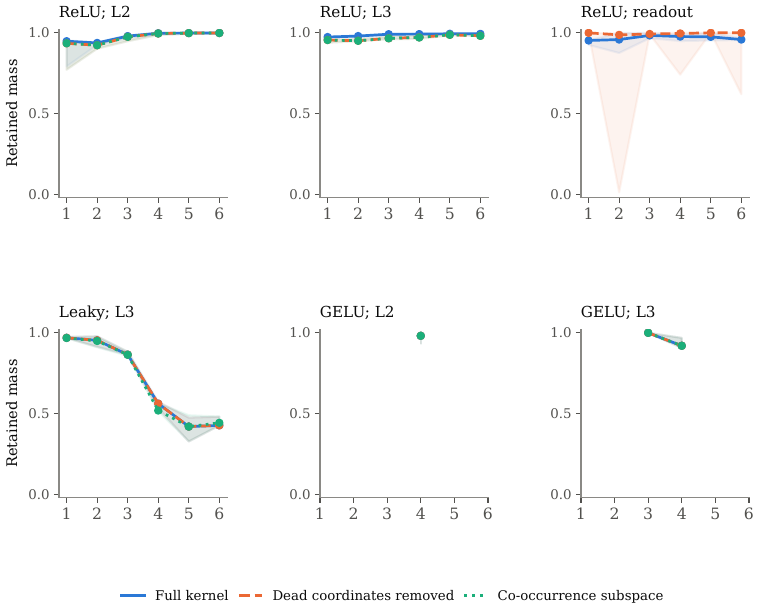}
\caption{Checkpoint persistence in width-128 MNIST MLPs: retained mass. Pair indices 1--6 denote HA$\to$Int, Int$\to$100, 100$\to$250, 250$\to$500, HA$\to$500 and Int$\to$500; HA and Int are high accuracy and interpolation. Curves show medians and ranges across three seeds. Empty spaces have no retention value; unresolved component splits are omitted. GELU kernels remain cutoff-dependent.}\label{fig:persistence}
\end{figure}

\begin{figure}[!htbp]\centering
\includegraphics[width=4.583in]{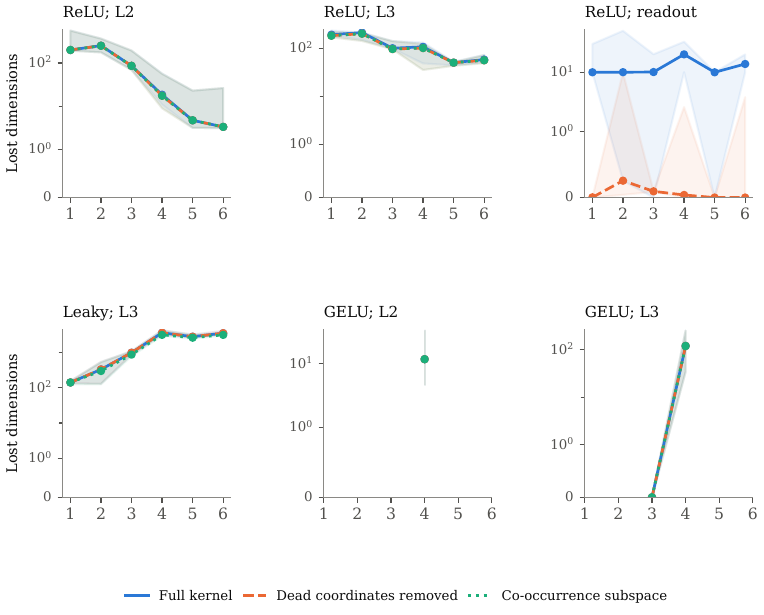}
\caption{Effective lost dimension, with the checkpoint pairs and conventions of Figure~\ref{fig:persistence}. The vertical scale is linear near zero and logarithmic above it.}\label{fig:persistence-lost}
\end{figure}

\begin{figure}[!htbp]\centering
\includegraphics[width=4.620in]{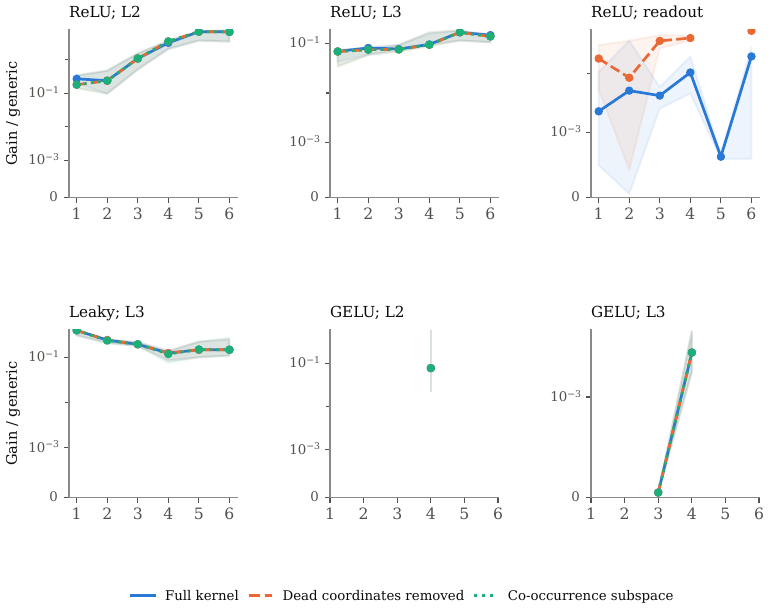}
\caption{Output gain of lost components relative to generic directions, with the conventions of Figure~\ref{fig:persistence}.}\label{fig:persistence-gain}
\end{figure}

\begin{figure}[!htbp]\centering
\includegraphics[width=4.710in]{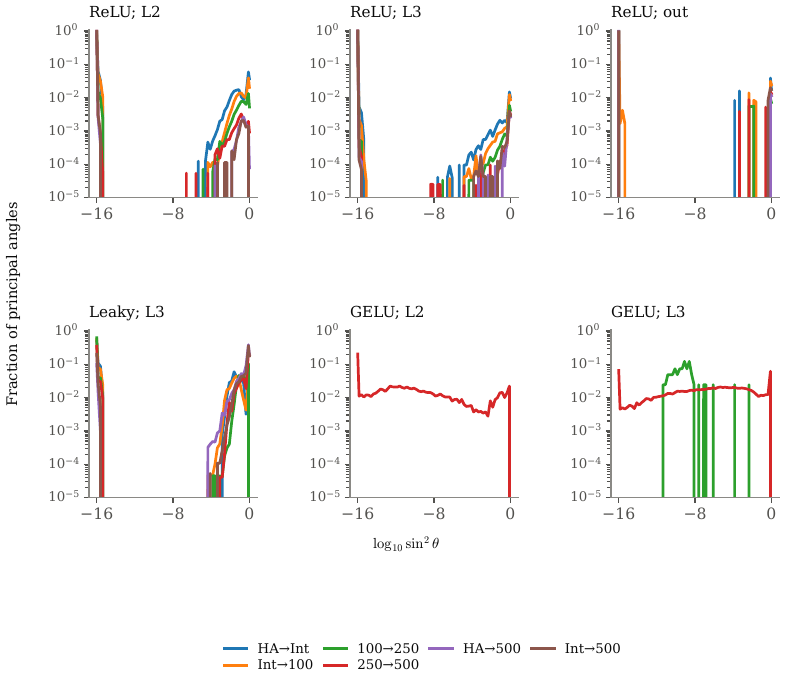}
\caption{Principal-angle distributions for the same checkpoint pairs, normalized within each seed and then averaged. ReLU retains a large aligned component, while leaky ReLU rotates more substantially late in training.}\label{fig:persistence-angles}
\end{figure}

Figures~\ref{fig:persistence}--\ref{fig:persistence-angles} compare all measured fitted-stage pairs in the reference ReLU, leaky-ReLU and GELU models. They separate retained mass, effective lost dimension and output sensitivity of the part that becomes visible. The principal-angle distributions show that ReLU's geometric loss is concentrated in a small part of the space. Many lost directions also have weak downstream output sensitivity, although a few late L2 comparisons have larger gain than generic directions. Geometric retention and functional exposure are therefore distinct measurements.

The activation comparisons qualify the persistence claim. Leaky-ReLU kernels rotate more substantially late in training, whereas GELU's late numerical kernels retain substantial projected mass but lack a stable exact-rank interpretation. These results show that strong ReLU persistence is a property of the learned geometry, not a consequence of the definition of a kernel.

\FloatBarrier

\begin{figure}[!htbp]\centering
\includegraphics[width=6.096in]{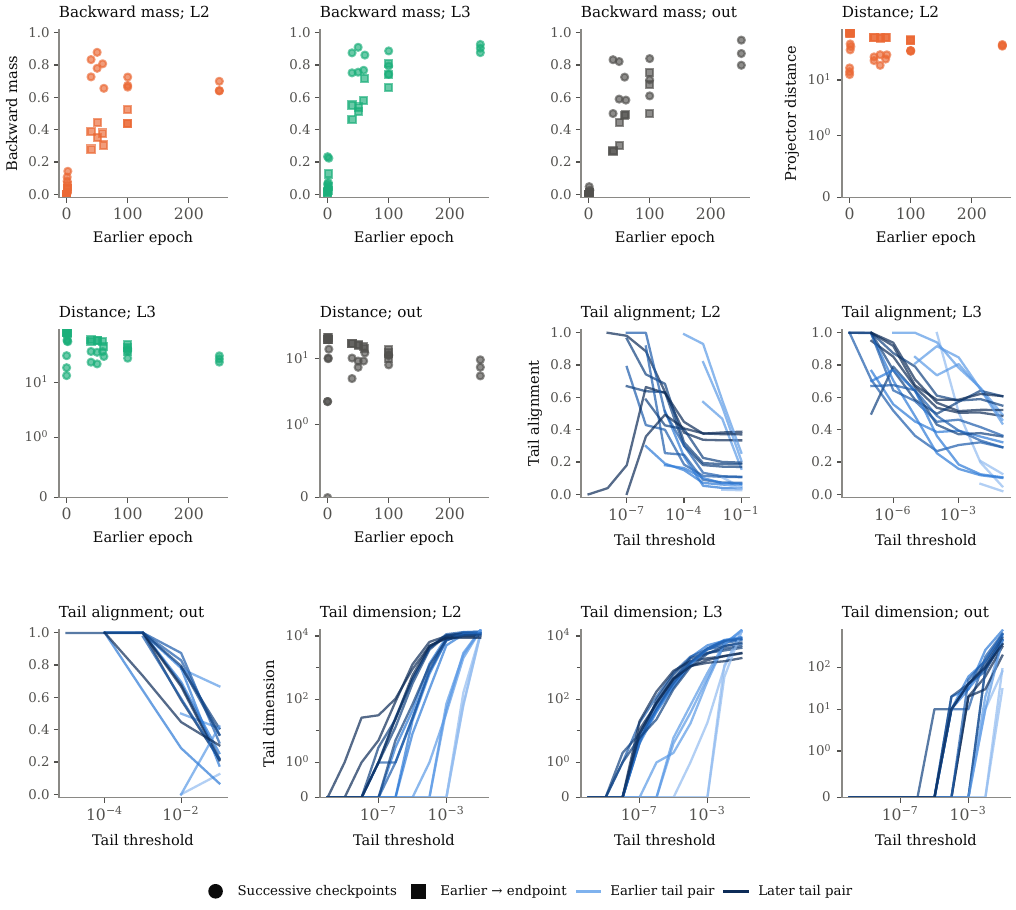}
\caption{Growth and spectral-tail alignment in the width-128 ReLU MLP, all three seeds. Panels 1--3 show how much of the later kernel was already present earlier; panels 4--6 show the projector distance, for every recorded checkpoint pair. Panels 7--9 and 10--12 show the positive spectral tail, excluding the current numerical kernel: its mean squared projection into the next kernel and its dimension, over the complete threshold grid on $J/\sqrt{s}$. Each line is one seed and consecutive checkpoint pair; darker lines denote later pairs. Alignment is projected mass, not an exact intersection dimension.}\label{fig:persistence-tails}
\end{figure}

Figure~\ref{fig:persistence-tails} places this alignment alongside dimension growth. The later kernel can contain many new directions even when it retains most of the earlier one. Approximate directions with smaller singular values also tend to align more strongly with the next checkpoint's kernel. These are projected-mass comparisons, not counts of directions proved to become exactly invariant. All comparisons use checkpoint endpoints; they do not specify how the kernel moves continuously between them. Parent--sibling intersections are reported with the connectivity application in Appendix~\ref{app:lmc}.

\FloatBarrier

\begin{figure}[!htbp]\centering
\includegraphics[width=6.081in]{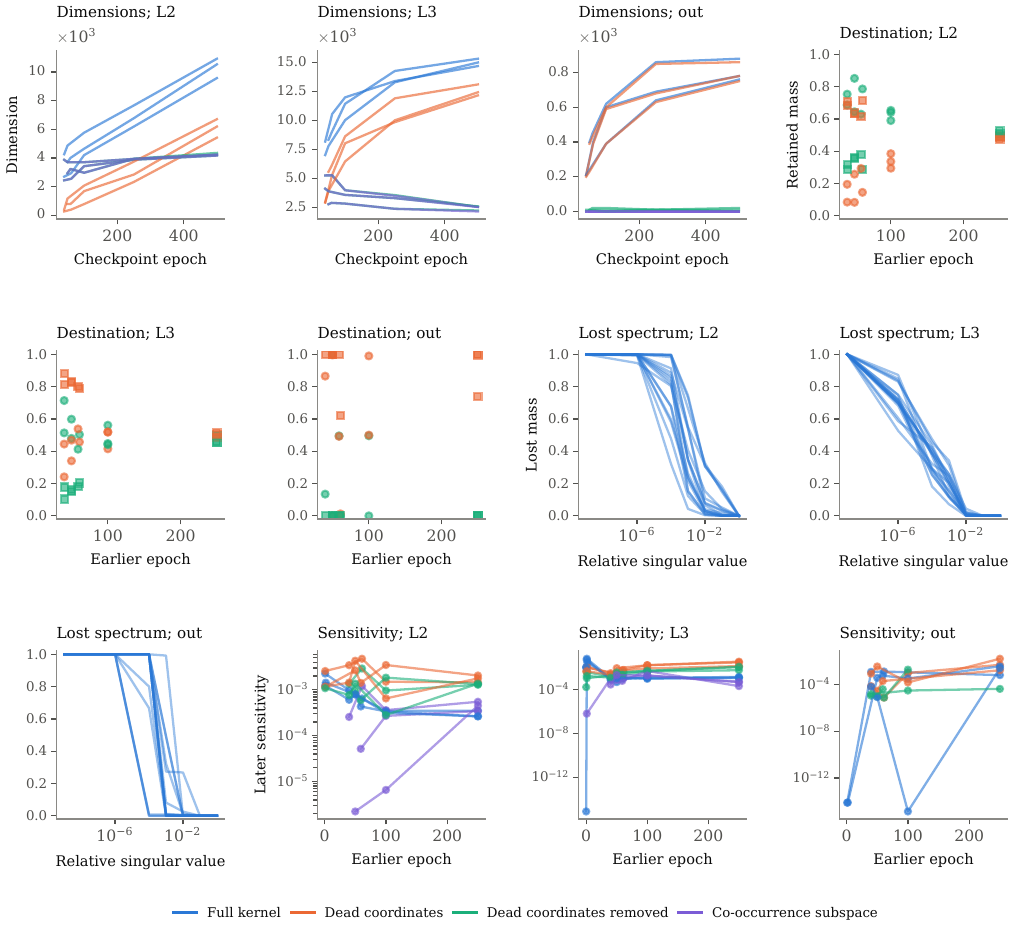}
\caption{Further ReLU persistence diagnostics, all measured layers, pairs and three seeds. Panels 1--3: dimensions of the full kernel and its component spaces. Panels 4--6: the earlier kernel with dead coordinates removed, split into mass retained in later non-dead directions (green) and later dead coordinates (orange); circles denote consecutive pairs and squares comparisons ending at epoch 500. Panels 7--9: cumulative lost mass carried by later singular directions above each relative singular-value threshold. Panels 10--12: median later sensitivity of 32 sampled directions from the earlier kernel (blue) and its positive tails at thresholds $10^{-3}$ (orange), $10^{-4}$ (green) and $10^{-6}$ (purple). Missing curves denote empty or unresolved spaces.}\label{fig:persistence-diagnostics}
\end{figure}

The component diagnostics in Figure~\ref{fig:persistence-diagnostics} distinguish two ways a direction can survive: it can remain invisible on active coordinates, or become supported on coordinates that are dead later. They also locate the lost mass within the later Jacobian spectrum and report the later sensitivity of sampled kernel and tail directions. For GELU, Table~\ref{tab:persistence-gelu-tau} records how the numerical dead-coordinate split changes with the derivative threshold; the unresolved co-occurrence split limits what can be concluded about its component persistence.

\begin{table}[!htbp]\centering
\caption{GELU numerical dead-coordinate sensitivity at epoch 500, width 128. A unit is counted when its derivative magnitude is at most $\tau$ on every analysis image. Coordinate dimensions are seed ranges. The last columns count seeds whose dead-coordinate and co-occurrence projector splits resolve, out of three. The co-occurrence split remains unresolved at the endpoint.}\label{tab:persistence-gelu-tau}
\small\begin{tabular}{llrrr}\toprule
Layer & $\tau$ & Coordinate dimension & Dead split & Deficit split\\\midrule
2 & $10^{-9}$ & 258--774 & 3/3 & 0/3\\
3 & $10^{-9}$ & 256--768 & 3/3 & 0/3\\
2 & $10^{-6}$ & 774--1806 & 0/3 & 0/3\\
3 & $10^{-6}$ & 768--1792 & 1/3 & 0/3\\
2 & $10^{-3}$ & 2064--3096 & 0/3 & 0/3\\
3 & $10^{-3}$ & 2624--3282 & 0/3 & 0/3\\
\bottomrule\end{tabular}\end{table}\FloatBarrier

\FloatBarrier
\section{From function to loss, extended version}
\label{app:applications}
This section contains an extended discussion and additional preliminary observations regarding the topic of  Section~\ref{sec:discussion}: how functional invariance relates to loss invariance.

\subsection{Functional invariance and loss geometry}
\label{app:hessian}
The loss compresses a network's outputs to one scalar. A region with constant outputs therefore has constant data loss, but a constant loss can also arise while outputs move along a loss level set. We first make this distinction explicit for MSE, then measure how much of the numerical Hessian and GGN kernels is captured by the constructed compatible layerwise invariant subspaces.

\subsubsection{What functional invariance implies}
Let $Y=(y_i)_{i=1}^s$ be the target outputs, let $e=F_X(\theta_0)-Y$ be the residual at the base network, and use the data loss $\mathcal L(\theta)=\|F_X(\theta)-Y\|_X^2$. The associated inner product is $\langle U,V\rangle_X=s^{-1}\sum_i U_i^\top V_i$. For an edit $w$, expanding the square gives
\begin{equation}
 \mathcal L(\theta_0+w)-\mathcal L(\theta_0)
 =2\langle e,\Delta F_X(w)\rangle_X+\|\Delta F_X(w)\|_X^2.
 \label{eq:application-loss-change}
\end{equation}
Thus an exact ocean preserves the loss. If the outputs vary by at most $\varepsilon$, the absolute loss change is at most $2\|e\|_X\varepsilon+\varepsilon^2$. At exact interpolation, $e=0$, and the loss change is precisely the square of the functional change. This explains why visually flat loss landscapes can contain visible functional variation.

For differentiable loss and network maps, exact constancy of $\mathcal L(\theta_0+tv)$ near $t=0$ implies
\[
 0=\left.\frac{d}{dt}\mathcal L(\theta_0+tv)\right|_{t=0}
   =\nabla_\theta\mathcal L(\theta_0)^\top v
   =\nabla_F\mathcal L(F_X(\theta_0))^\top J_Xv.
\]
Functionally invisible directions satisfy this condition automatically. Other directions can move the outputs tangentially to a loss level set. For squared loss, the level set at value $c>0$ is $\{F:\|F-Y\|_X=\sqrt c\}$; a finite output trajectory on this sphere preserves loss while changing the function. At $c=0$ the level set reduces to $F=Y$. For softmax cross-entropy, let $\pi_i=\operatorname{softmax}(f_i)$ and write $u_i$ for a first-order logit change. Its first-order loss-preservation condition is $s^{-1}\sum_i(\pi_i-y_i)^\top u_i=0$, compared with $\langle e,u\rangle_X=0$ for MSE. When the respective output gradient is nonzero, each defines a hyperplane of dimension $sd-1$.  Adding $a_i\mathbf1$ to each sample's logit vector leaves its probabilities and cross-entropy unchanged at every amplitude. Such output transformations need not all be realizable by parameter edits. 

At a twice differentiable base point, write $H=\nabla^2\mathcal L(\theta_0)$ for the loss Hessian and $G=(2/s)J_X^\top J_X$ for its Gauss--Newton part. With $a$ indexing output coordinates,
\begin{equation}
 H=G+R,\qquad
 R=\frac{2}{s}\sum_{i=1}^s\sum_{a=1}^d e_{ia}\nabla^2 F_{ia}(\theta_0).
 \label{eq:application-hessian}
\end{equation}
Every exact ocean direction annihilates $J_X$ and hence $G$. It also belongs to $\ker H$ when $e=0$. With nonzero residual, constancy along a straight direction $v$ implies $v^\top Hv=0$, but can leave $Hv\ne0$: the Hessian may couple that direction to directions outside the ocean. For ReLU networks these statements apply in the interior of an activation chamber. The numerical comparisons below retain the signed Hessian spectrum and define near-nullity using absolute eigenvalues.

\subsubsection{Constructed-ocean coverage of numerical kernels}
We evaluate depth-3, width-64 ReLU MLPs trained on MNIST, at epoch 500, using three seeds and the fixed 10,000-example analysis set. Each network has $p=25{,}418$ parameters. The Hessian uses sample-averaged, output-summed MSE with one-hot targets and no regularization. The comparison space $M$ is the bottom-up layerwise invariant subspace after imposing compatibility, constructed independently of the curvature eigendecompositions and validated against the original-network Jacobian.

For $T\in\{H,G\}$, let $Q_{T,\tau}$ contain orthonormal eigenvectors with $|\lambda_T|\le\tau$, and let $d_{T,\tau}$ be their number. With an orthonormal basis $Q_M$ of the constructed space, define
\[
 C_T(\tau)=\frac{\|Q_M^\top Q_{T,\tau}\|_F^2}{d_{T,\tau}}.
\]
This is the mean squared projection of a target-band direction onto $M$. It measures average alignment, rather than intersection dimension or subspace inclusion. An isotropic random space of the same dimension has expected coverage $\dim M/p$.

At $\tau=10^{-14}$, Hessian coverage exceeds $1-10^{-8}$ in every seed, and GGN coverage is $94.34$--$96.21\%$ (Table~\ref{tab:kernel-coverage-dimensions}). These are numerical-kernel measurements: the saved float64 eigensolver has a noise floor of order $10^{-14}$, so classification at this scale cannot resolve exact algebraic zeros. At the looser cutoff $10^{-12}$, Hessian coverage remains $99.90$--$99.94\%$. We retain both signs of the Hessian spectrum when forming each band. The GGN is positive semidefinite up to roundoff.

\begin{table}[!htbp]\centering
\caption{Dimensions and coverage at numerical cutoff $10^{-14}$; $d_H,d_G$ count the numerical-kernel dimensions. The small Hessian projection deficits are retained for reproducibility and do not measure eigensolver accuracy. The isotropic random-space expectation is $\dim M/p$, with $p=25{,}418$.}
\label{tab:kernel-coverage-dimensions}
\small\begin{tabular}{crrrrrr}\toprule
Seed & $\dim M$ & $d_H$ & $d_G$ & Hessian (\%) & GGN (\%) & Random (\%)\\\midrule
0 & 3,619 & 3,200 & 3,836 & $100-9.5423\!\times\!10^{-10}$ & 94.34 & 14.2\\
1 & 4,573 & 4,116 & 4,753 & $100-2.2455\!\times\!10^{-10}$ & 96.21 & 18.0\\
2 & 4,072 & 3,434 & 4,281 & $100-2.7936\!\times\!10^{-9}$ & 95.06 & 16.0\\
\bottomrule\end{tabular}
\end{table}

\begin{figure}[!htbp]\centering
\includegraphics[width=3.074in]{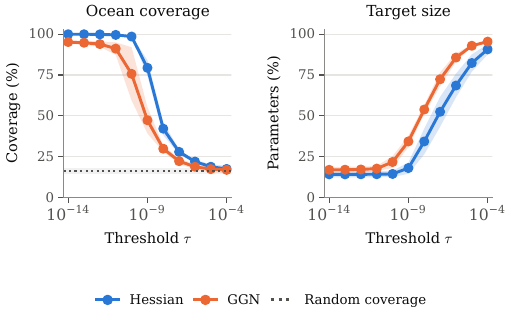}
\caption{Coverage of Hessian and GGN bands $|\lambda|\le\tau$ by the fixed constructed layerwise invariant subspace (left), and target-band dimension as a percentage of all parameters (right). Lines and shading show mean $\pm$ sample SD across three seeds. The dotted reference is dimension-matched random coverage. Cutoffs near $10^{-14}$ approach numerical resolution.}
\label{fig:fixed-ocean-coverage}
\end{figure}

\paragraph{Sensitivity to the target threshold.}
Figure~\ref{fig:fixed-ocean-coverage} increases the spectral threshold while retaining exactly the same constructed space at every threshold. The target bands grow to include small but nonzero curvature, and their coverage decreases. The right panel shows that growing denominator explicitly. These curves do not use approximate or enlarged oceans. The plotted mean and sample SD summarize the three seeds; they do not quantify uncertainty in numerical rank near the eigensolver floor.

\subsubsection{Finite-distance behavior of curvature bands}
\begin{figure}[!htbp]\centering
\includegraphics[width=3.165in]{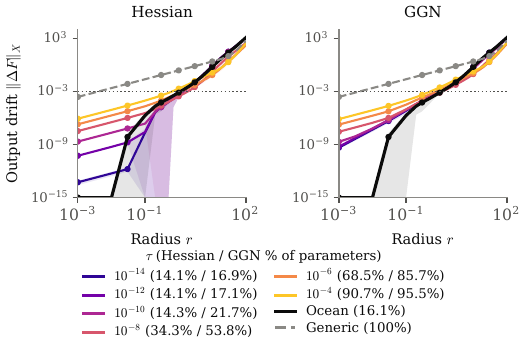}
\caption{Cumulative Hessian and GGN bands $|\lambda_T|\le\tau$ in width-64 ReLU MNIST MLPs at epoch 500. Curves show mean $\pm$ sample SD of the three seed medians, each over 256 sampled directions. Legend percentages give mean dimensions relative to all parameters (Hessian / GGN). Black: constructed ocean; dashed gray: generic directions. The dotted line marks output drift $10^{-3}$.}\label{fig:curvature-cumulative-rays}
\end{figure}

Figure~\ref{fig:curvature-cumulative-rays} tests the original network along random directions in the cumulative Hessian and GGN bands, using the protocol in Appendix~\ref{app:methods-curvature-rays}. The numerical kernels have visibility radii comparable to the constructed ocean. Enlarging a band changes both its dimension and its mixture of directions, so the finite-radius response need not vary monotonically with the eigenvalue cutoff. At output tolerance $10^{-3}$, the last safe tested radius of the median is 1 in every seed for the ocean, 1.5 for the Hessian band at $10^{-8}$, and 1 for the corresponding GGN band. At $10^{-4}$, these radii fall to 0.5 for the Hessian and 0.3--0.5 for the GGN, while generic directions reach only 0.003. These are sampled median responses, rather than uniform guarantees over either band.

\subsection{Linear mode connectivity}
\label{app:lmc}
Linear mode connectivity (LMC) asks whether the straight parameter segment between two trained networks maintains low loss. In this preliminary study, we examine whether shared ocean structure helps explain why networks descended from a common training checkpoint can remain functionally close.

\subsubsection{Loss and functional connectivity}
We study MNIST ReLU MLPs with three hidden layers of width 128 and the MNIST CNN. Each parent checkpoint is copied into two children, which inherit its parameters and Adam state and continue to epoch 500 with different minibatch orders. We compare these siblings with independently trained networks, using the protocol in Appendix~\ref{app:methods-lmc}.

For endpoint parameters $\theta_A,\theta_B$, let $\theta_\alpha=(1-\alpha)\theta_A+\alpha\theta_B$, with $\alpha\in[0,1]$, and let $F_A,F_B$ be their outputs on the evaluation set $X$. Loss connectivity concerns the largest loss excess along the segment above the chord joining its endpoint losses. To measure functional connectivity, we use the mean distance to the endpoint functions,
\[
 E(\alpha)=\tfrac12\bigl(\|F_X(\theta_\alpha)-F_A\|_X+
 \|F_X(\theta_\alpha)-F_B\|_X\bigr).
\]
Smaller values mean closer functions; even at an endpoint, this quantity includes half the difference between the endpoint functions. The additional chord-deviation measurement in Figure~\ref{fig:lmc-segments} measures departure from linear interpolation in output space, as defined in Appendix~\ref{app:methods-lmc}.

\begin{figure}[!htbp]\centering
\begin{minipage}[t]{.49\linewidth}\centering
\textbf{MLP}\par\smallskip
\includegraphics[width=\linewidth]{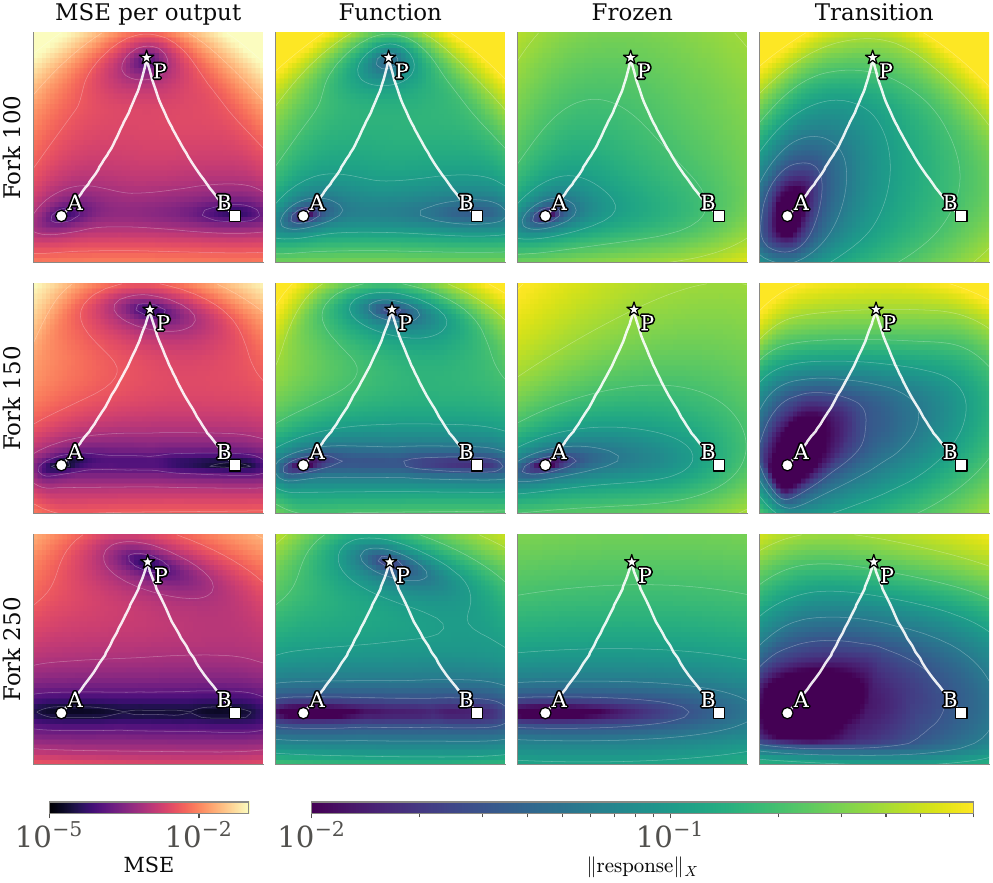}
\end{minipage}\hfill
\begin{minipage}[t]{.49\linewidth}\centering
\textbf{CNN}\par\smallskip
\includegraphics[width=\linewidth]{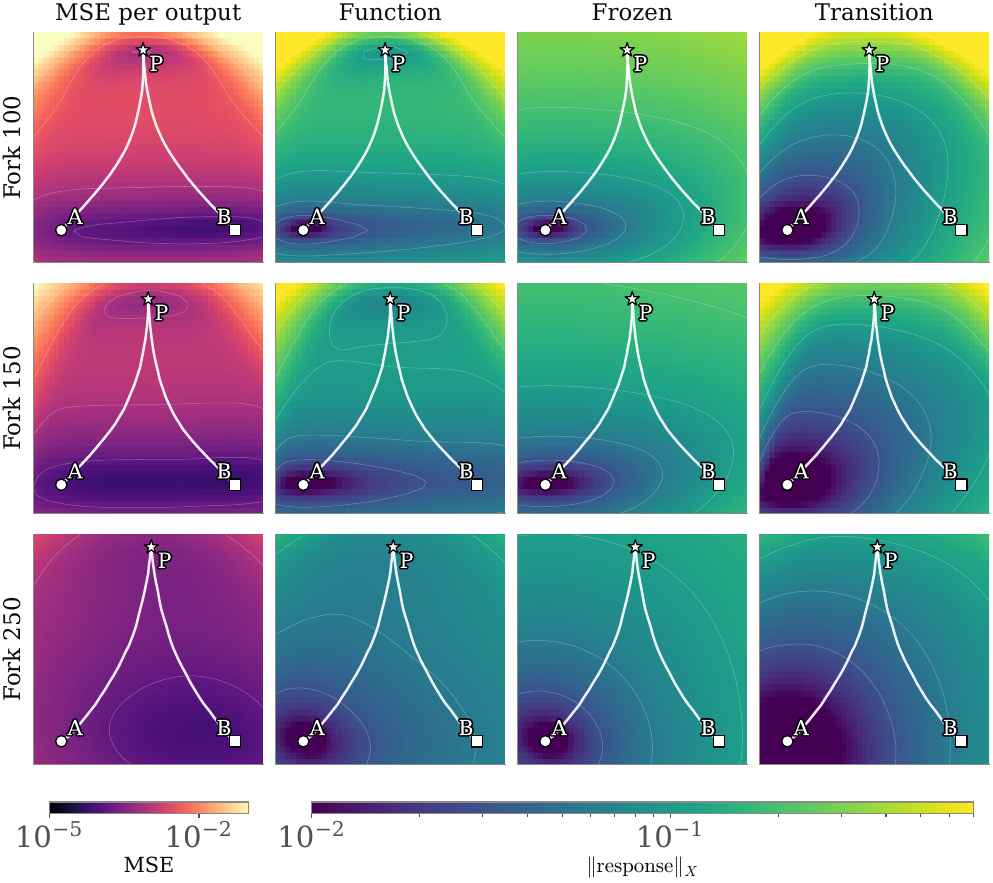}
\end{minipage}
\caption{Planes through parent $P$ and sibling endpoints $A,B$, seed 0, with forks at epochs 100, 150 and 250 and endpoints at epoch 500. Columns show MSE per output, functional change from $A$, frozen response and activation transition. White curves are projected training trajectories. The MLP planes use 10,000 analysis examples; the CNN planes use 2,000. Functional norms and color scales match across architectures.}\label{fig:lmc-landscape-full}\label{fig:lmc-landscape-cnn}
\end{figure}

Most sibling pairs maintain low loss, while the independent-network segments have much larger barriers. Sibling functions also remain closer along their connection, especially after later forks. On the test set, stable loss and classification error coexist with more visible output variation. The parent--sibling planes in Figure~\ref{fig:lmc-landscape-full} illustrate the same distinction between a low-loss region and functional invariance.
\begin{figure}[!htbp]\centering
\includegraphics[width=6.291in]{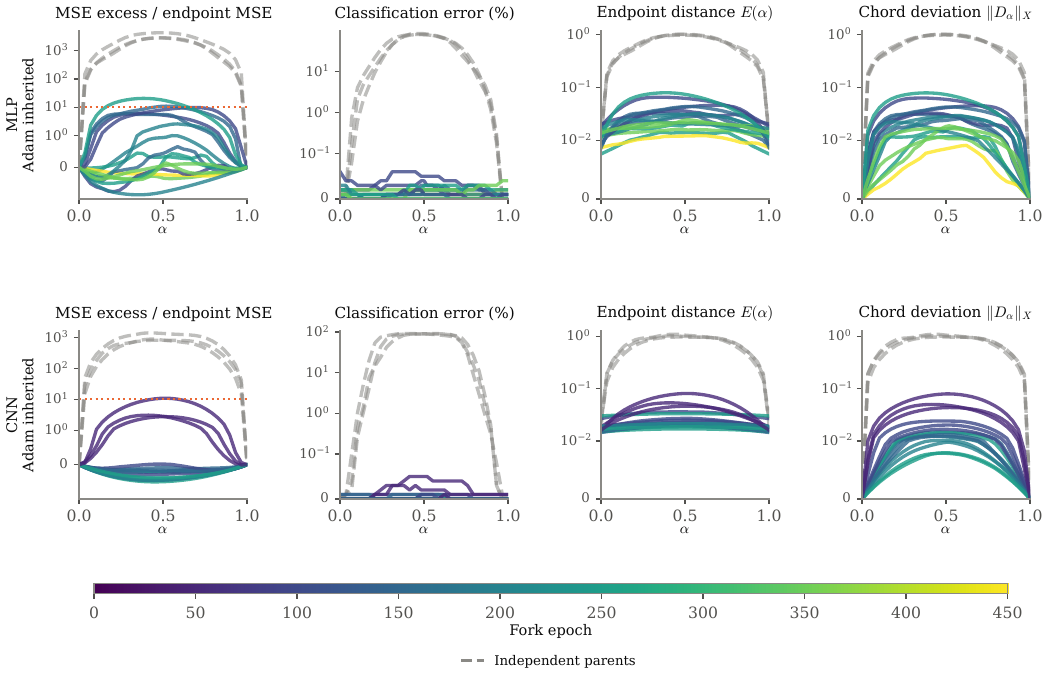}
\caption{Sibling connectivity on 10,000 training examples: MLPs above CNNs. Colors identify fork epochs for 16 MLP and 15 CNN sibling pairs; gray dashed curves show independently trained pairs. Columns show loss excess relative to the endpoint chord, normalized by mean endpoint loss; classification error; endpoint functional distance $E(\alpha)$; and output chord deviation. The dotted loss-connectivity threshold is 10.}\label{fig:lmc-segments}\label{fig:lmc-loss}
\end{figure}
\begin{figure}[!htbp]\centering
\includegraphics[width=6.291in]{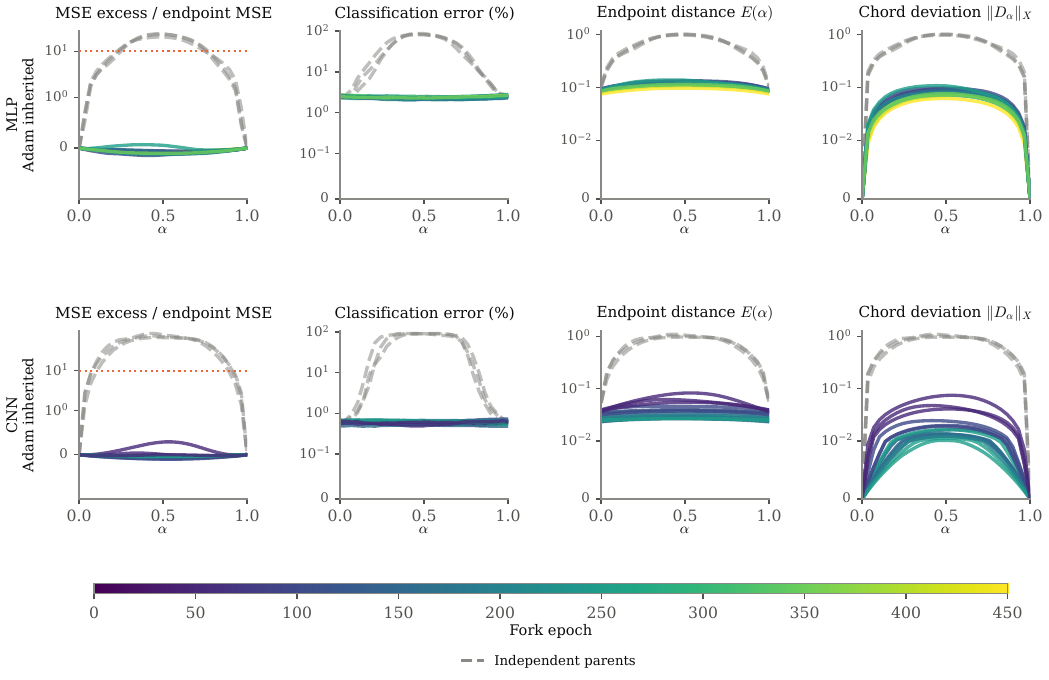}
\caption{The same connections on the separate 10,000-example test set, in the format of Figure~\ref{fig:lmc-segments}. Low loss and error can coexist with appreciable functional variation.}\label{fig:lmc-segments-test}
\end{figure}

\FloatBarrier
\subsubsection{Parent--sibling ocean persistence}
\label{app:sibling-persistence}
The persistence results above suggest that siblings inherit substantial ocean structure. We ask whether they inherit \emph{the same} directions. We measure the intersections of parent and child atomic kernels in CNNs, using a principal-angle tolerance of $0.1^\circ$ and residual checks.

Table~\ref{tab:lmc-cnn-persistence} sums dimensions over all parameter layers and divides the sibling intersection by the mean of the two child dimensions. This shared fraction generally increases with later forks. The children also retain a substantial fraction of their parent's invariant directions, establishing common structure inherited through their separate training trajectories.
\begin{table}[!htbp]\centering\small
\caption{CNN parent--sibling persistence, mean $\pm$ sample SD across three seeds (percent). Dimensions are summed over all layers before normalization. Sibling sharing divides the child-intersection dimension by the mean child dimension; parent retention divides the parent--two-child intersection by the parent dimension. $n_P$ counts seeds with a nonempty parent space, used for retention.}\label{tab:lmc-cnn-persistence}
\begin{tabular}{rrrr}\toprule
Fork epoch & Sibling sharing & Parent retention & $n_P$\\\midrule
50 & $60.0\pm7.0$ & 100.0 & 1 \\
100 & $64.9\pm5.5$ & $88.6\pm11.8$ & 3 \\
150 & $73.0\pm4.6$ & $88.8\pm5.3$ & 3 \\
200 & $74.8\pm5.8$ & $82.1\pm2.6$ & 3 \\
250 & $74.9\pm8.4$ & $78.0\pm5.7$ & 3 \\\bottomrule\end{tabular}\end{table}

\subsubsection{Sharing of constructed endpoint subspaces}
\label{app:lmc-shared}
We compare the complete compatible layerwise invariant subspaces at sibling endpoints in three seeds per architecture: width-128 MLPs forked at epoch 350 and CNNs forked at epoch 200, with all children evaluated at epoch 500. The different fork times mean that this is not an architecture-only comparison. Let $Q_A,Q_B$ be orthonormal bases of the endpoint spaces, with dimensions $k_A,k_B$, and let $P_A=Q_AQ_A^\top$, $P_B=Q_BQ_B^\top$. Their continuous sharing fraction is
\[
 s_{AB}=\frac{\|Q_A^\top Q_B\|_F^2}{\min(k_A,k_B)}.
\]
This measures how much of the smaller space projects into the other, including partial alignments. No principal-angle filtering is applied. It is not the dimension of an intersection or a fraction of all parameters.

The endpoint spaces overlap strongly in both architectures (Table~\ref{tab:lmc-constructed-sharing}). This aggregate masks a layer-specific exception: MLP seed 1's L1 components have mean squared cosine $0.273$ and median principal angle $61.2^\circ$; L1 components are zero in the other two seeds. The remaining MLP components and the nonempty CNN late-layer components are more closely aligned.

\begin{table}[!htbp]\centering
\caption{Constructed sibling-space sharing and displacement projections, mean $\pm$ sample SD across three seeds. Projection entries are norm fractions; sequential projections measure continuous overlap.}
\label{tab:lmc-constructed-sharing}
\small\setlength{\tabcolsep}{4pt}
\begin{tabular}{lccccc}\toprule
Model & $100s_{AB}$ & $\|P_Ad\|/\|d\|$ & $\|P_Bd\|/\|d\|$ & $\|P_BP_Ad\|/\|d\|$ & $\|P_AP_Bd\|/\|d\|$\\\midrule
MLP & $93.9 \pm 6.4$ & $0.386 \pm 0.045$ & $0.382 \pm 0.046$ & $0.330 \pm 0.019$ & $0.330 \pm 0.019$\\
CNN & $85.0 \pm 6.0$ & $0.487 \pm 0.103$ & $0.503 \pm 0.106$ & $0.444 \pm 0.124$ & $0.445 \pm 0.123$\\\bottomrule\end{tabular}\end{table}

\begin{figure}[!htbp]\centering
\includegraphics[width=4.585in]{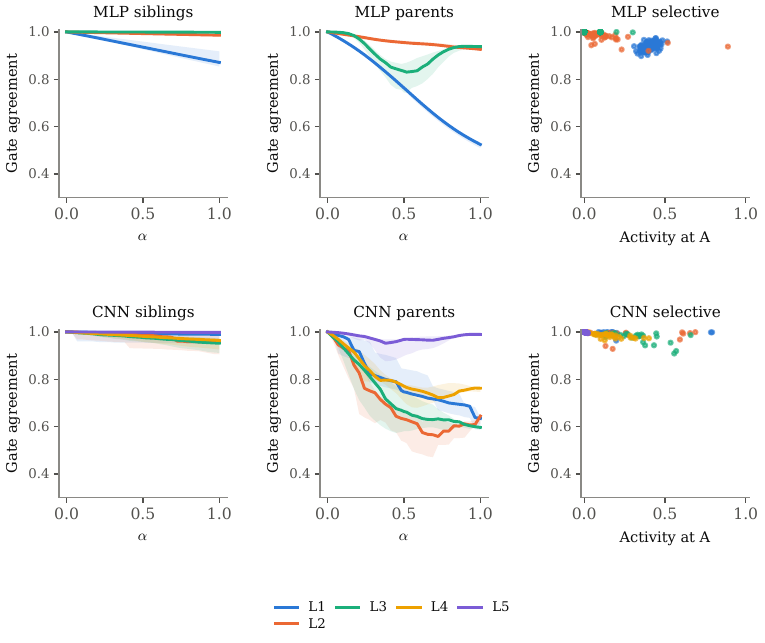}
\caption{Gate stability, MLP (first three panels) and CNN (last three). Sibling and independent-parent curves give median agreement with endpoint $A$ over all inherited-state pairs or independent pairs; bands span pairs. Selective-unit panels show activity at $A$ against midpoint gate agreement for seed 0, fork 250. Colors identify layers.}\label{fig:lmc-gates}\label{fig:lmc-selective}
\end{figure}

We also observe an intriguing pattern: siblings preserve a high fraction of activation gates along their connection, substantially more than independent networks (Figure~\ref{fig:lmc-gates}). Explaining how training produces this shared gate structure, and how it contributes to functional connectivity, is left to future work.
\FloatBarrier

\section{Additional material and plots}
\label{app:additional-material}
This section discusses the potential applications of Flat Oceans, sets out the open questions raised by the paper, then collects the ocean x-ray catalog and predictive-performance tables.

\subsection{Applications}
\label{app:finite-invariance-uses}

We give procedures to identify and construct a large region in which parameters can change while the network's outputs on an existing dataset remain controlled. Given reference parameters $\theta_0\in\mathbb R^p$, an orthonormal basis $Q\in\mathbb R^{p\times k}$ and a radius $r>0$ define such a region when
\begin{equation}
\mathcal O_{Q,r}=\{\theta_0+Qu:u\in\mathbb R^k,\ \|u\|_2\le r\},
\qquad
\sup_{\|u\|_2\le r}\|F_X(\theta_0+Qu)-F_X(\theta_0)\|_X\le\varepsilon.
\label{eq:operational-ocean}
\end{equation}
Both the dimension $k$ and the radius $r$ describe the available functional freedom. The applications below describe how this freedom could be used. We leave the development of algorithms in the specific areas to future work. 

\paragraph{Continual learning and adaptation.}
Several continual learning algorithms seek to optimize a new-task objective while preserving performance on old tasks. Orthogonal Gradient Descent \citep{farajtabar2020ogd} uses output-gradient constraints, and Elastic Weight Consolidation \citep{kirkpatrick2017ewc} uses Fisher-weighted penalties to discourage changes to important parameters. The flat ocean construction provides a finite region with explicit output-preservation control. Gradient Projection Memory's activation projections \citep{saha2021gpm} are closely related to our feature-null component. Our framework adds downstream and sample-pairing freedom and compatibility across layers, potentially enlarging the space available to fit new tasks.

\paragraph{Bayesian inference.}
Local Laplace approximations use curvature, often approximated by the GGN, to estimate posterior uncertainty \citep{daxberger2021laplace,immer2021linearization}. The flat directions are particularly consequential: weak likelihood curvature permits greater parameter uncertainty, yet small eigenvalues can be difficult to resolve accurately and are sensitive to numerical errors and curvature approximations. With an isotropic Gaussian prior of precision $\alpha>0$, a GGN eigenvalue $\lambda$ gives approximate posterior variance $(\lambda+\alpha)^{-1}$. Errors in small $\lambda$ can therefore substantially distort uncertainty when the prior precision is also small. Moreover, local curvature alone does not determine how far a flat direction extends. Flat oceans supply complementary information about finite regions over which the likelihood remains constant or controlled, potentially improving the quality of posterior approximations for neural network based models. 

\paragraph{Model editing and compression.}
Let $C:\mathbb R^p\to\mathbb R$ be an objective favoring a desired parameter property, such as smaller norms or sparsity. We can search for that property within the output-preserving region by solving
\begin{equation}
\min_{u\in\mathbb R^k,\ \|u\|_2\le r} C(\theta_0+Qu).
\label{eq:ocean-secondary-objective}
\end{equation}
Every feasible edit satisfies the output constraint in Equation~\ref{eq:operational-ocean}. For example, curvature-based pruning already exploits local loss information \citep{lecun1989brain}. The convexity of flat oceans makes it possible to run optimization programs over them.

\paragraph{Network analysis.}
The atomic decomposition identifies how features, downstream propagation and sample pairing create redundancy. The frozen/activation-remainder decomposition separates mixed-weight interactions from activation changes, providing a tool for studying finite updates even outside invariant regions.

\subsection{Open questions and future directions}
\label{app:future-work}
The present theory describes the invariant structure at a given network and dataset. Explaining how training produces this structure is the next central question. The observations also suggest applications and several extensions of the geometric theory.

\paragraph{Emergence, persistence and accretion.}
Training often enlarges the invariant subspaces while retaining many earlier directions (Appendix~\ref{app:persistence}). A dynamical account should explain how feature spans, downstream maps and their samplewise pairings evolve to create new atomic directions. It should also explain why existing directions survive later updates, so that growth can occur by accretion. Following the dimensions alone is insufficient: a theory must track the orientations of these spaces and the constraints that permit their composition.

\paragraph{Why the frozen approximation improves.}
The frozen network is defined at every checkpoint, but its accuracy over finite edits changes with training. The observed reduction of the activation remainder raises a question. Possible contributions include more stable ReLU gates, smaller local activation curvature for smooth networks, and weaker downstream propagation of activation residuals. Relating these mechanisms to learning extends existing analyses of gates and active paths \citep{lakshminarayanan2020paths}.

\paragraph{Training data and representation compression.}
In the matched comparisons of Appendix~\ref{app:training-data}, the full-data model has larger hidden atomic kernels than a model trained on a small subset, even when both are evaluated on that subset. Why can learning from more data produce more functional freedom? One question is whether additional data encourage compression of the feature and downstream spans, or change their pairings so that fewer independent output constraints remain. This concerns representations learned by different networks. For a fixed network, adding evaluation samples can only remove exact invariant directions. A dynamical explanation must account for both effects and the nonmonotone trends within the reduced-data training sweep.

\paragraph{Invariant subspace extent and unseen inputs.} The uniform norm bounds on the activation remainder can be much looser than measured radii. A sharper account should retain residual directions, downstream suppression and cancellations to characterize the extent of each invariant subspace at a prescribed tolerance. It must also connect sampled ray lengths to a guarantee over the whole invariant subspace ball. Extending finite-dataset certificates to unseen inputs requires controlling new feature and downstream constraints; the empirical transfers in Appendix~\ref{app:data-dependence} provide a starting point.

\paragraph{Maximum dimensions and efficient invariant subspace tests.}
Determining the maximum ocean dimension for nonlinear networks of arbitrary depth remains open, both for layerwise constructions and for general coupled invariant subspaces. Even the deep-linear subspace optimization requires a finite, efficient method for selecting target orientations. The general polynomial coefficient test is exact but becomes expensive with dimension and depth. Exploiting network structure to certify or construct large invariant subspaces without enumerating all coefficient constraints is an important algorithmic problem.

\paragraph{Flat symmetries across architectures.}
A systematic study should characterize flat symmetries in the full range of frozen network architectural structures. The convolutional and LayerNorm constructions here provide starting points for attention, residual connections, parameter sharing, normalization and other structured blocks. The aim is to identify which architectural identities create exact flat directions, which spaces arise from learned representations on a dataset, and when these directions compose across different block types.

\paragraph{Exceptional smooth cancellations.}
Appendix~\ref{app:smooth-exact} conjectures that, for tanh, sigmoid and softplus MLPs, the direction space of every exact local ocean is a frozen invariant subspace outside a Lebesgue-null set of data and base parameters. The one-unit calculation establishes only a special case. A general proof must exclude cancellations among units, paths and samples that persist at every order. The measure-zero claim concerns exceptional data and base parameters; the intrinsic volume of an ocean in its own affine hull is a different quantity. The reciprocal-affine counterexample shows why the activation assumptions matter.

\subsection{Ocean x-rays}
\label{app:xray}

The catalog collects the available two-dimensional parameter slices across ReLU and leaky-ReLU MLPs, GELU MLPs, CNNs, width and depth sweeps, and several direction constructions. Each Cartesian map shows the output response to two simultaneous parameter edits. Its angle--radius companion shows the same plane over a wider range of distances. The color scale is shared across the catalog, and each panel retains its own parameter scale.

White contours separate sampled points with unchanged gates from points with changed gates; black dashed contours mark output change $10^{-3}$. Their separation shows that gate changes can precede visible functional change. Smooth-activation panels have no ReLU chamber boundary. All views are measurements in selected planes; they do not certify the radius of an entire parameter subspace. The parent--sibling planes are reported separately in Appendix~\ref{app:lmc}.

\begin{figure}[!htbp]\centering
\includegraphics[width=6.367in]{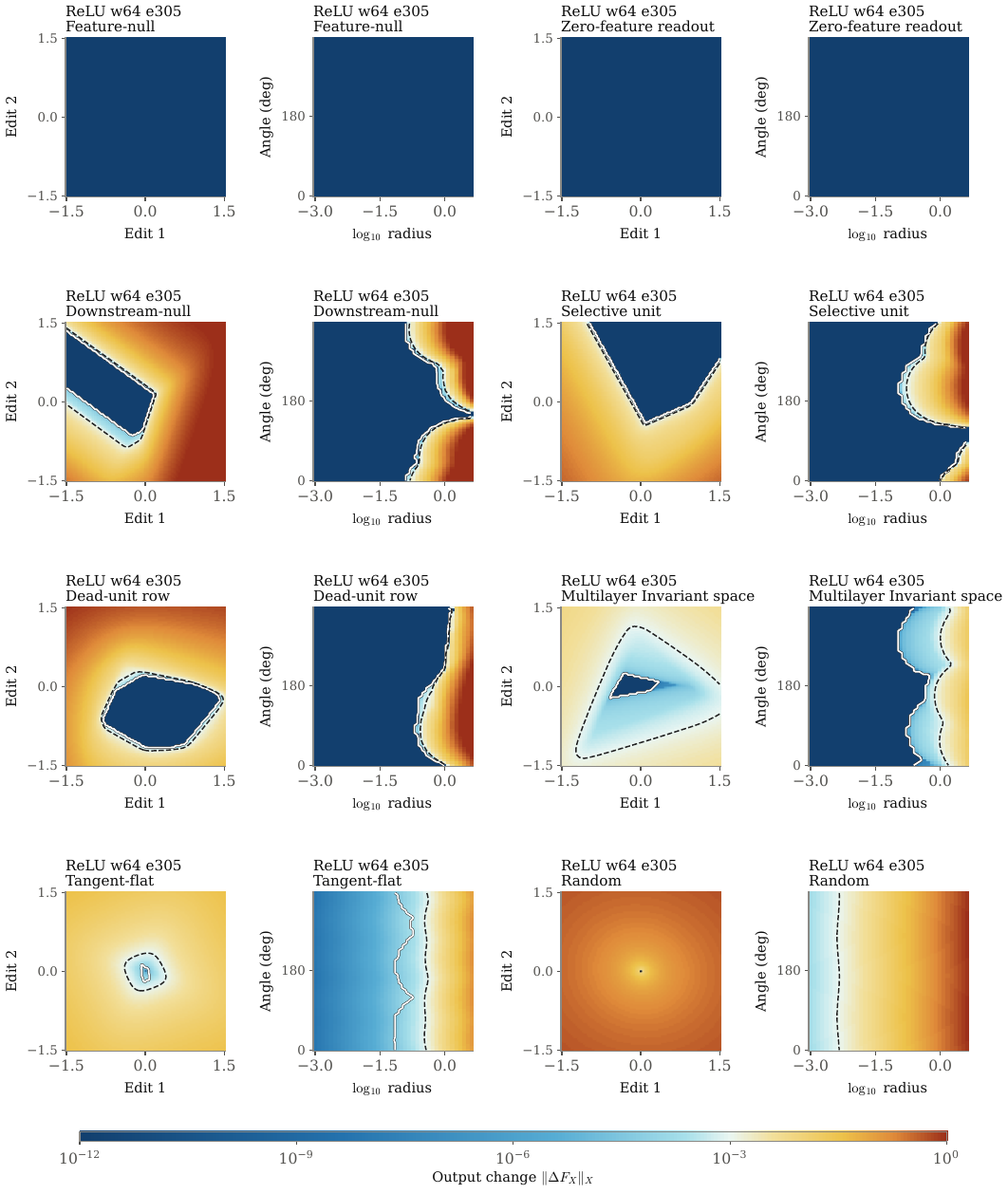}
\caption{Ocean x-ray catalog, page 1. All panels use MNIST and seed 0; titles give width (w), hidden depth (d), and checkpoint epoch (e). Adjacent Cartesian and angle--radius panels describe the same two-dimensional parameter plane. The final seven angle--radius panels of the catalog compare selective directions in width and depth sweeps; width sweeps have depth three and depth sweeps have width 128. White contours separate the sampled zero-gate-change region from changed gates, and black dashed contours mark output change $10^{-3}$. All panels share the color scale; values below $10^{-12}$ use its lowest color. See the text for the scope of these sampled views.}\label{fig:xray-catalogue-1}
\end{figure}
\FloatBarrier\begin{figure}[!htbp]\centering
\includegraphics[width=6.662in]{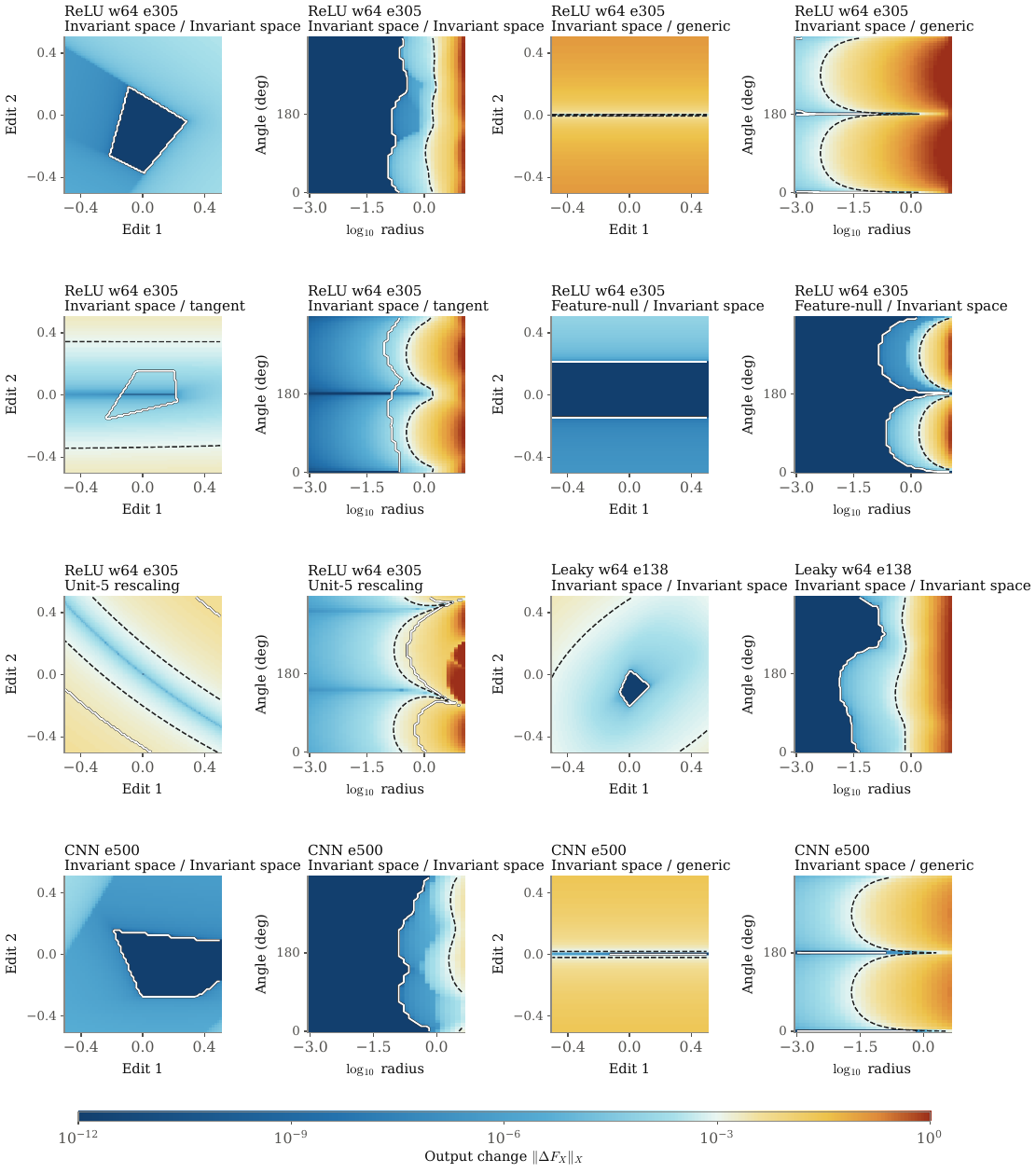}
\caption{Ocean x-ray catalog, continued. Conventions and color scale as in Figure~\ref{fig:xray-catalogue-1}.}\label{fig:xray-catalogue-2}
\end{figure}
\FloatBarrier\begin{figure}[!htbp]\centering
\includegraphics[width=6.156in]{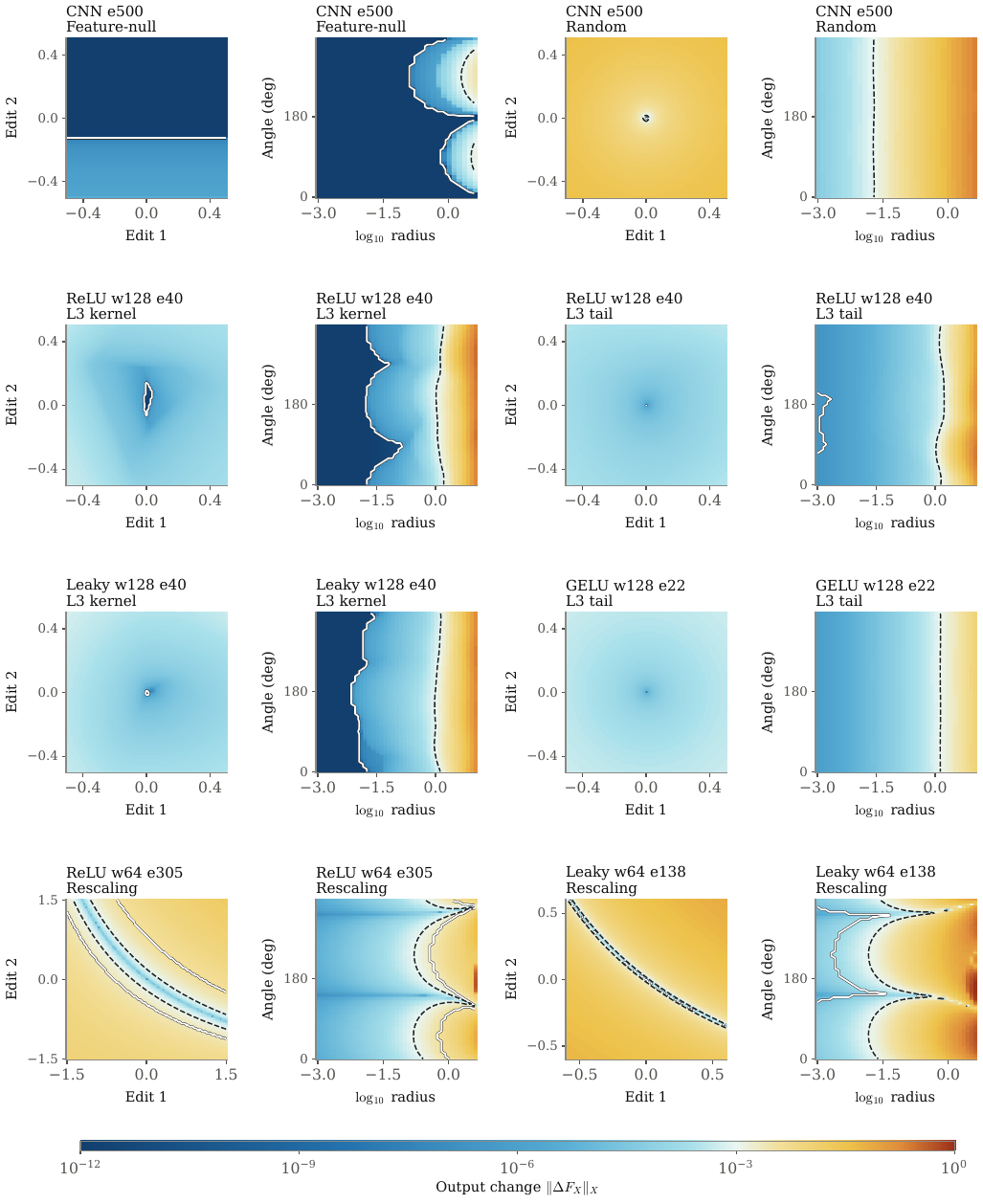}
\caption{Ocean x-ray catalog, continued. Conventions and color scale as in Figure~\ref{fig:xray-catalogue-1}.}\label{fig:xray-catalogue-3}
\end{figure}
\FloatBarrier\begin{figure}[!htbp]\centering
\includegraphics[width=6.156in]{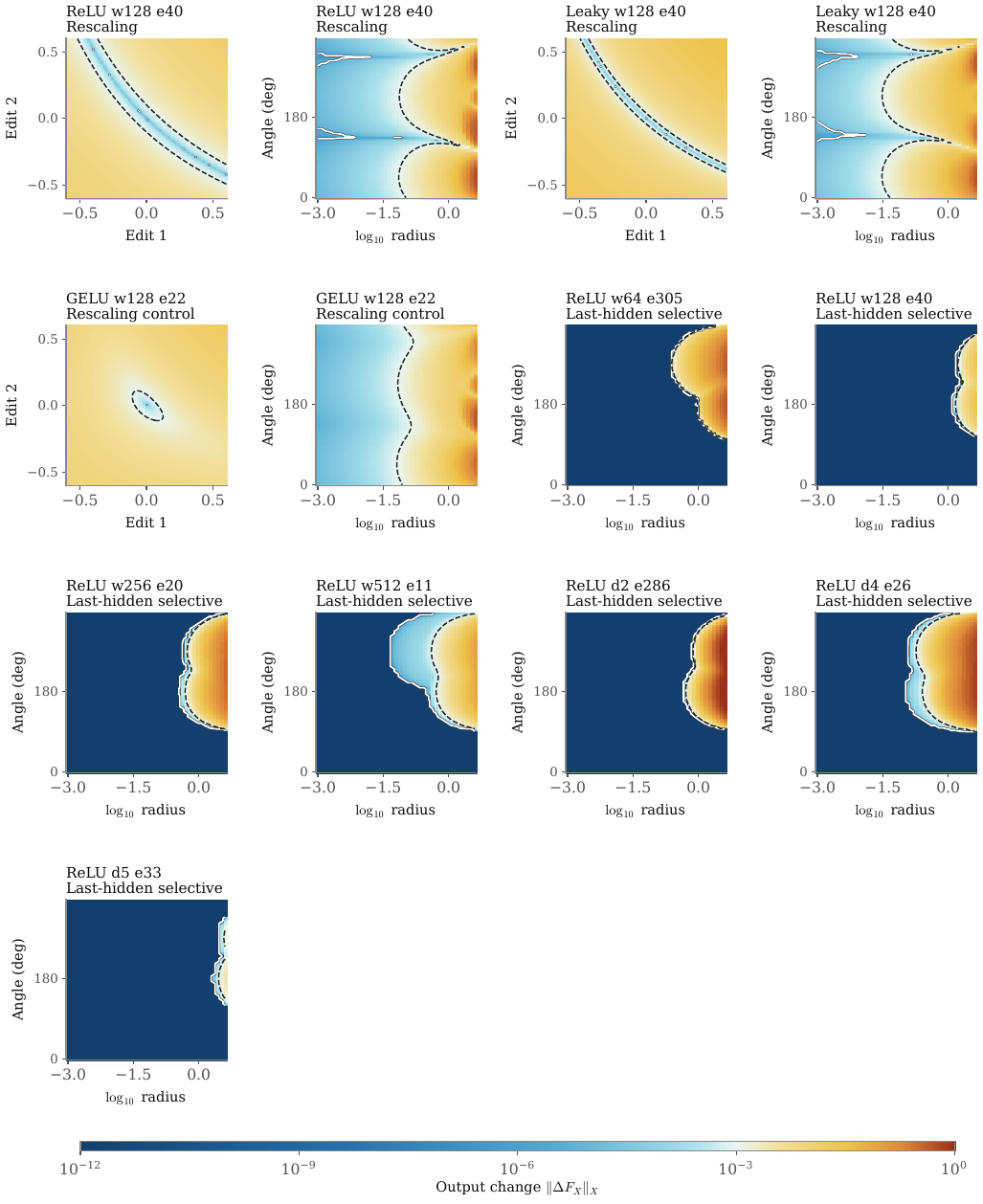}
\caption{Ocean x-ray catalog, continued. Conventions and color scale as in Figure~\ref{fig:xray-catalogue-1}.}\label{fig:xray-catalogue-4}
\end{figure}
\FloatBarrier\FloatBarrier

\FloatBarrier

\subsection{Training and test performance}
\label{app:performance}
The tables report accuracy at declared analysis stages and the intermediate checkpoints used for descriptive comparisons. Values come from the saved training histories at the actual checkpoint epoch, rather than from the earlier epoch at which a milestone was first crossed. Ranges use only available seeds, whose count is shown. Unreached milestones are omitted; no endpoint substitutes for them. Width-128, depth-three ReLU is the reference for both scaling grids. The identity model's recorded matched comparisons retain their own checkpoint epochs.

The raw/PCA, normalization, activation and dataset comparisons use the same reporting convention. Reduced-data tables report accuracy on the fitted subset and the full test set. Margin-regularized models additionally report unregularized training MSE and parameter norms, since an apparent radius change can accompany a change in fitting or scale. ViT uses cross-entropy and records online augmented training accuracy, as stated in Table~\ref{tab:native-vit-training}.

\begin{table}[!htbp]\centering
\caption{Reference and activations: train and test accuracy at declared analysis stages and descriptive checkpoints. Entries are ranges across the available seeds; epochs may differ by seed. Unreached milestones have no row.}\label{tab:perf-0-0}
\small\setlength{\tabcolsep}{4pt}
\begin{tabular}{llrrrr}\toprule
Model & Stage & Epoch & Seeds & Train (\%) & Test (\%)\\\midrule
ReLU, width 128 & Init. & 0 & 3 & 6.082--9.805 & 6.150--9.860\\
 & Early fit & 1--2 & 3 & 90.433--93.617 & 90.200--93.000\\
 & High accuracy & 40--50 & 3 & 99.900--99.927 & 97.310--97.650\\
 & Interpolation & 50--61 & 3 & 99.945--99.953 & 97.350--97.620\\
 & Epoch 50 & 50 & 3 & 99.927--99.945 & 97.310--97.590\\
 & Epoch 100 & 100 & 3 & 99.963--99.982 & 97.440--97.480\\
 & Epoch 150 & 150 & 3 & 99.968--99.998 & 97.450--97.680\\
 & Epoch 200 & 200 & 3 & 99.957--99.990 & 97.440--97.530\\
 & Epoch 250 & 250 & 3 & 99.978--100.000 & 97.380--97.580\\
 & Endpoint & 500 & 3 & 99.987--99.998 & 97.300--97.640\\
\midrule
Leaky ReLU, width 128 & Init. & 0 & 3 & 6.420--10.210 & 6.390--10.150\\
 & Early fit & 1--2 & 3 & 90.972--93.843 & 90.800--93.050\\
 & High accuracy & 40--48 & 3 & 99.903--99.915 & 97.810--97.860\\
 & Interpolation & 66--81 & 3 & 99.962--99.968 & 97.820--97.870\\
 & Epoch 50 & 50 & 3 & 99.863--99.928 & 97.540--97.810\\
 & Epoch 100 & 100 & 3 & 99.860--99.977 & 97.560--97.760\\
 & Epoch 150 & 150 & 3 & 99.832--99.995 & 97.020--97.920\\
 & Epoch 200 & 200 & 3 & 99.988--100.000 & 97.780--97.850\\
 & Epoch 250 & 250 & 3 & 99.992--99.998 & 97.800--97.890\\
 & Endpoint & 500 & 3 & 99.998--100.000 & 97.730--97.860\\
\midrule
GELU, width 128 & Init. & 0 & 3 & 6.612--10.407 & 6.530--10.710\\
 & Early fit & 1 & 3 & 90.667--92.163 & 90.580--91.880\\
 & High accuracy & 20--23 & 3 & 99.902--99.920 & 97.670--97.930\\
 & Interpolation & 41--42 & 3 & 99.982--99.990 & 97.710--97.780\\
 & Epoch 50 & 50 & 3 & 99.993 & 97.680--97.920\\
 & Epoch 100 & 100 & 3 & 99.998--100.000 & 97.740--97.810\\
 & Epoch 150 & 150 & 3 & 100.000 & 97.660--97.730\\
 & Epoch 200 & 200 & 3 & 99.978--100.000 & 97.590--97.780\\
 & Epoch 250 & 250 & 3 & 100.000 & 97.490--97.820\\
 & Endpoint & 500 & 3 & 99.968--100.000 & 97.520--97.660\\
\bottomrule\end{tabular}\end{table}

\FloatBarrier
\begin{table}[!htbp]\centering
\caption{Reference and activations: train and test accuracy at declared analysis stages and descriptive checkpoints. Entries are ranges across the available seeds; epochs may differ by seed. Unreached milestones have no row. Identity checkpoints follow its recorded comparison; its own high-accuracy and interpolation milestones are not attained.}\label{tab:perf-0-3}
\small\setlength{\tabcolsep}{4pt}
\begin{tabular}{llrrrr}\toprule
Model & Stage & Epoch & Seeds & Train (\%) & Test (\%)\\\midrule
Identity, width 128 & Init. & 0 & 3 & 10.770--14.468 & 10.630--14.330\\
 & Epoch 50 & 50 & 3 & 85.227--85.492 & 85.940--86.030\\
 & Epoch 100 & 100 & 3 & 85.113--85.173 & 85.410--85.890\\
 & Epoch 150 & 150 & 3 & 85.128--85.202 & 85.730--85.920\\
 & Epoch 200 & 200 & 3 & 85.053--85.583 & 85.880--86.340\\
 & Epoch 250 & 250 & 3 & 85.270--85.312 & 85.880--85.990\\
 & Endpoint & 500 & 3 & 85.125--85.397 & 85.770--86.210\\
\midrule
Leaky ReLU, width 64 & Init. & 0 & 3 & 8.305--11.187 & 8.220--11.150\\
 & Early fit & 2 & 3 & 90.197--91.633 & 90.300--91.550\\
 & High accuracy & 138--285 & 3 & 99.900--99.902 & 97.100--97.390\\
 & Interpolation & 251--468 & 3 & 99.927--99.950 & 97.260--97.440\\
 & Epoch 50 & 50 & 3 & 99.668--99.740 & 97.130--97.300\\
 & Epoch 100 & 100 & 3 & 99.825--99.877 & 97.330--97.480\\
 & Epoch 150 & 150 & 3 & 99.803--99.900 & 97.120--97.490\\
 & Epoch 200 & 200 & 3 & 99.867--99.915 & 97.180--97.380\\
 & Epoch 250 & 250 & 3 & 99.880--99.932 & 97.170--97.410\\
 & Endpoint & 500 & 3 & 99.917--99.968 & 97.120--97.300\\
\bottomrule\end{tabular}\end{table}

\FloatBarrier
\begin{table}[!htbp]\centering
\caption{MLP scaling: train and test accuracy at declared analysis stages and descriptive checkpoints. Entries are ranges across the available seeds; epochs may differ by seed. Unreached milestones have no row.}\label{tab:perf-1-0}
\small\setlength{\tabcolsep}{4pt}
\begin{tabular}{llrrrr}\toprule
Model & Stage & Epoch & Seeds & Train (\%) & Test (\%)\\\midrule
Width 32 & Init. & 0 & 3 & 6.262--13.848 & 5.820--14.450\\
 & Early fit & 5 & 3 & 92.838--94.217 & 92.470--93.690\\
 & Epoch 50 & 50 & 3 & 99.027--99.187 & 95.780--96.060\\
 & Epoch 100 & 100 & 3 & 99.212--99.472 & 95.540--95.720\\
 & Epoch 150 & 150 & 3 & 99.348--99.513 & 95.310--95.770\\
 & Epoch 200 & 200 & 3 & 99.385--99.502 & 95.110--95.630\\
 & Epoch 250 & 250 & 3 & 99.382--99.612 & 95.000--95.730\\
 & Endpoint & 500 & 3 & 99.485--99.635 & 95.030--95.760\\
\midrule
Width 64 & Init. & 0 & 3 & 8.943--11.245 & 9.150--11.510\\
 & Early fit & 2 & 3 & 90.062--91.297 & 89.980--91.300\\
 & High accuracy & 188--313 & 3 & 99.900--99.910 & 96.710--96.910\\
 & Interpolation & 364 & 1 & 99.940 & 96.870\\
 & Epoch 50 & 50 & 3 & 99.687--99.725 & 96.610--97.050\\
 & Epoch 100 & 100 & 3 & 99.753--99.850 & 96.890--96.970\\
 & Epoch 150 & 150 & 3 & 99.688--99.873 & 96.760--96.880\\
 & Epoch 200 & 200 & 3 & 99.857--99.907 & 96.900--97.190\\
 & Epoch 250 & 250 & 3 & 99.822--99.890 & 96.770--97.000\\
 & Endpoint & 500 & 3 & 99.897--99.960 & 96.740--96.870\\
\midrule
Width 256 & Init. & 0 & 3 & 9.365--11.188 & 9.840--11.590\\
 & Early fit & 1 & 3 & 93.267--93.405 & 92.750--92.890\\
 & High accuracy & 16--25 & 3 & 99.902--99.933 & 97.870--97.940\\
 & Interpolation & 31--47 & 3 & 99.978--99.982 & 97.900--98.100\\
 & Epoch 50 & 50 & 3 & 99.982--99.987 & 97.920--98.060\\
 & Epoch 100 & 100 & 3 & 99.973--99.992 & 97.750--98.020\\
 & Epoch 150 & 150 & 3 & 99.975--99.992 & 97.760--97.950\\
 & Epoch 200 & 200 & 3 & 99.987--100.000 & 97.870--98.060\\
 & Epoch 250 & 250 & 3 & 99.995--99.997 & 97.700--98.170\\
 & Endpoint & 500 & 3 & 99.978--99.992 & 97.560--97.790\\
\bottomrule\end{tabular}\end{table}

\FloatBarrier
\begin{table}[!htbp]\centering
\caption{MLP scaling: train and test accuracy at declared analysis stages and descriptive checkpoints. Entries are ranges across the available seeds; epochs may differ by seed. Unreached milestones have no row.}\label{tab:perf-1-3}
\small\setlength{\tabcolsep}{4pt}
\begin{tabular}{llrrrr}\toprule
Model & Stage & Epoch & Seeds & Train (\%) & Test (\%)\\\midrule
Width 512 & Init. & 0 & 3 & 7.437--9.718 & 7.080--9.690\\
 & Early fit & 1 & 3 & 95.055--95.240 & 94.190--94.590\\
 & High accuracy & 10--11 & 3 & 99.902--99.928 & 97.900--98.080\\
 & Interpolation & 34--36 & 3 & 99.988--99.998 & 98.120--98.360\\
 & Epoch 50 & 50 & 3 & 99.995 & 98.140--98.280\\
 & Epoch 100 & 100 & 3 & 99.977--99.998 & 98.000--98.240\\
 & Epoch 150 & 150 & 3 & 99.993--99.997 & 97.890--98.050\\
 & Epoch 200 & 200 & 3 & 99.985--99.998 & 97.860--98.030\\
 & Epoch 250 & 250 & 3 & 99.957--100.000 & 97.820--97.930\\
 & Endpoint & 500 & 3 & 99.993--99.997 & 97.640--97.710\\
\midrule
Depth 2 & Init. & 0 & 3 & 9.343--15.093 & 9.770--14.800\\
 & Early fit & 1--2 & 3 & 90.627--93.575 & 90.070--93.070\\
 & High accuracy & 85--286 & 3 & 99.900--99.910 & 96.620--97.230\\
 & Epoch 50 & 50 & 3 & 99.790--99.858 & 97.410--97.530\\
 & Epoch 100 & 100 & 3 & 99.855--99.917 & 97.100--97.180\\
 & Epoch 150 & 150 & 3 & 99.863--99.923 & 96.910--96.990\\
 & Epoch 200 & 200 & 3 & 99.877--99.920 & 96.690--96.820\\
 & Epoch 250 & 250 & 3 & 99.878--99.925 & 96.530--96.800\\
 & Endpoint & 500 & 3 & 99.900--99.928 & 96.260--96.810\\
\midrule
Depth 4 & Init. & 0 & 3 & 9.162--11.103 & 9.240--11.960\\
 & Early fit & 1 & 3 & 90.617--91.978 & 90.630--91.610\\
 & High accuracy & 26--28 & 3 & 99.900--99.913 & 97.390--97.720\\
 & Interpolation & 36--55 & 3 & 99.947--99.977 & 97.510--97.650\\
 & Epoch 50 & 50 & 3 & 99.923--99.972 & 97.420--97.800\\
 & Epoch 100 & 100 & 3 & 99.382--99.975 & 96.560--97.770\\
 & Epoch 150 & 150 & 3 & 99.992--99.997 & 97.510--97.860\\
 & Epoch 200 & 200 & 3 & 99.982--99.992 & 97.380--97.940\\
 & Epoch 250 & 250 & 3 & 99.995--99.997 & 97.720--97.790\\
 & Endpoint & 500 & 3 & 99.973--99.997 & 97.450--97.630\\
\bottomrule\end{tabular}\end{table}

\FloatBarrier
\begin{table}[!htbp]\centering
\caption{MLP scaling: train and test accuracy at declared analysis stages and descriptive checkpoints. Entries are ranges across the available seeds; epochs may differ by seed. Unreached milestones have no row.}\label{tab:perf-1-6}
\small\setlength{\tabcolsep}{4pt}
\begin{tabular}{llrrrr}\toprule
Model & Stage & Epoch & Seeds & Train (\%) & Test (\%)\\\midrule
Depth 5 & Init. & 0 & 3 & 8.527--11.118 & 8.810--11.410\\
 & Early fit & 1 & 3 & 91.198--91.833 & 90.710--91.630\\
 & High accuracy & 33--40 & 3 & 99.902--99.908 & 97.220--97.470\\
 & Interpolation & 48--82 & 3 & 99.962--99.977 & 97.370--97.550\\
 & Epoch 50 & 50 & 3 & 99.837--99.932 & 97.350--97.520\\
 & Epoch 100 & 100 & 3 & 99.950--99.982 & 97.610--97.710\\
 & Epoch 150 & 150 & 3 & 99.942--99.993 & 97.560--97.770\\
 & Epoch 200 & 200 & 3 & 99.865--99.993 & 97.570--97.680\\
 & Epoch 250 & 250 & 3 & 99.818--99.978 & 97.400--97.600\\
 & Endpoint & 500 & 3 & 99.992--99.993 & 97.740--97.810\\
\bottomrule\end{tabular}\end{table}

\FloatBarrier
\begin{table}[!htbp]\centering
\caption{Convolution and normalization: train and test accuracy at declared analysis stages and descriptive checkpoints. Entries are ranges across the available seeds; epochs may differ by seed. Unreached milestones have no row.}\label{tab:perf-2-0}
\small\setlength{\tabcolsep}{4pt}
\begin{tabular}{llrrrr}\toprule
Model & Stage & Epoch & Seeds & Train (\%) & Test (\%)\\\midrule
MNIST CNN, base & Init. & 0 & 3 & 9.068--9.890 & 8.970--9.640\\
 & Early fit & 1 & 3 & 96.607--96.872 & 96.940--97.030\\
 & High accuracy & 65--68 & 3 & 99.902--99.905 & 99.450--99.560\\
 & Interpolation & 281--385 & 3 & 99.998--100.000 & 99.360--99.480\\
 & Epoch 50 & 50 & 3 & 99.792--99.815 & 99.430--99.530\\
 & Epoch 100 & 100 & 3 & 99.962--99.977 & 99.450--99.540\\
 & Epoch 150 & 150 & 3 & 99.990--99.995 & 99.420--99.460\\
 & Epoch 200 & 200 & 3 & 99.997--99.998 & 99.420--99.520\\
 & Epoch 250 & 250 & 3 & 99.988--100.000 & 99.450--99.530\\
 & Endpoint & 500 & 3 & 99.998--100.000 & 99.370--99.490\\
\midrule
MNIST CNN, $0.5\times$ & Init. & 0 & 3 & 8.618--11.785 & 8.480--10.640\\
 & Early fit & 1 & 3 & 94.940--96.513 & 95.300--96.730\\
 & High accuracy & 384--446 & 3 & 99.900 & 99.000--99.360\\
 & Epoch 50 & 50 & 3 & 99.292--99.435 & 99.210--99.250\\
 & Epoch 100 & 100 & 3 & 99.565--99.603 & 99.300--99.420\\
 & Epoch 150 & 150 & 3 & 99.663--99.717 & 99.280--99.410\\
 & Epoch 200 & 200 & 3 & 99.715--99.788 & 99.300--99.410\\
 & Epoch 250 & 250 & 3 & 99.758--99.808 & 99.260--99.440\\
 & Endpoint & 500 & 3 & 99.903--99.923 & 99.030--99.200\\
\midrule
MNIST CNN, $2\times$ & Init. & 0 & 3 & 8.827--9.973 & 7.900--9.910\\
 & Early fit & 1 & 3 & 97.388--97.697 & 97.620--97.970\\
 & High accuracy & 31--35 & 3 & 99.903--99.917 & 99.480--99.580\\
 & Interpolation & 106--130 & 3 & 100.000 & 99.470--99.570\\
 & Epoch 50 & 50 & 3 & 99.980--99.985 & 99.510--99.590\\
 & Epoch 100 & 100 & 3 & 99.998--100.000 & 99.500--99.590\\
 & Epoch 150 & 150 & 3 & 100.000 & 99.440--99.610\\
 & Epoch 200 & 200 & 3 & 100.000 & 99.450--99.570\\
 & Epoch 250 & 250 & 3 & 100.000 & 99.490--99.570\\
 & Endpoint & 500 & 3 & 100.000 & 99.450--99.490\\
\bottomrule\end{tabular}\end{table}

\FloatBarrier
\begin{table}[!htbp]\centering
\caption{Convolution and normalization: train and test accuracy at declared analysis stages and descriptive checkpoints. Entries are ranges across the available seeds; epochs may differ by seed. Unreached milestones have no row.}\label{tab:perf-2-3}
\small\setlength{\tabcolsep}{4pt}
\begin{tabular}{llrrrr}\toprule
Model & Stage & Epoch & Seeds & Train (\%) & Test (\%)\\\midrule
BatchNorm, width 64 & Init. & 0 & 3 & 10.462--10.750 & 11.000--11.220\\
 & Early fit & 2 & 3 & 91.823--92.240 & 92.000--92.260\\
 & High accuracy & 51--56 & 3 & 99.905--99.910 & 97.300--97.620\\
 & Interpolation & 177--199 & 3 & 99.970--99.988 & 97.290--97.430\\
 & Epoch 50 & 50 & 3 & 99.873--99.897 & 97.290--97.660\\
 & Epoch 100 & 100 & 3 & 99.948--99.973 & 97.280--97.500\\
 & Epoch 150 & 150 & 3 & 99.967--99.983 & 97.440--97.490\\
 & Epoch 200 & 200 & 3 & 99.973--99.987 & 97.370--97.530\\
 & Epoch 250 & 250 & 3 & 99.982--99.985 & 97.290--97.440\\
 & Endpoint & 500 & 3 & 99.993--99.997 & 97.480--97.510\\
\bottomrule\end{tabular}\end{table}

\FloatBarrier
\begin{table}[!htbp]\centering
\caption{Input and dataset comparisons: train and test accuracy at declared analysis stages and descriptive checkpoints. Entries are ranges across the available seeds; epochs may differ by seed. Unreached milestones have no row.}\label{tab:perf-3-0}
\small\setlength{\tabcolsep}{4pt}
\begin{tabular}{llrrrr}\toprule
Model & Stage & Epoch & Seeds & Train (\%) & Test (\%)\\\midrule
Raw MNIST, width 64 & Init. & 0 & 3 & 10.075--13.660 & 10.130--13.130\\
 & Early fit & 1--2 & 3 & 91.455--93.277 & 91.730--92.930\\
 & High accuracy & 90--125 & 3 & 99.900--99.912 & 97.180--97.390\\
 & Interpolation & 224--379 & 3 & 99.952--99.955 & 97.270--97.360\\
 & Epoch 50 & 50 & 3 & 99.752--99.775 & 97.120--97.360\\
 & Epoch 100 & 100 & 3 & 99.833--99.877 & 97.130--97.190\\
 & Epoch 150 & 150 & 3 & 99.838--99.912 & 97.160--97.350\\
 & Epoch 200 & 200 & 3 & 99.882--99.948 & 97.190--97.370\\
 & Epoch 250 & 250 & 3 & 99.922--99.945 & 97.130--97.380\\
 & Endpoint & 500 & 3 & 99.957--99.975 & 97.020--97.460\\
\midrule
SVHN MLP, baseline & Init. & 0 & 3 & 9.044--10.077 & 8.812--9.496\\
 & Early fit & 40 & 3 & 90.026--90.555 & 80.059--80.374\\
 & Epoch 50 & 50 & 3 & 90.830--91.524 & 79.955--80.186\\
 & Epoch 100 & 100 & 3 & 92.585--93.177 & 79.168--79.940\\
 & Epoch 150 & 150 & 3 & 93.169--93.819 & 78.688--79.402\\
 & Epoch 200 & 200 & 3 & 93.295--94.103 & 78.350--79.007\\
 & Epoch 250 & 250 & 3 & 93.700--94.181 & 78.196--78.669\\
 & Endpoint & 500 & 3 & 94.080--94.788 & 77.704--77.812\\
\midrule
SVHN MLP, revised & Init. & 0 & 3 & 9.325--10.249 & 9.342--9.715\\
 & Early fit & 40--50 & 3 & 90.429--90.468 & 83.559--83.924\\
 & Epoch 50 & 50 & 3 & 90.429--90.630 & 83.559--84.139\\
 & Epoch 100 & 100 & 3 & 91.352--91.465 & 83.805--84.684\\
 & Epoch 150 & 150 & 3 & 91.670--91.793 & 83.897--84.542\\
 & Epoch 200 & 200 & 3 & 91.586--92.054 & 84.519--84.842\\
 & Epoch 250 & 250 & 3 & 91.819--91.917 & 84.404--84.753\\
 & Endpoint & 500 & 3 & 91.963--92.351 & 84.588--84.669\\
 & Validation selected & 371--494 & 3 & 91.960--92.235 & 84.315--84.577\\
\bottomrule\end{tabular}\end{table}

\FloatBarrier
\begin{table}[!htbp]\centering
\caption{Input and dataset comparisons: train and test accuracy at declared analysis stages and descriptive checkpoints. Entries are ranges across the available seeds; epochs may differ by seed. Unreached milestones have no row.}\label{tab:perf-3-3}
\small\setlength{\tabcolsep}{4pt}
\begin{tabular}{llrrrr}\toprule
Model & Stage & Epoch & Seeds & Train (\%) & Test (\%)\\\midrule
SVHN CNN & Init. & 0 & 3 & 7.707--13.379 & 8.032--12.957\\
 & Early fit & 5 & 3 & 91.178--92.904 & 90.124--91.841\\
 & High accuracy & 113--116 & 3 & 99.903--99.906 & 92.594--92.720\\
 & Epoch 50 & 50 & 3 & 99.244--99.428 & 94.000--94.111\\
 & Epoch 100 & 100 & 3 & 99.835--99.868 & 92.744--93.009\\
 & Epoch 150 & 150 & 3 & 99.947--99.951 & 92.286--92.482\\
 & Epoch 200 & 200 & 3 & 99.969--99.974 & 92.029--92.133\\
 & Epoch 250 & 250 & 3 & 99.977--99.978 & 91.683--91.829\\
 & Endpoint & 500 & 3 & 99.984--99.988 & 91.268--91.299\\
\bottomrule\end{tabular}\end{table}

\FloatBarrier
\begin{table}[!htbp]\centering
\caption{Reduced training samples: train and test accuracy at declared analysis stages and descriptive checkpoints. Entries are ranges across the available seeds; epochs may differ by seed. Unreached milestones have no row.}\label{tab:perf-4-0}
\small\setlength{\tabcolsep}{4pt}
\begin{tabular}{llrrrr}\toprule
Model & Stage & Epoch & Seeds & Train (\%) & Test (\%)\\\midrule
MNIST, $N=1{,}000$ & Init. & 0 & 3 & 6.600--11.200 & 6.150--9.860\\
 & Early fit & 20 & 3 & 94.000--97.600 & 64.610--70.220\\
 & High accuracy & 31--39 & 3 & 99.900 & 71.400--75.310\\
 & Interpolation & 184--209 & 3 & 100.000 & 74.060--77.210\\
 & Epoch 50 & 50 & 3 & 100.000 & 72.580--76.610\\
 & Epoch 100 & 100 & 3 & 100.000 & 74.130--77.340\\
 & Epoch 150 & 150 & 3 & 100.000 & 74.170--77.230\\
 & Epoch 200 & 200 & 3 & 100.000 & 74.070--77.150\\
 & Epoch 250 & 250 & 3 & 100.000 & 73.920--77.170\\
 & Endpoint & 500 & 3 & 100.000 & 73.790--77.040\\
\midrule
MNIST, $N=2{,}000$ & Init. & 0 & 3 & 6.350--10.750 & 6.150--9.860\\
 & Early fit & 20 & 3 & 97.550--98.650 & 77.090--81.450\\
 & High accuracy & 29--39 & 3 & 99.900 & 81.210--84.570\\
 & Interpolation & 212--238 & 3 & 100.000 & 82.230--83.920\\
 & Epoch 50 & 50 & 3 & 99.950--100.000 & 82.560--85.100\\
 & Epoch 100 & 100 & 3 & 100.000 & 83.250--84.710\\
 & Epoch 150 & 150 & 3 & 100.000 & 82.810--84.370\\
 & Epoch 200 & 200 & 3 & 100.000 & 82.460--83.930\\
 & Epoch 250 & 250 & 3 & 100.000 & 82.190--83.820\\
 & Endpoint & 500 & 3 & 100.000 & 82.220--84.390\\
\midrule
MNIST, $N=5{,}000$ & Init. & 0 & 3 & 6.260--10.300 & 6.150--9.860\\
 & Early fit & 10 & 3 & 93.740--95.380 & 84.790--87.190\\
 & High accuracy & 32--46 & 3 & 99.900 & 91.130--91.480\\
 & Interpolation & 335--476 & 3 & 100.000 & 91.450--92.480\\
 & Epoch 50 & 50 & 3 & 99.940--100.000 & 91.130--91.740\\
 & Epoch 100 & 100 & 3 & 100.000 & 90.480--91.340\\
 & Epoch 150 & 150 & 3 & 100.000 & 90.010--91.020\\
 & Epoch 200 & 200 & 3 & 100.000 & 89.720--91.300\\
 & Epoch 250 & 250 & 3 & 100.000 & 90.190--91.730\\
 & Endpoint & 500 & 3 & 100.000 & 91.770--93.050\\
\bottomrule\end{tabular}\end{table}

\FloatBarrier
\begin{table}[!htbp]\centering
\caption{Controlled training: train and test accuracy at declared analysis stages and descriptive checkpoints. Entries are ranges across the available seeds; epochs may differ by seed. Unreached milestones have no row.}\label{tab:perf-5-0}
\small\setlength{\tabcolsep}{4pt}
\begin{tabular}{llrrrr}\toprule
Model & Stage & Epoch & Seeds & Train (\%) & Test (\%)\\\midrule
Margin $\lambda=0.01$ & Init. & 0 & 3 & 8.943--11.245 & 9.150--11.510\\
 & Early fit & 2--5 & 3 & 90.418--95.242 & 90.270--94.730\\
 & Epoch 50 & 50 & 3 & 99.385--99.462 & 96.580--96.860\\
 & Epoch 100 & 100 & 3 & 99.512--99.538 & 96.380--96.540\\
 & Epoch 150 & 150 & 3 & 99.437--99.582 & 96.190--96.580\\
 & Epoch 200 & 200 & 3 & 99.540--99.633 & 96.370--96.480\\
 & Epoch 250 & 250 & 3 & 99.575--99.650 & 96.100--96.590\\
 & Endpoint & 500 & 3 & 99.672--99.713 & 96.140--96.380\\
\midrule
Margin $\lambda=0.1$ & Init. & 0 & 3 & 8.943--11.245 & 9.150--11.510\\
 & Early fit & 2--5 & 3 & 90.583--93.767 & 90.680--93.420\\
 & Epoch 50 & 50 & 3 & 98.510--98.707 & 95.950--96.300\\
 & Epoch 100 & 100 & 3 & 98.735--98.935 & 95.660--96.030\\
 & Epoch 150 & 150 & 3 & 98.825--98.982 & 95.720--95.990\\
 & Epoch 200 & 200 & 3 & 98.892--99.045 & 95.520--95.940\\
 & Epoch 250 & 250 & 3 & 98.917--99.058 & 95.570--95.840\\
 & Endpoint & 500 & 3 & 99.062--99.105 & 95.150--95.710\\
\bottomrule\end{tabular}\end{table}

\FloatBarrier
\begin{table}[!htbp]\centering
\caption{Fitting and parameter scale at epoch 500 for the width-64 control and interventions. Ranges over three seeds. Training MSE excludes the regularization term.}\label{tab:perf-controlled-scale}
\small\begin{tabular}{lrrrr}\toprule
Arm & Train (\%) & Test (\%) & Training MSE & $\|\theta\|_2$\\\midrule
Control & 99.897--99.960 & 96.740--96.870 & 7.5e-05--0.00028 & 54.2--54.3\\
Margin $\lambda=0.01$ & 99.672--99.713 & 96.140--96.380 & 0.00048--0.00053 & 91.6--92.1\\
Margin $\lambda=0.1$ & 99.062--99.105 & 95.150--95.710 & 0.0017--0.0021 & 136.2--140.3\\
\bottomrule\end{tabular}\end{table}
\FloatBarrier
\begin{table}[!htbp]\centering
\caption{ViT fine-tuning, all recorded epochs, three seeds. Train accuracy is online over augmented minibatches; validation and test accuracy use the end-of-epoch network. Entries are seed ranges in percent.}\label{tab:perf-native-history}
\small\begin{tabular}{rrrr}\toprule
Epoch & Train (\%) & Validation (\%) & Test (\%)\\\midrule
1 & 54.048--55.856 & 86.400--87.067 & 85.952--86.238\\
2 & 91.321--91.826 & 91.644--92.311 & 91.635--92.587\\
3 & 95.730--95.886 & 93.111--94.178 & 92.778--93.048\\
4 & 97.562--97.616 & 92.800--93.956 & 92.619--93.667\\
5 & 98.697--98.835 & 93.244--94.711 & 93.429--94.492\\
6 & 99.207--99.321 & 93.689--94.222 & 93.762--93.952\\
7 & 99.670--99.700 & 94.178--94.844 & 94.365--94.952\\
8 & 99.850--99.892 & 94.978--95.333 & 94.952--95.397\\
9 & 99.946--99.994 & 95.067--95.289 & 94.857--95.270\\
10 & 99.982--99.988 & 94.800--95.644 & 95.000--95.397\\
11 & 99.988--99.994 & 94.889--95.511 & 95.032--95.317\\
12 & 100.000 & 94.889--95.467 & 95.048--95.333\\
\bottomrule\end{tabular}\end{table}
\FloatBarrier
\begin{table}[!htbp]\centering
\caption{Train/test accuracy of all inherited-Adam sibling endpoints shown in the connectivity comparison. Ranges pool both children across the available seeds; endpoints are at epoch 500.}\label{tab:perf-lmc-endpoints}
\small\begin{tabular}{llrrr}\toprule
Model & Spawn epoch & Children & Train (\%) & Test (\%)\\\midrule
MLP & 100 & 6 & 99.963--99.992 & 97.330--97.580\\
MLP & 150 & 6 & 99.987--99.998 & 97.360--97.680\\
MLP & 200 & 6 & 99.973--99.995 & 97.480--97.660\\
MLP & 250 & 6 & 99.987--99.997 & 97.220--97.660\\
MLP & 350 & 6 & 99.968--100.000 & 97.290--97.490\\
MLP & 450 & 2 & 99.987--99.995 & 97.450--97.550\\
CNN & 50 & 6 & 99.998--100.000 & 99.370--99.430\\
CNN & 100 & 6 & 99.995--100.000 & 99.280--99.450\\
CNN & 150 & 6 & 99.998--100.000 & 99.340--99.520\\
CNN & 200 & 6 & 99.998--100.000 & 99.330--99.440\\
CNN & 250 & 6 & 99.997--100.000 & 99.330--99.470\\
\bottomrule\end{tabular}\end{table}

\FloatBarrier

\end{document}